\pdfoutput=1
\documentclass[12pt]{cmuthesis}

\usepackage[backref,pageanchor=true,plainpages=false, pdfpagelabels, bookmarks, bookmarksnumbered]{hyperref}
\usepackage{xcolor}
\usepackage{color}
\hypersetup{colorlinks, linkcolor={blue!50!black}, citecolor={blue!50!black}, urlcolor={blue!50!black}}
\usepackage[utf8]{inputenc}
\usepackage[activate={true,nocompatibility},final,tracking=true,kerning=true,spacing=true,factor=1100,stretch=10,shrink=10]{microtype}
\usepackage{textcomp}
\usepackage{epigraph} 
\usepackage{times}
\usepackage{latexsym}
\usepackage{tabularx}
\usepackage{graphicx}
\graphicspath{ {images/} }
\usepackage{amsmath}
\usepackage{lmodern}
\usepackage{caption}
\usepackage[titletoc,title]{appendix}
\usepackage{soul} 
\usepackage{caption}
\usepackage{amsthm}        
\usepackage{amssymb}
\usepackage{makecell}
\usepackage{array,multirow}
\usepackage{booktabs}
\usepackage{CJKutf8}
\usepackage{makecell} 
\RequirePackage{filecontents}
\usepackage[strict]{changepage}
\usepackage{subcaption}
\usepackage{etoolbox}
\usepackage{fullpage}
\usepackage[numbers,sort]{natbib}
\theoremstyle{definition}
\newtheorem{definition}{Definition}[section]   
\theoremstyle{remark}

\usepackage[T1,T5]{fontenc}
\newcolumntype{b}{X}
\newcolumntype{t}{>{\hsize=.15\hsize}X}
\newcolumntype{y}{>{\hsize=.25\hsize}X}
\newcolumntype{s}{>{\hsize=.85\hsize}X}
\newcolumntype{a}{>{\hsize=1.05\hsize}X}
\newcolumntype{u}{>{\hsize=1.2\hsize}X}
\newcolumntype{l}{>{\hsize=1.5\hsize}X}
\newcolumntype{w}{>{\centering\hsize=2\hsize}X}
\newcolumntype{v}{>{\hsize=2.5\hsize}X}
\newcolumntype{z}{>{\centering\hsize=6\hsize}X}
\usepackage{textcomp}

\newenvironment{myindentpar}[1]{\begin{list}{}{\setlength{\leftmargin}{#1}}\item[]}
 {\end{list}}
\usepackage[ruled,vlined]{algorithm2e}
\usepackage{amsmath}
\usepackage{amssymb}
\usepackage{fancyhdr}
\usepackage[english]{babel}
\usepackage{amsthm}
\usepackage{mathtools}
\usepackage{bbm}

\usepackage{verbatim}
\usepackage{listings}

\newtheorem*{namedtheorem}{\theoremname}
\newcommand{\theoremname}{testing}

\theoremstyle{definition}

\usepackage[letterpaper,twoside,vscale=.8,hscale=.75,nomarginpar,hmarginratio=1:1]{geometry}
\AtBeginDocument{
  \let\oldlabel\label
  \let\oldref\ref
}
\newcommand{\addlabelprefix}[1]{%
  \renewcommand{\label}[1]{\oldlabel{#1-##1}}
  \renewcommand{\ref}[1]{\oldref{#1-##1}}
}
\newcommand{\removelabelprefix}{%
  \renewcommand{\label}{\oldlabel}
  \renewcommand{\ref}{\oldref}
}
\usepackage{upquote}

\usepackage[center,sc]{titlesec}
\usepackage{tocloft}
\usepackage{lettrine}   
\newcommand{\hsp}{\kern 1pt}    
\usepackage{fancyhdr}
\usepackage{xpatch}
\makeatletter
\xpatchcmd{\@endpart}{\vfil\newpage}{}{}{}
\xpatchcmd{\@endpart}{\newpage}{}{}{}
\makeatother

\begin{document} 
\frontmatter

\pagestyle{empty}

\title{{\bf Massively Multilingual \\Text Translation for\\ Low-Resource Languages
}}
\author{Zhong Zhou}
\date{CMU-LTI-23-014}
\Year{2023}
\trnumber{}

\committee{
  \vspace{6pt}
  \textit{Alexander Waibel}$^\dagger$$^\ddag$ (Chair) \\
  \textit{Alon Lavie}$^\dagger$ \\
  \textit{Graham Neubig}$^\dagger$ \\ 
  \textit{Jan Niehues}$^\ddag$ \\
  \vspace{6pt}
  $^\dagger$ Carnegie Mellon University \\
  $^\ddag$ Karlsruhe Institute of Technology \\
}

\support{}
\disclaimer{}


\maketitle
\thispagestyle{empty}

\begin{dedication} 
\centering
  To God.
\end{dedication}

\begin{quotepages}

\newlength\longest
\null\vfill
\settowidth\longest{\itshape In this job, not in some other, God looks for faithfulness.}
\centering
\parbox{\longest}{%
\centering
  \raggedright{
This job has been given to me to do. \\
Therefore, it is a gift. \\
Therefore, it is a privilege. \\
Therefore, it is an offering I may make to God. \\
Therefore, it is to be done gladly, if it is done for Him. \\
Here, not somewhere else, I may learn God’s way. \\
In this job, not in some other, God looks for faithfulness. \\ 
  \par\bigskip}
  \raggedleft\textit{Elisabeth Elliot}\par}
\vfill\vfill
\end{quotepages}

\pagestyle{plain} 

\begin{abstract}\label{big:abstract}
\vspace{6pt}
Translation into severely low-resource languages has both the cultural goal of saving and reviving those languages and the humanitarian goal of assisting the everyday needs of local communities that are accelerated by the recent COVID-19 pandemic. 
In many humanitarian efforts, translation into severely low-resource languages often does not require a universal translation engine, but a dedicated \textit{text}-\textit{specific} translation engine. 
For example, healthcare records, hygienic procedures, government communication, emergency procedures and religious texts are all limited texts. 
While generic translation engines for all languages do not exist, translation of multilingually known limited texts into new, low-resource languages may be possible and reduce human translation effort. 
We attempt to leverage translation resources from rich-resource languages to efficiently produce best possible translation quality for well \textit{known texts}, which are available in multiple languages, in a new, low-resource language. 

To achieve this efficiency, we translate a closed text that is known in advance and available in multiple source languages into a new and low-resource language. 
Despite the challenges of little data and few human experts, we build methods to promote cross-lingual transfer, leverage paraphrase diversity, address the variable-binding problem, measure language similarity, build efficient active learning algorithms for learning seed sentences, activate knowledge in large pretrained models and produce quality translation with as small as a few hundreds lines of low-resource data. Working with extremely small data, we demonstrate that it is possible to produce useful translations for machines to work alongside human translators to expedite the translation process, which is exactly the goal of this thesis. 

To reach this goal, we argue that in translating a closed text into low-resource languages, generalization to out-of-domain texts is not necessary, but generalization to new languages is. Performance gain comes from massive source parallelism by careful choice of
close-by language families, style-consistent corpus-level paraphrases within the same language and strategic adaptation of existing large pretrained multilingual models to the domain first and then to the language. Such performance gain makes it possible for machine translation systems to collaborate with human translators to expedite the translation process into new, low-resource languages.

\end{abstract}

\begin{acknowledgments}
\vspace{6pt}
Many thanks to my advisor and all my committee members and mentors who has helped me grow over the years. This thesis is completed through much help and support from our scientific community that I am deeply grateful to. 

I would like to thank my advisor Alex Waibel. Alex is an insightful researcher, an experienced entrepreneur and a proficient writer. I am grateful that he shares my research goal. Alex has generously funded the entire work, has helped me with academic writing and has purchased multiple machines for our research over the years. His support is pivotal in this thesis. 

I am thankful to my committee members and my collaborators, Alon Lavie, Graham Neubig, Jan Niehues, Matthias Sperber, and Mark Bean. I am deeply grateful to Alon's sharing of wisdom on growing scientific career on Machine Translation evaluation, Graham's help with learning time management and prioritization skills, Jan's insights in prioritizing on key results in paper writing, Matthias' consistent support and Mark's expertise in Quechuan languages. 

I am deeply grateful to the support of Carolyn Rosé and Jamie Callan. Both of them share wisdom and strengthen me to grow and mature in research. I will pass on their kindness and wisdom to others.

We stand on the shoulder of giants, and I would like to thank all brilliant minds in the fields who help me grow over the years. I want to thank Ramayya Krishnan, Tom Mitchell, Alan Black, Rita Singh, Lori Levin, Uri Alon, David Mortensen, Rema Padman, Rahul Telang, Brian K. Kovak, David Choi, Steven Shreve, Larry Wassermann and many others for their advice and mentorship. 

I am so thankful to Angela Lusk, Jonny Cagwin and Suzie Laurich-McIntyre. They helped me learn and grow over the years. They are exemplary in their kindness, wisdom, character, strength and their dedication to student well-being. 
I want to thank Kevin Haworth, Keely Austin, Michael Laudenbach, Rose Chang, Laura DeLuca, Elizabeth Dietrich, Leah Yacknin-Dawson, Christian Hallstein and Jessica Hsu for improving my academic writing skills. I want to thank Stacey Young, Mary Jo Bensasi, Jessica Majuire for their kindness and help. 

Thanks to family and friends especially Christian and Shirley Hallstein, Paul and Sharon Johnston, Krissy Geffel, Ong Ai Boon, Jimmy and Valerie Williams, Cammie Dunaway, Linda and Fred Griffin, Shannon Libengood, Kristen Emrick, Jessica Hsu, Abigail Holizna, Janice Turner, Bri Saleone, Michael and Denise Danko, Jeff Bergeson, Brenda Miller, Liam Ain Levi, Jamie Ian Joel, Ryan Len Reese and the Schöpfle family, and many who help me grow.      
\end{acknowledgments}

\begin{publications}
\vspace{6pt}
Parts of this thesis have previously appeared in the following publications: 

\begin{enumerate}

\item{Zhong Zhou, Matthias Sperber, and Alex Waibel. Massively parallel cross-lingual learning
in low-resource target language translation. In \textit{Proceedings of the 3rd conference on
Machine Translation. Association for Computational Linguistics}, 2018.}
\item{Zhong Zhou, Matthias Sperber, and Alex Waibel. Paraphrases as foreign languages in
multilingual neural machine translation. In \textit{Proceedings of the Student Research Workshop
at the 56th Annual Meeting of the Association for Computational Linguistics}, 2019.} 
\item{Zhong Zhou and Alex Waibel. Family of origin and family of choice: Massively parallel
lexiconized iterative pretraining for severely low resource text-based translation. In \textit{Proceedings of the 3rd Workshop on Research in Computational Typology and Multilingual NLP in
the 20th Conference of the North American Chapter of the Association for Computational
Linguistics on Human Language Technologies}, 2021. }
\item{Zhong Zhou and Alex Waibel. Active learning for massively parallel translation of constrained text into low resource languages. In \textit{Proceedings of the 4th Workshop on Technologies
for Machine Translation of Low Resource Languages in the 18th Biennial Machine Translation Summit}, 2021.}
\item{Zhong Zhou, Jan Niehues, and Alex Waibel. Train global, tailor local: Minimalist multilingual translation into endangered languages. In \textit{Proceedings of the 6th Workshop on
Technologies for Machine Translation of Low-Resource Languages (LoResMT) of the 17th
Conference of the European Chapter of the Association for Computational Linguistics},
2023.}
\end{enumerate} 
\end{publications}

\tableofcontents
\preto\listoffigures{%
  \clearpage
  \csname phantomsection\endcsname
  \addcontentsline{toc}{section}{\listfigurename}%
}
\preto\listoftables{%
  \clearpage
  \csname phantomsection\endcsname
  \addcontentsline{toc}{section}{\listtablename}%
}

\clearpage
\removelabelprefix 
All figures, graphs and visuals in this thesis except photographs are created by the author. 
All photographs except those in Figure 1.5, Figure 2.3, and Figure 7.1 are taken by Mark Bean in Peru, reproduced with permission. Figure 1.5, Figure 2.3, and Figure 7.1 are provided by Mark Bean, reproduced with permission. Photographs with people are included with the permission of the Quechuan language communities involved. 
In all photograph captions, "Panao" refers to Panao Quechua, "Sihuas" refers to Sihuas Quechua and "Margos" refers to Margos-Yarowilca-Lauricocha Quechua. For permissions, please contact the author. 

\listoffigures
\listoftables

\mainmatter

\chapter{Introduction}\label{big:intro}
\addlabelprefix{0}

\epigraph{``To have another language is to possess a second soul.''}{\textit{Charlemagne}}

\begin{figure*}[t]
  \centering
  \includegraphics[width=.9\linewidth]{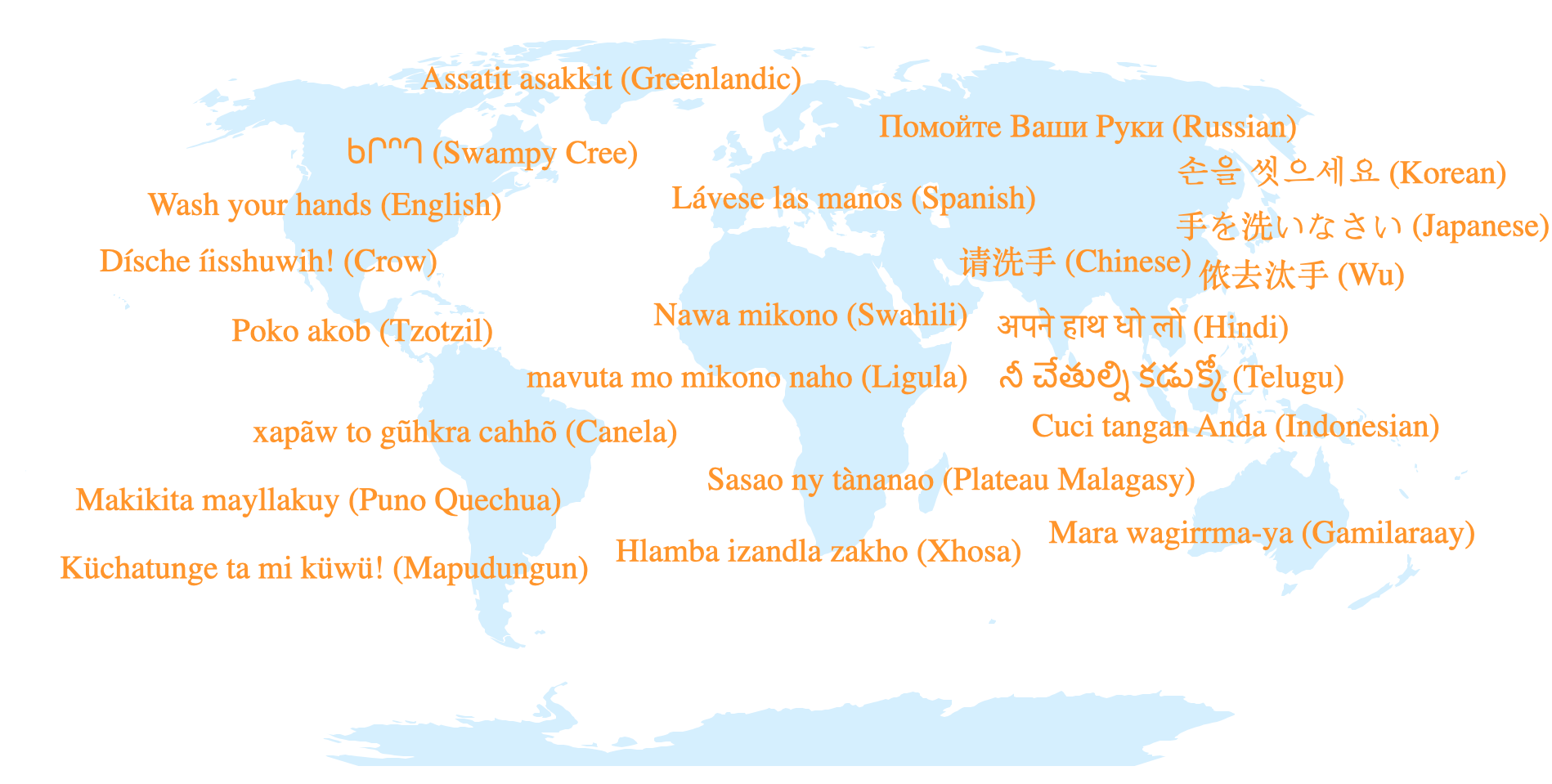}
  \caption{``Wash your hands'' in world languages \citep{eberhard2021ethnologue}.} \label{fig:handwashing}
\end{figure*}

\lettrine{T}{ranslation into severely low resource languages} has both the cultural goal of saving low-resource languages and the humanitarian goal of assisting the everyday needs of local communities that are accelerated by the recent COVID-19 pandemic. 
In many humanitarian efforts, translation into severely low resource languages often does not require a universal translation engine, but a dedicated \textit{text}-\textit{specific} translation engine. 
For example, healthcare records, hygienic procedures, government communication, emergency procedures and religious texts are all limited texts. 
Translation of limited texts have many real-world applications. One such application is the translation of 
water, sanitation and
hygiene (WASH) guidelines to protect
Indian tribal children against
waterborne diseases and more recently COVID-19 infections, introducing
earthquake preparedness techniques to Indonesian
tribal groups living near volcanoes
and delivering information to the disabled or the
elderly in low-resource language communities in Uganda 
\cite{reddy2017water, mulumba2014perceptions, anastasiou2010translating, perry2017treasure}.
These are useful examples of
translating a closed text known in advance
to the severely low-resource language. 
We show ``wash your hands'' in many languages in Figure~\ref{fig:handwashing}. 

While generic translation engines for all languages do not exist, translation of multilingually known limited texts into new, low-resource languages may be possible and reduce human translation effort. 
We attempt to leverage translation resources from rich-resource languages to efficiently produce best possible translation quality for well \textit{known texts}, which are available in multiple languages, for a new, severely low-resource language. 

To achieve better efficiency, we translate a closed text that is known in advance into
a new and severely low-resource language by leveraging massive
source parallelism. In other words, we make use of
available translations in many known source languages
to produce a good translation in severely low-resource language.
The source languages may be rich-resource, but may include other low-resource languages that have slightly more data or human expertise. 

In this problem setup, given a text that is multilingually available, we are interested in translating it into a new, low-resource language. The problem has three unique aspects that are different from traditional Machine Translation (MT) problems: 
\begin{enumerate} 
\item 
Our text is closed, not arbitrary as in traditional MT problems. 
\item 
Our text has multiple source languages with complete text translations while traditional MT is typically single-source.
\item 
Our text has little to no translation in the target low-resource language, while traditional MT assumes abundant data. 
\end{enumerate}

\section{Thesis Statement}
Machine Translation focuses on building a practical solution for human communication across cultures \cite{hutchins2001machine, popel2020transforming, van2003last, martinez2003human}. Translation into severely low-resource languages is ultimately an extremely small data problem. 
The methods that usually work in big data settings may not work for severely low-resource scenarios. Many large and evolved models may not perform as well as small and simple models in cases with little to no data. To overcome this challenge of extremely small data, our focus is to build a practical text-based Machine Translation engine that joins forces with human translators and uses minimal resources to expedite the translation process into severely low-resource languages. 

\begin{myindentpar}{1.5cm}
    \noindent \textbf{\MakeUppercase{Thesis Statement}}
    \textit{In translating a multilingually known limited texts 
    into a new, low-resource language,
    we argue that generalization to out-of-domain texts is not necessary, but
    generalization to new languages is necessary.
    Performance gain comes from massive source parallelism through the following:
    1) close-by language families, 2) style-consistent corpus-level paraphrases within the same language, 3) carefully-constructed linguistic closeness, 4) selective choice of active learning methods, and 5) strategic adaptation of existing large pretrained multilingual models to the domain first and then to the language. Such performance gain makes it possible for machine translation systems to collaborate with human translators to expedite the translation process into new, low-resource languages.
    }
\end{myindentpar}  

While the industry trend is to move towards bigger models with bigger data, our approach uses fewer languages, smaller data and minimal expert efforts. Given that expert efforts are mostly compensated or measured by the number of translations or edits,  
this saves computation power and resources. Therefore, this saves time and money, while improving translation performance. Saving time and money helps low-resource language communities to thrive on limited resources. 
Furthermore, with adequate scheduling 
and suitable model training on the whole text of multiple
source languages, it is possible to build a
sufficiently good translation model based on little
data in a new and severely low-resource language. 
This does not mean that we can build a universal interlingua
using such small data as the good translation results produced
by learning about the text may not generalize to other texts.
Nevertheless, generalization to other texts is desirable but not
necessary in our goal of producing practical and high quality translation
of the given closed text in our research problem. In this thesis, instead of focusing on translating any text into any language, we focus on the practical goal of translating a given text into a new, low-resource language. 

\begin{figure*}[t]
  \centering
  \includegraphics[width=0.9\linewidth]{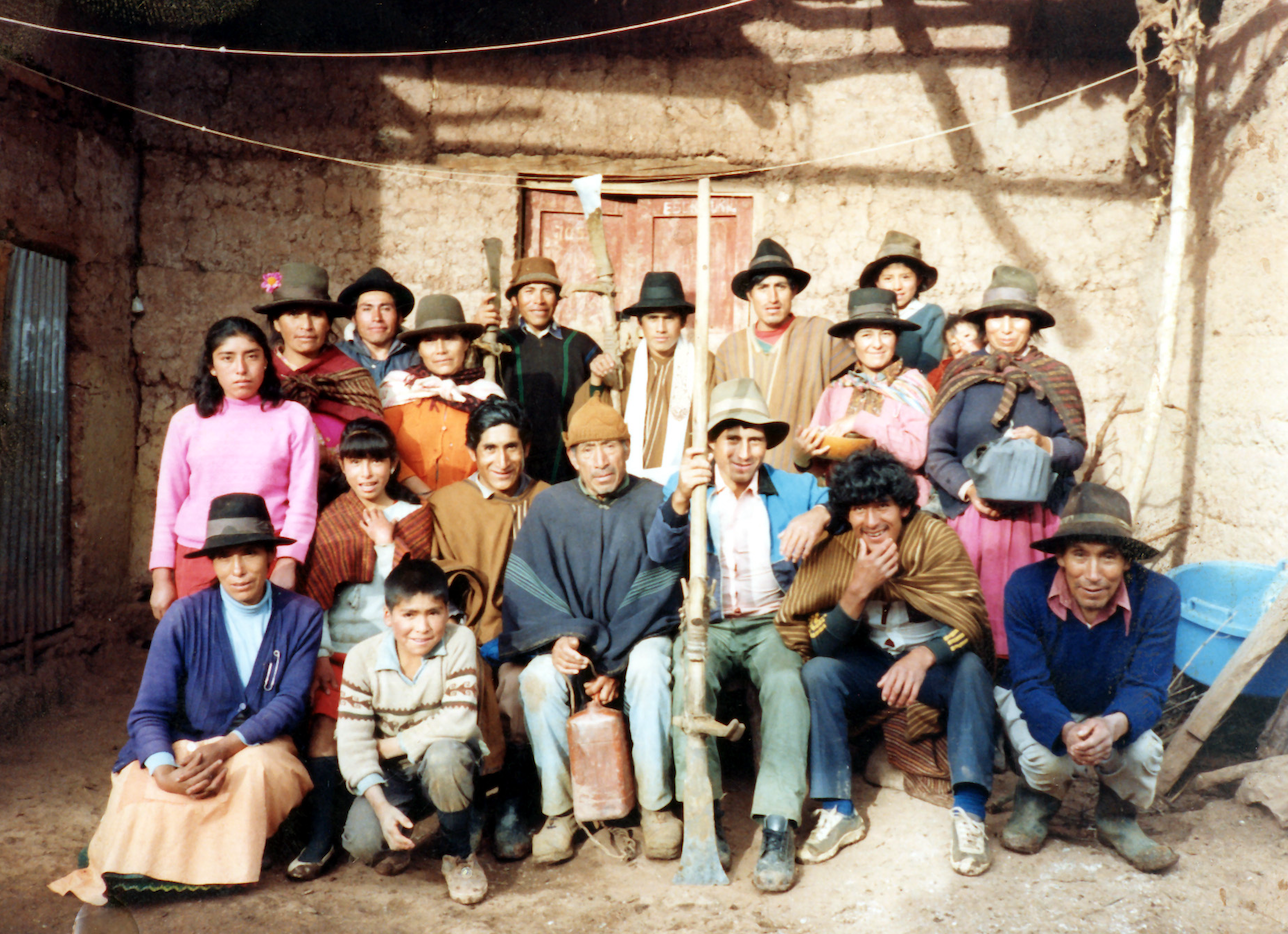}
    \caption{Quechuan language community in Peru. Photograph by Mark Bean. }
    \label{fig:quechuan_community}  
\end{figure*}

\section{Thesis Overview}
\begin{figure*}[t]
  \centering
  \includegraphics[width=1\linewidth]{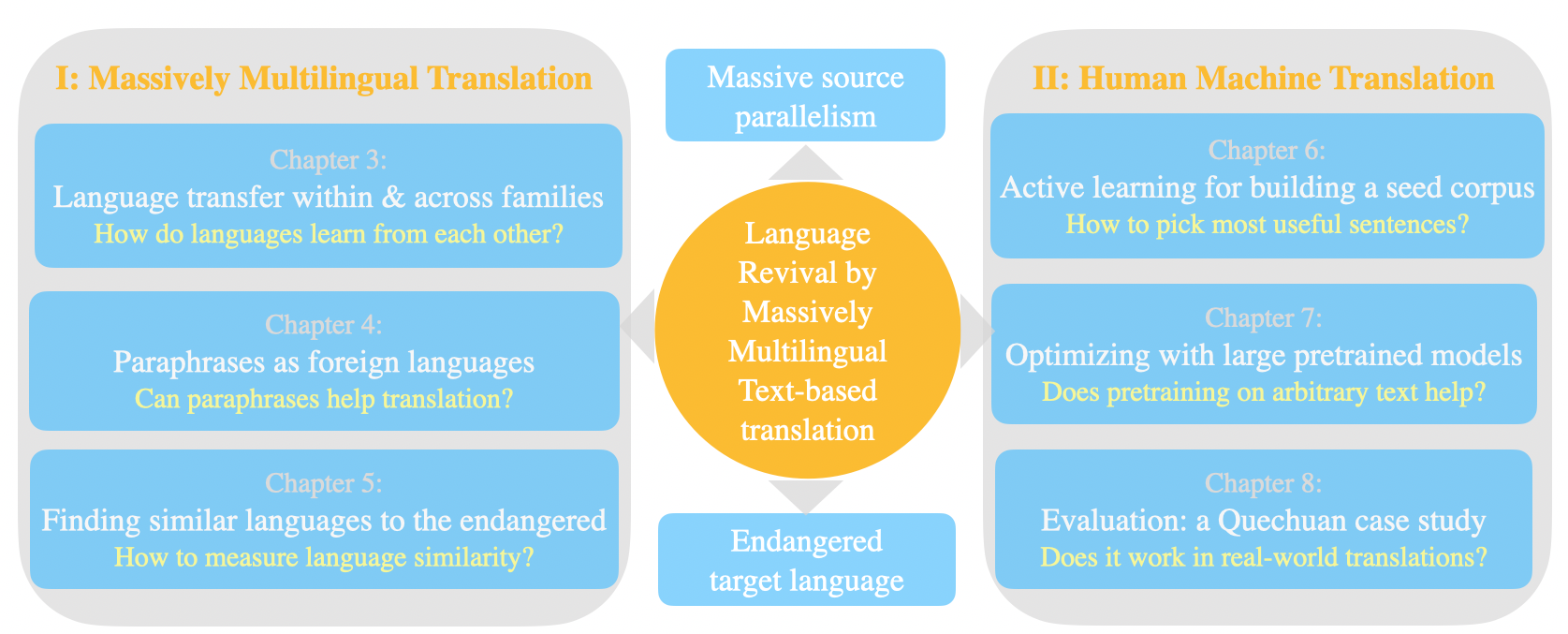}
    \caption{Overview of the work done as part of this thesis. }
    \label{fig:overview}  
\end{figure*}

\removelabelprefix 
To realize the goal of expediting the translation process of a multi-source text into new, low-resource languages, we show an overview of the work done as part of this thesis and how they are contributing to our main goal in Figure~\ref{fig:overview}. 
Following this introduction, we examine related existing research work in Chapter \ref{big:lit}. After literature review, we then present this thesis in two main parts: massively multilingual translation (Chapter \ref{big:family}, Chapter \ref{big:paraphrase} and Chapter \ref{big:ipml}), and human machine translation (Chapter \ref{big:active}, Chapter \ref{big:large} and Chapter \ref{big:confidence}). 
We focus on a case study in Quechuan language family for applying the methods built in this thesis to the real-world translation in Chapter \ref{big:confidence}. 
All these chapters contribute to the main goal of this thesis: translation of a multilingually known limited text into a new, low-resource language. 
To conclude, Chapter \ref{big:conclude} summarizes our main contributions and explores future research directions and opportunities in this research space.  

We focus on two parts of this thesis: massively multilingual translation (Chapter \ref{big:family}, Chapter \ref{big:paraphrase} and Chapter \ref{big:ipml}), and human machine translation (Chapter \ref{big:active}, Chapter \ref{big:large} and Chapter \ref{big:confidence}) and give an overview of these two parts below. 
\begin{itemize}
\item{\textbf{Massively multilingual translation} (Part \ref{part:part1})}: We examine how source parallelism benefits translation of a given text into new, low-resource languages through multilingual training. In Chapter \ref{big:family}, we build cross-lingual transfer both within a given language family and also across different language families. We find that in practice, training on two closely related language families or equivalently around ten closely related languages is often enough for translating to a given low-resource language. We also propose an order-preserving named entity translation mechanism to resolve the variable binding problem and produce high quality lexiconized translations under severely low-resource scenarios. When data of similar languages is not available, we may leverage different translations of the same text in the same language as described in Chapter \ref{big:paraphrase}. We treat paraphrases as foreign languages, and train on corpus-level paraphrases to improve translation performance. We find that our multi-paraphrase translation models improve performance better than multilingual models and improves the sparsity issue of rare word translation as well as diversity in lexical choice. When paraphrase information is also not available, we may build our own linguistic distance metric based on translation distortion, fertility and performance. In Chapter \ref{big:ipml}, we propose a method, \textit{Iteratively Pretrained Multilingual Order-preserving Lexiconized Transformer}, to train on low-resource language data. We push the limit by using only $\sim$1,000 lines ($\sim$3.5\% of the entire text) to translate the whole text and achieve good translation performance. 

\begin{figure*}[t]
  \centering
  \includegraphics[width=0.9\linewidth]{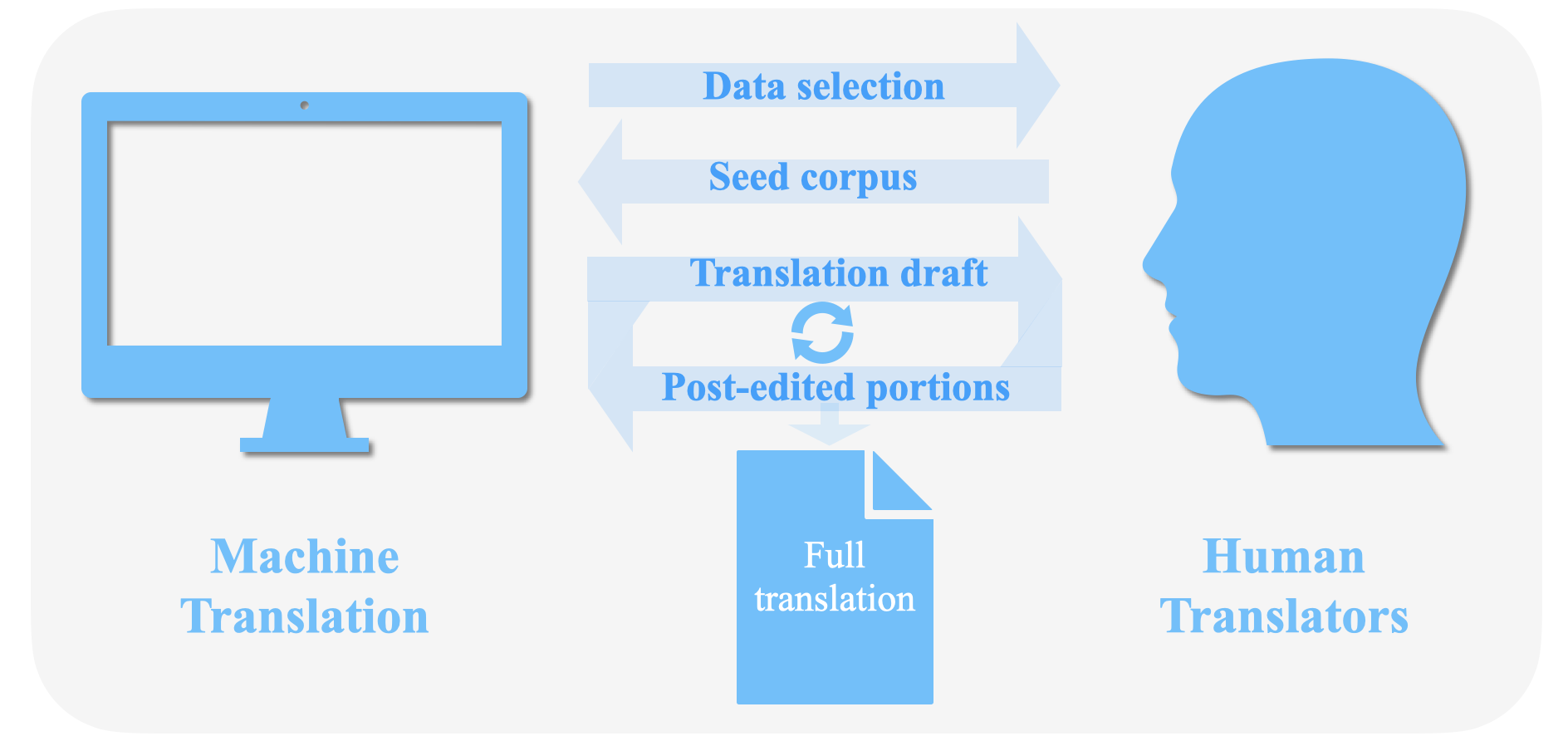}
    \caption{Human machine translation process. }
    \label{fig:hmtoverview}  
\end{figure*}

\item{\textbf{Human Machine translation} (Part \ref{part:part2})}: Having examined source parallelism, we build a human machine translation workflow algorithm for machine translation systems to collaborate with human translators to expedite the process.  
Our proposed human machine translation is not to
replace the human translators with machine translation systems, but instead,
to get the best of both worlds
as shown in Figure \ref{fig:hmtoverview}. In our translation process, human translators are informed by machine sentence ranking through active learning to produce a seed corpus. Machine systems then use this seed corpus to produce a full translation draft. Human translators post-edit the draft, and feed new data to machines each time they finish post-editing a portion of the text. In each iteration, machines produce better and better drafts with new data, and human translators find it easier and faster to post-edit. Together they complete the translation of the whole text into a severely low-resource 
language. We first develop various active learning methods on known languages and transfer ranking to the new, low-resource language in Chapter \ref{big:active}. Secondly, we activate the knowledge of large multilingual models by proposing multilingual and multi-stage adaptations through different training schedules in Chapter \ref{big:large}; we find that adapting pretrained models to the domain and then to the low-resource language works best. Thirdly, we aggregate scores from 115 languages to provide a universal ranking and increase robustness by the \textit{relaxed memoization} method. In Chapter \ref{big:confidence}, having examined both source parallelism and human machine translation workflow, we evaluate our work in all previous chapters by translating it into practical use in a case study in Quechuan language family in extensive collaboration with Mark Bean and his translation group with in-depth knowledge of various Quechuan languages. We find that machine translation performance is significantly positively correlated with language similarity. The more connected a language is, the higher the performance. Moreover, we find that decluttering poorly-connected languages improves translation score. Based on this finding, we show effectiveness of our models through good results in translation into a new, low-resource language called Sihuas Quechua. 
\end{itemize}

Finally, we summarize our contributions and discuss future research opportunities in translating into new, low-resource languages in Chapter \ref{big:conclude}.  

\section{Practical Goal Setting}
Having understood the overview of this thesis, we would like to clarify the goal of this work  
and how to make it practical and attainable given the real-life constraints of low-resource languages. 

In essence, our goal is to minimize human translation and post-editing  efforts required to generate a full publishable-standard translation of a given text. We aim to minimize human translators' efforts in both the translation process of the seed corpus sentences and the post-editing process of the subsequent iterations. In other words, we want to make translation and post-editing work easier and smaller for human translators through automation. 

To realize this goal, ideally we want to hire a large number of human translators, measure and compare the resources (time and money) used to translate the same text into a target low-resource language that does not have any translations of the text. To show that this thesis has achieved the goal of reducing human translation efforts, optimally we need to measure and compare time taken and money spent in two translation scenarios: 
\begin{enumerate}
\item 
Baseline: total time taken and total amount of money paid when human translators translate the text to meet publishable standards.
\item 
Total time taken and total amount of money paid when human translators work with machine translation systems to translate and post-edit the text to meet publishable standards through the human machine translation algorithm introduced in this thesis.
\end{enumerate} 

However, this ideal solution is unrealistic especially in large translation projects. Large translation projects in real-life usually takes decades, if not centuries of work. For example, in the case of Bible Translation, the average project cost of a complete written Bible is \$937,446 and the average time of completing a Bible translation with sufficient resources is 15.8 years \citep{sagamore2022trans}. This is clearly not realistic and beyond the scope of this thesis. 

To set more practical goals, we treat the translation process of the seed corpus and the post-editing process of the subsequent iterations separately, and use the following two sub-goals as the proxy sub-goals for minimizing time and money used for the translation project:
\begin{enumerate}
\item
To minimize human translation effort at the translation process of the seed corpus, we optimize and minimize the amount of sentences to be used to construct seed corpus as a proxy. 
\item
To minimize human post-editing process of the subsequent iterations, we maximize the quality and utility of MT-generated translation of the full text and optimize translation efficiency. 
\end{enumerate}

To understand why we choose these two sub-goals as proxies, we first need to understand that most of the human translators are paid by the number of words they translate or post-edit \citep{bloodgood2014bucking, eck2008developing, tomanek2009semi}. The rate they translate a word is different from the rate they post-edit a word \citep{eck2008developing}. Therefore, the total human translation cost is entirely tied to the number of translated words and post-edits human translators need to perform to meet publishable standards. Firstly, minimizing the amount of sentences used to construct seed corpus directly reduces the number of words human translators need to translate. Therefore, it serves as a good proxy for minimizing human translation efforts of the seed corpus. Secondly, maximizing translation performance of the MT-system minimizes the number of edits required for generating a full publishable-standard translation of the given text \citep{bentivogli2015evaluation, shih2021re, bentivogli2018machine, turchi2017continuous}. Since time and money saved is directly linked with the number of edits saved, maximizing translation performance is an appropriate proxy sub-goal for 
minimizing human post-editing processes. 

Using the proxy sub-goals, we transform our goal of minimizing human translation efforts required to generate a full translation of the given text into 
two practical proxy sub-goals as the following: 
\begin{enumerate}
\item 
Optimizing and minimizing the amount of sentences used to construct seed corpus.
\item 
Maximizing the quality and utility of MT-generated translations of the full text and optimizing translation efficiency.
\end{enumerate}

The first sub-goal of minimizing the seed corpus serves as a proxy in Chapter \ref{big:active} to minimize the human translation efforts in the creation of the seed corpus, while the second sub-goal of maximizing translation performance serves as a proxy in Chapter \ref{big:large} to minimize human translation efforts in the post-editing process during the subsequent iterations. 

To measure translation performance, our primary automatic metric in this thesis is chrF \citep{popovic2015chrf}. We choose chrF for accuracy, fluency and expressive power in morphologically-rich languages \citep{papineni2002bleu}. We use the metric chrF for the beginning and the conclusion of this thesis to motivate and summarize our main contributions of this paper. In addition to our main evaluation metric chrF, in order to have a comprehensive understanding of each chapter, we give a wide range of evaluation metrics to supplement our understanding of translation performance through BLEU , characTER, COMET score, and BERTscore \citep{wang2016character, post-2018-call, zhang2019bertscore, stewart-etal-2020-comet, rei2021mt}. Moreover, we explore human evaluation methods in addition to automatic evaluation methods. For detailed analysis, we prioritize BLEU in Massively multilingual translation (Part \ref{part:part1}) as we mainly work with European languages while we prioritize chrF in Human Machine translation (Part \ref{part:part2}) as we mainly work with morphologically-rich low-resource languages. Overall, we will use our metric chrF to conclude this thesis. 

\begin{figure*}[t]
  \centering
  \includegraphics[width=0.8\linewidth]{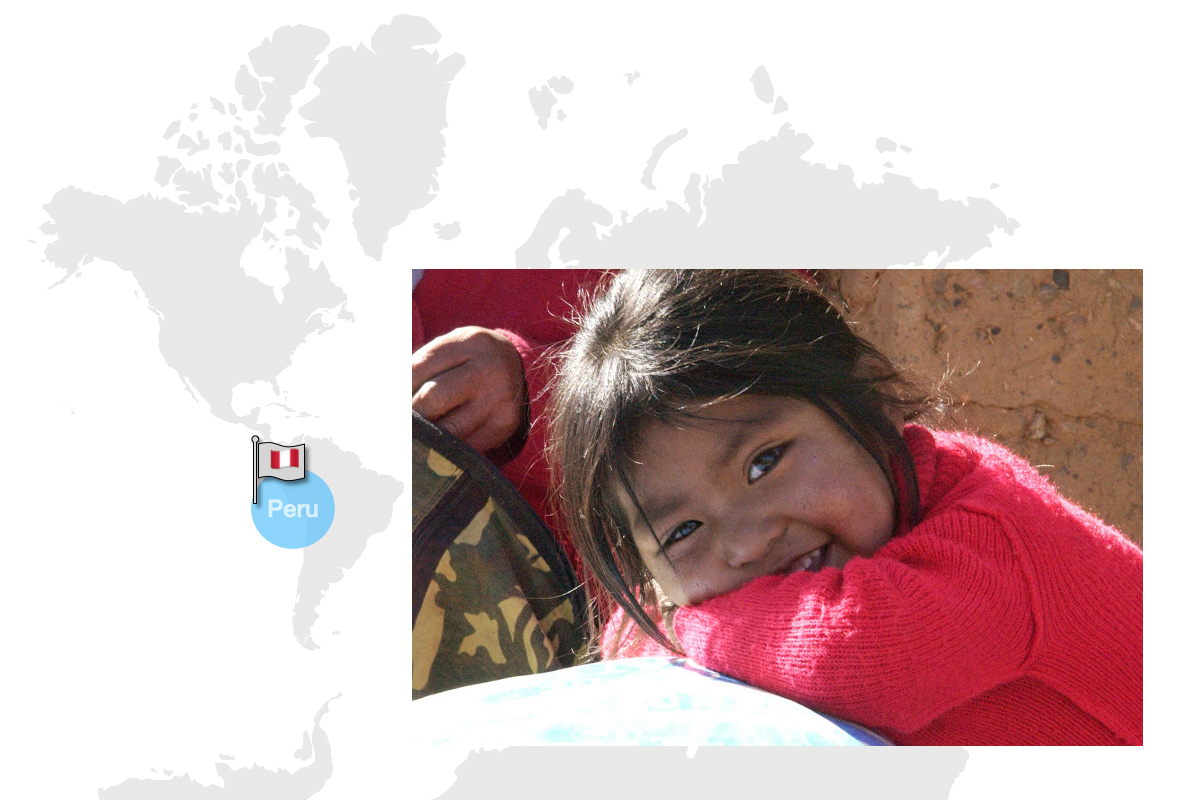}
    \caption{A practical application of this thesis in Peru. Photograph provided by Mark Bean. }
    \label{fig:peru_location}  
\end{figure*}

\section{Thesis in Practice}

As we have discussed in the Thesis Overview, we integrate this thesis into real-world scenarios by examining a case study in Quechuan language family in Chapter \ref{big:confidence}. This provides practical insights in understanding how to use this thesis for practical situations. We will discuss the case study in Quechuan language family in detail at the end of our thesis. However, we would like to give a glimpse of this case study before we dive into the details of the different chapters of this thesis, so that we have a more intuitive understanding of the problem we are solving. 

The Quechuan language family, also called Runasimi (``people's language'') by local communities, is a varied group of languages covering a wide region of Peruvian Andes in South America, extending from Colombia, Ecuador, Peru, Bolivia to northwest Argentina \citep{luykx2016communicative, howard2011quechua}. Its history traces back to the Inca Empire and there is broad spectrum of sociolinguistic diversity among Quechuan languages throughout the Spanish colonial history of the area \citep{durston2007pastoral}. Many are low-resource languages. 

We make practical use of findings of this thesis to translate into a few low-resource languages in Quechua in Chapter \ref{big:confidence}. We provide extensive analysis to answer the following questions: 
\begin{enumerate}
\item Under what conditions does our method work? 
\item Under what conditions does our method fail to work? 
\item When it works, does it provide real help to human translators working in the field? 
\item When it does not work, are there ways we can improve user experience for human translators? 
\end{enumerate}
Through the extensive analysis to answer the above questions, we find that machine translation performance is significantly positively correlated with language similarity. The more connected a language is, the better it is to translate into this language. Using this result, we employ our thesis in practice and achieve good performance for translation into a new, low-resource language called Sihuas Quechua. We give a glimpse of this thesis in practice before we dive into the details of each chapter, and we will look closely at this Quechuan case study after the detailed discussion of this thesis in Chapter \ref{big:confidence}. Through this case study, we show a few opportunities for further research that this thesis presents. 

\section{How to Read This Thesis}  
Given the overview, we aim to make this thesis accessible and inclusive to readers coming from diverse research backgrounds for the many potential collaborative opportunities in this research space. 

Following this introduction, we give an overview of the research landscape by examining related existing research done in the space of translation into low-resource languages in Chapter \ref{big:lit}. For Chapter \ref{big:family} to Chapter \ref{big:confidence}, each chapter is relatively self-contained and readers can read them not in sequence. However, we suggest to read Chapter \ref{big:family} and \ref{big:paraphrase} first before Chapter \ref{big:ipml} to gain an understanding what could be done with language closeness and why it is important to measure language closeness when such information is not available. We also recommend to read Chapter \ref{big:active} and Chapter \ref{big:large} together as most of the work in these two chapters have previously appeared in publications together. Lastly, we recommend reading Chapter \ref{big:confidence} after having completed all the previous chapters as this chapter leads us from the comfort zone of academic research to enter the real-world translation process, which is exactly the place our work is most impactful.            
\removelabelprefix

\chapter{Literature Review}\label{big:lit}
\addlabelprefix{1}                           

\epigraph{``Every act of communication is a miracle of translation.''}{\textit{Ken Liu}}

\lettrine{T}{o understand the existing literature} in the space of translation into low-resource languages, 
we focus on examining different aspects of our translation task through related research works. We only show related works that are relevant to all parts of this thesis in this chapter. For related works that are only relevant to a particular chapter, we will introduce them in the specific chapter directly. 

In this chapter,  we show related works that are closely connected to all parts of this thesis in three 
broad categories: low-resource languages, machine translation, and translation in practice. These three categories are intricately linked to each other. Firstly, related works in ``low-resource languages'' show the broader impact of empowering low-resource language communities and facilitating information dissemination in these communities. Secondly, related works in ``Machine Translation'' demonstrate the state-of-the-art tools researchers have built that could be applied in the severely low-resource translation, including massively multilingual translation and large pretrained multilingual models. Finally, related works in ``translation in practice'' bridge the two worlds by putting MT tools into practical use of real-world translation. More specifically, we show the related works in the areas of human machine translation, active learning and post-editing. 

Having examined related works in low-resource languages, machine translation, and translation in practice, we conclude this chapter by introducing our research framework 
that is built based on existing literature, 
including the toolkits we use, our core datasets and evaluation metrics we employ. 

\section{Low-Resource Languages}

\subsection{Information Dissemination}
Interactive Natural Language Processing (NLP)
systems are classified into information assimilation,
dissemination, and dialogue
\cite{bird2020decolonising, ranzato2015sequence, waibel2008spoken, waibel1989modularity, waibel2016communicating}.
Assimilation and dissemination have very broad definition and our definition below
assumes the reference point of rich-resource communities and discusses them
only in the context of rich and low-resource languages.
\textit{Information assimilation} involves information
flow from low-resource to rich-resource
language communities while \textit{information dissemination}
involves information flow from rich-resource to low-resource language communities. 
It helps low-resource language communities to make better-informed and autonomous decisions. 
Taken together, they allow \textit{dialogue} and interaction of
different groups at eye level.
Most research is on information assimilation with
examples span from urgent earthquake detection
to infectious disease surveillance
\cite{berard2020multilingual, earle2012twitter, brownstein2008surveillance}.
Few work on dissemination is on information dissemination with examples
ranging from introducing disaster prevention techniques
to delivering information to the disabled and the
elderly in low-resource language communities 
\cite{ostling2017neural, zoph2016transfer, anastasopoulos2017spoken, adams2017cross, bansal2017weakly}.
Information dissemination is challenging because there is
little low-resource language data, much less parallel corpus, little funding, and few
human experts in training and evaluation
\cite{duong2016attentional, anastasopoulos2017spoken}.
Some low-resource languages have no
formal writing systems \cite{adams2017cross, bansal2017weakly}.
Note that in a broader definition of dissemination,
the field is largely under-researched even if the focus is not low-resource.
However, our discussion focuses largely on dissemination in 
the severely low-resource scenarios. 

\subsection{Low-Resource Languages}
A language is alive when many people speaks it 
and dies when no one speaks it. 
There are $\sim$7,139 languages in the world, 
unequally distributed across the world, 
with drastic differences in their number of speakers 
and vitality \citep{grenoble1998endangered, payne1999endangered, moseley2008encyclopedia}. 
More than half of these languages will die in the 
next 80 years \citep{austin2011cambridge, eberhard2021ethnologue}. 
Though there are cases where a language dies through war, genocide, natural disasters or infectious diseases, 
most languages dies while the speakers do not; the speakers 
either voluntarily or are forced to speak another mainstream 
language as part of the endangerment process that is deeply rooted 
in political, historical, social and economical reasons \citep{thomason2015endangered, maffi2002endangered}. 

A language needs attention when it is spoken by enough people that it could survive under favorable
conditions but few or no children are learning it
\citep{crystal2002language, kincade1991decline, wurm2001atlas}. 
Such languages may survive and thrive if they gain prestige, power and visibility \citep{crystal2002language}. 
Language preservation is therefore an intricate and complex matter that invites many different views \citep{ladefoged1992another, romaine2007preserving, hale1992endangered, austin2007endangered, dorian1993response}. 

\begin{figure*}[t]
  \centering
  \includegraphics[width=0.7\linewidth]{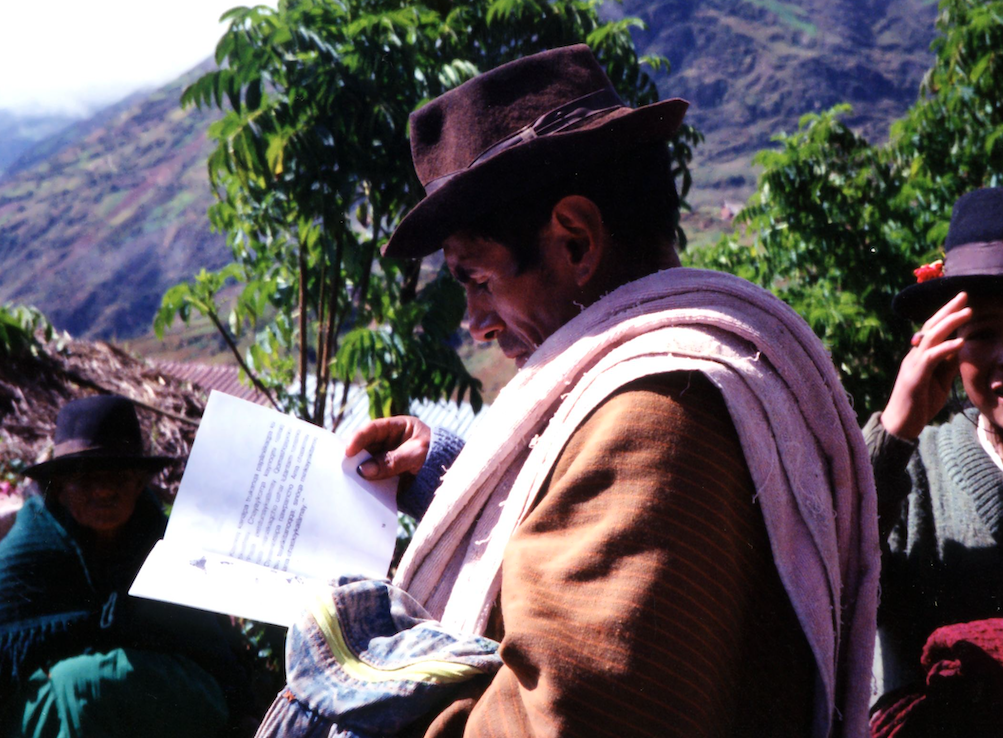}
    \caption{A native man in Peru reading translated text. Photograph by Mark Bean. }
    \label{fig:peru_read}  
\end{figure*}

\section{Machine Translation}

\subsection{Massively Multilingual Translation}
Machine polyglotism, training machines to be proficient
in many languages, is a new paradigm of multilingual NMT
\cite{johnson2017google, ha2016toward, firat2016multi, zoph2016multi, dong2015multi, gillick2016multilingual, al2013polyglot, tsvetkov2016polyglot, zhou2018paraphrases, aharoni2019massively, kocmi2020exploring}. 
The objective is to translate from any of the input languages to any of the 
output languages \cite {firat2016multi}.
Many multilingual NMT systems involve multiple encoders
and decoders \cite{ha2016toward}, and it is hard to combine
attention for quadratic language pairs bypassing
quadratic attention mechanisms
\cite{firat2016multi}.
In multi-source scenarios, multiple
encoders share a combined attention
mechanism \cite{zoph2016multi}. In
multi-target scenarios, every decoder
handles its own attention
with parameter sharing \cite{dong2015multi}.
Attention combination schemes include simple
combination and hierarchical combination
\cite{libovicky2017attention}.

Attentional Neural Machine Translation (NMT) is trained directly in an end-to-end
system and has flourished recently \cite{wu2016google, sennrich2016neural, ling2015character}.
The state-of-the-art of multilingual NMT is adding
source and target language labels in training
a universal model with a single attention scheme, and
Byte-Pair Encoding (BPE) is used at preprocessing stage \cite{ha2016toward}.
This method is elegant in its simplicity and its advancement in
low-resource language translation as well as zero-shot translation using
pivot-based translation scheme \cite{johnson2017google}.
However, these works have training data that increases
quadratically with the number of languages
\cite{dong2015multi, gillick2016multilingual},
rendering training on massively parallel corpora difficult.

\subsection{Large Pretrained Multilingual Models} 
As discussed before, the state-of-the-art multilingual machine translation systems 
translates from many source languages to many target languages 
\cite{johnson2017google, ha2016toward, firat2016multi, zoph2016multi, zhou2018paraphrases}. The bottleneck in building such systems lies in the limitations of computing power, 
data, and machines. The training data quadratically increase 
with the number of languages, rendering training on many languages 
difficult. Companies equipped with vast multilingual datasets and 
substantial computational capabilities have built and released some large 
pretrained multilingual models for researchers to work on 
\citep{liu2020multilingual, tang2020multilingual}.
These are very helpful for researchers to build on 
this pretrained model to further work in multilingual 
translation. M2M100, released by Facebook, is trained in 100 languages 
\citep{fan2021beyond, schwenk2019ccmatrix, el2019massive} and covers 
a few low-resource languages. DeltaLM, released by Microsoft, is pre-trained in 6 TB of
multilingual data from CC100, CC-Net, and Wikipedia in 
100 languages \citep{deltalm} and allows for flexible input languages. 
NLLB (No Language Left Behind), released by Meta AI, 
is trained on 200 languages \citep{costa2022no}. 
With more advances in the future, there will be 
models covering more languages. 

\subsection{Low-Resource Machine Translation}
Recent advances have succeeded in building multilingual
methods to translate from multiple rich-resource
languages to a new, low-resource language
\citep{johnson2017google, ha2016toward, firat2016multi, zhou2018massively, zhou2018paraphrases}.
Many have demonstrated good transfer learning to low-resource languages
as few as $\sim$4,000 lines \citep{lin2020pre, qi2018and}
and $\sim$1,000 lines \citep{zhou2021family} of data.
Transliteration can replace parallel corpora in translation \citep{karakanta2018neural}.
Many use simulated low-resource languages, some use real low-resource languages \citep{joshi2020state, wang2022expanding, blasi2021systematic}.

In addition to researching and training on extremely small data, 
some researchers work on zero-shot learning
\citep{neubig2018rapid, pham2019improving, philip2020monolingual, karakanta2018neural, zhang2020improving, chen2022towards, chen2021zero}. 
Zero-shot translation in severely low-resource settings
exploits the massive multilingualism, cross-lingual
transfer, pretraining, iterative back-translation
and freezing sub-networks
\cite{lauscher2020zero, nooralahzadeh2020zero, wang2020negative, li2020learn, pfeiffer2020mad, baziotis2020language, chronopoulou2020reusing, lin2020pre, thompson2018freezing, wei2020iterative, dou2020dynamic}.
However, zero-shot learning is volatile
and unstable, and is not suitable for tasks that requires high level of accuracy \citep{rios2020subword}, which is crucial in our translation goal. 
Therefore, instead of zero-shot learning, we choose to use extremely small data instead. 

\section{Translation in Practice}

\subsection{Human and Machine Translation}
Machine translation began about the same time as the first
computer \citep{hirschberg2015advances, popel2020transforming}.
Over the years,
human translators have different
reactions to machine translation advances,
mixed with doubt or fear \citep{hutchins2001machine}.
Some researchers study human translation taxonomy
for machine to better assist human
translation and post-editing efforts \citep{carl2011taxonomy, denkowski2015machine}.
Human translators benefit from machine assistance
as human individual bias and translation capacity limitations
are compensated for by large-scale
machine translation
\citep{koehn2009interactive, li2014comparison, savoldi2021gender, bowker2002computer, bowker2010computer, koehn2009process}.
On the other hand, machine translation
benefits from professional human translators'
context-relevant and culturally-appropriate
translation and post-editing efforts \citep{hutchins2001machine}.
Severely low-resource translation is a fitting ground for
close human machine collaboration
\citep{zong2018research, carl2011taxonomy, martinez2003human}.

\subsection{Active Learning}
Active learning has long been used
in machine translation
\citep{settles2012active, ambati2012active, eck2005low, haffari2009activeb, gonzalez2012active, miura2016selecting, gangadharaiah2009active, hu2021phrase}.
Random sampling and data selection through active learning
has been surprisingly powerful 
\citep{kendall1938randomness, knuth19913, clarkson1989applications, sennrich2015improving, hoang2018iterative, he2016dual, gu2018meta}. Some researchers choose to train
on simpler data with shorter sentences
through curriculum learning and
active learning \citep{eck2005low, platanios2019competence}.
The mathematician Donald Knuth uses the population
of Menlo Park to illustrate the value of random sampling
\citep{knuth19913}. The population of Menlo Park is approximately
the same as the number of Bible verses ($\sim$31,000).
Knuth claims that random sampling across the
city is likely to produce a better set of unbiased,
reliable and independent statistical samples than
choosing people working in the same building.

There is extensive research to beat random sampling by methods based on 
entropy \citep{koneru2022cost}, coverage and
uncertainty \citep{peris2018active, zhao2020active},
clustering \citep{haffari2009activea, gangadharaiah2009active},
consensus \citep{haffari2009activeb}, syntactic parsing
\citep{miura2016selecting}, mixture based on density and diversity
\citep{koneru2022cost, ambati2011multi, dou2020dynamic}, and learning to learn active learning strategies \citep{liu2018learning}.

\subsection{Post-editing}
Human post-editing helps neural MT systems to produce publishable materials \citep{denkowski2015machine}. Large MT systems often have issues with under-prediction, over-prediction, repetition, mislabelled data, and hallucination \citep{marcus2022deep,bommasani2021opportunities}. The state-of-the-art MateCat Tool uses translation memories (TMs) as a reference point to search for matches, exact or approximate, of the current TM segment for translation \citep{federico2014matecat}. It successfully integrates suggestions found TMs and helps human translators and provides a simple and useful interface for human translators to finish the post-editing process \citep{federico2014matecat}. Other tools include PET, CATaLog Online, and iterative interfaces \citep{sperber2016optimizing, aziz-etal-2012-pet, pal2016catalog}. Many evaluate these tools for productivity and quality gain \citep{guerberof2009productivity, o2014towards, vieira2011review, guerberof2013professional}. 

Some researchers also work on automatic post-editing to improve MT performance through online learning approach and neural programmer-interpreter method \citep{font2006automating, vu2018automatic}. The post-editing feedback is often used to  build adaptive MT systems in real time \citep{denkowski2014real}. 

In addition, there is also research down for confidence measure on the translations on how much post-editing is required. One such example uses the human-targeted Translation Edit Rate (HTER), on both the word, and the character level \citep{snover2006study, wang2016character}. 

\section{Research Framework}

After reviewing relevant literature on language revival, machine translation and translation in practice, We introduce a few tool kits that we use in this thesis, the core datasets that we work with, and the evaluation metrics we use. 

\subsection{Tools}

We use several natural language processing and translation tools for our translation systems. They are listed below. 

\begin{enumerate}
    \item{\emph{Fairseq}: a neural sequence learning framework on Pytorch for machine translation, language modelling and other generation jobs \citep{ott2019fairseq}. Fairseq allows us to work with a few large pretrained multilingual models and finetune those models to customize to our translation tasks. Fairseq is licensed under the MIT License. }
    \item{\emph{OpenNMT}: an open-source Neural Machine Translation (NMT) system \cite{klein2017opennmt}. It is very flexible with many severely low-resource languages. OpenNMT-py is the PyTorch version of the OpenNMT project. OpenNMT/OpenNMT-py is licensed under the MIT License. }
    \item{\emph{Sockeye}: an open-source neural machine translation framework on Pytorch \citep{hieber-etal-2020-sockeye}. Sockeye is licensed under the Apache License 2.0. } 
    \item{\emph{Googletrans}: a free and simple python library for Google Translate API. This serves as a way to partially evaluate languages that our team do not speak or cannot find native speakers in. When we evaluate the Google's English translation of our system translations into low-resource languages, we are aware that the final outputs introduce another layer of potential error from Google's system. This is therefore used with caution. This does not compete with automatic evaluation methods, and is not comparable with real human qualitative evaluation, but it serves as an imperfect way to understand the overall performance of the system. Googletrans is licensed under the MIT License. }
    \item{\emph{COMET}: a neural framework for MT evaluation \citep{rei2020comet}. COMET offers a neural way of training multilingual MT evaluation models that attains levels of correlation with human evaluation for some of the rich-resource languages. COMET is licensed under the Apache License 2.0. } 
    \item{\emph{MT Telescope}: a toolkit for comparing translations from different MT systems \citep{rei2021mt}. We use MT telescope to compare across multiple translation systems using a myriad of metrics including COMET above. MT-Telescope is licensed under the Apache License 2.0. }   
    \item{\emph{Compare-MT}: a toolkit for comparing outputs from multiple systems for language generation tasks including MT, summarization, dialog generation and others \citep{neubig2019compare}. Compare-mt is licensed under the BSD 3-Clause ``New'' or ``Revised'' License. }
    \item{\emph{Fast\_align}: a light-weight, quick, and unsupervised word aligner \citep{dyer2013simple}. Fast\_align is licensed under the Apache License 2.0. }
    \item{FastText: an open-source, free, simple framework that helps with learning text representations and text classifiers \citep{joulin2017bag, bojanowski2017enriching}. FastText is MIT-licensed. }
    \item{\emph{SentencePiece}: a light-weight and unsupervised text tokenizer and detokenizer for neural generation systems with a fixed vocabulary size \citep{}. SentencePiece is licensed under the Apache License 2.0. }
    \item{\emph{Subword-nmt}: a simple word segmenter \citep{sennrich2016neural}. This is uses in our system as an alternative tool to SentencePiece. Subword-nmt is licensed under the MIT License. }
    \item{\emph{KenLM}: a Language Model (LM) toolkit \citep{heafield2011kenlm, heafield2013scalable}. KenLM is released mostly under the GNU LGPL license, and is distributed under the GNU Lesser General Public License 2.1. }
    \item{\emph{NLTK}: a natural language processing platform that works seamlessly with python codes \citep{loper2002nltk}. We use the LM from NLTK alongside KenLM above. We also use the automatic evaluation metrics from NLTK for MT evaluation. NLTK is licensed under the Apache License 2.0. } 
    \item{\emph{Moses}: an open-source toolkit for Statistical Machine Translation (SMT) \citep{koehn2007moses}. Since we are working the severely low-resource scenario, our low-resource data is extremely small. Statistical models which are often based in frequency of words is very helpful in text processing. One such feature we use often in our research is truecaser, where the first word of every sentence is normalized to its most frequent form. This prevents a named entity at the beginning of the sentence to be lowercased. Moses is licensed under the GNU LGPL. }
    \item{\emph{Lang2vec}: a lightweight library for querying the URIEL typological database, a carefully structured resource on language typology and universals \citep{littell2017uriel, malaviya2017learning}. Though not every low-resource language we are working on is represented by Lang2vec, it does cover a broad set of languages. It focuses on the typology and various linguistic features of each language. Lang2vec is licensed under Creative Commons Attribution Share Alike 4.0 International. } 
\end{enumerate} 

\subsection{Data}
Our main text is the Bible in multiple languages \citep{mayer2014creating}. Existing research classifies world languages into Resource 0 to 5, with 0 having the lowest resource and 5 having the highest  \citep{joshi2020state}. We choose a few target languages as severely low-resource languages, both real and hypothetical, ranging from Resource 0 to 5. Each severely low-resource language seed corpus contains a extremely small portion (as low as $\sim$3\%) the text, while all other languages have full text. Our goal is to translate the rest of the text into the severely low-resource language.

\subsection{Baseline Systems}
In our setup we have the new, low-resource language as the target language, and we have a few neighboring languages as the source languages that are either in the same linguistic language family or geographically close to facilitate linguistic transfer. In effect, we have a few source languages with full translations of the text and a new and low-resource language that has an extremely small seed corpus. 
We use the state-of-the-art multilingual machine translation systems including transformers prepending both source and target language labels to each source sentence with BPE \citep{johnson2017google, ha2016toward, sennrich2016neural, kudo2018sentencepiece}. For precise translation for all named entities, we use an existing method of \textit{order-preserving named entity translation} by masking each named entity with ordered \texttt{\_\_NE}s using a parallel multilingual lexicon table in all available languages \citep{zhou2021family, wu2018creating}. We use BPE in our models \citep{sennrich2016neural, kudo2018sentencepiece}. 

\subsection{Automatic Evaluation}
We use chrF as our main metric for introduction and conclusion of the key summary of the contribution of this thesis \citep{popovic2015chrf}. We choose chrF for accuracy, fluency and expressive power in morphologically-rich languages \citep{papineni2002bleu}. Additionally, we use a complete set of automatic evaluation metrics to supplement our understanding of translation performance: chrF, characTER, BLEU, COMET score, and BERTscore \citep{popovic2015chrf, wang2016character, post-2018-call, zhang2019bertscore, stewart-etal-2020-comet, rei2021mt}. For detailed analysis, we prioritize BLEU in Massively multilingual translation (Part 1) as we mainly work with European languages while we prioritize chrF in Human Machine translation (Part 2) as we mainly work with morphologically-rich low-resource languages. And we will use our overall metric chrF to conclude this thesis.    
\removelabelprefix

\part{Massively Multilingual Translation}\label{part:part1}

In the first part of this thesis, we explore how source parallelism benefits translation of a given text into new, low-resource languages through multilingual training. In Chapter \ref{big:family}, we build cross-lingual transfer both within a given language family and also across different language families. We also propose an order-preserving lexiconized machine translation model to resolve the variable binding problem, producing high quality lexiconized translations under severely low-resource scenarios. In Chapter \ref{big:paraphrase}, we treat paraphrases as foreign languages, propose a multi-paraphrase translation model which trains on corpus-level paraphrases to improve translation performance. We find that our multi-paraphrase translation models improve performance better than multilingual models and improve the sparsity issue of rare word translation as well as diversity in lexical choice. 
In Chapter \ref{big:ipml}, we build our own linguistic distance metric based on translation distortion, fertility and performance. We propose a method, \textit{Iteratively Pretrained Multilingual Order-preserving Lexiconized Transformer} (IPML), to train on the low-resource language data. 
We use only $\sim$1,000 lines ($\sim$3.5\% of the entire text) to translate the whole text and achieve good translation performance. 
     
\chapter{Language Transfer within and across Families}\label{big:family}
\addlabelprefix{2}
\epigraph{``Learning, learned people knew, was a multilingual enterprise.''}{\textit{Michael D. Gordin}}

\lettrine{H}{aving examined closely the existing literature} in space of translation into low-resource languages, we investigate intra-family and inter-family transfer in translating a multi-source, closed text into a new, low-resource language.

\section{Introduction} \label{introduction}
We work on translation 
from a rich-resource language 
to a low-resource language. There is usually 
little low-resource
language data, and much less parallel data available
\cite{duong2016attentional, anastasopoulos2017spoken}. Despite the challenges of little data
and few human experts, 
it has many 
useful applications. Applications include translating 
water, sanitation and
hygiene (WASH) guidelines to protect 
Indian tribal children against 
waterborne diseases, introducing 
earthquake preparedness techniques to Indonesian 
tribal groups living near volcanoes 
and delivering information to the disabled or the
elderly in low-resource language communities 
\cite{reddy2017water, barrett2005support, anastasiou2010translating, perry2017treasure}.
We show part of the lyrics of a hand-washing song in a few low-resource languages in Table~\ref{table:wash} \citep{thampi2020s}. 
These are useful examples of 
translating a closed text known in advance 
to the low-resource language. 

There are three main challenges. 
Firstly, most previous works research individual languages instead of collective 
families. Cross-lingual
impacts and similarities are very helpful 
when there is little data
in low-resource languages 
\cite{shoemark2016towards, sapir1921languages, odlin1989language, cenoz2001effect, toral2018level, de1997pursuit, hermans2003cross, specia2016shared}.
Secondly, 
most of the multilingual translation works assume
the same amount of training data for all 
languages. In the low-resource case,
it is important to exploit low or partial data
in low-resource languages to produce high
quality translations. 
The third issue is the variable-binding
problem that is common in neural systems, 
where ``John calls Mary''
is treated the same way as ``Mary calls John'' 
\cite{fodor1988connectionism, graves2014neural}.
It is more challenging when both ``Mary'' 
and ``John'' are rare words. Solving 
the binding problem is crucial because the 
mistakes in 
named entities change the meaning of 
the translation. It is especially challenging
in the low-resource case because
many words are rare words.

\begin{figure*}[t]
  \centering
  \includegraphics[width=0.7\linewidth]{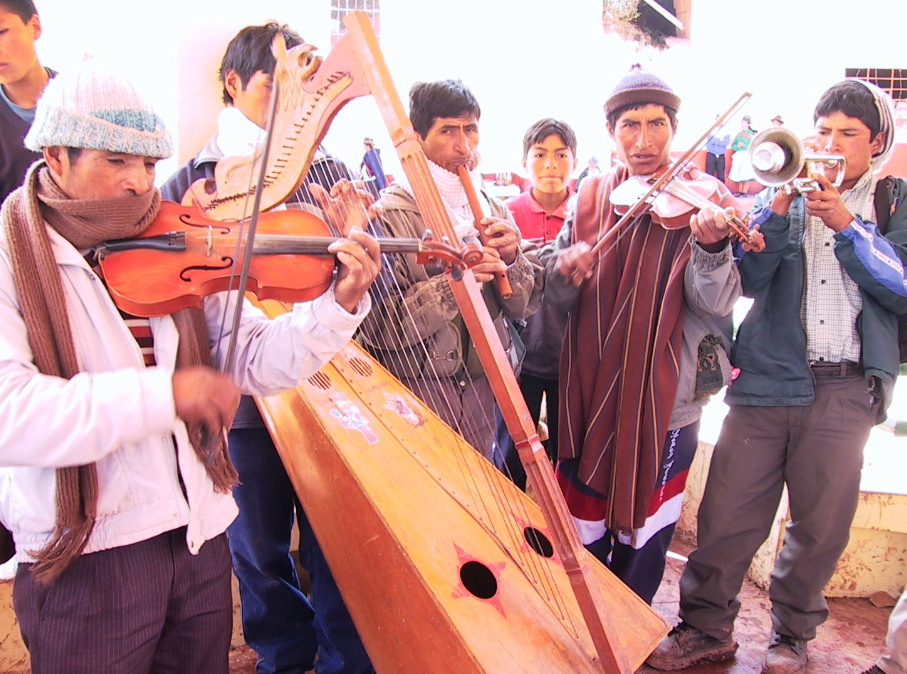}
    \caption{A low-resource language community in Peru gathering to celebrate together. Photograph by Mark Bean. }
    \label{fig:peru_read}  
\end{figure*}

Our contribution\footnote{The material in this chapter was originally 
published as ``Massively Parallel Cross-Lingual Learning in Low-Resource Target 
Language Translation'' in WMT, 2018 \citep{zhou2018massively}.} in addressing these issues 
is three-fold, extending from multi-source
multi-target attentional models. 
Firstly, to examine intra-family and
inter-family influences, we add source and target 
language family labels in training.
Training on multiple families improves
BLEU scores significantly; moreover, we find training on two 
neighboring families closest to the 
low-resource language gives reasonably
good BLEU scores, and we define neighboring families closely in
Section~\ref{proposedextension}. Secondly, we conduct an 
ablation study to explore
how generalization changes with 
different amounts of data and
find that we only need a small amount of
low-resource language data to produce 
reasonably good BLEU scores. We use full data	
except for the ablation	study.
Finally, to address the variable-binding 
problem, we
build a parallel lexicon table across 
twenty-three European
languages and devise a novel order-preserving
named entity translation method. Our method works
in translation of any text with a fixed set of
named entities known in advance. Our methods work best
when we require high accuracy and when 
there are many long sentences and out-of-vocabulary words, including
many words that only occur once (hapax legomenon) \cite{jehn1993hapax, mardaga2012hapax}. 
Our goal is to minimize manual labor, but
not to fully automate the manual process to ensure
the correct translation of named entities and their
ordering. 

\begin{figure*}[t]
  \centering
  \includegraphics[width=.9\linewidth]{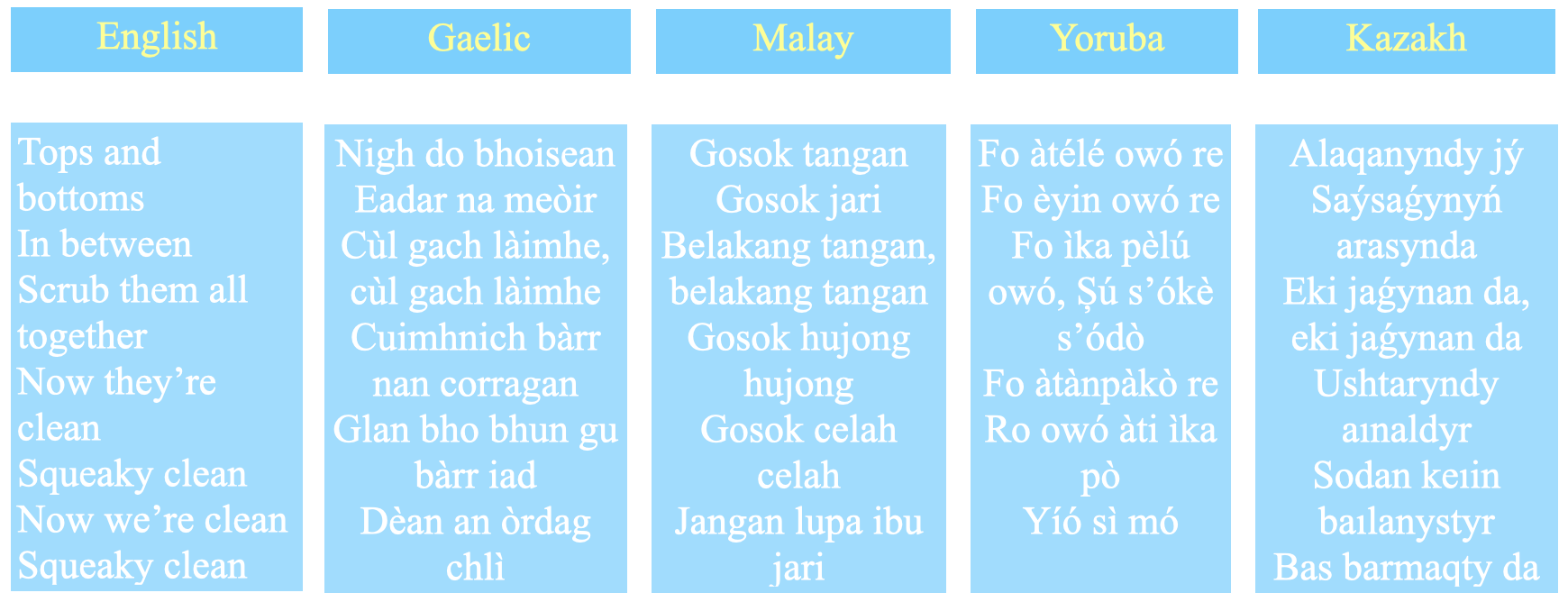}
    \caption{An example of a hand-washing song that is translated into a few languages \citep{thampi2020s}. }
    \label{table:wash}  
\end{figure*}

\section{Related Work} \label{relatedwork}

\subsection{Sub-word Level Machine Translation}
Many MT systems lack robustness with
out-of-vocabulary words (\textit{OOV}s) \cite{wu2016google}.
Most \textit{OOV}s are treated as unknowns (\textit{\$UNK}s) uniformly,
even though they are semantically important and different
\cite{ling2015character, sennrich2016neural}.
To tackle the \textit{OOV} problem,
researchers work on byte-level
\cite{gillick2016multilingual}
and character-level models
\cite{ling2015character, chung2016character}.
Many character-level models do not work as well
as word-level models, and do not produce optimal alignments
\cite{tiedemann2012character}.
As a result, many researchers shift to sub-word level modeling between
character-level and word-level. Many uses BPE which iteratively learns
subword units and balances sequence length and expressiveness
with robustness \cite{sennrich2016neural}.

\subsection{Lexiconized Machine Translation}
Many work with lexicons and named entities in MT
\cite{nguyen2017improving, wang2017sogou, arthur2016incorporating}.
Some create a separate
character-level named entity model
and mark all named entities as \textit{\$TERM}s
to train \cite{wang2017sogou}.
This method learns people's names well
but does not improve
BLEU scores \cite{wang2017sogou}. It is
time-consuming and adds to the system complexity.
Others build lexicon translation seamlessly with attentional MT
by using an affine transformation
of attentional weights
\cite{nguyen2017improving, arthur2016incorporating}.
Some embed crosslingual lexicons
into the same vector space for transfer \cite{duong2017multilingual}.

\begin{table*}[t]
  \small
  \centering
  \begin{tabularx}{\columnwidth}{p{1.6cm} | X}
    \toprule
    Families & Languages \\
    \midrule 
    Germanic & German (de) Danish (dn) Dutch (dt) Norwegian (no) Swedish (sw) English (en) \\
    Slavic & Croatian (cr) Czech (cz) Polish (po) Russian (ru) Ukrainian (ur) Bulgarian (bg) \\
    Romance & Spanish (es) French (fr) Italian (it) Portuguese (po) Romanian (ro) \\
    Albanian & Albanian (ab) \\
    Hellenic & Greek (gk) \\
    Italic & Latin (ln) [descendants: Romance languages] \\
    Uralic & Finnish (fn) Hungarian (hg) \\
    Celtic & Welsh (ws) \\
    \bottomrule
  \end{tabularx}
  \caption{Language families. Language codes are in parentheses.}
  \label{table:family}
\end{table*}

\section{Translation System}

\subsection{Baseline Translation System} \label{baseline}
Our baseline is multi-source multi-target attentional model within
one language family through adding
source and target language labels with a single
unified attentional scheme, with BPE used at the preprocessing stage.
The source and target vocabulary are not shared.

\subsection{Proposed Extensions}\label{proposedextension}
We present our methods in solving three
issues relevant to translation into low-resource language as
our proposed extensions.

\subsubsection{Language Families and Cross-lingual Learning}
Cross-lingual and cross-cultural influences and similarities
are important in linguistics
\cite{shoemark2016towards, levin1998interlingua, sapir1921languages, odlin1989language, cenoz2001effect, toral2018level, de1997pursuit, hermans2003cross, specia2016shared}.
The English word, ``Beleaguer'' originates
from the Dutch word ``belegeren''; ``fidget''
originates from the Nordic word ``fikja''.
English and Dutch belong to the same family
and their proximity has effect on
each other
\cite{harding1988classification, ross2006language}.
Furthermore, languages that do not belong
to the same family affect each other too
\cite{sapir1921languages, ammon2001dominance, toral2018level}. ``Somatic'' stems from the Greek
word ``soma''; ``\begin{CJK}{UTF8}{min}広告\end{CJK}'' (Japanese), ``\begin{CJK}{UTF8}{}\CJKfamily{mj}광고\end{CJK}''(Korean), ``Qu{\h{a}}ng c{\'a}o''(Vietnamese) are closely related to the Traditional Chinese word ``\begin{CJK*}{UTF8}{bsmi}{\CJKfamily{bkai}廣告}\end{CJK*}''. Indeed, many cross-lingual similarities are present.

In this work, we use the language phylogenetic tree as the measure of closeness of languages and language families \cite{petroni2008language}. The distance measure of language families is the collective of all of the component languages. Language families that are next to each other in the language phylogenetic tree are treated as neighboring families in our work, like Germanic family and Romance family. In our discussion, we will often refer to closely related families in the language phylogenetic tree as neighboring families.

We prepend the source and target family labels, in addition to the source and target language labels to the source sentence to increase translation performance. For example, all French-to-English translation pairs are prepended with four labels, the source and target family labels (\texttt{\_\_opt\_family\_src\_romance} \texttt{\_\_opt\_family\_tgt\_germanic}) and the source and target languages labels (\texttt{\_\_opt\_src\_fr} \texttt{\_\_opt\_tgt\_en}). 
In Section~\ref{experiments}, we examine intra-family and inter-family effects more closely.
\begin{table}[t]
    \small
    \centering
    \begin{tabularx}{\columnwidth}{X|XXXXXX}
      \toprule
      Language & German & Danish & Dutch & English & Norwegian & Swedish \\
      \midrule
      German & N.A. & 37.5 & 43.4 & 45.1 & 41.1 & 35.8\\
      Danish & 39.0 & N.A. & 37.1 & 41.1 & 42.6 & 37.4\\
      Dutch & 43.5 & 36.3 & N.A. & 45.1 & 39.0 & 34.3\\
      English & 40.4 & 34.5 & 41.1 & N.A. & 37.1 & 34.0\\
      Norwegian & 40.5 & 42.7 & 40.4 & 42.8 & N.A. & 40.6\\
      Swedish & 39.4 & 38.9 & 37.5 & 40.4 & 43.0 & N.A.\\
      \bottomrule
    \end{tabularx}
    \caption{(Baseline model) Germanic family multi-source multi-target translation.
      Each row represents source, each column represents target.
    }
    \label{table:germanic}
  \end{table}

\subsubsection{Ablation Study on Target Training data}
  To achieve high information transfer from
  rich-resource language to low-resource target language,
  we would like to find out how much target training
  data is needed to produce reasonably good performance.
  We vary the amount of low-resource training data to
  examine how to achieve reasonably good BLEU score using limited low-resource data.
  In the era of Statistical Machine Translation (SMT), researchers have
  worked on data sampling and sorting measures
  \cite{eck2005low, axelrod2011domain}.

  To rigorously determine how much low-resource
  target language is needed for reasonably good results,
  we do a range of control experiments by drawing samples from the
  low-resource language data
  randomly with replacement and duplicate
  them if necessary to ensure all experiments carry
  the same number of training sentences.
  We keep the amount
  of training data in rich-resource languages
  the same, and vary
  the amount of training data in low-resource language
  to conduct rigorous control experiments.
  Our data selection process is different
  from prior research in that only the
  low-resource training data is reduced,
  simulating the real world scenario of having
  little data in low-resource language. By
  comparing results from control
  experiments, we determine how much
  low-resource data is needed.

\subsubsection{Order-preserving Lexiconized Model}
  The variable-binding problem is an inherent issue in connectionist architectures \cite{fodor1988connectionism, graves2014neural}. ``John calls Mary''
  is not equivalent to ``Mary calls John'', but neural networks
  cannot distinguish the two easily
  \cite{fodor1988connectionism, graves2014neural}.
  The failure of traditional neural models to distinguish
  the subject and the object of a sentence is detrimental.
  For example, in the narration ``John told his son Ryan to
  help David, the brother of Mary'',
  it is a serious mistake if we reverse John and
  Ryan's father-son relationships or
  confuse Ryan's and David's relationships with Mary.

  While many machine translation systems can successfully solve the Variable Binding Problem when there is abundant training data and there are a few named entities in a sentence, the Variable Binding Problem for severely low-resource scenarios remains very relevant even today due to the following reasons. Firstly, when there are many occurrences of hapax legomenon in a text, meaning when there are many words that only appear in the text once, it is almost impossible to translate an unseen named entity that never appears in the training data but appears in the test data. In the Bible, for instance, there are $\sim$1,480 words that only occur once. These words are very hard to translate. Additionally, if a text contains extremely long sentences with many rare named entities, it is hard for MT systems to translate all names correctly even if they appear in the training data. For example, the Bible dataset contain many sentences that are extremely long and has more than 10 named entities with an order that is very meaningful and cannot be mistranslated. Therefore, the Variable Binding Problem for severely low-resource languages is still relevant today. A successful solution to the Variable Binding Problem for low-resource languages is very useful when we aim for high accuracy when we translate long sentences with many unseen words, among which many only occur in the text once and never appear in the training data. 

  Variable Binding Problem is still relevant for hapax legomena, which are words that only appear in the text once. This means that a name may never appear in the training data, and only appear once in test data.  the Bible has more than 2000 named entities that are hapaxes. 

  In our work, we focus mainly on text with
  a fixed set of named entities known in advance. We assume that experts help to translate
  a given list of named entities into low-resource language first before
  attempting to translate any text. Under this assumption, we
  propose an order-preserving named entity translation mechanism.
  Our solution is to first create a parallel lexicon table for
  all twenty-three European languages using a seed English lexicon table
  and fast-aligning it with the rest \cite{dyer2013simple}.
  Instead
  of using \textit{\$UNK}s to replace the named entities, we
  use \textit{\$NE}s to distinguish them from the other unknowns.
  We also sequentially tag named entities in a sentence
  as \textit{\$NE1}, \textit{\$NE2}, \ldots, to preserve their ordering.
  For every sentence pair in the multilingual training,
  we build a target named entity decoding dictionary by using all target
  lexicons from our lexicon table that matches with those appeared in the
  source sentence. During the evaluation stage, we
  replace all the numbered \textit{\$NE}s
  using the target named entity decoding dictionary
  to present our final translation.
  This method improves translation accuracy greatly and preserves the order.

      \begin{figure}[t]
    \centering
    \includegraphics[width=6.1in]{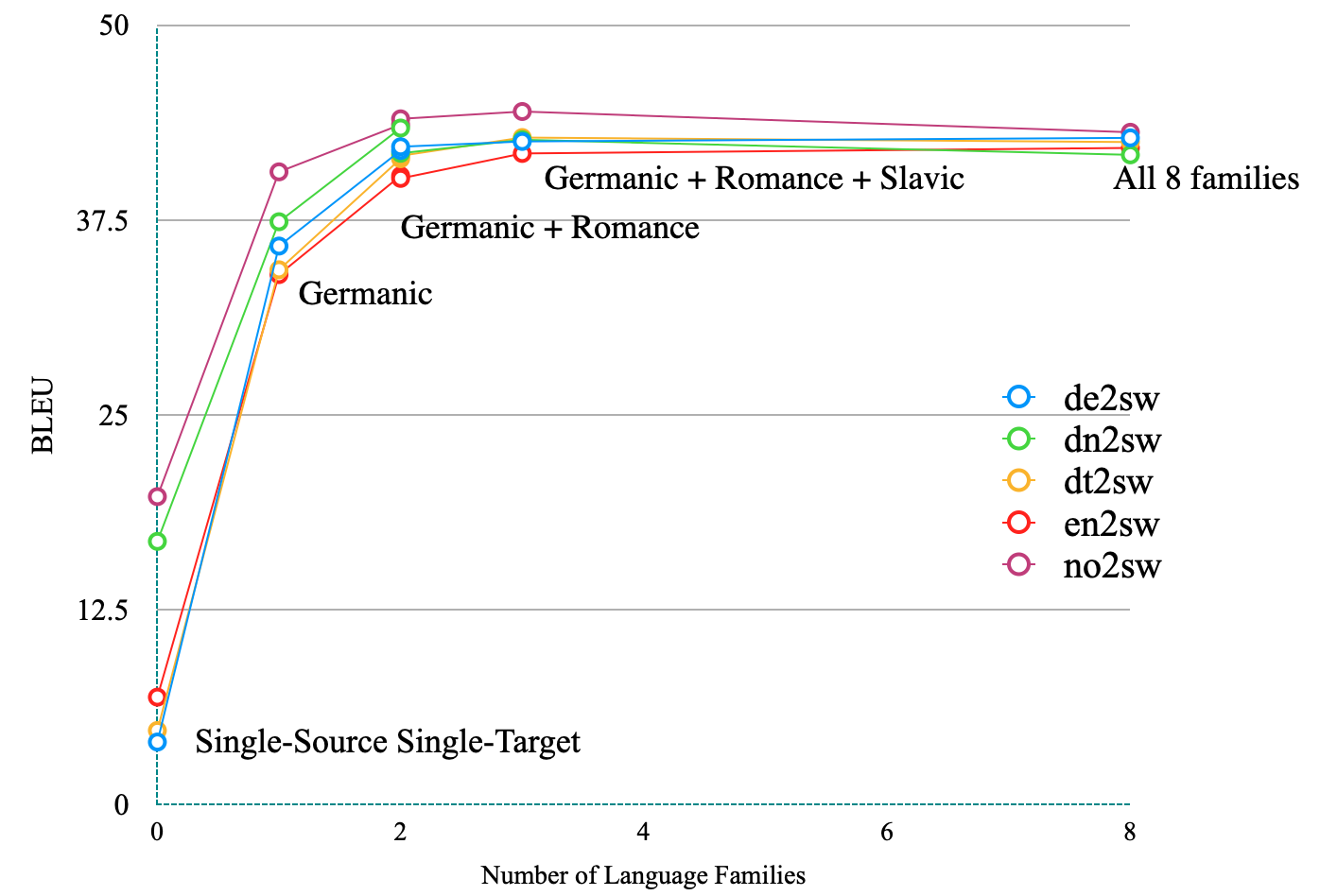}
    \caption{Intra-family and inter-family effects on BLEU scores with respect to increasing addition of language families. }
    \label{fig:multi_family1}
  \end{figure}
      \begin{figure}[t]
    \centering
    \includegraphics[width=6.1in]{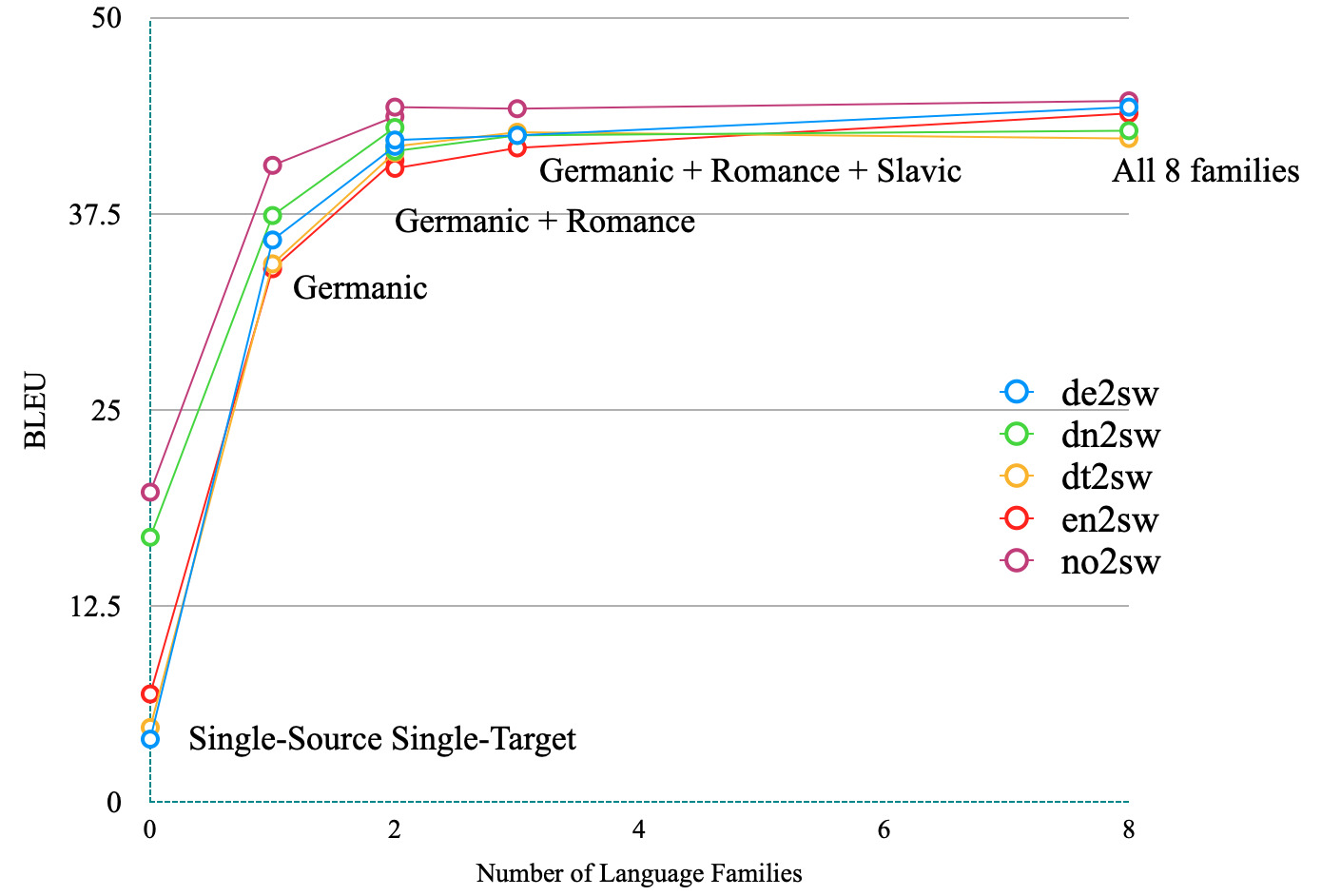}
    \caption{Effects of adding family labels on BLEU scores with respect to increasing addition of language families. }
    \label{fig:multi_family2}
  \end{figure}

As a result of our contribution, the experts only need
  to translate a few lexicons and a small amount of low-resource text before passing
  the task to our system to obtain good results. Post-editing and minor changes
  may be required to achieve 100\% accuracy before the releasing the
  translation to the low-resource language communities.

\section{Experiments}\label{experiments}
Having examined our baseline translation system and the proposed extensions, we are ready to discuss our experimental setups and premises.

Our main goal is translating a multilingually known limited text into a new, low-resource language. There are three premises in our research that differ from traditional MT work: 
\begin{enumerate} 
\item 
Our text is closed, not arbitrary as in traditional MT problems. 
\item 
Our text has multiple source languages with complete text translations while traditional MT is typically single-source.
\item 
Our text has little to no translation in the target low-resource language, while traditional MT assumes abundant data. 
\end{enumerate}

With these three premises, we build a multilingual model\footnote{In Chapter 5-9, all multilingual models are multilingual transformer-based. In Chapter 3-4, all models are multilingual LSTM-based model. This is because our research in Chapter 3-4 is done before transformer comes about. } from scratch\footnote{We train these models from scratch because we want to assume severely low-resource situation where there is no additional resource except for a little data from the low-resource language. Since all models are trained in one stage, there is no need to finetune. Results from this chapter serve as a baseline for Chapter 6-9 where we use multi-stage adaptation using large pretrained models.}, exclusively on limited massively multilingual data, in translation into a new, low-resource language. Training across all language pairs is done jointly to maximize interlingual transfer. 

Thus, we explore interlingual transfer strategies within and across language families in extremely low-training-data setups. Our ablation study aims to find out how to minimize the low-resource training data to achieve sufficiently good performance; our interlingual transfer studies the transfer within and across language families; and our ordered named entities translation mechanism aim to produce highly accurate named entity translations even when such words do not appear in the training data. 

\subsection{Data}
  We choose the Bible corpus as a test ground for
  our proposed extensions because the Bible is the most
  translated text that exists and is freely
  accessible.
  Though it has limitations, it has fewer
  copyright issues like most of literary works
  that are translated into many languages do.
  The Bible provides massively parallel high-quality
  translations into many
  severely low-resource languages and is rapidly growing in language coverage.  
  There are many research works done using the Bible
  \cite{naaijer1993parallel, mayer2014creating, scannell2006machine, dufter2018universal, resnik1999bible, chan2001encyclopaedia, banchs2011semantic, christodouloupoulos2015massively, beale2005document}.
  Unlike many past research works where
  only New Testament is used \cite{dufter2018universal},
  we use both Old Testament and New Testament in our Bible corpus.
  We align all Bible verses across different languages.

  We train our proposed model on twenty-three European languages
  across eight families on a parallel Bible corpus.
  For our purpose, we treat Swedish as our hypothetical
  low-resource target language, English as our
  rich-resource language in the single-source single-target case and
  all other Germanic languages as our rich-resource languages
  in the multi-source multi-target case.

  Firstly, we present our data and training parameters. Secondly, we add
  family tags in different configurations of language families showing
  intra-family and inter-family effects.
  Thirdly, we conduct an ablation study and plot the generalization curves
  by varying the amount of training data in Swedish, and
  we show that training on one fifth of the data give reasonably good BLEU scores.
  Lastly, we devise an order-preserving lexicon translation method by
  building a parallel lexicon table across twenty-three
  European languages and tagging named entities in order.

  \subsection{Training Parameters}
  We clean and align the Bible in twenty-three European languages
  in Table~\ref{table:family}. We randomly sample the
  training, validation and test sets according to the 0.75, 0.15, 0.10 ratio.
  Our training set contains 23K verses, but is massively parallel. In our
  control experiments, we also use the experiment training on the WMT'14 French-English dataset
  together with French and English Bibles to compare with
  our results. Note that our WMT baseline contains French and English Bibles
  in addition to the WMT'14 data, and is used to contrast our results with
  the effect of increasing data.

  In all our experiments, we use a minibatch size of 64, dropout rate of 0.3,
  4 RNN layers of size 1000, a word vector size of 600, learning rate of 0.8
  across all LSTM-based multilingual experiments. For single-source single-target translation, we use 2 RNN layers of
  size 500, a word vector size of 500, and learning rate of 1.0. All learning
  rates are decaying at the rate of 0.7 if the validation score is not improving
  or it is past epoch 9. We use SGD as our learning algorithm.
  We build our code based on OpenNMT \cite{klein2017opennmt}.
  For the ablation study, we train on BLEU scores directly
  until the \textit{Generalization Loss} (\textit{GL}) exceeds a threshold of $\alpha = 0.1$
  \cite{prechelt1998early}. \textit{GL} at epoch $t$ is defined as
  $GL(t) = 100 ( 1- \frac{E_{val}^t}{E_{opt}^t} )$,
  modified by us to suit our objective using BLEU scores \cite{prechelt1998early}.
  $E_{val}^t$ is the validation score at
  epoch $t$ and $E_{opt}^t$ is the optimal score up to epoch $t$.
  We evaluate our models using both BLEU scores \cite{papineni2002bleu} and qualitative evaluation.

\section{Results}
Having understood our data, training parameters and experimental setup, we investigate our results from three perspectives: interlingual transfer within and across families, ablation study and order-preserving lexiconized translation. 

  \subsection{Interlingual Transfer Within and Across Families}
  \begin{table}[t]
    \small
    \centering
    \begin{tabularx}{\columnwidth}{X|XXXXXX}
      \toprule
      Experiment & S & G & GS & GR & 3F & 8F  \\
      \midrule
      de2sw & 4.0 & 35.8 & 42.0 & 42.2 & 42.5 & 42.8 \\ 
      dn2sw & 16.9 & 37.4 & 43.4 & 41.8 & 42.7 & 41.7 \\
      dt2sw & 4.8 & 34.3 & 41.4 & 41.6 & 42.8 & 42.5 \\ 
      en2sw & 6.9 & 34.0 & 40.3 & 40.2 & 41.8 & 42.1 \\ 
      no2sw & 16.8 & 40.6 & 43.6 & 44.0 & 44.5 & 43.1 \\
      \bottomrule
    \end{tabularx}
    \caption{Inter-family and intra-family effects on BLEU scores with respect to increasing addition of language families. \\
      S: single-source single-target model.
      \\G: training on Germanic family.
      \\GS: training on Germanic, Slavic family.
      \\GR: training on Germanic, Romance family.
      \\3F: training on Germanic, Slavic, Romance family.
      \\8F: training on all 8 European families together.
    }
    \label{table:summary}
  \end{table}
  \begin{table}[t]
    \small
    \centering
    \begin{tabularx}{\columnwidth}{X|XXXXXX}
      \toprule
      Experiment & S & G & GSl & GRl & 3Fl & 8Fl  \\
      \midrule
      de2sw & 4.0 & 35.8 & 41.8 & 42.2 & 42.5 & 44.3 \\ 
      dn2sw & 16.9 & 37.4 & 43.0 & 41.5 & 42.5 & 42.8 \\ 
      dt2sw & 4.8 & 34.3 & 41.4 & 41.8 & 42.7 & 42.3 \\ 
      en2sw & 6.9 & 34.0 & 40.9 & 40.4 & 41.7 & 43.9 \\ 
      no2sw & 16.8 & 40.6 & 43.7 & 44.3 & 44.2 & 44.7 \\ 
      \bottomrule
    \end{tabularx}
    \caption{Effects of adding family labels on BLEU scores with respect to increasing addition of language families. \\
      S and G: same as in Table~\ref{table:summary}.
      \\GSl: Germanic, Slavic family with family labels.
      \\GRl: Germanic, Romance family with family labels.
      \\3Fl: Germanic, Slavic, Romance family with family labels.
      \\8Fl: all 8 European families together with family labels
    }
    \label{table:summary_label}
  \end{table}

  We first investigate intra-family
  and inter-family influences and the
  effects of adding family labels.
  We use full training data
  in this subsection. 

  \textbf{\ul{Languages have varying closeness to each other:}}
  Single-source single-target translations of different languages in
  Germanic family to Swedish show huge differences in BLEU scores as
  shown in Table~\ref{table:summary}. These differences are well
  aligned with the multi-source multi-target results. Norwegian-Swedish
  and Danish-Swedish translations have much higher BLEU scores
  than the rest. This hints that Norwegian and Danish are closer to Swedish
  than the rest in the neural representation.

  \textbf{\ul{Multi-source multi-target translation improves greatly from single-source
      single-target translation:}} English-Swedish
  single-source single-target translation gives a low BLEU score
  of 6.9 as shown in Table~\ref{table:summary},
  which is understandable as our dataset is very small.
  BLEU score for English-Swedish translation
  improves greatly to 34.0 in the multi-source multi-target
  model training on Germanic family as shown in
  Table~\ref{table:germanic}. In this work, we treat the Germanic multi-source
  multi-target model as our baseline model. We present only
  relevant columns important for cross-lingual learning and translation
  into low-resource language here.

  \textbf{\ul{Adding languages from other families into training improves
      translation quality within each family greatly:}} English-Swedish translation's BLEU score
  improves significantly from 34.0 to 40.3 training on Germanic and Slavic families,
  and 40.2 training on Germanic and Romance
  families as shown in Table~\ref{table:summary}.
  After we add all three families in training,
  BLEU score for English-Swedish translation increases further to
  41.8 in Table~\ref{table:summary}. Finally,
  after we add all eight families,
  BLEU score for English-Swedish translation
  increases to 42.1 in Table~\ref{table:summary}.
  \begin{figure}[t]
    \centering
    \includegraphics[width=6.1in]{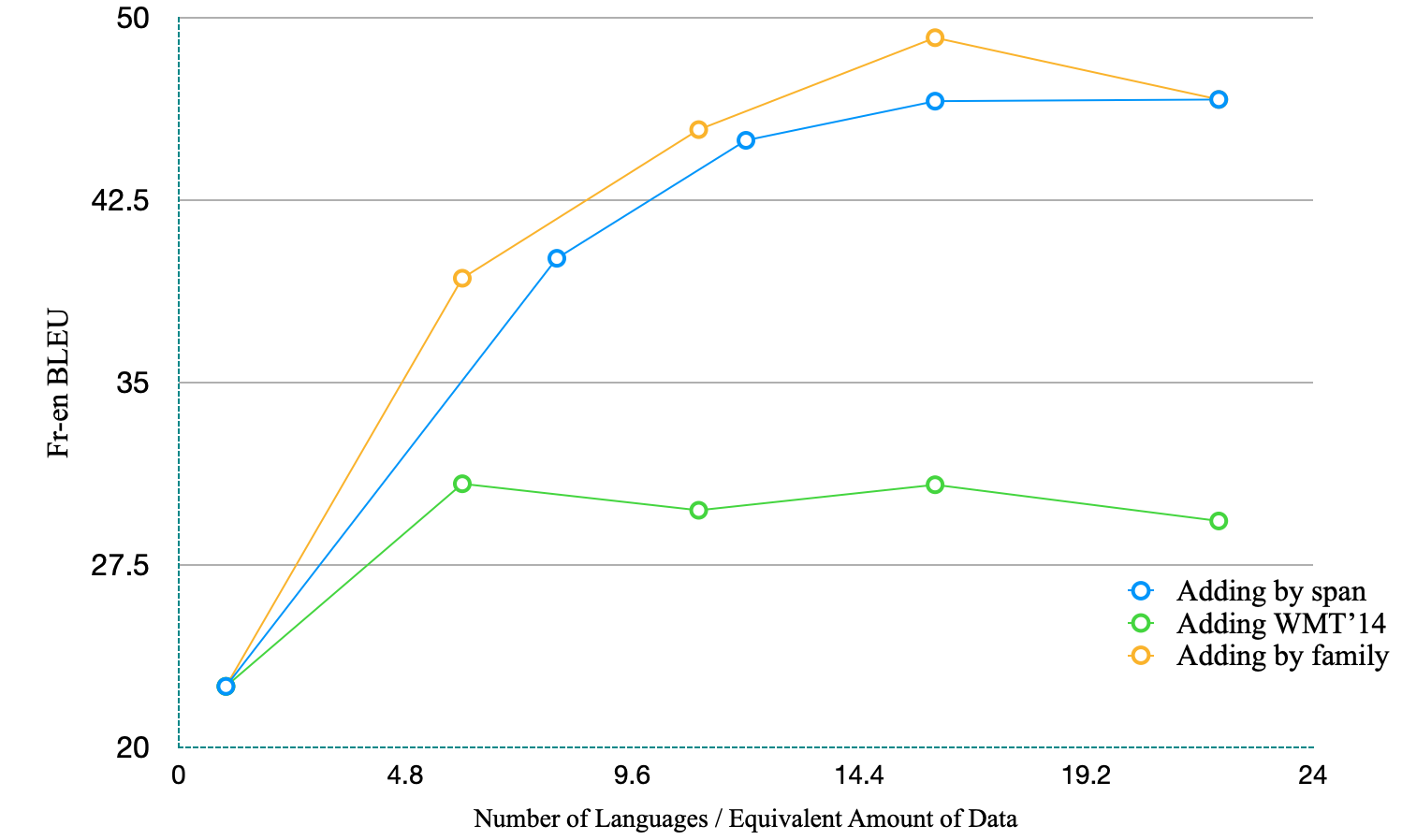}
    \caption{Comparison of different ways of increasing training Data in French-English translation. \\
      Family: Adding data from other languages based on the family unit
      \\WMT'14: Adding WMT'14 data as control experiment
      \\Sparse: Adding data from other languages that spans the eight European families
    }
    \label{fig:adding}
  \end{figure}
  
  \textbf{\ul{A Plateau is observed after adding more than one
      neighboring family:}} A plateau is observed when we plot
  Table~\ref{table:summary} in Figure~\ref{fig:multi_family1}.
  The increase in BLEU scores after adding two families is much milder
  than that of the first addition of a neighboring family. This
  hints that using unlimited number of languages to train may not
  be necessary.

  \textbf{\ul{Adding family labels increases translation performance:}} 
      We observe in Table~\ref{table:summary_label} that
  BLEU scores for most language pairs improve with the addition of
  family labels. Training on eight language families,
  we achieve a BLEU score of 43.9 for English-Swedish
  translation, +9.9 above the Germanic baseline. Indeed,
  the more families we have, the more helpful it is to distinguish them.
    
    \begin{table*}[t]
    \small
    \centering
    \begin{tabularx}{\textwidth}{X|XXXXXXXXXX}
      \toprule
      Data & 0.1 & 0.2 & 0.3 & 0.4 & 0.5 & 0.6 & 0.7 & 0.8 & 0.9 & 1 \\
      \midrule
      \#w &53589 &107262 &161332 &214185 &268228 &322116 &375439 &429470 &483440 &538030 \\
      log\#w &4.73 &5.03 &5.21 &5.33&5.43 &5.51 &5.57 &5.63 &5.68 &5.73 \\
      \midrule
      en2sw &25.2 &30.6 &32.9 &32.7 &34.2 &34.2 &33.8 &33.6 &34.3 &34.9 \\
      de2sw &26.5 &33.4 &34.8 &35.7 &36.7 &36.5 &37.1 &37.1 &36.4 &37.5 \\
      dn2sw &27.2 &34.8 &35.8 &37.1 &37.6 &37.1 &38.5 &38.0 &37.4 &38.4 \\
      dt2sw &26.1 &32.5 &34.2 &34.9 &36.0 &35.8 &36.0 &35.7 &35.8 &36.6 \\
      no2sw &27.7 &36.9 &37.9 &39.5 &39.4 &39.2 &41.3 &40.8 &39.2 &40.5 \\
      \bottomrule
    \end{tabularx}
    \caption{Ablation Study on Germanic Family. \#w is the word count of unique sentences in Swedish data. }
    \label{table:ablation}
  \end{table*}
  
  \textbf{\ul{Training on two neighboring families nearest to the
      low-resource language gives better result than training on languages
      that are further apart:}} Our observation of the
  plateau hints that training on two neighboring
  families nearest to the low-resource language
  is good enough as shown in Table~\ref{table:summary}.
  Before jumping to conclusion, we compare results of adding languages
  by family with that of adding languages by random samples that
  span all eight families, defined as the following.
  \theoremstyle{definition}
  \begin{definition}[Language Spanning]
    A set of languages spans a set of families when it contains
    at least one language from each family.
  \end{definition}
  In Figure~\ref{fig:adding}, we conduct a few experiments
  on French-English translation using
  different ways of adding training data. Let
  \textit{family addition} describe the addition of training data
  through adding close-by language families
  based on the unit of family; let
  \textit{sparse addition} describe
  the addition of training data through adding language sets that spans eight
  language families. In sparse addition, languages are further apart as each may
  represent a different family. We find that
  family addition gives better generalization
  than that of sparse addition.
  It strengthens our earlier results that
  training on two families closest
  to our low-resource language is a
  reliable way to reach good generalization.

  \textbf{\ul{Generalization is not merely an effect of
      increasing amount of data:}} In Figure~\ref{fig:adding}, we compare all methods of adding
  languages against a WMT'14 curve by
  using equivalent amount of
  WMT'14 French-English data in each experiment.
  The WMT'14 curve serve as our benchmark
  of observing the effect of increasing data,
  we observe that our addition of other
  languages improve BLEU scores much sharply
  than the increase in the benchmark, showing that
  our generalization is not merely an
  effect of increasing data. We also observe
  that though increase WMT'14 data initially increases
  BLEU score, it reaches a plateau and adding more
  WMT'14 data does not increase performance from
  very early point.
  
   \subsection{Ablation Study on Target Training Data}
  We use full training data from all rich-resource languages,
  and we vary the amount of training data in Swedish, our low-resource language,
  spanning from one tenth to full length uniformly.
  We duplicate the subset to ensure all training sets,
  though having a different number of unique
  sentences, have the same number of total sentences.

\begin{figure}[t]
    \centering
    \includegraphics[width=6.1in]{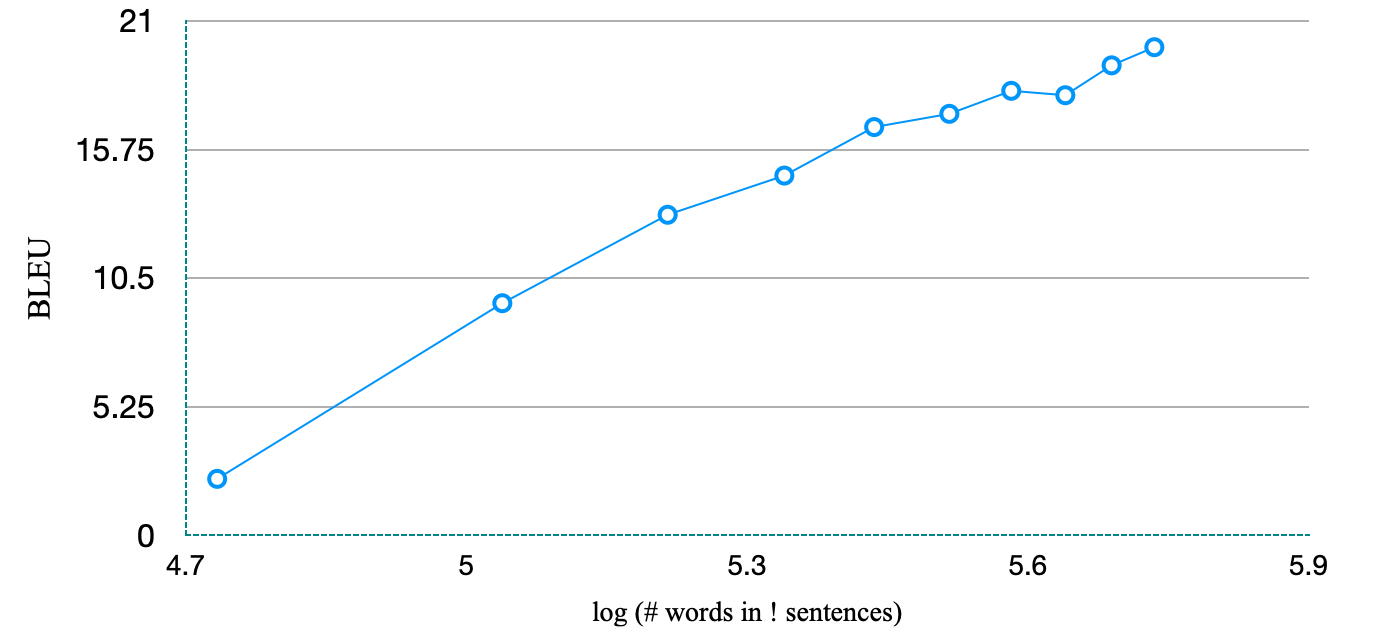}
    \caption{Single-source single-target English-Swedish BLEU plots against increasing amount of Swedish data.}
    \label{fig:curve_single}
  \end{figure}

\begin{figure}[t]
    \centering
    \includegraphics[width=6.1in]{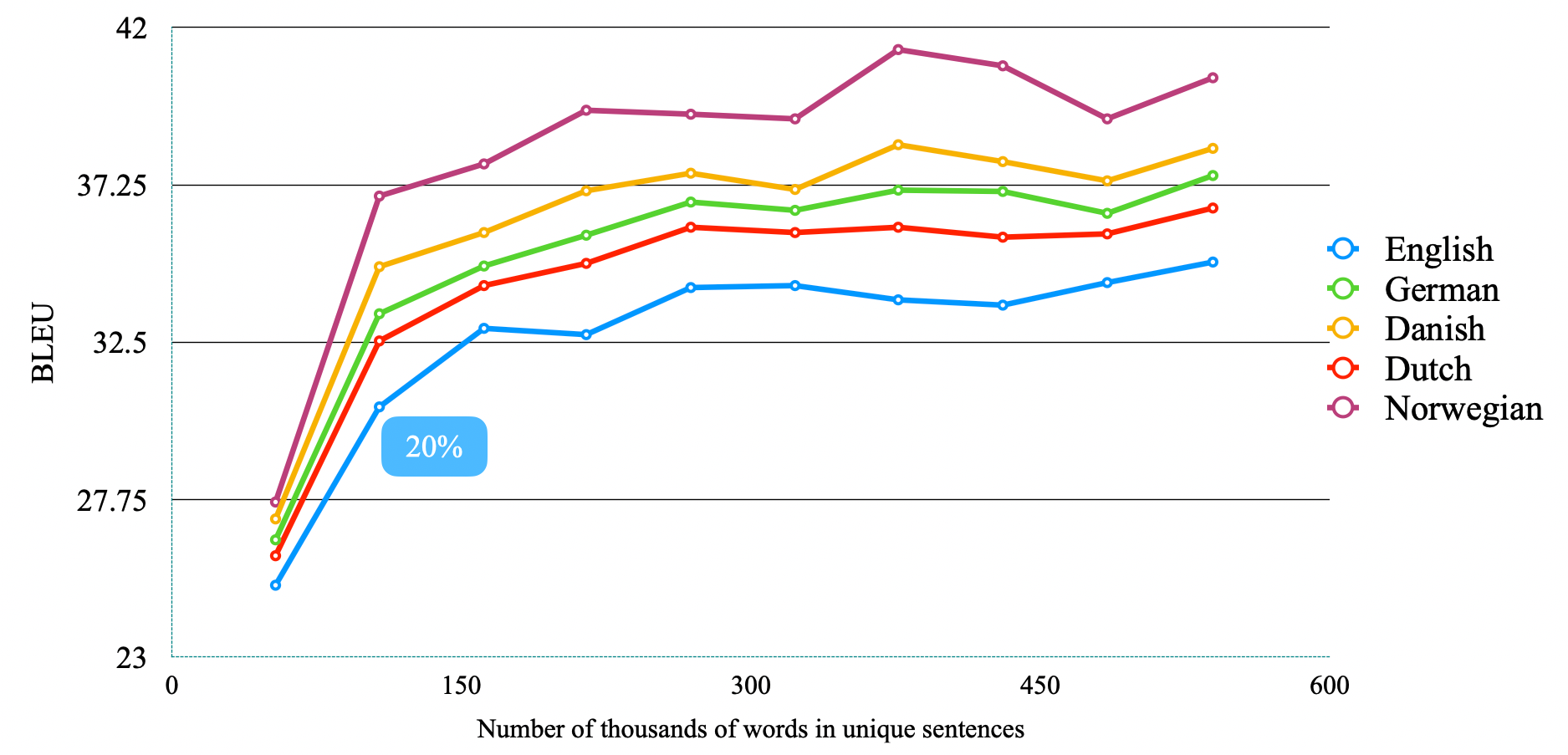}
    \caption{Multi-source multi-target
      Germanic-family-trained BLEU plots
      against increasing amount of Swedish
      data.}
    \label{fig:curve_multi}
  \end{figure}

  \textbf{\ul{Power-law relationship is observed between the performance and the
      amount of training data in low-resource language:}}
  Figure~\ref{fig:curve_multi} shows how BLEU scores
  vary logarithmically with the number of
  unique sentences in the low-resource
  training data. It follows a linear pattern
  for single-source single-target
  translation from English to Swedish
  as shown in Figure~\ref{fig:curve_single}.
  We also observe a linear pattern for the multi-source
  multi-target case, though more uneven
  in Figure~\ref{fig:curve_multi}.
  The linear pattern with BLEU scores against
  the logarithmic data shows the power-law relationship between
  the performance in translation and
  the amount of low-resource training data.
  Similar power-law relationships are also found in past research and contemporary literature \cite{turchi2008learning, hestness2017deep}.

  \textbf{\ul{We find one fifth of data is enough for sufficiently good translation performance:}} 
  For the multi-source multi-target case, we find that using one fifth
  of the low-resource training data is sufficient for good translation performance. It gives reasonably good BLEU scores
  as shown in Figure~\ref{fig:curve_multi}.
  This is helpful when we have
  little low-resource data.
  For translation into the low-resource language, experts only need to translate a small amount of seed
  data before passing it to our system\footnote{Note that
    using nine tenth of random samples yields higher
    performance than using full data, but it may not be generalized
    to other datasets.}.

  \subsection{Order-preserving Lexiconized Model}
  We devise a mechanism to build
  a parallel lexicon table across twenty-three European
  languages using very little data and zero manual work.
  A few lexicon examples are shown in Table~\ref{table:lexicon_examples}.
  We first extract named entities from the English Bible
  \cite{manning2014stanford} and combine them with English biblically named entities from multiple
  sources \cite{easton1897eastons, nave1903nave, smith1967smith, hitchcock1874hitchcock, rice2015people}.
  Secondly, we carefully automate the filtering process to obtain a clean English
  lexicon list. Using this list as the seed, we build a parallel lexicon
  table across all twenty-three languages through
  fast-aligning \cite{dyer2013simple}.
  The final parallel lexicon table has 2916 named entities. In
  the translation task into low-resource language,
  we assume that the experts first translate these lexicon entries,
  and then translate approximately one fifth random sentences
  before we train our model. If necessary, the experts evaluate
  and correct translations before releasing the final
  translations to the low-resource language
  community. We aim to reduce human effort in post-editing and increase
  machine accuracy. After labeling named entities in each sentence pair in
  order, we train and obtain good translation results.

 \begin{table}[t]
    \small
    \centering
      \begin{tabularx}{\columnwidth}{XXXXXX}
      \toprule
      English & German & Czech & Spanish & Finnish & Swedish \\
      \midrule
      Joseph&Joseph&Jozef&Jos{\'e}&Joseph&Josef \\ 
      Peter&Petrus&Petr&Pedro&Pietari&Petrus \\ 
      Zion&Zion&Sion&Sion&Zionin&Sion \\ 
      John&Johannes&Jan&Juan&Johannes&Johannes \\ 
      Egypt&{\"A}gypten&Egyptsk{\'e}&Egipto&Egyptin&Egyptens \\ 
      Noah&Noah&No{\'e}&No{\'e}&Noa&Noa \\ 
      \bottomrule
    \end{tabularx}
    \caption{A few examples from the parallel lexicon table.}
    \label{table:lexicon_examples}
  \end{table}
  \begin{table}[t]
    \small
    \centering
    \begin{tabularx}{\columnwidth}{X|XXXX}
      \toprule
      Experiment & G & OG & OG1 & OGM  \\
      \midrule
      de2sw & 35.8 & 36.6 & 36.6 & 36.9 \\ 
      dn2sw & 37.4 & 37.0 & 37.2 & 36.9  \\ 
      dt2sw & 34.3 & 35.8 & 35.6 & 35.9 \\ 
      en2sw & 34.0 & 33.6 & 33.9 & 33.4  \\
      no2sw & 40.6 & 41.2 & 41.0 & 41.4  \\
      \bottomrule
    \end{tabularx}
    \caption{Summary of order-preserving lexicon translation. \\
      G: training on Germanic family without using order-preserving method. \\
      OG: order-preserving lexicon translation.
      \\OG1: OG translation using lexicons with frequency 1.
      \\OGM: OG translation using lexicons with manual selection.
    }
    \label{table:summary_lexicon}
  \end{table}

  \textbf{\ul{We observe 60.6\% accuracy in human evaluation where our translations are parallel to human translations:}} In Table~\ref{table:lexicon_results1}, we show some examples of machine
  translated text, we also show the expected correct translations for comparison. Not only the named entities are correctly mapped, but also
  the ordering of the subject and the object is preserved.
  In a subset of our test set, we conduct
  human evaluation on 320 English-Swedish results to
  rate the translations into three categories:
  accurate (parallel to human translation), almost accurate
  (needing minor corrections) and inaccurate. More precisely,
  each sentence is evaluated using three criteria: correct set
  of named entities, correct positioning of named entities, and accurate
  meaning of overall translation.
  If a sentence achieves all three, then it is termed as accurate;
  if either a name entity is missing or its position is
  wrong, then it is termed as almost accurate (needing
  minor correction); if the meaning of the sentence
  is entirely wrong, then it is inaccurate.
  Our results are 60.6\% accurate, 33.8\% needing minor corrections, and 5.6\% inaccurate. Though human evaluation
  carries bias and the sample is small,
  it does give us perspective on the performance of our model.

  \begin{table*}[th]
    \small
    \centering
    \begin{tabularx}{\columnwidth}{X|X|X|X|X}
      \toprule
      Source Sentence
      & Model Translation without Order Preservation (Before)
      & Model Translation with Order Preservation (After)
      & Reference
      & Frequency of Named Entities\\
      \midrule
      And \emph{Noah} fathered three sons, \emph{Shem}, \emph{Ham}, and \emph{Japheth}.
      & Och \emph{Noa} f{\"o}dde tre s{\"o}ner, \emph{Sem}, \emph{Ham} och \emph{Jafet}.
      & Och \emph{Noa} f{\"o}dde tre s{\"o}ner, \emph{Sem} \emph{Ham} och \emph{Jafet}
      & Och \emph{Noa} f{\"o}dde tre s{\"o}ner: \emph{Sem}, \emph{Ham} och \emph{Jafet}.
      & \emph{Noah}: 58, \emph{Shem}: 18, \emph{Ham}: 17, \emph{Japheth}: 11
      \\
      \midrule
      And \emph{Saul} spoke to his son \emph{Jonathan}, and to all his servants, to kill \emph{David}.
      & Och \emph{Saul} sade till \emph{Jonatan}, hans son, och alla hans tj{\"a}nare, s{\aa} att de skulle d{\"o}da \emph{David}.
      & Och \emph{Saul} talade till sin son \emph{Jonatan} och alla hans tj{\"a}nare f{\"o}r att d{\"o}da \emph{David}
      & Och \emph{Saul} talade med sin son \emph{Jonatan} och med alla sina tj{\"a}nare om att d{\"o}da \emph{David}
      & \emph{Saul}: 424, \emph{Jonathan}: 121, \emph{David}: 1134
      \\
      \midrule
      And they killed \emph{Parshandatha}, and \emph{Dalphon}, and \emph{Aspatha}, and \emph{Poratha}, and \emph{Adalia}, and \emph{Aridatha}, and \emph{Parmashta}, and \emph{Arisai}, and \emph{Aridai}, and \emph{Vajezatha},
      & Och de dr{\"a}pte \emph{Kedak}, \emph{Ir-Fittim}, \emph{Aquila}, \emph{d{\"o}rrvaktarna}, \emph{Amarja}, \emph{Bered}, \emph{vidare Bet-Hadt}, \emph{Berota}, \emph{Gat-Rimmon},
      & Och de dr{\"a}pte \emph{Parsandata} \emph{Dalefon} och \emph{Aspata} \emph{Porata} \emph{Adalja} \emph{Aridata} \emph{Parmasta} \emph{Arisai} \emph{Aridai} \emph{Vajsata}
      & Och \emph{Parsandata}, \emph{Dalefon}, \emph{Aspata}, \emph{Porata}, \emph{Adalja}, \emph{Aridata}, \emph{Parmasta}, \emph{Arisai}, \emph{Aridai} och \emph{Vajsata},
      &\emph{Parshandatha}: 1, \emph{Dalphon}: 1, \emph{Aspatha}: 1, \emph{Poratha}: 1, \emph{Adalia}: 1, \emph{Aridatha}: 1, \emph{Parmashta}: 1, \emph{Arisai}: 1, \emph{Aridai}: 1, \emph{Vajezatha}: 1
      \\
      \bottomrule
    \end{tabularx}
    \caption{Examples of order-preserving lexicon-aware translation for English to Swedish. The frequency of the named entities are the number of occurrences each named entity appears in the whole dataset; for example, all named entities in the last sentence do not appear in the training data.}
    \label{table:lexicon_results1}
  \end{table*}

  \textbf{\ul{Order-preservation performs well especially when the named entities
      are rare words:}} In Table~\ref{table:lexicon_results1}, the model
  without order-preservation lexiconized treatment
  performs well when named entities
  are common words, but fails to predict
  the correct set of named entities and
  their ordering when named entities are rare words.
  The last column shows the
  number of occurrences of each named entity. For the last example,
  there are many named entities that only occur in data once,
  which means that they never appear in training and
  only appear in the test set. The model without
  order-preservation lexiconized treatment
  predicts the wrong set of
  named entities with the wrong ordering. Our lexiconized
  order-preserving MT model, on the contrary, performs
  well at both the head and tail of the distribution,
  predicts the right set of named entities with the right ordering.

  \textbf{\ul{Prediction with longer sentences and many named entities
      are handled well:}} In Table~\ref{table:lexicon_results1},
  we see that the model
  without order-preservation lexiconized treatment
  performs well with short sentences
  and few named entities in a sentence.
  But as the number of the name entities per sentence increases,
  especially when the name entities are rare unknowns
  as discussed before, normal model cannot make correct
  prediction of the right set of name entities
  with the correct ordering in Table~\ref{table:lexicon_results1}.
  Our lexiconized order-preserving model,
  on the contrary, gives very high accuracy when there are
  many named entities in the sentence
  and maintains their correct ordering.

  \textbf{\ul{Trimming the lexicon list that keeps the tail helps to increase BLEU scores:}} Different from most of the previous lexiconized model works where BLEU scores never increase \cite{wang2017sogou},
  our BLEU scores show minor improvements. BLEU score for German-Swedish translation increases from 35.8
  to 36.6 in Table~\ref{table:summary_lexicon}.
  As an attempt to increase our BLEU scores even further,
  we conduct two more experiments. In one setting, we keep only the tail of the lexicon table
  that occur in the Bible once. In another setting, we keep only a manual selection of lexicons.
  Note that this is the only place where manual work is involved and is not essential.
  There are minor improvements in BLEU scores in both cases.

  \textbf{\ul{33.8\% of the translations require minor corrections: }}
  The sentence length for these translations that require minor corrections
  is often longer. We notice that
  some have repetitions that do not affect meaning, but need to be trimmed. Some
  have the under-prediction problem where certain named entities in the source sentence
  never appear; in this case, missing named entities need to be added.
  Some have minor issues with plurality and tense.
  Typically, sentences with longer sentence length
  and more complicated named entity relationships require minor corrections to
  achieve high translation quality.

\section{Conclusion and Future Directions} \label{conclusion}
  We present our order-preserving translation system for cross-lingual
  learning in European languages. We examine
  three issues that are important to translation into
  low-resource language: the lack of low-resource data, effective
  cross-lingual transfer, and the variable-binding problem.

  Firstly, we add the source and the target family labels
  in training and examined intra-family and inter-family effects.
  We find that training on multiple families,
  more specifically, training on two neighboring
  families nearest to the low-resource language
  improves BLEU scores to a reasonably good level.
  Secondly, 
  we devise a
  rigorous ablation study and
  show that we only need one fifth of the
  low-resource target data to
  produce sufficiently good translation performance.
  Thirdly, to address
  the variable-binding problem, we build a
  parallel lexicon table across twenty-three European
  languages and design a novel order-preserving
  named entity translation method by tagging named entities
  in each sentence in order. We achieve
  reasonably good
  quantitative and qualitative improvements
  in a preliminary study.

  The order-preserving named entity translation labels
  named entities in order. Since there are relatively less number of
  long sentences with many named entities than short sentences with
  few named entities, underprediction of
  named entities in long sentences may occur.
  To seek solution to the underprediction
  problem, we are looking at randomized
  labeling of the named entities. Moreover, our
  order-preserving named entity translation method works well
  with a fixed pool of named entities in any static
  document known in advance. This is due to our unique use cases for
  applications like translating water, sanitation and hygiene (WASH)
  guidelines written in the introduction. We devise our method
  to ensure high accuracy targeting translating named entities in static
  document known in advance.
  However, researchers may need to translate dynamic document to low-resource
  language in real-time. We are actively researching
  into the dynamic timely named entity discovery with high accuracy.

  We are actively extending our work to cover more world languages,
  more diverse domains, and more varied sets of datasets to show our
  methods are generalizable. Since our experiments shown in this work
  are using European languages, we are also interested
  on non-European languages like Arabic, Indian, Chinese, Indonesian and many others to
  show that our model is widely generalizable. We also expect to discover
  interesting research ideas exploring a wider universe of linguistically
  dissimilar languages.

  Our work is helpful for translation into low-resource language,
  where human translators only need to translate a few lexicons and
  a partial set of data before passing it to
  our system. Human translators may also be needed during post-editing
  before a fully accurate translation is released. Our future goal is to
  minimize the human correction efforts and to present high quality
  translation timely.

  We would also like to work on real world low-resource tribal languages
  where there is no or little training data. Translation using
  limited resources and data
  in these tribal groups that fits with the
  culture-specific rules
  will be very important \cite{levin1998interlingua}.
  Real world low-resource
  languages call for cultural-aware translation. 
  Careful research with real world low-resource languages
  will create great value and long impact for the low-resource language communities.
  
  One situation when we experience real-world limits is that we may have little 
  information of nearby languages, not to say translation of the multi-source text 
  that is available in nearby languages. In those situations, we still aim to make 
  the best of all the resources that we can find about the low-resource languages. In real-world 
  situations, sometimes we do have multiple translators working on the same text, producing 
  different paraphrases of the text in the same language, rather than in other languages. 
  In the next chapter, we will explore ways to leverage paraphrases within the same language 
  for translation of a multi-source, closed text into a new, low-resource language.                                                                            
\removelabelprefix

\chapter{Paraphrases as Foreign Languages}\label{big:paraphrase}
\addlabelprefix{3}
\epigraph{``Learning another language is not only learning different words for the same things, but learning another way to think about things.''}{\textit{Flora Lewis}}

\lettrine{H}{aving demonstrated successful} language transfer within and across families in the translation of a multi-source text into 
a new, low-resource languages, we would like to make full use of all the information that we can 
find in the low-resource language.

However, we face many translation cases where there is little information or incomplete information of nearby languages, not to mention translation of the given text in those nearby languages. 
In those situations, we might have translators translating the same text into the low-resource languages in different translating and 
writing styles due to quality control reasons. Therefore, paraphrases within the same language could be a by-product of the human translation process, and is another resource we could use for translation into low-resource languages. 
In other words, we are researching situations where there is more information in paraphrases within the same language, rather than across languages. We would like to use this as a new angle to make full use of all existing information that we can find in the target low-resource language. By using all information of paraphrases within the same language, we contribute to the main thesis of translation into low-resource languages by maximizing our use of the limited resources. 
In this chapter, we seek to leverage paraphrases for translation of a multi-source text into a new, low-resource language. 

\begin{figure}[t]
  \centering
  \begin{subfigure}[t]{.7\textwidth}
    \centering
    \includegraphics[width=4.8in]{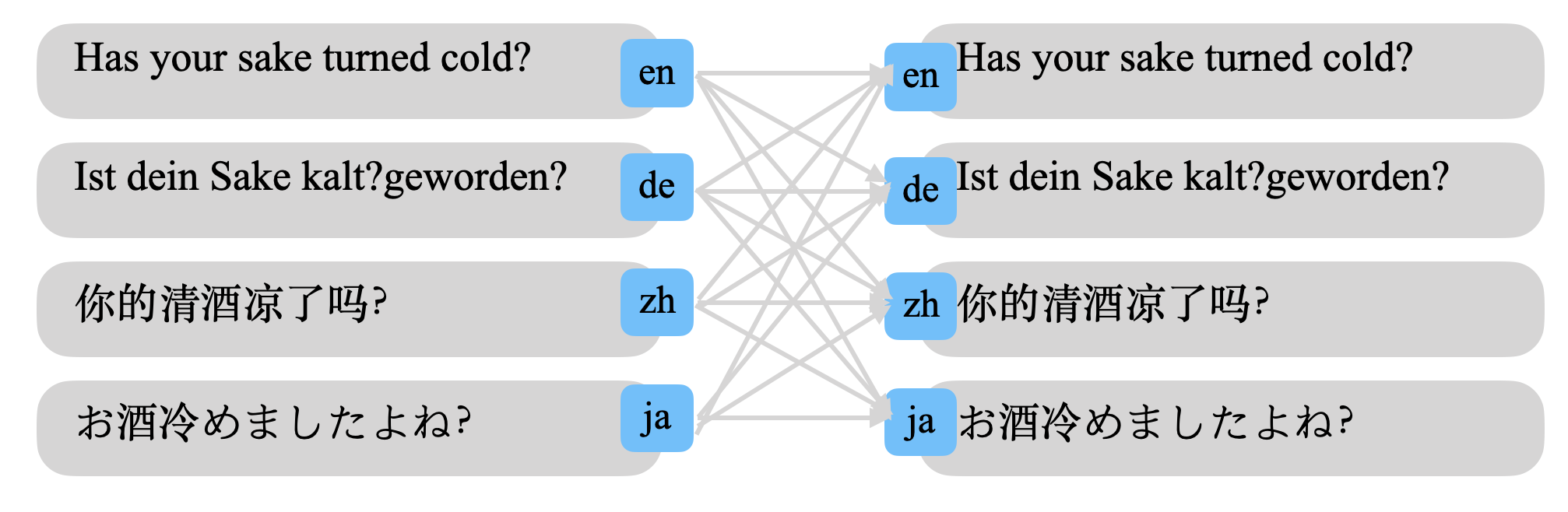}
   \caption{Multilingual model}
   \label{fig:complete}
\end{subfigure}
\begin{subfigure}[t]{.7\textwidth}
  \centering
  \includegraphics[width=4.8in]{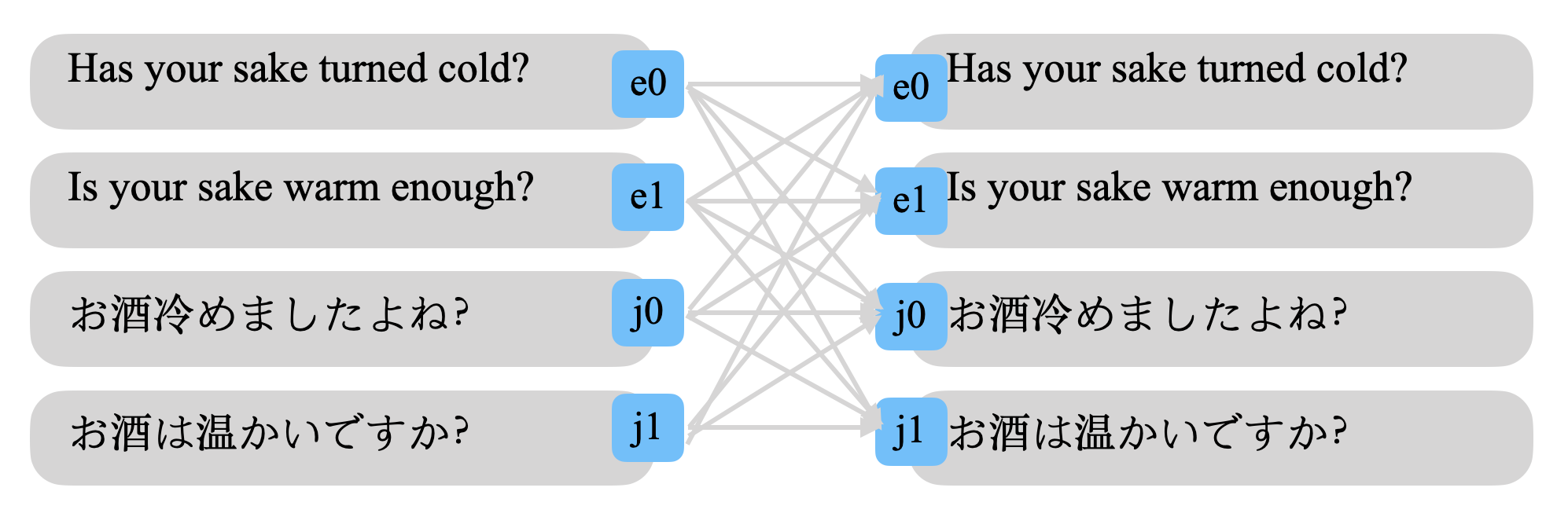}
  \caption{Multi-paraphrase model}
  \label{fig:star}
\end{subfigure}
\caption[Two configurations of translation paths]{Translation Paths in (a) multilingual model (b) multi-paraphrase model. Both form almost a complete bipartite graph. }
   \label{fig:new_graph}
\end{figure}

\section{Introduction} \label{introduction}

Paraphrases, rewordings of texts with  
preserved semantics, are often used to 
improve generalization and the sparsity issue in translation
\cite{callison2006improved, fader2013paraphrase, ganitkevitch2013ppdb, narayan2017split, sekizawa2017improving}.
Unlike previous works that use paraphrases at the word/phrase 
level, we research different translations of the 
whole corpus that are consistent in structure, content, coherence and style 
as paraphrases at the corpus level; we refer to paraphrases
as the different translation versions of the
same corpus. We train paraphrases in the style of multilingual translation systems \cite{johnson2017google, ha2016toward, firat2016multi, zoph2016multi, dong2015multi, gillick2016multilingual, al2013polyglot, tsvetkov2016polyglot}.
Implicit parameter sharing 
enables multilingual translation systems to learn across 
languages and achieve better generalization
\cite{johnson2017google}. 
Training on closely 
related languages are shown to
improve translation 
\cite{zhou2018massively}. We view 
paraphrases as an extreme case of closely related 
languages and view multilingual data as paraphrases 
in different languages. Paraphrases can 
differ randomly or systematically as each carries 
the translator's unique style. 

We treat paraphrases as foreign languages,
and train a unified model\footnote{The material in this chapter was originally 
published in SRW at ACL, 2019 \citep{zhou2018paraphrases}.} on paraphrase-labeled data
with a shared attention in the style of multilingual translation systems. 
Similar to multilingual translation systems' objective of translating
from any of the $N$ input languages to any of the
$M$ output languages \cite {firat2016multi}, multi-paraphrase 
translation systems aim to translate from any of the $N$ input paraphrases to any of the
$M$ output paraphrases in Figure~\ref{fig:new_graph}. In Figure~\ref{fig:new_graph}, we see different expressions of a host
showing courtesy to a guest to ask whether sake (a type of alcohol drink that is normally served warm in Asia)
needs to be warmed.  In Table~\ref{table:bible}, we show a few examples of parallel paraphrasing data in the Bible corpus. Different translators' styles
give rise to rich parallel paraphrasing
data, covering wide range of domains. In Table~\ref{table:if}, we also show 
some paraphrasing examples from the modern poetry dataset, which we are considering for future research. 

  \begin{table*}[t]
    \small
    \centering              
    \begin{tabularx}{\textwidth}{p{4.0cm}p{3.5cm}p{3.5cm}p{3.5cm}} 
      \toprule
      Source Sentence
      & Translation 1
      & Translation 2
      & Translation 3 \\ \midrule
      My little horse must think it queer, to stop without a farmhouse near, between the woods and frozen lake, the darkest evening of the year. ``Stopping by woods on a snowy evening'', \textit{Robert Frost}.
      &
         \begin{CJK*}{UTF8}{gkai}
        我的小马该满腹狐疑： 
        附近没农舍，为什么歇息, 
        在树林和冰湖之间的地方，
        在这一年中最暗的黃昏里. 
      \end{CJK*} Translated by \textit{An tu}. 
      &
       \begin{CJK*}{UTF8}{gkai}
        我的小马一定颇惊讶：
        四望不见有什么农家，
        偏是一年最暗的黄昏，
        寒林和冰湖之间停下. 
      \end{CJK*} Translated by \textit{Guangzhong Yu}. 
      &
       \begin{CJK*}{UTF8}{gkai}
        我的小马肯定觉得奇怪,
        附近没有房子却停下来,
        在林子和结冰的湖水间,
        一年中最为黑暗的夜晚.
\end{CJK*}
        Translated by \textit{Tiejun Yang}. 
      \\\midrule
      Two roads diverged in a wood, and I-- I took the one less traveled by, and that has made all the difference. ``The road not taken'', \textit{Robert Frost}
      &
       \begin{CJK*}{UTF8}{gkai}
      林子里有两条路，朝着两个方向，
      而我-- 我走上一条更少人迹的路，
      于是带来完全不同的一番景象.
      \end{CJK*} Translated by \textit{Ping Fang}. 
      &
        \begin{CJK*}{UTF8}{gkai}
          一片树林里分出两条路，
          而我选了人迹更少的一条，
          从此决定了我一生的道路.
      \end{CJK*} Translated by \textit{Zixing Gu}. 
      &
         \begin{CJK*}{UTF8}{gkai}
          林中两路, 岔道傍徨,
          我选一路, 人迹稀疏,　
          人生道路, 全不一样. 
      \end{CJK*} Translated by \textit{Ming Lei}. 
      \\\bottomrule
    \end{tabularx}
    \caption{Examples of parallel paraphrasing data in English-Chinese poetry translation. }
    \label{table:chinese}
  \end{table*}

Indeed,
we go
beyond the traditional MT learning of one-to-one
mapping between the source and the target text;
instead,
we exploit the many-to-many mappings between the source
and target text through training on paraphrases that are consistent
to each other at the corpus level.
Our method achieves high translation performance and gives
interesting findings. 
The differences between our work and the prior works are mainly the following.

Unlike previous works that use paraphrases at the word or phrase level,
we use paraphrases at the entire corpus level to improve
translation performance. We use different translations of the whole training
data consistent in structure as paraphrases of the
full training data. Unlike most of the multilingual MT works that uses data from
multiple languages, we use paraphrases as foreign languages in
a single-source single-target MT system training only on data from the source
and the target languages. 

Our work is meaningful because in translation into severely low-resource languages, we may have little information/data or incomplete information/data of nearby languages, not to mention translation of the text in those nearby languages. In those situations, for quality control reasons, we might have translators translating the same text into the low-resource languages in different translating and writing styles. This process provides us with paraphrases within the same language as a by-product of the human translation process. Indeed, this is another resource we could use for translation into low-resource languages. In other words, we are interested in researching into situations where there is more information in paraphrases within the same language, rather than across languages. We would like to use this as a new angle to make full use of all existing information that we can find in the target low-resource language. By using all information of paraphrases within the same language, we contribute to the main thesis of translation into low-resource languages by maximizing our use of the limited resources. When information is scarce, paraphrases within a given language becomes a powerful resource for increasing lexical diversity, improving sparsity and improving translation performance. 
Therefore, it is meaningful to leverage paraphrases for translation of a multi-source text into a new, low-resource language. 

Additionally, our work is also meaningful in contributing findings in harnessing paraphrases in MT as the following: 
\begin{enumerate}
\item{Our multi-paraphrase MT\footnote{In Chapter 5-9, all multilingual models are multilingual transformer-based. In Chapter 3-4, all models are multilingual LSTM-based model. This is because our research in Chapter 3-4 is done before transformer comes about.} results
  show significant improvements in BLEU scores
  over all baselines. }
\item{Our paraphrase-exploiting MT\footnote{We train these models from scratch because we want to assume severely low-resource situation where there is no additional resource except for a little data from the low-resource language. Though this chapter's main focus is on paraphrases, we make design decisions based on our goal in this thesis. } uses only two languages, the source
  and the target languages, and achieves higher BLEUs than the multi-source and
  multi-target MT that incorporates more languages. }
\item{We find that adding the source paraphrases helps better
  than adding the target paraphrases.}
\item{We find that adding paraphrases at both the source and the target
  sides is better than adding at either side. } 
\item{We also find that adding
  paraphrases with additional multilingual data yields mixed performance; its performance is better than training on language families alone, but is worse than training on both the source and target paraphrases without language families.} 
\item{Adding paraphrases improves
  the sparsity issue of rare word translation
  and diversity in lexical choice.}
\end{enumerate}

\section{Related Work} \label{relatedwork}

\subsection{Paraphrasing}
Many works generate and harness paraphrases
\cite{barzilay2001extracting, pang2003syntax, callison2005scaling, mallinson2017paraphrasing, ganitkevitch2013ppdb, brad2017neural, quirk2004monolingual, madnani2012re, suzuki2017building, hasan2016neural}.
Some are on question and answer 
\cite{fader2013paraphrase, dong2017learning}, evaluation of translation \cite{zhou2006re} 
and translation  
\cite{narayan2017split, sekizawa2017improving}.
Past research includes
paraphrasing unknown
words/phrases/sub-sentences \cite{callison2006improved, narayan2017split, sekizawa2017improving, fadaee2017data}.
These approaches are similar in transforming the 
difficult sparsity problem of rare words prediction and long sentence
translation into a simpler problem with known words and
short sentence translation.
It is worthwhile to contrast paraphrasing that diversifies
data, with knowledge distillation that benefits from making data
more consistent \cite{gu2017non}.

Our work is different
in that we exploit paraphrases at the corpus
level, rather than at the word or phrase level. 

\subsection{Multilingual Attentional Translation Models}
As discussed before, machine polyglotism which trains 
machines to translate  
any of the $N$ input languages to any of the
$M$ output languages
is a new paradigm in multilingual translation models \cite{firat2016multi, zoph2016multi, dong2015multi, gillick2016multilingual, al2013polyglot, tsvetkov2016polyglot}. An interesting work is training 
a universal model with a shared attention mechanism
with the source and target language labels and
Byte-Pair Encoding (BPE)
\cite{johnson2017google, ha2016toward}.
Many multilingual translation systems involve multiple encoders
and decoders \cite{ha2016toward}, and it is hard to combine
attention for quadratic language pairs bypassing
quadratic attention mechanisms
\cite{firat2016multi}. An interesting work is training 
a universal model with a shared attention mechanism
with the source and target language labels and
Byte-Pair Encoding (BPE)
\cite{johnson2017google, ha2016toward}.
This method is elegant in its 
simplicity and its 
advancement in
low-resource language 
translation and zero-shot 
translation using
pivot-based translation mechanism 
\cite{johnson2017google, firat2016multi}. 

\begin{figure}[t]
  \small
  \centering
  \includegraphics[width=5.1in]{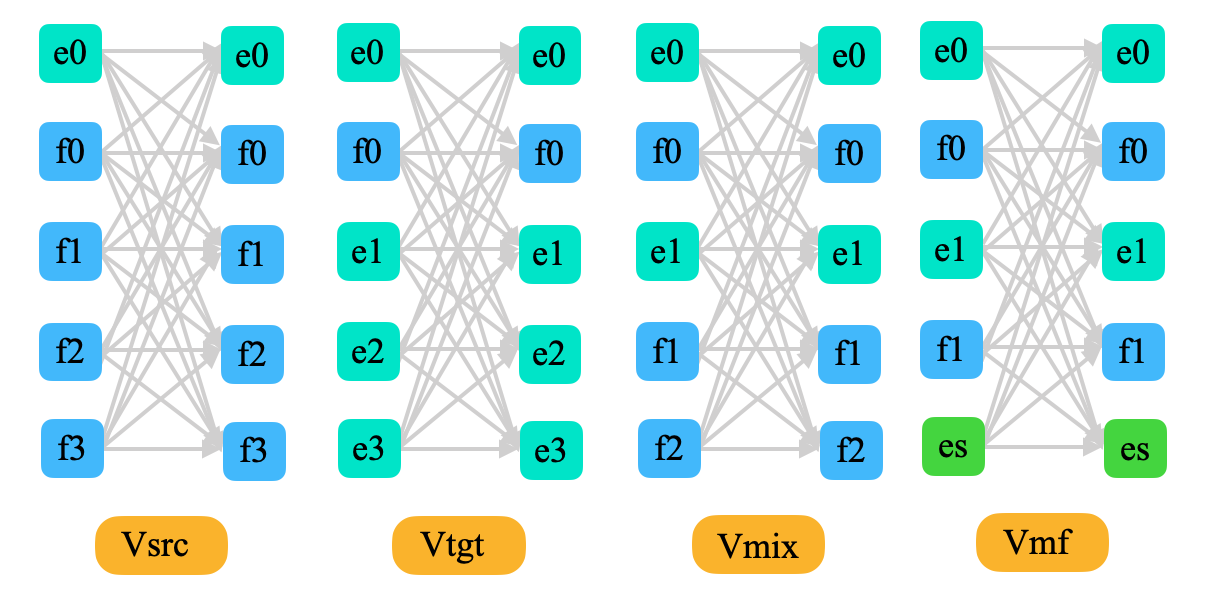}
  \caption{Examples of different ways of adding 5 paraphrases. \texttt{e[?n]} and \texttt{f[?n]} refers to different English and French paraphrases, \texttt{es} refers to the Spanish (an example member of Romance family) data. We always evaluate the translation path from \texttt{f0} to \texttt{e0}. }
\end{figure}

Unlike previous works, our parallelism
is across paraphrases, not across
languages.
In other words, we achieve higher translation
performance in the single-source single-target paraphrase-exploiting
MT systems than that of the multilingual MT systems.
  
\section{Models} \label{methodology}
We have four baseline models. Two are single-source
single-target attentional models, the other two are multilingual
MT models with a shared attention
\cite{johnson2017google, ha2016toward}.
In Figure~\ref{fig:new_graph}, we show an example of multilingual attentional
MT. Translating from all 4 languages to each other, we have 12 translation paths.
For each translation path, we
label the source sentence with the source and target language tags.
Translating from \begin{CJK*}{UTF8}{gbsn}``你的清酒凉了吗?''\end{CJK*}
to ``Has your sake turned cold?'', we label
the source sentence with \texttt{\_\_opt\_src\_zh \_\_opt\_tgt\_en}.
More details are in Section~\ref{experiments}.

In multi-paraphrase model, all source sentences
are labeled with the paraphrase tags.
For example, in French-to-English translation,
a source sentence may be tagged with \texttt{\_\_opt\_src\_f1 \_\_opt\_tgt\_e0},
denoting that it is translating from version ``f1'' of French data to
version ``e0'' of English data. 
In Figure~\ref{fig:new_graph},
we show 2 Japanese and 2 English paraphrases.
Translating from all 4
paraphrases to each other ($N=M=4$),
we have 12 translation paths as $N\times(N-1)=12$.
For each translation path, we
label the source sentence with the source and target
paraphrase tags. For
the translation path from
\begin{CJK}{UTF8}{min}``お酒冷めましたよね?''\end{CJK}
to ``Has your sake turned cold?'', we label
the source sentence with \texttt{\_\_opt\_src\_j1 \_\_opt\_tgt\_e0} in Figure~\ref{fig:new_graph}.
Paraphrases of the same translation path carry the same labels.
Our paraphrasing data is at the corpus level, and
we train a unified MT model with a shared attention.
Unlike the
paraphrasing sentences in Figure~\ref{fig:new_graph},
We show this example
with only one sentence, it is similar when the training data contains many sentences.  
All sentences in the same paraphrase path share the same labels.  
  
  \begin{table}[t]
  \small
  \centering
  \begin{tabularx}{\columnwidth}{X|XXXXXX}
    \toprule
    data & 1 & 6 & 11 & 16 & 22 & 24\\
    \midrule
    \textit{\texttt{WMT}} & 22.5 & 30.8 & 29.8 & 30.8 & 29.3 & - \\
    \textit{\texttt{Family}} & 22.5 & 39.3 & 45.4 & 49.2 & 46.6 & - \\
    \textit{\texttt{Vmix}} & 22.5 & 44.8 & 50.8 & 53.3 & 55.4 & 57.2 \\ 
    \textit{\texttt{Vmf}} & - & - & 49.3 & - & - & - \\
    \bottomrule
  \end{tabularx}
  \caption{Comparison of adding a mix of the source paraphrases and the target paraphrases against the baselines. All acronyms including data are explained in Section~\ref{experimentsbase}.
  }
  \label{table:bleu}
\end{table}
\begin{table}[t]
  \small
  \centering
  \begin{tabularx}{\columnwidth}{X|XXXX}
    \toprule
    Data & 1 & 6 & 11 & 13  \\
    \midrule
    \textit{\texttt{Vsrc}} & 22.5 & 41.4 & 48.9 & 48.8 \\ 
    \textit{\texttt{Vtgt}} & 22.5 & 40.5 & 47.0 & 47.4 \\ 
    \bottomrule
  \end{tabularx}
  \caption{Comparison of adding source paraphrases and adding target paraphrases.
    All acronyms including data are explained in Section~\ref{experimentsbase}.
  }
  \label{table:srcVStarget}
\end{table}
  
 \begin{table}[t]
  \small
  \centering
  \begin{tabularx}{\columnwidth}{X|XXXXX} 
    \midrule
    data & 6 & 11 & 16 & 22 & 24\\
    \midrule
    Entropy & 5.6569 & 5.6973 & 5.6980 & 5.7341 & 5.7130\\ 
    Bootstrap 95\% CI &(5.6564, 5.6574)&(5.6967, 5.6979)&(5.6975, 5.6986)&(5.7336, 5.7346)&(5.7125, 5.7135)\\
    \textit{\texttt{WMT}} & - & 5.7412 & 5.5746 & 5.6351 & - \\
    \bottomrule
  \end{tabularx}
  \caption{Entropy increases with the number of paraphrase corpora in \textit{\texttt{Vmix}}. The 95\% confidence interval is calculated via bootstrap resampling with replacement.
  }
  \label{table:entropy}
\end{table}
\begin{table}[t]
  \small
  \centering
  \begin{tabularx}{\columnwidth}{X|XXXXX} 
    \toprule
    data & 6 & 11 & 16 & 22 & 24\\ 
    \midrule
    F1(freq1) & 0.43 & 0.54 & 0.57 & 0.58 & 0.62 \\
    \textit{\texttt{WMT}} & - & 0.00 & 0.01 & 0.01 & -  \\
    \bottomrule
  \end{tabularx}
  \caption{F1 score of frequency 1 bucket increases with the number of paraphrase corpora in \textit{\texttt{Vmix}}, showing training on paraphrases improves the sparsity at tail and the rare word problem.}
  \label{table:f1}
\end{table}
\begin{figure}[t]
  \small
  \centering
  \includegraphics[width=5in]{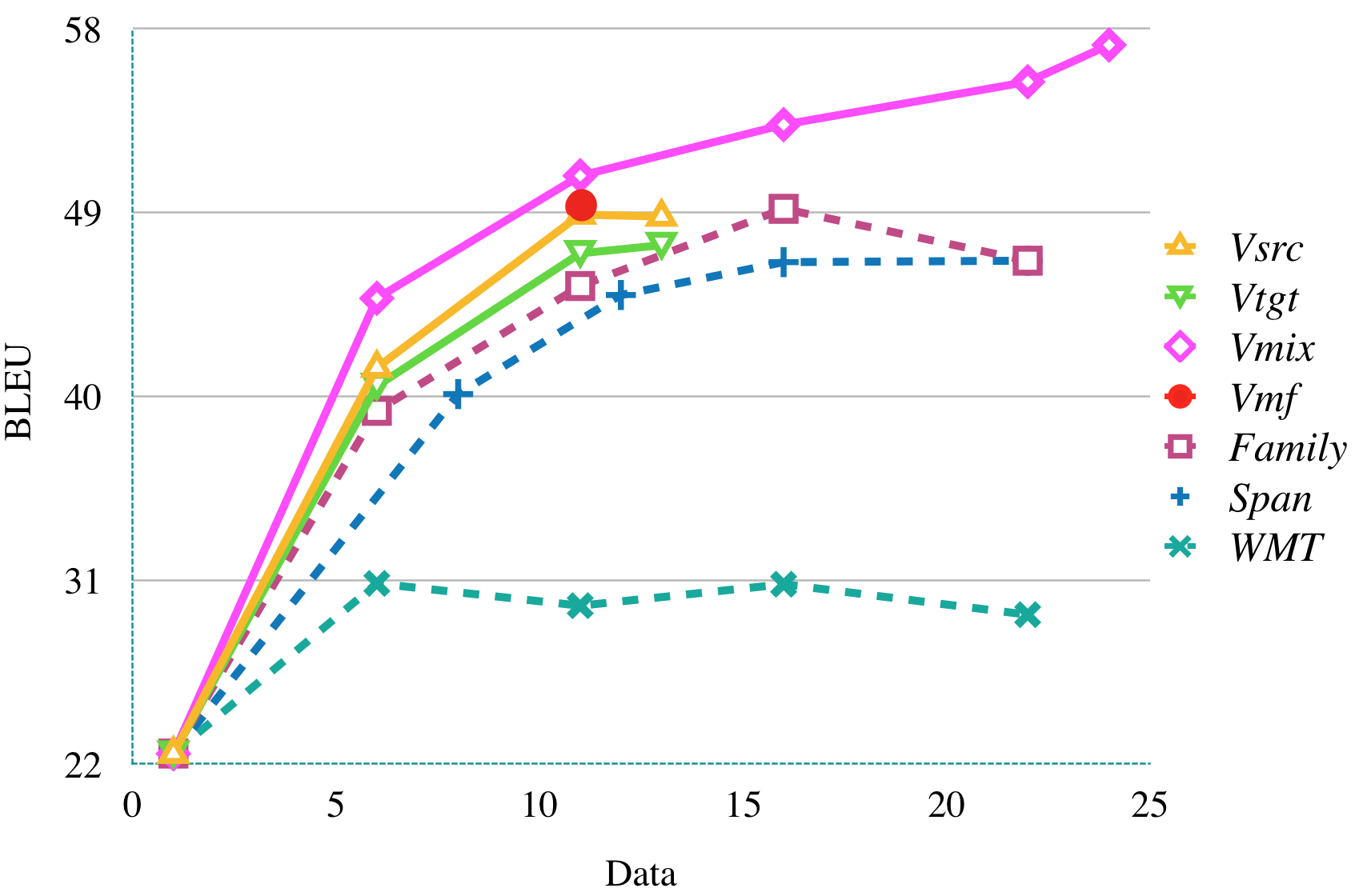}
  \caption{BLEU plots showing the effects of different ways of adding training data in French-to-English Translation. All acronyms including data are explained in Section~\ref{experimentsbase}. 
    }
  \label{fig:multi_family1}
\end{figure}

  \begin{table*}[t]
  \small
  \centering
    \begin{tabularx}{\textwidth}{X|X|X}
      \toprule
    Source Sentence
    & Machine Translation
    & Reference\\
    \midrule
    Comme de l'eau fra{\^i}che pour une personne fatigu{\'e}, Ainsi est une bonne nouvelle venant d'une terre lointaine.
      & As cold waters to a thirsty soul, so is good news from a distant land.
      & Like cold waters to a weary soul, so is a good report from a far country.
    \\
    \midrule
        Lorsque tu seras invit{\'e} par quelqu'un {\`a} des noces, ne te mets pas {\`a} la premi{\`e}re place, de peur qu'il n'y ait parmi les invit{\'e}s une personne plus consid{\'e}rable que toi,
      & When you are invited to one to the wedding, do not be to the first place, lest any one be called greater than you.
      & When you are invited by anyone to wedding feasts, do not recline at the chief seat lest one more honorable than you be invited by him,
        \\
        \midrule
    Car chaque arbre se conna{\^i}t {\`a} son fruit. On ne cueille pas des figues sur des {\'e}pines, et l'on ne vendange pas des raisins sur des ronces.
      & For each tree is known by its own fruit. For from thorns they do not gather figs, nor do they gather grapes from a bramble bush.
      & For each tree is known from its own fruit. For they do not gather figs from thorns, nor do they gather grapes from a bramble bush.
    \\
    \midrule
    Nous la c{\^o}toy{\^a}mes avec peine, et nous arriv{\^a}mes {\`a} un lieu nomm{\'e} Beaux Ports, pr{\`e}s duquel {\'e}tait la ville de Las{\'e}e.
      &  And coasting along it with difficulty, we came to a certain place called Fair Havens, near to which was the city of Lasea.
      & And coasting along it with difficulty, we came to a certain place named Fair Havens, near to which was a city , Lasea.
    \\
    \midrule
    Vous tous qui avez soif, venez aux eaux, Même celui qui n'a pas d'argent! Venez, achetez et mangez, Venez, achetez du vin et du lait, sans argent, sans rien payer!
    &Come, all you thirsty ones, come to the waters; come, buy and eat. Come, buy for wine, and for nothing, for without money.
    & Ho, everyone who thirsts, come to the water; and he who has no silver, come buy grain and eat. Yes, come buy grain, wine and milk without silver and with no price.
    \\
    \midrule
    Oui , vous sortirez avec joie , Et vous serez conduits en paix ; Les montagnes et les collines éclateront d'allégresse devant vous , Et tous les arbres de la campagne battront des mains .
    & When you go out with joy , you shall go in peace ; the mountains shall rejoice before you , and the trees of the field shall strike all the trees of the field .
    & For you shall go out with joy and be led out with peace . The mountains and the hills shall break out into song before you , and all the trees of the field shall clap the palm .
    \\
    \midrule
  \end{tabularx}
  \caption{Examples of French-to-English translation trained using 12 French paraphrases and 12 English paraphrases.}
  \label{table:french}
  \end{table*}

    \begin{table*}[ht]
    \small
    \centering
    \begin{tabularx}{\textwidth}{p{2.9cm} |X} \toprule
      English &
      If you can fill the unforgiving minute
      with sixty seconds' worth of distance run,
      yours is the Earth and everything that's in it,
      and—which is more—you'll be a Man, my son!
      ``if'', \textit{Rudyard Kipling}. \\\midrule
      \multirow{4}{*}{German}
      &
      Wenn du in unverzeihlicher Minute
      Sechzig Minuten lang verzeihen kannst:
      Dein ist die Welt—und alles was darin ist—
      Und was noch mehr ist—dann bist du ein Mensch!
      Translation by \textit{Anja Hauptmann}.  \\\cline{2-2}
      &
      Wenn du erfüllst die herzlose Minute
      Mit tiefstem Sinn, empfange deinen Lohn:
      Dein ist die Welt mit jedem Attribute,
      Und mehr noch: dann bist du ein Mensch, mein Sohn!
      Translation by \textit{Izzy Cartwell}.  \\\cline{2-2}
      &
      Füllst jede unerbittliche Minute
      Mit sechzig sinnvollen Sekunden an;
      Dein ist die Erde dann mit allem Gute,
      Und was noch mehr, mein Sohn: Du bist ein Mann!
      Translation by \textit{Lothar Sauer}. \\\midrule
      \multirow{4}{*}{Chinese}
      &
      \begin{CJK*}{UTF8}{gkai}
        若胸有激雷, 而能面如平湖, 则山川丘壑, 天地万物皆与尔共, 吾儿终成人也！
      \end{CJK*} Translation by \textit{Anonymous}.
        \\\cline{2-2}
      &
        \begin{CJK*}{UTF8}{gkai}
        如果你能惜时如金利用每一分钟不可追回的光阴；那么，你的修为就会如天地般博大，并拥有了属于自己的世界，更重要的是：孩子，你成为了真正顶天立地之人！
        \end{CJK*} Translation by \textit{Anonymous}.
        \\\cline{2-2}
      &
          \begin{CJK*}{UTF8}{gkai}
            假如你能把每一分宝贵的光阴,
化作六十秒的奋斗—-你就拥有了整个世界, 最重要的是——你就成了一个真正的人，我的孩子！
          \end{CJK*} Translation by \textit{Shan Li}.
          \\
          \midrule
      \multirow{4}{*}{Portuguese}
      &
      Se você puder preencher o valor do inclemente minuto perdido
      com os sessenta segundos ganhos numa longa corrida,
      sua será a Terra, junto com tudo que nela existe,
      e—mais importante—você será um Homem, meu filho!
      Translation by \textit{Dascomb Barddal}. \\\cline{2-2}
      &
      Pairando numa esfera acima deste plano,
      Sem receares jamais que os erros te retomem,
      Quando já nada houver em ti que seja humano,
      Alegra-te, meu filho, então serás um homem!...
      Translation by \textit{Féliz Bermudes}.\\\cline{2-2}
      &
      Se és capaz de dar, segundo por segundo,
      ao minuto fatal todo valor e brilho.
      Tua é a Terra com tudo o que existe no mundo,
      e—o que ainda é muito mais—és um Homem, meu filho!
      Translation by \textit{Guilherme de Almeida}. \\\bottomrule
    \end{tabularx}
    \caption{Examples of parallel paraphrasing data with German, Chinese, and Portuguese paraphrases of the English poem ``If'' by Rudyard Kipling.}
    \label{table:if}
  \end{table*}
  \begin{table*}[ht]
    \small
    \centering
    \begin{tabularx}{\columnwidth}{p{2.9cm} |X}
      \toprule
      \multirow{4}{*}{English}
      & Consider the lilies, how they grow: they neither toil nor spin, yet I tell you, even Solomon in all his glory was not arrayed like one of these. \textit{English Standard Version}.\\\cline{2-2}
      & Look how the wild flowers grow! They don't work hard to make their clothes. But I tell you Solomon with all his wealth wasn't as well clothed as one of these flowers. \textit{Contemporary English Version}.\\\cline{2-2}
      & Consider how the wild flowers grow. They do not labor or spin. Yet I tell you, not even Solomon in all his splendor was dressed like one of these.  \textit{New International Version}.
      \\
      \midrule
      \multirow{4}{*}{French}
      & Considérez les lis! Ils poussent sans se fatiguer à tisser des vêtements. Et pourtant, je vous l’assure, le roi Salomon lui-même, dans toute sa gloire, n’a jamais été aussi bien vêtu que l’un d’eux! \textit{La Bible du Semeur}.\\\cline{2-2}
      & Considérez comment croissent les lis: ils ne travaillent ni ne filent; cependant je vous dis que Salomon même, dans toute sa gloire, n'a pas été vêtu comme l'un d'eux. \textit{Louis Segond}. \\\cline{2-2}
      & Observez comment poussent les plus belles fleurs: elles ne travaillent pas et ne tissent pas; cependant je vous dis que Salomon lui-même, dans toute sa gloire, n'a pas eu d’aussi belles tenues que l'une d'elles. \textit{Segond 21}.
      \\
      \midrule
      \multirow{4}{*}{Tagalog}
      & Wariin ninyo ang mga lirio, kung paano silang nagsisilaki: hindi nangagpapagal, o nangagsusulid man; gayon ma'y sinasabi ko sa inyo, Kahit si Salomon man, sa buong kaluwalhatian niya, ay hindi nakapaggayak na gaya ng isa sa mga ito. \textit{Ang Biblia 1978}. \\\cline{2-2}
      & Isipin ninyo ang mga liryo kung papaano sila lumalaki. Hindi sila nagpapagal o nag-iikid. Gayunman, sinasabi ko sa inyo: Maging si Solomon, sa kaniyang buong kaluwalhatian ay hindi nadamitan ng tulad sa isa sa mga ito. \textit{Ang Salita ng Diyos}. \\\cline{2-2}
      & Tingnan ninyo ang mga bulaklak sa parang kung paano sila lumalago. Hindi sila nagtatrabaho ni humahabi man. Ngunit sinasabi ko sa inyo, kahit si Solomon sa kanyang karangyaan ay hindi nakapagdamit ng singganda ng isa sa mga bulaklak na ito. \textit{Magandang Balita Biblia}.
      \\
      \midrule
      \multirow{4}{*}{Spanish}
      &	Considerad los lirios, cómo crecen; no trabajan ni hilan; pero os digo que ni Salomón en toda su gloria se vistió como uno de éstos. \textit{La Biblia de las Américas}. \\\cline{2-2}
      & Fíjense cómo crecen los lirios. No trabajan ni hilan; sin embargo, les digo que ni siquiera Salomón, con todo su esplendor, se vestía como uno de ellos. \textit{Nueva Biblia al Día}. \\\cline{2-2}
      &	Aprendan de las flores del campo: no trabajan para hacerse sus vestidos y, sin embargo, les aseguro que ni el rey Salomón, con todas sus riquezas, se vistió tan bien como ellas. \textit{Traducción en lenguaje actual}. 
      \\
      \bottomrule
    \end{tabularx}
    \caption{Examples of parallel paraphrasing data with English, French, Tagalog and Spanish paraphrases in Bible translation. }
    \label{table:bible}
  \end{table*}
  
\section{Experiments}\label{experiments}
\subsection{Data}\label{experimentsdata}
Our main data is the French-to-English Bible corpus
\cite{mayer2014creating}, containing 12 versions
of the English Bible and 12 versions of the French Bible
\footnote{We considered the open subtitles
  with different scripts of
  the same movie in the same language; they
  covers many topics, but they are noisy and
  only differ in interjections.
  We also considered the poetry dataset where a poem
  like ``If'' by Rudyard Kipling
  is translated many times, 
  by various people into the same language,
  but the data is small.}.
We translate from French to English. 
Since these 24 translation versions are consistent in structure,
we refer to them as paraphrases at corpus level.
In our work, each paraphrase refers to
each translation version of whole Bible corpus.
To understand our setup, if we use all
12 French paraphrases and all 12 English paraphrases
so there are 24 paraphrases in total, i.e., $N=M=24$,
we have 552 translation paths because $N\times(N-1)=552$.

To prepare data for this experimental setup, we clean and align 24
paraphrases of the Bible corpus because the original corpus contains missing or extra verses for different
paraphrases. After cleaning and aligning all 24 paraphrases of the Bible corpus, we randomly sample the training, validation and test
sets according to the 0.75, 0.15, 0.10 ratio.
Our training set contains only 23K verses,
but is massively parallel across paraphrases.

For all experiments, we choose a specific English corpus
as \texttt{e0} and a specific French corpus as \texttt{f0} which we evaluate across all
experiments to ensure consistency in comparison, and 
we evaluate all translation performance from \texttt{f0} to \texttt{e0}.
The reason we choose this particular language pair is because this language pair from \texttt{f0} to \texttt{e0} is the language pair that is part of the multilingual baseline that we are comparing our results with. We want to have the same language pair to ensure consistency when we compare all experiments against the baselines. This way of evaluation is necessary for our comparison with the baselines but it is also a limitation of our research which we will discuss at the end of this chapter. 

\subsection{Training Parameters}\label{experimentspara}
  In all our experiments,
we use a minibatch size of 64,
dropout rate of 0.3,
4 RNN layers of size 1000,
a word vector size of 600,
number of epochs of 13,
a learning rate of 0.8 that decays at
the rate of 0.7 if the validation score
is not improving
or it is past epoch 9
across all LSTM-based experiments. Byte-Pair Encoding (BPE) is used at preprocessing stage \cite{ha2016toward}.
Our code is built on OpenNMT \cite{klein2017opennmt} and we evaluate our models using
BLEU scores \cite{papineni2002bleu}, entropy \cite{shannon1951prediction},
F-measure and qualitative evaluation. 

\subsection{Baselines}\label{experimentsbase}
We introduce a few acronyms for our four baselines to describe the experiments
in Table~\ref{table:srcVStarget}, Table~\ref{table:bleu}
and Figure~\ref{fig:multi_family1}. Firstly, we have two single-source
single-target attentional MT models, \textit{\texttt{Single}} and
\textit{\texttt{WMT}}. \textit{\texttt{Single}} trains
on \texttt{f0} and \texttt{e0} and gives a BLEU of 22.5,
the starting point for all curves in Figure~\ref{fig:multi_family1}.
\textit{\texttt{WMT}} adds the
out-domain WMT'14 French-to-English data
on top of \texttt{f0} and \texttt{e0}; it serves as
a weak baseline that helps us to evaluate all
experiments' performance
discounting the effect of increasing data.

Moreover, we have two multilingual baselines\footnote{
  For multilingual baselines, we use the
  additional Bible corpus in
  22 European languages that are cleaned and
  aligned to each other.} built on multilingual
attentional MT, \textit{\texttt{Family}}
and \textit{\texttt{Span}} \cite{zhou2018massively}. 
\textit{\texttt{Family}} refers to the multilingual
baseline by adding one language family at a time,
where on top of the French corpus \texttt{f0}
and the English corpus \texttt{e0}, we add up to 20
other European languages.
\textit{\texttt{Span}} refers
to the multilingual
baseline by adding one \textit{span} at a time,
where a span is a set of languages
that contains at least one language
from all the families in the data; in
other words, span is a sparse
representation of all the families.
Both \textit{\texttt{Family}}
and \textit{\texttt{Span}}
trains on the Bible in 22 Europeans languages trained using
multilingual MT.
Since \textit{\texttt{Span}} is always sub-optimal
to \textit{\texttt{Family}} in our results,
we only show numerical results
for \textit{\texttt{Family}} in Table~\ref{table:srcVStarget}
and Table~\ref{table:bleu}, and we plot both \textit{\texttt{Family}}
and \textit{\texttt{Span}} in Figure~\ref{fig:multi_family1}.
The two multilingual baselines are strong baselines while
the \textit{\texttt{WMT}} baseline is a weak baseline that helps us to evaluate all
experiments' performance
discounting the effect of increasing data.
All baseline results are taken
from a research work which uses the grid of
(1, 6, 11, 16, 22) for the number
of languages or equivalent number of
unique sentences and we follow the same in Figure~\ref{fig:multi_family1}
\cite{zhou2018massively}.
All experiments for each grid point carry the same number of unique sentences.

Furthermore, \textit{\texttt{Vsrc}}
refers to adding more source (English) paraphrases,
and \textit{\texttt{Vtgt}} refers to
adding more target (French) paraphrases.
\textit{\texttt{Vmix}} refers to
adding both the source and the target paraphrases.
\textit{\texttt{Vmf}} refers to combining
\textit{\texttt{Vmix}} with additional multilingual data;
note that only \textit{\texttt{Vmf}},
\textit{\texttt{Family}} and \textit{\texttt{Span}} 
use languages other than French and English,
all other experiments use only English and French. For the x-axis,
data refers to the number of paraphrase corpora for
\textit{\texttt{Vsrc}}, \textit{\texttt{Vtgt}}, \textit{\texttt{Vmix}}; data refers
to the number of languages for \textit{\texttt{Family}}; data refers to
and the equivalent number of unique training sentences compared to other training curves for
\textit{\texttt{WMT}} and \textit{\texttt{Vmf}}.

\section{Results}
\textbf{\ul{Training on paraphrases gives better performance than all baselines:}} The
translation performance of training
on 22 paraphrases, i.e., 11 English paraphrases and 11 French paraphrases, 
achieves a BLEU	score of 55.4, which is +32.9 above the \textit{\texttt{Single}} baseline,
+8.8 above the \textit{\texttt{Family}} baseline,
and +26.1 above the \textit{\texttt{WMT}} baseline. 
Note that the \textit{\texttt{Family}} baseline
uses the grid of (1, 6, 11, 16, 22) for number of languages,
we continue to use this grid for our results on number of
paraphrases,
which explains why we pick 22 as an example here. 
The highest BLEU 57.2 is achieved when we train on 24
paraphrases, i.e., 12 English paraphrases
and 12 French paraphrases. 

\textbf{\ul{Adding the source paraphrases boosts translation performance more than adding the target paraphrases:}} The
translation performance	of adding
the source paraphrases is higher than that of adding
the target paraphrases. Adding the source
paraphrases diversifies the data,
exposes the model to more rare words,
and enables better generalization. 
Take the experiments training on 13 paraphrases for example,
training on the source (i.e., 12 French paraphrases and the English paraphrase \texttt{e0}) gives a BLEU score of 48.8,
which has a gain of +1.4 over 47.4, the
BLEU score of training on the target (i.e., 12 English paraphrases and the
French paraphrase \texttt{f0}).
This suggests that adding
the source paraphrases is more effective than adding
the target paraphrases.

\textbf{\ul{Adding paraphrases from both sides is better than adding paraphrases from either side:}}
The curve of adding paraphrases from both the source and the target sides
is higher than both the
curve of adding the target paraphrases and
the curve of adding the source paraphrases.
Training on 11 paraphrases from
both sides, i.e., a total of 22 paraphrases achieves a BLEU score of 50.8,
which is +3.8 higher than that of
training on the target
side only and +1.9 higher than that of
training on the source side only. 
The advantage of combining both sides is that we can
combine paraphrases from both the source and
the target to reach 24 paraphrases in total to achieve a BLEU score
of 57.2.

\textbf{\ul{Adding both paraphrases and language families yields mixed performance:}}
We conduct one more experiment combining the source and target paraphrases together with
additional multilingual data. This is the only experiment on paraphrases 
where we use multilingual data
other than only French and English data. The BLEU score is 49.3, 
higher than training on families alone, in fact,
it is higher than training on eight European families altogether. 
However, it is lower than training on English and French paraphrases alone.
Indeed, adding
paraphrases as foreign languages is effective,
however, when there is a lack of
data, mixing the paraphrases with
multilingual data is helpful. 

\textbf{\ul{Adding paraphrases increases entropy and diversity in lexical choice, and improves the
    sparsity issue of rare words: }} We
use bootstrap resampling and construct 95\% confidence intervals for entropies \cite{shannon1951prediction}
of all models of \textit{\texttt{Vmix}}, i.e., models adding
paraphrases at both the source and the target sides. We find that
the more paraphrases, the higher
the entropy, the more diversity
in lexical choice as shown in Table~\ref{table:entropy}.
From the word F-measure shown in Table~\ref{table:f1},
we find that the more paraphrases, the better the model handles the
sparsity of rare words issue. 
Adding paraphrases not only achieves much higher
BLEU score than the \textit{\texttt{WMT}} baseline, but
also handles the sparsity issue
much better than the \textit{\texttt{WMT}} baseline. 

\textbf{\ul{Adding paraphrases helps rhetoric translation and increases expressiveness:}} Qualitative
evaluation shows many cases where rhetoric translation is
improved by training on diverse sets of paraphrases.
In Table~\ref{table:french},
Paraphrases help the model to
use a more contemporary synonym of ``silver'', ``money'',
which is more direct and
easier to understand.
Paraphrases simplifies the rhetorical or subtle
expressions, for example, our model uses ``rejoice'' to replace
``break out into song'', a personification device of mountains to describe joy, 
which captures the essence of the meaning being conveyed. 
However, we also observe
that the model wrongly translates ``clap the palm'' to ``strike''.
We find the quality of rhetorical translation ties closely with
the diversity of parallel paraphrases data. 
Indeed, the use of paraphrases to improve rhetoric translation is
a good future research question.
Please refer to the Table~\ref{table:french} for more qualitative examples.
  
\section{Conclusion} \label{conclusion}
We train on paraphrases as
foreign languages in the style of
multilingual translation systems. Adding
paraphrases improves translation quality, the rare word issue,
and diversity in lexical choice. 
Adding the source paraphrases helps more than adding the target
ones, while combining both boosts performance further.
Adding multilingual data to paraphrases yields mixed performance. 

Having understood our main contributions in this chapter, we need to understand a limitation of this research in our evaluation setup. Since we would like to measure how much improvement we have achieved above the multilingual baseline where there is a particular language pair (paraphrase pair) present, we can only choose this particular language pair (paraphrase pair) for evaluation across all experiments. Moreover, we want to explore using all of the existing paraphrases as alternative references in a multi-reference evaluation manner. However, we did not choose this setup, instead we chose to use a particular language pair (paraphrase pair) that is part of the multilingual baseline to ensure consistency in comparison. This is a necessary design decision, but is also a limitation to our research. In the future, we are interested in multi-reference evaluation on top of the existing evaluation mechanism. 

A natural question one may ask is that if we have
known target paraphrases available, is there still
a need to translate it in the first place?
Our answer is yes. Each person's writing
and translation style carries a unique signature,
which is exactly what the machine learns. For example,
if we have a text translated for adults, and
we need to paraphrase for adolescents, the machine
needs to learn the tone and style of adolescents.
In this setting, our finding is very useful for
training the translation of text with a specific 
style, and translation of text that is requires
a specific level of formality. 
One may also ask, 
how does our method compare with monolingual adaptation 
in the target language? Why do we do multi-paraphrase training if we can just adapt on monolingual data? This is because the nature of the low-resource scenarios in our use case. We have little to no data in the target low-resource language. As we move to the following chapters, we will use as low as $\sim$600 sentences in the target language.  The intention of this chapter is to contribute to the main thesis, which is to how to address machine translation and style adaptation under severely low-resource scenarios. Indeed, multi-paraphrase training provides better generalization and avoids overfitting than using monolingual adaptation on severely low-resource languages. 

We would like to explore the common structure
and terminology consistency 
across different paraphrases. Since structure and terminology 
are shared across paraphrases, we are interested in building 
an explicit representation of the paraphrases and extending our work 
for better translation, or translation with more explicit and more 
explainable hidden states, which is very important in all neural 
systems. 
All of these have immeasurable value for translation into low-resource languages. In low-resource situations, we might have translators translating the same text into the
low-resource languages in different translating and writing styles. Paraphrases within the
same language could be therefore a by-product of human translation process, and is another
resource we could use for translation into low-resource languages. By using all information of paraphrases within the same language, we contribute
to the main thesis of translation into low-resource languages by maximizing the use of our limited resources. 

We are interested in broadening our dataset in our future experiments. 
In addition to the parallel paraphrasing data as shown
in the Bible corpus in Table~\ref{table:bible}, 
we also show some paraphrasing examples
from the modern poetry dataset in Table~\ref{table:if}.
There are very few poems that are
translated multiple times into the same
language, we therefore need to train
on extremely small dataset.
Rhetoric in paraphrasing is important in poetry
dataset, which again depends on the training
paraphrases. The limited data issue is also
relevant to the low-resource setting.

We would like to effectively train on an extremely small
low-resource paraphrasing data. As discussed above about the 
potential research poetry dataset, dataset with multiple paraphrases 
is typically small and yet valuable. If we can train using extremely 
small amount of data, especially in the low-resource scenario, we would 
exploit the power of multi-paraphrase translation further. 

Cultural-aware paraphrasing and subtle expressions are
vital \cite{levin1998interlingua, larson1984meaning}. Rhetoric in paraphrasing is a 
very important tool. 
In Figure~\ref{fig:new_graph}, ``is your sake warm enough?'' in Asian culture 
is an implicit way of
saying ``would you like me to warm the sake for you?''.
We would like to model the culture-specific
subtlety through multi-paraphrase training. 

In this chapter and the previous chapter, we 
have explored ways to leverage information 
in paraphrases in the same 
language and treat them as foreign languages, as well as 
nearby languages within the same or nearby language families. 
However, in real-world scenarios, information of which language is close may not be available, 
and information of paraphrases within the same language might also not be available. 
In the next chapter, we focus on these information-scarce situations, and we research how to make best use of the available multi-source text, and build multiple metric spaces of language closeness statistically to facilitate training.

\removelabelprefix

\chapter{Building Language Family with Incomplete Information}\label{big:ipml}
\addlabelprefix{4}         
\epigraph{``The art of translation lies less in knowing the other language than in knowing your own.''}{\textit{Ned Rorem}}

\lettrine{I}{n the previous two chapters}, 
we have carefully examined ways to make the best use of information 
on nearby languages/ language families as well as different paraphrases within the same language 
for our task of translation of a multi-source text into a new, low-resource language. However, in real-world 
situations, such information may not be complete. We may not have access to multiple paraphrases within the same language and we may have limited information about which languages are close by and which language family the target language belongs to. In such situations, we can best make use of the existing multi-source text by 
statistically building metric spaces of language closeness to predict which languages are close to the target language. We can then use the synthetic language families that we built for multilingual training. 

\section{Introduction} \label{introduction}

\begin{figure*}[t]
  \centering
  \includegraphics[width=0.9\linewidth]{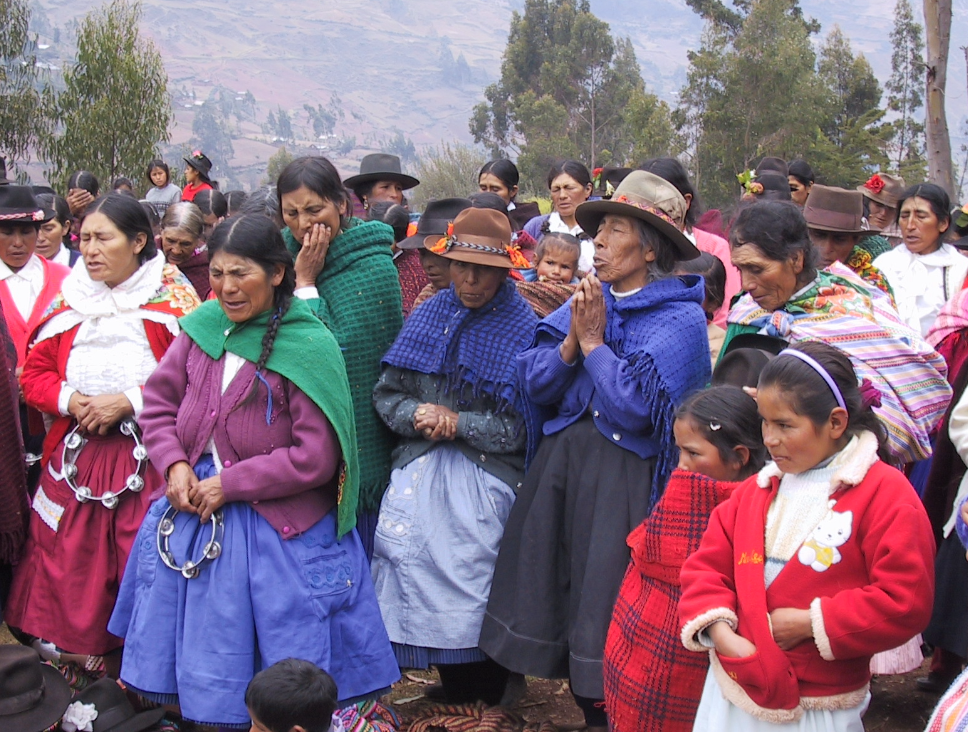}
    \caption{A Quechua-speaking community gathering in Peru. Photograph by Mark Bean. }
    \label{fig:peru_read}  
\end{figure*}

2020 is the year that we started the
life-saving hand washing practice globally.
Applications like translating water, sanitation,
and hygiene (WASH) guidelines into severely
low-resource languages are very impactful in
tribes like those in Papua New Guinea with
839 living languages \cite{gordon2005ethnologue, simons2017ethnologue}.
Translating humanitarian texts like WASH guidelines
with scarce data and expert help is key \cite{bird2020decolonising}. 

\begin{table}[t]
  \small
  \centering
  \begin{tabularx}{\columnwidth}{ss|ss} 
    \toprule
    \multicolumn{2}{w|}{Eastern Pokomchi} & \multicolumn{2}{w}{English} \\ 
    \midrule
    \textit{FAMD} & \textit{FAMP} & \textit{FAMD} & \textit{FAMP} \\ 
    \midrule
    Chuj* & Dadibi & Danish* & Dutch* \\ 
    Cakchiquel* & Thai & Norwegian* & Afrikaans* \\ 
    Guajajara* & Gumatj & Italian & Norwegian* \\ 
    Toba & Navajo & Afrikaans* & German* \\
    Myanmar & Cakchiquel* & Dutch* & Danish* \\
    Slovenský & Kanjobal & Portuguese & Spanish \\ 
    Latin & Guajajara* & French & Frisian* \\
    Ilokano & Mam* & German* & Italian \\ 
    Norwegian & Kim & Marshallese & French \\ 
    Russian & Chuj* & Frisian* & Portuguese \\ 
    \bottomrule
  \end{tabularx}
    \caption{Top ten languages closest to Eastern Pokomchi (left)
      and English (right) in ranking 124 source languages. \textit{FAMD} and \textit{FAMP} are two
      constructions of Family of Choice (\textit{FAMC}) by distortion and performance metrics respectively. 
      All are trained on $\sim$1,000 lines. 
      We star those in Family of Origin.}
    \label{table:ranking}
\end{table}

Inspired by such applications, we translate multilingually known texts into a severely low-resource language. In the translation process, we focus on five challenges that are not addressed previously\footnote{The material in this chapter was originally 
published in SIGTYP at NAACL, 2021 \citep{zhou2021family}.}.
Most multilingual transformer works that translate
into low-resource language
limit their training
data to available data in the same
or close-by language families 
or the researchers' intuitive discretion;
and are mostly limited to less than 30 languages
\cite{gu2018universal, zhou2018massively, zhu2020language}.
Instead, we examine ways to pick useful
source languages from 124 source languages in a principled fashion. 
Secondly, most works require at least
4,000 lines of low-resource data \cite{lin2020pre, qi2018and, zhou2018massively};
we use 
only $\sim$1,000 lines of low-resource data to simulate real-life situation
of having extremely small seed target translation.
Thirdly, many works use rich-resource languages as hypothetical
low-resource languages. 
Moreover, most works do not treat named entities separately; 
we add an order-preserving lexiconized component
for more accurate translation of named
entities. Finally, many multilingual works present final
results as sets of translations from all source languages;
we build a novel method to combine all translations into one. 

We have five contributions. Firstly, we rank the
124 source languages to determine their
closeness to the low-resource language and choose the top few. 
We call the linguistic definition of
language family \textit{Family of Origin} (FAMO), and
we call the empirical definition of higher-ranked 
languages using our metrics 
\textit{Family of Choice} (FAMC).
They often overlap, but may not coincide. 

Secondly, we build an
\textit{Iteratively Pretrained Multilingual Order-preserving Lexiconized Transformer} (IPML)
training on $\sim$1,000 lines
of low-resource data. 
Using iterative pretraining, 
we get a +23.9 BLEU increase over
a multilingual order-preserving lexiconized transformer
baseline (MLc)
using English as a hypothetical low-resource
language, and a +10.3 BLEU increase over our asymmetric baseline. 
Training with the low-resource
language on both the source and target sides
boosts translation into the target side.
Training on 1,093 lines from the book of Luke, 
we reach a 23.7 BLEU score testing on 30,022 lines of Bible.
We have a 42.8 BLEU score for Portuguese-English translation
on the medical EMEA dataset.

Thirdly, we use a real-life severely
low-resource Mayan language, Eastern Pokomchi,
a Class 0 language \cite{joshi2020state}
as one of our experiment
setups. In addition, we also use English
as a hypothetical low-resource language for easy 
evaluation. 

We also add an order-preserving lexiconized
component to translate named entities well. To
solve the variable-binding
problem to distinguish ``Ian calls Yi''
from ``Yi calls Ian'' \cite{fodor1988connectionism, graves2014neural, zhou2018massively},
we build a lexicon table for
2,939 Bible named entities in 124 source languages
including more than 66 severely low-resource languages. 

Finally, we combine translations from all source
languages by using a novel method. For every sentence,
we find the translation that is closest to
the translation cluster center.
The expected BLEU score of our combined translation
is higher than translation from any of the individual source. 

\section{Related Works} \label{relatedwork}
\subsection{Multilingual Pretraining}
Recent research on machine polyglotism involves
training machines to be adept in many languages
by adding language labels in the training data
with a single attention 
\cite{johnson2017google, ha2016toward, firat2016multi, gillick2016multilingual, zhou2018paraphrases}. Some explores data symmetry \cite{freitag2020complete, birch2008predicting, lin2019choosing}.
Zero-shot translation in severely low-resource settings
exploits the massive multilinguality, cross-lingual
transfer, pretraining, iterative back-translation
and freezing subnetworks
\cite{lauscher2020zero, nooralahzadeh2020zero, wang2020negative, li2020learn, pfeiffer2020mad, baziotis2020language, chronopoulou2020reusing, lin2020pre, thompson2018freezing, luong2014addressing, wei2020iterative, dou2020dynamic}.

\subsection{Linguistic Distance}
To construct linguistic distances
\cite{hajic2000machine, oncevay2020bridging, chung2020improving},
researchers explore
typological distance
\cite{chowdhury2020understanding, rama2012good, pienemann2005processability, svalberg1998english, hansen2012beginning, comrie2005world},
on World Atlas of Language Structures \cite{comrie2005world},
lexical distance on the Swadesh list \cite{huang2007towards},
normalized Levenshtein distance and Jaccard distance
\cite{serva2008indo, holman2008advances, adebara2020translating},
sonority distance \cite{parker2012sonority},
trade cost \cite{isphording2013costs},
structural distance on immigrants' fluency \cite{chiswick2005linguistic}
and spectral distance \cite{dubossarsky2020secret}.

\begin{figure}[t]
  \centering
  \begin{subfigure}[b]{0.7\textwidth}
  \centering
  \includegraphics[width=0.9\linewidth]{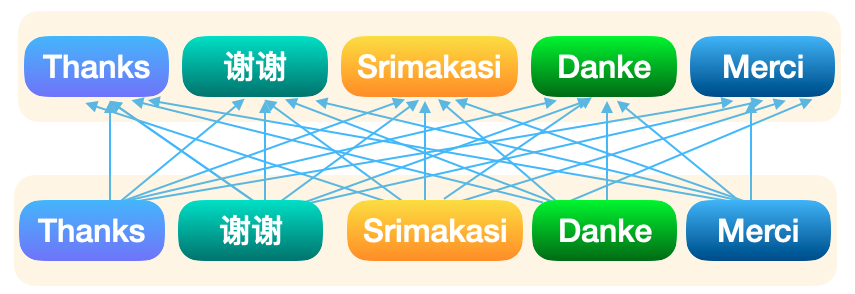}
   \caption{Complete graph configuration}
   \label{fig:complete}
\end{subfigure}
\begin{subfigure}[b]{0.7\textwidth}
  \centering
  \includegraphics[width=0.9\linewidth]{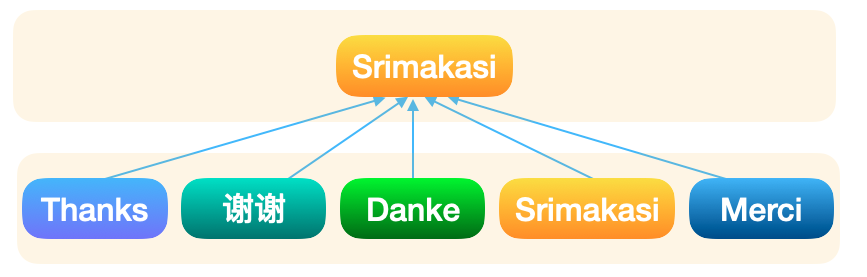}
  \caption{Star graph configuration}
  \label{fig:star}
\end{subfigure}
\caption[Two configurations of translation paths]{(a) Complete
  graph configuration of translation paths (Many-to-many)
  in an example of multilingual translation. (b)
  Star configuration of translation paths (Many-to-one) using Indonesian
  as the low-resource example.}
\label{fig:config}
\end{figure}

\section{Methodology}\label{method}

\subsection{Multilingual Order-preserving Lexiconized Transformer}
\subsubsection{Multilingual Transformer}
In training, each sentence is labeled with the source and target language label. For example, if we translate from Chuj (``ca'') to Cakchiquel (``ck''),
each source sentence is tagged with 
\texttt{\_\_opt\_src\_ca \_\_opt\_tgt\_ck}. A sample source
sentence is
``\texttt{\_\_opt\_src\_ca \_\_opt\_tgt\_ck} Tec'b'ejec e b'a mach ex tzeyac'och Jehová yipoc e c'ool''.

We train on Geforce RTX 2080 Ti
using $\sim$100 million parameters,
a 6-layer encoder and a 6-layer decoder
that are powered by 512 hidden states,
8 attention heads, 512 word vector size,
a dropout of 0.1, an attention
dropout of 0.1, 2,048 hidden transformer feed-forward units, a
batch size of 6,000, ``adam'' optimizer, ``noam'' decay method,
and a label smoothing of 0.1 and
a learning rate of 2.5 on OpenNMT \cite{klein2017opennmt, vaswani2017attention}.
After 190,000 steps, we validate based on BLEU score
with early stopping patience of 5. 

\subsubsection{Star Versus Complete Configuration}
We show two configurations of translation paths
in Figure~\ref{fig:config}: \textit{star}
graph (multi-source single-target) configuration and
\textit{complete} graph (multi-source multi-target) configuration.
The complete configuration data
increases quadratically with the number of languages 
while the star configuration data increases linearly. 

\subsubsection{Order-preserving Lexiconized transformer}
The variable binding problem issue is difficult in severely
low-resource scenario; most neural models cannot distinguish
the subject and the object of a
simple sentence like ``Fatma asks her sister Wati to
call Yi, the brother of Andika'',
especially when all named entities appear
once or never appear in training
\cite{fodor1988connectionism, graves2014neural}.
Recently, researchers use order-preserving lexiconized Neural
Machine Translation models where named entities are sequentially tagged
in a sentence as \texttt{\_\_NE}s  
\cite{zhou2018massively}. The previous example becomes 
``\texttt{\_\_NE0} asks her sister \texttt{\_\_NE1} to                         
call \texttt{\_\_NE2}, the brother of \texttt{\_\_NE3}''.

This method works under the assumption
of translating a closed text known in advance. 
Its success relies on good coverage of named entities.
To cover many named entities,
we build on existing research literature \cite{wu2018creating, zhou2018massively}
to construct a massively parallel 
lexicon	table that covers 2,939 named entities across 124
languages in our Bible database. Our lexicon table
is an expansion of the existing literature that covers 1,129 named
entities \cite{wu2018creating}. We add in 
1,810 named entities that are in the extreme end of the tail
occurring only once. We also include 66 more real-life severely low-resource
languages. 

For every sentence pair,
we build a target named entity decoding dictionary
by using all target
lexicons from the lexicon table
that match with those in the
source sentence. In severely low-resource setting,
our sequence tagging is larged based on
dictionary look-up; we also include lexicons
that are not in the dictionary but have small
edit distances with the source lexicons.
In evaluation, we
replace all the ordered \texttt{\_\_NE}s
using the target decoding dictionary
to obtain our final translation.

Let us translate ``Fatma asks her sister Wati to call Yi, the brother of Andika''
to Chinese and German. 
Our tagged source sentence that translates to Chinese is
``\texttt{\_\_opt\_src\_en} \texttt{\_\_opt\_tgt\_zh} \texttt{\_\_NE0}
asks her sister \texttt{\_\_NE1} to
call \texttt{\_\_NE2}, the brother of \texttt{\_\_NE3}'';
and we use \texttt{\_\_opt\_tgt\_de} for German.
The source dictionary is ``\texttt{\_\_NE0}: Fatma,
\texttt{\_\_NE1}: Wati, \texttt{\_\_NE2}: Yi,
\texttt{\_\_NE3}: Andika'' and we create the target dictionaries.
The Chinese output is ``\texttt{\_\_NE0}\begin{CJK*}{UTF8}{gbsn}叫她的姐妹\end{CJK*}\texttt{\_\_NE1}\begin{CJK*}{UTF8}{gbsn}去打电话给\end{CJK*}\texttt{\_\_NE3}\begin{CJK*}{UTF8}{gbsn}的兄弟\end{CJK*}\texttt{\_\_NE2}''
      and the German output is ``\texttt{\_\_NE0} bittet ihre Schwester \texttt{\_\_NE1} darum, \texttt{\_\_NE2}, den Bruder \texttt{\_\_NE3}, anzurufen''.
      We decode the named entities to get final translations. 

\begin{table*}[t]
    \footnotesize 
    \centering
      \begin{tabularx}{\columnwidth}{XXX}
      \toprule
      Source Sentence & IPML Translation & Reference \\
      \midrule
      En terwyl Hy langs die see van Galiléa loop, sien Hy Simon en Andréas, sy broer, besig om 'n net in die see uit te gooi; want hulle was vissers.
      &
      And as He drew near to the lake of Galilee, He Simon saw Andrew, and his brother, lying in the lake, for they were fishermen.
      &
      And walking along beside the Sea of Galilee, He saw Simon and his brother Andrew casting a small net in the sea; for they were fishers.
      \\
      \midrule
      En toe Hy daarvandaan 'n bietjie verder gaan, sien Hy Jakobus, die seun van Sebedéüs, en Johannes, sy broer, wat besig was om die nette in die skuit heel te maak.
      &
      And being in a distance, He saw James, the son of Zebedee, and John, his brother. who kept the nets in the boat.
      &
      And going forward from there a little, He saw James the son of Zebedee, and his brother John. And they were in the boat mending the nets.
      \\
      \midrule
      En verder Jakobus, die seun van Sebedéüs, en Johannes, die broer van Jakobus- aan hulle het Hy die bynaam Boanérges gegee, dit is, seuns van die donder-
      &
      And James the son of Zebedee, and John the brother of James; and He gave to them the name, which is called Boanerges, being of the voice.
      &
      And on James the son of Zebedee, and John the brother of James, He put on them the names Boanerges, which is, Sons of Thunder.
      \\
      \bottomrule
    \end{tabularx}
      \caption{Examples of Iteratively Pretrained Multilingual
        Order-preserving Lexiconized Transformer (IPML) translation
        from Afrikaans to English using \textit{FAMP}.
        We train on only 1,093 lines of English data. }
    \label{table:englishQualitative}
  \end{table*}

\subsection{Ranking Source Languages} \label{pathway} \label{dataranking} 
Existing works on 
translation from multiple source
languages into a single low-resource language
usually have at most 30 source
languages \cite{gu2018universal, zhou2018massively, zhu2020language}.
They are limited within the same or close-by
language families, or those with available data, or those
chosen based on the researchers' intuitive discretion. 
However, the set of related languages chosen as source languages under such scenarios could be incomplete. 
Instead, we examine ways to pick useful
source languages in a principled fashion
motivated by cross-lingual impacts
and similarities
\cite{shoemark2016towards, sapir1921languages, odlin1989language, cenoz2001effect, toral2018level, de1997pursuit, hermans2003cross, specia2016shared}.
We find that using many languages that are distant to the target
low-resource language may produce marginal improvements,
if not negative impact.
Indeed, existing literature on zero-shot translation
also suffers from the limitation of linguistic distance between the
source languages and the target language
\cite{lauscher2020zero, lin2020pre, pfeiffer2020mad}.
We therefore rank and select the top few source
languages that are closer to the target low-resource
language using the two metrics below.

We rank source languages according to their closeness to the
low-resource language. We construct
the Family of Choice (FAMC) by comparing different ways of
ranking linguistic distances
empirically based on the small low-resource data. 

To rank linguistic distances empirically based on the small low-resource data, we propose two different ways to construct FAMCs: one by distortion, the other by fertility. Distortion represents the amount of rearrangement work that needs to be done if we were to have a word-by-word translation of the source sentence. Fertility of a source word represents the number of target words produced by that source word. Ideally, if two languages are very close, we expect to see a one-on-one mapping between source and target words. Moreover, we also expect to do minimal rearrangement work if two languages are close. Therefore, at a higher level of understanding, we aim to rank languages by how high its probability of distortion equalling zero and fertility equalling one is. The higher the probability a source language has, the closer this source language is to the target language. With this understanding, let us explain the mathematical formulation of our distortion-based and fertility-based similarity measures in the following. 

Let $S_{s}$ and $S_{t}$ be the source and target
sentences, let $L_{s}$ be the source length,
let $P(S_{t} = s_{t} | s_{s}, l_{s})$ be the alignment
probability, let $F_{s}$ be the fertility of how many target
words a source word is aligned to, let $D_{t}$ be the distortion based on the fixed distance-based reordering model \cite{koehn2009statistical}. 

We first construct a word-replacement model based on aligning
the small amount of target
low-resource data with that of each source language
using \texttt{fast\_align} \cite{dyer2013simple}.
We replace every source word with the most probable
target word according to the product of the alignment 
probability and the probability of fertility equalling one and distortion equalling zero  
$P(F_{s} = 1, D_{t} = 0 | s_{t}, s_{s}, l_{s})$.
We choose a simple word-replacement model because 
we aim to work with around 1,000 lines of low-resource
data. For fast and efficient ranking on such small data, a word-replacement
model suits our purpose. 

We use two alternatives to create our FAMCs.
Our distortion measure is the probability of distortion equalling zero, 
$P(D_{t} = 0 | s_{t}, s_{s}, l_{s})$, 
aggregated over all words in a source language.
We use the distortion measure to rank
the source languages and obtain the distortion-based FAMC (\textit{FAMD});
we
use the translation BLEU scores of the
word-replacement model as another alternative
to build the performance-based FAMC (\textit{FAMP}).  
In Table~\ref{table:ranking}, we list the top ten
languages in FAMD and
FAMP for Eastern Pokomchi
and English. We use both alternatives to build FAMCs. 

To prepare for transformer training, we choose
the top ten languages neighboring our target low-resource
language in FAMD and FAMP.
We choose ten because
existing literature shows that
training with ten languages from two neighboring language
families is sufficient in producing
quality translation through cross-lingual transfer
\cite{zhou2018massively}. Since for
some low-resource languages, there may not be
ten languages in FAMO in our database, we add
languages from neighboring families to make
an expanded list denoted by \textit{FAMO$^+$}. 

\subsection{Iterative Pretraining}
Having ranked similar languages to the low-resource language, 
we choose a suitable set of source languages for training to translate 
into the low-resource language. For training purposes, 
we propose two stages of pretraining using multilingual 
order-preserving lexiconized transformers
on the complete and the star configuration.
We design iterative pretraining on symmetric data to
address catastrophic
forgetting that is common in training \cite{french1999catastrophic, kirkpatrick2017overcoming}.
The word ``iterative'' refers to the multi-stage pretraining that 
is focused on the complete configuration on all source languages first followed by adding the target low-resource language. 
Each stage of pretraining has 
symmetric data for optimization. We propose iterative pretraining 
to promote data symmetry, address 
catastrophic forgetting, and increase robustness. 

\begin{table*}[t]
    \footnotesize
    \centering
    \begin{tabular}{p{4.5cm}p{4.6cm}p{5cm}}
      \toprule
      Source Sentence & IPML Translation & Reference \\ \midrule
      Ket idi limmabas iti dinna ti baybay ti Galilea, nakitana ni Simon ken ni Andres a cabsatna, nga iwaywayatda ti iket iti baybay; ta dumadaclisda idi.
      &
      Eh noq ojik i rub{\textquotesingle}an i Jesús juntar i k{\textquotesingle}isa palaw i Galilea, xrilow reje i Simón ruch{\textquotesingle}ihil i Andres, re{\textquotesingle} i rutuut i k{\textquotesingle}isa palaw, ruum jinaj i k{\textquotesingle}isa palaw barco.
      &
      Noq k{\textquotesingle}ahchi{\textquotesingle} rik{\textquotesingle}iik i Jesús chi chii{\textquotesingle} i k{\textquotesingle}isa palaw ar Galilea, xrilow reje wach i Simón ruch{\textquotesingle}ihil i ruchaaq{\textquotesingle}, Andres rub{\textquotesingle}ihnaal. Re{\textquotesingle} keh aj karineel taqe, k{\textquotesingle}ahchi{\textquotesingle} kikutum qohoq i kiya{\textquotesingle}l pan palaw.
      \\ \midrule
      Ket idi nagna pay bassit nakitana ni Santiago nga anac ni Zebedeo ken ni Juan a cabsatna, nga addada idi iti barangayda, a tartarimaanenda dagiti iketda.
      &
      Eh noq ojik i rub{\textquotesingle}an i Jesús, xrilow i Jacobo, re{\textquotesingle} i Jacobo rak{\textquotesingle}uun i Zebedeo, re{\textquotesingle} Juan rub{\textquotesingle}ihnaal, ruch{\textquotesingle}ihil taqe i raj tahqaneel. eh xkikoj wo{\textquotesingle} wach chinaah i k{\textquotesingle}isa palaw.
      &
      Eh junk{\textquotesingle}aam-oq chik i xb{\textquotesingle}ehik reje i Jesús, xrilow kiwach i ki{\textquotesingle}ib{\textquotesingle} chi winaq kichaaq{\textquotesingle} kiib{\textquotesingle}, re{\textquotesingle} Jacobo, re{\textquotesingle} Juan, rak{\textquotesingle}uun taqe i Zebedeo. Eh wilkeeb{\textquotesingle} chupaam jinaj i barco, k{\textquotesingle}ahchi{\textquotesingle} kik{\textquotesingle}ojem wach i kiya{\textquotesingle}l b{\textquotesingle}amb{\textquotesingle}al kar.
      \\ \midrule
      Ket immasideg ni Jesus ket iniggamanna iti imana ket pinatacderna; ket pinanawan ti gorigor , ket nagservi cadacuada.
      &
      Eh re{\textquotesingle} Jesús xujil i koq riib{\textquotesingle}, xutz{\textquotesingle}a{\textquotesingle}j i koq chinaah i q{\textquotesingle}ab{\textquotesingle}. eh re{\textquotesingle} i kaq tz{\textquotesingle}a{\textquotesingle} chi riij. eh jumehq{\textquotesingle}iil xwuktik johtoq, re{\textquotesingle} chik i reh xutoq{\textquotesingle}aa{\textquotesingle} cho yej-anik kiwa{\textquotesingle}.
      &
      Eh re{\textquotesingle} i Jesús xujil i koq riib{\textquotesingle} ruuk{\textquotesingle} i yowaab{\textquotesingle}, xuchop chi q{\textquotesingle}ab{\textquotesingle}, xruksaj johtoq, eh jumehq{\textquotesingle}iil xik{\textquotesingle}ik i tz{\textquotesingle}a{\textquotesingle} chi riij. Eh re{\textquotesingle} chik i reh xutoq{\textquotesingle}aa{\textquotesingle} cho yej-anik kiwa{\textquotesingle}.
      \\ \bottomrule
    \end{tabular}
    \caption{Examples of Iteratively Pretrained Multilingual
      Order-preserving Lexiconized Transformer (IPML) translation
      from Ilokano to Eastern Pokomchi using \textit{FAMD}. We
      train on only 1,086 lines of Eastern Pokomchi data.}
    \label{table:pokomchiQualitative}
  \end{table*}
  
 \begin{table*}[t]
    \footnotesize
    \centering
    \begin{tabularx}{\columnwidth}{XXX}
      \toprule
      Source Sentence & IPML Translation & Reference \\ \midrule
      Caso detecte efeitos graves ou outros efeitos não mencionados neste folheto, informe o médico veterinário.
      &
      If you notice any side effects or other side effects not mentioned in this leaflet, please inform the vétérinaire.
      &
      If you notice any serious effects or other effects not mentioned in this leaflet, please inform your veterinarian. \\
      \midrule
      No tratamento de Bovinos com mais de 250 Kg de peso vivo, dividir a dose de forma a não administrar mais de 10 ml por local de injecção.
      &
      In the treatment of infants with more than 250 kg in vivo body weight, a the dose to not exceed 10 ml per injection.
      &
      For treatment of cattle over 250 kg body weight, divide the dose so that no more than 10 ml are injected at one site. \\
      \midrule
      No entanto, uma vez que é possível a ocorrência de efeitos secundários, qualquer tratamento que exceda as 1-2 semanas deve ser administrado sob supervisão veterinária regular.
      &
      However, because any of side effects is possible, any treatment that 1-5 weeks should be administered under regular supraveghere.
      &
      However, since side effects might occur, any treatment exceeding 1–2 weeks should be under regular veterinary supervision.
      \\
      \bottomrule
    \end{tabularx}
    \caption{Examples of IPML translation on medical EMEA dataset 
      from Portuguese to English using \textit{FAMO$^+$}. }
    \label{table:EMEAQualitative}
\end{table*} 
  
\subsubsection{Stage 1: Pretraining on Neighbors}\label{step1}
Firstly, we pretrain on the complete
graph configuration of translation paths using the
top ten languages neighboring our target low-resource
language in FAMD, 
FAMP, and FAMO$^+$ respectively. Low-resource data is excluded in training. 

We use the multilingual order-preserving
lexiconized transformer. Our vocabulary is the combination of the
vocabulary for the top ten languages together with the low-resource vocabulary built from the $\sim$1,000 lines.
The final model can translate from any
of the ten languages to each other.

\subsubsection{Stage 2: Adding Low-Resource Data} \label{step2}
We include the low-resource data
in the second stage of training. Since the low-resource
data covers $\sim3.5\%$ of the text while all the source
languages cover the whole text, the data is highly asymmetric.
To create symmetric data, we align the low-resource
data with the subset of data from all source languages.
As a result, all source languages in the second stage of training
have $\sim3.5\%$ of the text that is aligned with the
low-resource data. We therefore
create a complete graph configuration of training
paths using all the eleven languages.

Using the pretrained model from the previous stage,
we train on the complete
graph configuration of translation paths from all eleven
languages including our low-resource
language. The vocabulary used is the same as before.
We employ the multilingual order-preserving lexiconized transformer for pretraining.
The final model can translate from any of the eleven languages to each other.

\subsection{Final Training}\label{step3}
Finally, we focus on translating into the low-resource
language.
We use the symmetric data built 
from the second stage of pretraining.
However, instead of using the complete 
configuration, we use the
star configuration of translation
paths from the all source languages to the
low-resource language. All languages
have $\sim3.5\%$ of the text. 

Using the pretrained model from the second stage,
we employ the multilingual order-preserving
lexiconized transformer on the star
graph configuration. We use the same vocabulary
as before. 
The final trained model can
translate from any
of the ten source languages to the low-resource language.
Using the lexicon dictionaries, we decode the named
entities and obtain our final translations. 

\subsection{Combination of Translations}\label{step4}
We have multiple translations, one per each source language.
Combining all translations is
useful for both potential post-editing works and
systematic comparison of different
experiments especially when the sets of the source languages differ.

Our combination method assumes that we have the same text in
all source languages. For each sentence, we form a cluster of
translations from all source languages into the low-resource
language. Our goal is to find the translation that is closest
to the center of the cluster. We rank all translations according to how
centered this translation is with respect to other sentences by
summing all its similarities to the rest. The highest score is the
closest to the cluster center. We take the most centered
translation for each sentence and output our combined result.
The expected BLEU score of our combined translation
is higher than translation from	any of the individual
source language.

\section{Data}\label{data}
We use the Bible dataset and the medical
EMEA dataset \cite{mayer2014creating, tiedemann2012parallel}.
EMEA dataset is from the European Medicines Agency and
contains a lot of medical information that may be beneficial to the
low-resource communities.  
Our method can be
applied to other datasets like WASH guidelines.

For the Bible dataset,
we use 124 source languages with 31,103 lines
of data and
a target low-resource language with $\sim$1,000 lines ($\sim$3.5\%) of data.
We have two setups for the
target low-resource language.
One uses Eastern Pokomchi,
a Mayan language; 
the other uses English
as a hypothetical low-resource language \footnote{In Table~\ref{table:ranking} and Table~\ref{table:pokomchi}, Kanjobal is
Eastern Kanjobal, Mam is Northern Mam, Cuzco is Cuzco Quechua, Ayacucho is Ayacucho Quechua,
Bolivian is South Bolivian Quechua, and Huallaga is Huallaga Quechua.}.
We train
on only $\sim$1,000 lines of low-resource data
from the book of Luke and test on the 678 lines
from the book of Mark. Mark is
topically similar to Luke, but is written by a different author. 
For the first stage of pretraining, we use
80\%, 10\%, 10\% split for training, validation
and testing. For the second stage onwards, we
use 95\%, 5\% split of Luke for training and validation,
and 100\% of Mark for testing. 

Eastern Pokomchi
is Mayan, and English is Germanic. Since our database does not
have ten members of each family,
we use FAMO$^+$, the expanded version of FAMO. 
For English, we include five Germanic
languages and five Romance languages in FAMO$^+$;
for Eastern Pokomchi, we include five Mayan languages
and five Amerindian languages in FAMO$^+$. 
The Amerindian family is broadly
believed to be close to the Mayan family
by the linguistic community. 

We construct
FAMCs by comparing different ways of
ranking linguistic distances
empirically based on $\sim$1,000 lines of training data.
In Table~\ref{table:ranking}, we list the top ten
languages for Eastern Pokomchi
and English in FAMD and
FAMP respectively.

To imitate the real-life situation of having small
seed target translation data, 
we choose to use $\sim$1,000 lines ($\sim$3.5\%) of
low-resource data. We also include Eastern Pokomchi
in addition to using English as a hypothetical low-resource language.
Though data size
can be constrained to mimic severely low-resource
scenarios, much implicit information is still used for
the hypothetical low-resource language
that is actually rich-resource.
For example, implicit information like
English is Germanic is often used.
For real low-resource scenarios, the family information may have yet to be determined; the neighboring languages may be unknown or incomplete, and if they are known, they are highly likely to be low-resource too. We thus use Eastern Pokomchi as our real-life severely low resource language.

To understand Eastern Pokomchi, it is a Mayan language that is morphological rich, ergative and agglutinative \cite{aissen2017mayan, clemens2015ergativity}. It has been isolated from other languages historically and is seen by many as non-transparent, complex and obscure \cite{england2011grammar}. 

In addition to the Bible dataset, we work with the
medical EMEA dataset \cite{tiedemann2012parallel}. Using
English as a hypothetical language, we train on randomly sampled 
1,093 lines, and test on 678 lines of data. 
Since there are only 9 languages in Germanic and
Romance families in EMEA dataset, we include
a slavic language Polish in our FAMO$^+$ for experiments.

The EMEA dataset is less than ideal comparing with the Bible dataset.
The Bible dataset contains the same text for all source languages; however, the EMEA dataset does not
contain the same text. It is
built from similar documents but
has different parallel data for
each language pair.
Therefore, during test time, we do not
combine the translations from various source languages
in the EMEA dataset.

 \begin{table}[t]
  \centering
  \small
  \begin{tabularx}{\textwidth}{X|XXXXXX}
    \toprule
    Experiments & IPML & MLc & MLs & PMLc & PMLs & AML\\
    \midrule
    Pretrained & \checkmark & &  & \checkmark & \checkmark &  \\
    Iterative & \checkmark & & & & & \\
    Lexiconized & \checkmark & \checkmark & \checkmark & \checkmark & \checkmark & \checkmark \\
    Symmetrical & \checkmark & \checkmark & \checkmark & \checkmark & \checkmark \\
    Star & \checkmark & & \checkmark & & \checkmark & \\
    Complete & \checkmark & \checkmark & & \checkmark & & \checkmark \\
    \midrule
    Combined & 37.3 & 13.4 & 14.7 & 34.7 & 35.7 & 27.0 \\
    \midrule 
    German & 35.0 & 11.6 & 12.3 & 33.3 & 34.5 & 25.4 \\
    Danish & 36.0 & 12.5 & 12.4 & 33.3 & 34.2 & 26.2 \\
    Dutch & 35.6 & 11.5 & 11.1 & 32.3 & 33.7 & 25.0 \\
    Norwegian & 35.7 & 12.3 & 12.0 & 33.2 & 34.1 & 25.8 \\
    Swedish & 34.5 & 11.8 & 12.4 & 32.3 & 33.4 & 24.9 \\
    Spanish & 36.4 & 11.7 & 11.8 & 34.1 & 35.0 & 26.2 \\
    French & 35.3 & 10.8 & 10.8 & 33.1 & 34.0 & 25.8 \\
    Italian & 35.9 & 11.7 & 11.7 & 34.3 & 34.5 & 26.1 \\
    Portuguese & 31.5 & 9.6 & 10.1 & 30.0 & 30.4 & 23.1 \\
    Romanian & 34.6 & 11.3 & 12.1 & 32.3 & 33.2 & 25.0 \\
    \bottomrule
  \end{tabularx}
  \caption{
    Comparing our iteratively pretrained
    multilingual order-preserving lexiconized transformer (IPML)
    with the baselines training on 1,093 lines of English data
    in \textit{FAMO$^+$}. We checkmark the key components used
    in each experiments and explain all the baselines in
    details in Section \ref{results}.
  }
    \label{table:englishCompare}
\end{table}

\section{Results}\label{results}

\begin{figure*}[t]
  \centering
  \includegraphics[width=0.7\linewidth]{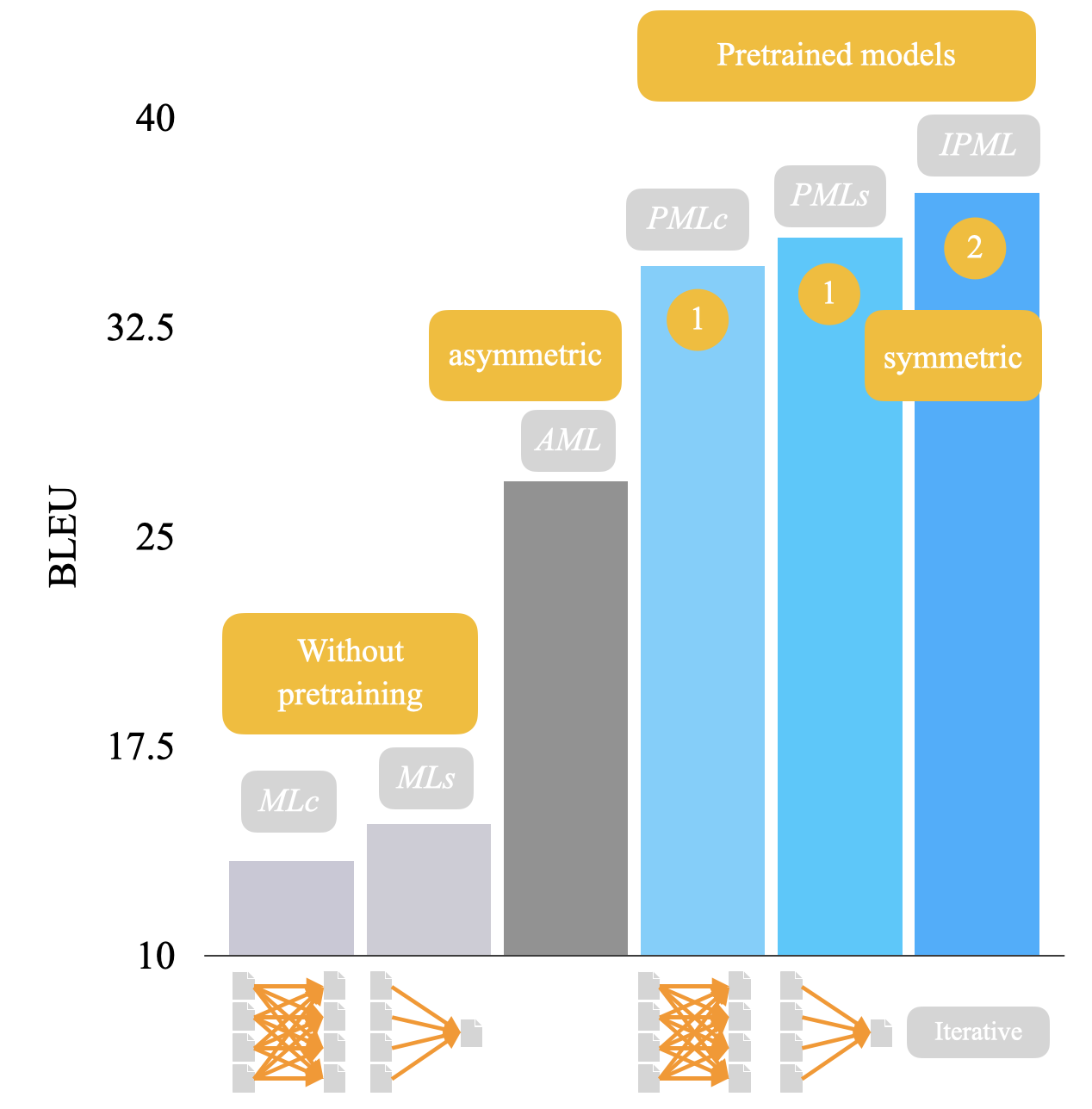}
    \caption{Comparing our method with different baselines for translation into English as a hypothetical low-resource language using $\sim$1,000 lines of data. }
    \label{fig:ipml1}  
\end{figure*}

We summarize our results in Figure~\ref{fig:ipml1} and we show our result is generalizable to medical domains by looking at EMEA datasets which we will focus on at the end of the results section.  

In Figure~\ref{fig:ipml1}, we compare our iteratively pretrained multilingual
order-preserving lexiconized transformer (IPML)
with five baselines in Table~\ref{table:englishCompare}. 
\textit{MLc} is a baseline model of
multilingual order-preserving
lexiconized transformer
training on complete configuration;
in other words, we skip the first stage of
pretraining and train on the second stage
in Section \ref{step2} only.
\textit{MLs} is a baseline model
of multilingual order-preserving lexiconized
transformer training on star configuration;
in other words, we skip both steps of
pretraining and train on the final stage in
Section \ref{step3} only.
\textit{PMLc} is a baseline
model of pretrained multilingual
order-preserving lexiconized transformer
training on complete configuration;
in other words, we skip the final stage
of training after completing both stages of
pretraining.
\textit{PMLs} is a baseline model of
pretrained multilingual order-preserving lexiconized
transformer training on star configuration;
in other words, after the first stage of pretraining,
we skip the second stage
of pretraining and proceed to the final training
directly. Finally,
\textit{AML} is a baseline model of
multilingual order-preserving lexiconized transformer
on asymmetric data. We replicate the $\sim$1,000
lines of the low-resource data
till it matches the training size of
other source languages; we train on
the complete graph configuration using 
eleven languages. Though the number of low-resource
training lines is the same as others,
information is highly asymmetric. 

\begin{table}[t]
  \centering
  \small
  \begin{tabularx}{\textwidth}{XXXXXX} 
    \toprule
    \multicolumn{6}{c}{Input Language Family} \\
    \midrule
    \multicolumn{2}{c|}{By Linguistics} & \multicolumn{2}{c|}{By Distortion} & \multicolumn{2}{c}{By Performance} \\
    \midrule
    \multicolumn{2}{c|}{\textit{FAMO$^+$}} & \multicolumn{2}{c|}{\textit{FAMD}} & \multicolumn{2}{c}{\textit{FAMP}} \\
    \midrule
    Source & BLEU & Source & BLEU & Source & BLEU \\
    \midrule
    Combined & 37.3 & Combined & 38.3 & Combined & 39.4 \\
    \midrule
    German & 35.0 & German & 36.7 & German & 37.6 \\
    Danish & 36.0 & Danish & 37.1 & Danish & 37.5 \\
    Dutch & 35.6 & Dutch & 35.6 & Dutch & 36.7 \\
    Norwegian & 35.7 & Norwegian & 36.9 & Norwegian & 37.1 \\
    Swedish & 34.5 & Afrikaans & 38.3 & Afrikaans & 39.3 \\
    Spanish & 36.4 & Marshallese & 34.7 & Spanish & 38.4 \\
    French & 35.3 & French & 36.0 & French & 36.6 \\
    Italian & 35.9 & Italian & 36.9 & Italian & 37.7 \\
    Portuguese & 31.5 & Portuguese & 32.9 & Portuguese & 33.1 \\
    Romanian & 34.6 & Frisian & 36.1 & Frisian & 36.9 \\
    \bottomrule
    \end{tabularx}
  \caption{Performance of Iteratively Pretrained Multilingual Order-preserving Lexiconized Transformer (IPML)
    training for English on \textit{FAMO$^+$}, \textit{FAMD} and \textit{FAMP}. We train on only 1,093 lines of English data.}
    \label{table:english}
\end{table}

Pretraining is key as IPML beats the two baselines that
skip pretraining in Table~\ref{table:englishCompare}.
Using English as a hypothetical low-resource language training on
FAMO$^+$, combined translation
improves from 13.4 (MLc) and 14.7 (MLs) to 37.3 (IPML)
with iterative pretraining. Training with the low-resource
language on both the source and the target sides
boosts translation into the target side.
Star configuration has a slight
advantage over complete configuration as it
gives priority to translation
into the low-resource language.
Iterative pretraining with BLEU score 37.3
has an edge over one stage of
pretraining with scores 34.7 (PMLc) and 35.7 (PMLs).  

All three pretrained models on symmetric data,
IPML, PMLc and PMLs, beat asymmetric baseline AML. 
In Table~\ref{table:englishCompare}, IPML has
a +10.3 BLEU increase over our asymmetric baseline
on combined translation using English
as a hypothetical low-resource language training on FAMO$^+$.
All four use the same amount of data, but differ in
training strategies and data configuration.
In severely low-resource scenarios, effective
training strategies on symmetric data improve translation greatly.

\begin{table}[t] 
  \centering
  \small
  \begin{tabularx}{\textwidth}{XXXXXX}
    \toprule
    \multicolumn{6}{c}{Input Language Family} \\
    \midrule
    \multicolumn{2}{c|}{By Linguistics} & \multicolumn{2}{c|}{By Distortion} & \multicolumn{2}{c}{By Performance} \\
    \midrule
    \multicolumn{2}{c|}{\textit{FAMO$^+$}} & \multicolumn{2}{c|}{\textit{FAMD}} & \multicolumn{2}{c}{\textit{FAMP}} \\
    \midrule
    Source & BLEU & Source & BLEU & Source & BLEU \\
    \midrule
    Combined & 23.0 & Combined & 23.1 & Combined & 22.2 \\
    \midrule
    Chuj & 21.8 & Chuj & 21.9 & Chuj & 21.6 \\ 
    Cakchiquel & 22.2 & Cakchiquel & 22.1 & Cakchiquel & 21.3 \\ 
    Guajajara & 19.7 & Guajajara & 19.1 & Guajajara & 18.8 \\
    Mam & 22.2 & Russian & 22.2 & Mam & 21.7 \\ 
    Kanjobal & 21.9 & Toba & 21.9 & Kanjobal & 21.4 \\ 
    Cuzco & 22.3 & Myanmar & 19.1 & Thai & 21.8 \\
    Ayacucho & 21.6 & Slovenský & 22.1 & Dadibi & 19.8 \\
    Bolivian & 22.2 & Latin & 21.9 & Gumatj & 19.1 \\
    Huallaga & 22.2 & Ilokano & 22.5 & Navajo & 21.3 \\
    Aymara & 21.5 & Norwegian & 22.6 & Kim & 21.5 \\
    \bottomrule
  \end{tabularx}
  \caption{Performance of Iteratively Pretrained Multilingual Order-preserving Lexiconized Transformer (IPML)
    training for Eastern Pokomchi on \textit{FAMO$^+$}, \textit{FAMD} and \textit{FAMP}. We train on only 1,086 lines of Eastern Pokomchi data.}
    \label{table:pokomchi}
\end{table}

We show IPML results training on different
sets of source languages in FAMO$^+$, FAMD, and FAMP,
for English and Eastern Pokomchi in
Table~\ref{table:english} and \ref{table:pokomchi}.
FAMP performs the best for translation into English while both 
FAMP and FAMD outperforms
FAMO$^+$ as shown in Table~\ref{table:english}.
FAMD performs best for translation into Eastern
Pokomchi as shown in Table~\ref{table:pokomchi}. Afrikaans has the
highest score for English's FAMD and FAMP, outperforming Dutch,
German or French. A reason may be that Afrikaans
is the youngest language in the Germanic family
with many lexical and syntactic borrowings from English
and multiple close neighbors of English \cite{gordon2005ethnologue}. 

Note that we are not focusing on comparing FAMO$^+$ with FAMCs, our focus is what we could do to boost translation performance when there is incomplete information of which languages are close to our given low-resource language. 
When language family information is limited or incomplete, 
i.e., when FAMO$^+$ is limited or incomplete, 
constructing FAMC to determine neighbors
is very useful in translation. 

\begin{table}[t]

\parbox{.45\linewidth}{
  \centering
  \small
  \begin{tabularx}{.45\columnwidth}{XX}
    \toprule
    Source & BLEU \\
    \midrule
    Combined & N.A. \\
    \midrule
    German & 34.8 \\
    Danish & 37.7 \\
    Dutch & 39.7 \\
    Swedish & 37.7 \\    
    Spanish & 42.8 \\ 
    French & 41.6 \\
    Italian & 39.2 \\
    Portuguese & 42.8 \\
    Romanian & 40.0 \\    
    Polish & 34.1 \\
    \bottomrule
    \end{tabularx}
  \caption{IPML Performance on the EMEA dataset trained on only 1,093 lines of English data.}
    \label{table:EMEA}
}
\hfill
\parbox{.45\linewidth}{
  \centering
  \small
  \begin{tabularx}{.45\columnwidth}{XX}
    \toprule
    Source & BLEU \\
    \midrule 
    Combined & 23.7 \\
    \midrule
    German & 21.6 \\
    Danish & 22.9 \\
    Dutch & 21.2 \\
    Norwegian & 21.3 \\
    Swedish & 19.9 \\
    Spanish & 22.9 \\
    French & 22.3 \\
    Italian & 21.8 \\
    Portuguese & 20.7 \\
    Romanian & 16.3 \\
    \bottomrule
    \end{tabularx}
  \caption{IPML Performance on the entire Bible excluding $\sim$1k lines of training and validation data.}
  \label{table:bible}
}
\end{table}

Comparing Eastern Pokomchi results
with English results, we see that translation into real-life
severely low-resource languages is more
difficult than translation into hypothetical ones.
The combined score is 38.3
for English in Table~\ref{table:english}
and 23.1 for Eastern Pokomchi on FAMD
in Table~\ref{table:pokomchi}. Eastern Pokomchi
has ejective consonants which makes
tokenization process difficult.
It is agglutinative, morphologically
rich and ergative just like Basque \cite{aissen2017mayan, clemens2015ergativity}. It is complex, unique and nontransparent to the
outsider \cite{england2011grammar}.
Indeed, translation into real severely
low-resource languages is difficult. 

We are curious of how our model trained
on $\sim$1,000 lines of data performs on the rest of the Bible.
In other words, we would like to know how IPML performs
if we train on $\sim$3.5\% of the
Bible and test on $\sim$96.5\% of the Bible. In
Table~\ref{table:bible}, training on 1,093 lines from
the book of Luke,
we achieve a BLEU score of 23.7 for IPML
using FAMP in English \footnote{A previous version of this work
shows higher BLEU scores with random sampling. Since
active learning is not the focus of this section, 
we show all results training on the book of Luke in this paper.
For further results in active learning, please refer to our 
follow-up work in the next section \cite{zhou2021active}.}. 

We show qualitative examples 
in Table~\ref{table:englishQualitative}
and \ref{table:pokomchiQualitative}.
The source content is translated well
overall and there are a few places for improvement
in Table~\ref{table:englishQualitative}.
The words ``fishermen'' and ``fishers'' are paraphrases of
the same concept. IPML predicts
the correct concept though it is penalized by BLEU.

Infusing the order-preserving lexiconized component
to our training greatly improves qualitative evaluation. But it 
does not affect BLEU much as BLEU has its limitations
in severely low-resource scenarios. This is why all
experiments include the lexiconized component in training. 
The BLEU comparison in our paper also
applies to the comparison of all experiments without the
order-preserving lexiconized component. This is important
in real-life situations when a low-resource lexicon list
is not available, or has to be invented. For example, a
person growing up in a local village in Papua New Guinea
may have met many people named ``Bosai'' or ``Kaura'', but 
may have never met a person named ``Matthew'', 
and we may need to create a lexicon
word in the low-resource language for ``Matthew''
possibly through phonetics.

We also see good results with the medical
EMEA dataset. Treating English as a hypothetical
low-resource language, we train on only 1,093 lines
of English data. For Portuguese-English translation,
we obtain a BLEU score of 42.8 while the rest of
languages all obtain BLEU scores above 34 in
Table~\ref{table:EMEA} and Table~\ref{table:EMEAQualitative}.
In Table~\ref{table:EMEAQualitative}, we see that our
translation is very good, though a few words are
carried from the source language including ``vétérinaire''.
This is mainly because our $\sim$1,000 lines contain
very small vocabulary; however, by carrying the
source word over, key information is preserved. 

\section{Conclusion}\label{conclusion}
We use $\sim$1,000 lines of low-resource data
to translate a closed text that is known in advance to
a severely low-resource language by leveraging massive
source parallelism.
We present two metrics to rank the
124 source languages and construct FAMCs.  
We build an
iteratively pretrained multilingual
order-preserving lexiconized transformer and
combine translations from all source languages
into one by using our centric measure.
Moreover, we add a multilingual order-preserving lexiconized 
component to translate the named entities
accurately. We build a massively parallel lexicon table for
2,939 Bible named entities in 124 source languages,
covering more than 66 severely low-resource languages.
Our good result for the medical EMEA dataset shows that
our method is useful for other datasets and applications. 

Our final result can also serve as a ranking
measure for linguistic distances
though it is much more expensive
in terms of time and resources.
In the future, we would like to explore more metrics that are
fast and efficient in ranking linguistic distances to
the severely low-resource language.

%

Having examined ways to build linguistic measure and build metric spaces of language closeness that are suitable for multilingual training, when language 
family information is not complete, we have established a proven way to 
train multilingual models for the translation task of a multi-source text 
into a new, low-resource language. We have shown success translating the whole text 
using about 1,000 lines of training data as our seed corpus. 
Our next question is, what is the optimal 
1,000 lines to produce as seed corpus to optimize machine translation performance? 
In other words, is there a way to use active learning to determine the most effective way 
to build seed corpus to optimize for machine translation? This is what we would like to address in the next chapter.

\removelabelprefix

\part{Human Machine Translation}\label{part:part2} 
Having examined source parallelism in the first part of the thesis (Chapter \ref{big:family}, Chapter \ref{big:paraphrase} and Chapter \ref{big:ipml}), we build a human machine translation workflow algorithm for machine translation systems to collaborate with human translators to expedite the translation process of a multi-source text into new, low-resource languages (Chapter \ref{big:active}, Chapter \ref{big:large} and Chapter \ref{big:confidence}). Our proposed human machine translation is not to
replace the human translators with machine translation systems, but instead,
to get the best of both worlds and to expedite the translation process. In our translation process, human translators are informed by machine sentence ranking through active learning to produce a seed corpus. Machine systems then use this seed corpus to produce a full translation draft. Human translators post-edit the draft, and feed new data to machines each time they finish post-editing a portion of the text. In each iteration, machines produce better and better drafts with new data, and human translators find it easier and faster to post-edit. Together they complete the translation of the whole text into an severely low-resource language. 

In Chapter \ref{big:active}, we first develop various active learning methods on known languages and transfer ranking to the new, low-resource language. In Chapter \ref{big:large}, we activate the knowledge of large multilingual models by proposing multilingual and multi-stage adaptations through different training schedules; we find that adapting pretrained models to the domain and then to the low-resource language works best. Thirdly, we aggregate scores from 115 languages to provide a universal ranking and increase robustness by the \textit{relaxed memoization} method. Having examined both source parallelism and human machine translation workflow, we evaluate our work in all previous chapters by translating academic progress to the real-world translation process in a case study in Quechuan language family in Chapter \ref{big:confidence}. We collaborate extensively with a translation group with in-depth knowledge of various Quechuan languages and focus on evaluation. We find that machine translation performance is significantly positively correlated with language similarity. The more connected a language is, the easier it is to translate into it. In addition, we find that decluttering poorly-connected languages improves translation score. Using this finding, we achieve good results in translating into a new, low-resource language called Sihuas Quechua.   
\chapter{Active Learning for Building a Seed Corpus}\label{big:active}
\addlabelprefix{5}
\epigraph{``Tell me and I forget. \\Teach me and I remember. \\Involve me and I learn.''}{\textit{Benjamin Franklin}}

\lettrine{H}{aving examined interlingual transfer} both within and across different language families, paraphrases within the same language, and the thorough way of building linguistic measure and building metric spaces of language closeness, we have established a proven way to use a small seed corpus (around 1,000 lines of low-resource language data) to complete our translation task of translating a multi-source text into a new, low-resource language. In this chapter, we examine the optimal way to build seed corpus to best help with machine translation. We incorporate active learning to learn sentence ranking in the low-resource language and inform human translators which set of sentences to translate first, in order to produce the most effective and useful seed corpus. 

\begin{figure}
  \centering
  \includegraphics[width=0.8\linewidth]{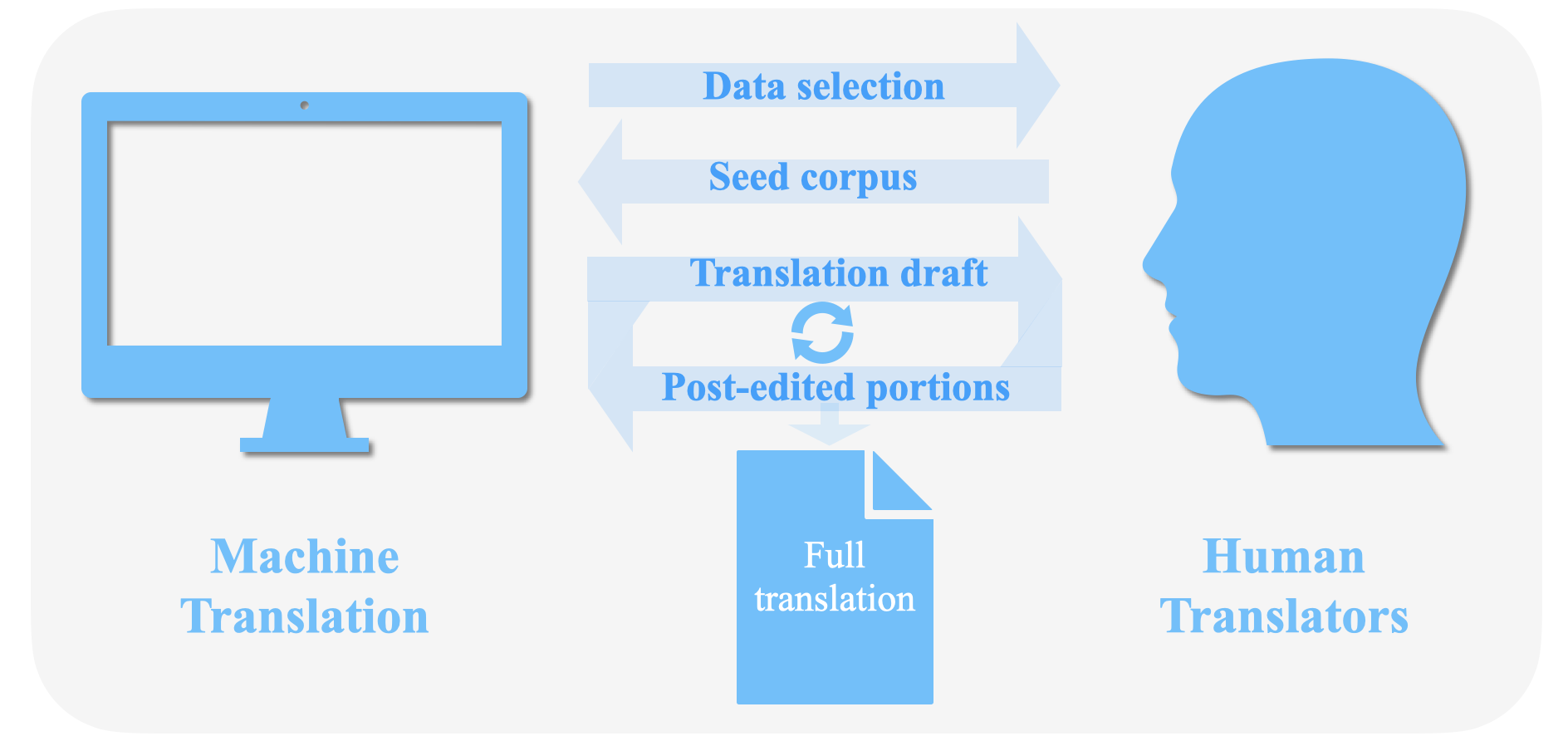}
  \caption{Translation workflow for severely low-resource languages.}
  \label{fig:workflow}
\end{figure}

\begin{figure*}[t]
  \centering
  \includegraphics[width=.9\linewidth]{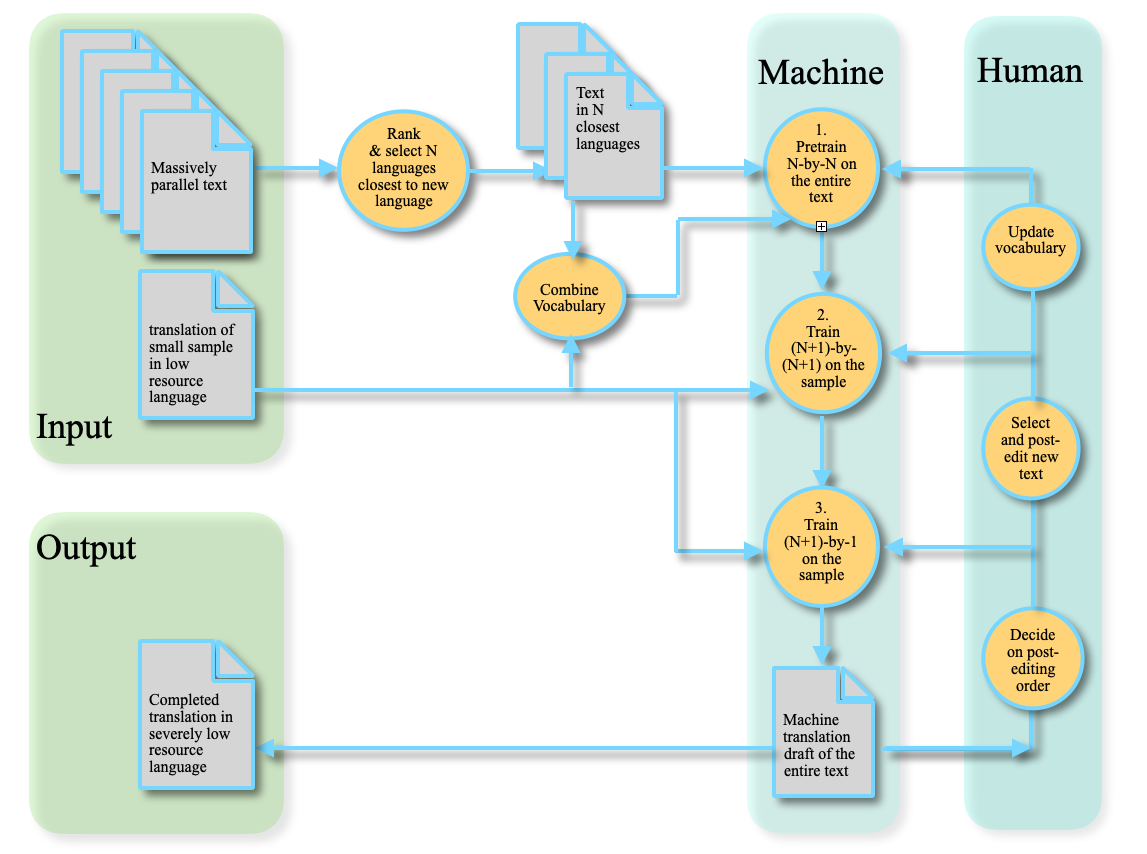}
  \caption{Proposed joint human machine translation sequence for a given closed text. }
  \label{fig:hmt_algo}
\end{figure*}

\section{Introduction} \label{introduction}
Machine translation has flourished ever since
the first computer was made
\citep{hirschberg2015advances, popel2020transforming}.
Over the years, human translation is assisted by machine
translation to 
remove human bias and translation 
capacity limitations
\citep{koehn2009interactive, li2014comparison, savoldi2021gender, bowker2002computer, bowker2010computer, koehn2009process}.
By learning human translation taxonomy
and post-editing styles,
machine translation borrows many ideas
from human translation to improve performance
through active learning 
\citep{settles2012active, carl2011taxonomy, denkowski2015machine}.
As discussed in Chapter 1, we propose a workflow\footnote{The material in this chapter was originally 
published in LoresMT at MT Summit, 2021 \citep{zhou2021active} and LoResMT at ACL, 2023 \citep{zhou2023train}.} to bring human
translation and machine translation to work together
seamlessly in translation of a closed text
into a severely low-resource language as shown in
Figure~\ref{fig:workflow} and Figure~\ref{fig:hmt_algo} and Algorithm~\ref{algo:proposedtrans}. With our human machine translation framework, we are interested in translating a given text that has many existing translations in different languages into a severely low-resource language well. 

In this translation framework, human translators are informed by machine data selection process through active learning to produce a seed corpus. Machine systems then use this seed corpus to produce a full translation draft. Human translators post-edit the draft, and feed new data to machines each time the newly post-edited text iteratively. In each iteration, machines produce better and better drafts with new data, and human translators find it easier and faster to post-edit. Together they complete the translation of the whole text into a severely low-resource language.

As discussed in Chapter 1, our overall goal in this thesis is to minimize human translation and post-editing efforts required to generate a full publishable-standard translation of the given text. Ideally, we want to hire a large number of human translators to measure and compare the resources (time and money) used to translate the same text into a target low-resource language that does not have any translations of the text with and without our help. However, this ideal solution is unrealistic especially in large translation projects. This is why we transform our goal of minimizing human translation efforts required to generate a full translation of the given text into 
two practical proxy sub-goals as the following: 
\begin{enumerate}
\item 
Optimizing and minimizing the amount of sentences to be used to construct seed corpus.
\item 
Maximizing the quality and utility of MT-generated translation of the full text and optimizing translation efficiency.
\end{enumerate} 

The first sub-goal of minimizing the seed corpus serves as a proxy in this chapter to minimize the human translation efforts in the creation of the seed corpus, while the second sub-goal of maximizing translation performance serves as a proxy in the next chapter to minimize human translation efforts in the post-editing process during the subsequent iterations. 

With our work in the first part of the thesis, we minimize our seed corpus size to $\sim$3\% of the text. In other words, we use $\sim$3\% of the text to translate $\sim$97\% of the text \citep{zhou2021family, zhou2021active, zhou2018massively}. In this chapter, we are interested in finding out which $\sim$3\% of the text is to be translated first to build a seed corpus to best improve the translation performance of the first translation draft of the machine translation system. This will help to minimize the post-editing efforts through the iterations afterwards. 

\begin{algorithm*}[t]
\small
 \KwIn{A text of $N$ lines consisting multiple books/portions, parallel in $L$ source languages}  
 \KwOut{A full translation in the target low-resource language, $l'$}
  0. Initialize translation size, $n = 0$, vocabulary size, $v = 0$, vocabulary update size, $\triangle v = 0$ \;
  1. Using Active Learning to choose $S$ ($\sim$1,000) sentences with vocabulary size $v_S$ for human translators to produce the seed corpus, update $n = S$, $v = v_S$ \;
  2. Rank and pick a family of close-by languages by linguistic, distortion or performance metric \;
  \While{$n < N$} {
    \If{$\triangle v > 0 $ } {
  3. Pretrain on the full texts of neighboring languages \;
  }
  4. Train on the $n$ sentences of all languages in multi-source multi-target configuration \;
  5. Train on the $n$ sentences of all languages in multi-source single-target configuration \;
  6. Combine translations from all source languages using the centeredness measure \; 
  7. Review all books/portions of the translation draft \;
  8. Pick a book/portion with $n'$ lines and $v'$ more vocabulary \;
  9. Complete human post-editing of the portion chosen, $v = v + v'$, $n = n + n'$, $\triangle v = v'$ \;
  }
 return full translation co-produced by human (Step 1, 7-9) and machine (Step 0, 2-6) translation \;
  \caption{Proposed joint human machine translation sequence for a given closed text.}
  \label{algo:proposedtrans}
\end{algorithm*}

To find which $\sim$3\% of the text is to be translated first to optimize performance, we examine how this process of construction methods of such seed corpora is completed in past. 
Historically, this is 
mostly determined by field linguists'
experiential and intuitive discretion. 
Many human translators
employ a portion-based strategy when translating large texts.
For example, translation of the book ``The Little Prince''
may be divided into smaller tasks
of translating 27 chapters, or
even smaller translation units like a few consecutive pages.
Each translation unit contains
consecutive sentences.
Consequently, machine translation often uses
seed corpora that are chosen based on human
translators' preferences, but may not be optimal
for machine translation. For optimal machine translation, 
researchers have yet to examine various 
Active Learning (AL) methods to improve accuracy
and effectiveness in building better optimized 
seed corpora so as to minimize the initial human
effort. 

To solve this problem, we propose explainable and 
robust active learning methods that
perform as well as or better than random sampling; 
we transfer methods learned on data of known
languages to the new, severely low-resource language. Our contribution is two-fold: 1. in addition 
to random sampling, we develop 14 active learning methods on known 
languages and transfer ranking to the new, severely low-resource 
language; 2. we also 
aggregate scores from 115 languages to provide a universal 
ranking and increase robustness by \textit{relaxed memoization} method. 

In this work, we aim to answer this question: when field linguists have
limited time and resources, which lines would be
given priority? Given a closed text, we propose
that it would beneficial if field linguists translate
$\sim$1,000 lines chosen based on active learning ranking first, getting the first
machine translated draft of the whole text, and then
post-edit to obtain final translation of each portion
iteratively as shown in Algorithm \ref{algo:proposedtrans}.
We recognize that the portion-based translation is very helpful in
producing quality translation with
formality, cohesion and contextual relevance.
Thus, our proposed way is not to
replace the portion-based approach, but instead,
to get the best of both worlds and
to expedite the translation process as shown in
Figure \ref{fig:hmt_algo}. 

\subsection{Translation Workflow}
In our translation workflow, human translators are informed by machine sentence ranking through active learning to produce a seed corpus. Machine systems then use this seed corpus to produce a full translation draft. Human translators post-edit the draft, and feed new data to machines each time they finish post-editing a portion of the text. In each iteration, machines produce better and better drafts with new data, and human translators find it easier and faster to post-edit. Together they complete the translation of the whole text into an severely low-resource 
language (Figure~\ref{fig:workflow}).  

\begin{figure}
  \centering
  \includegraphics[width=0.6\linewidth]{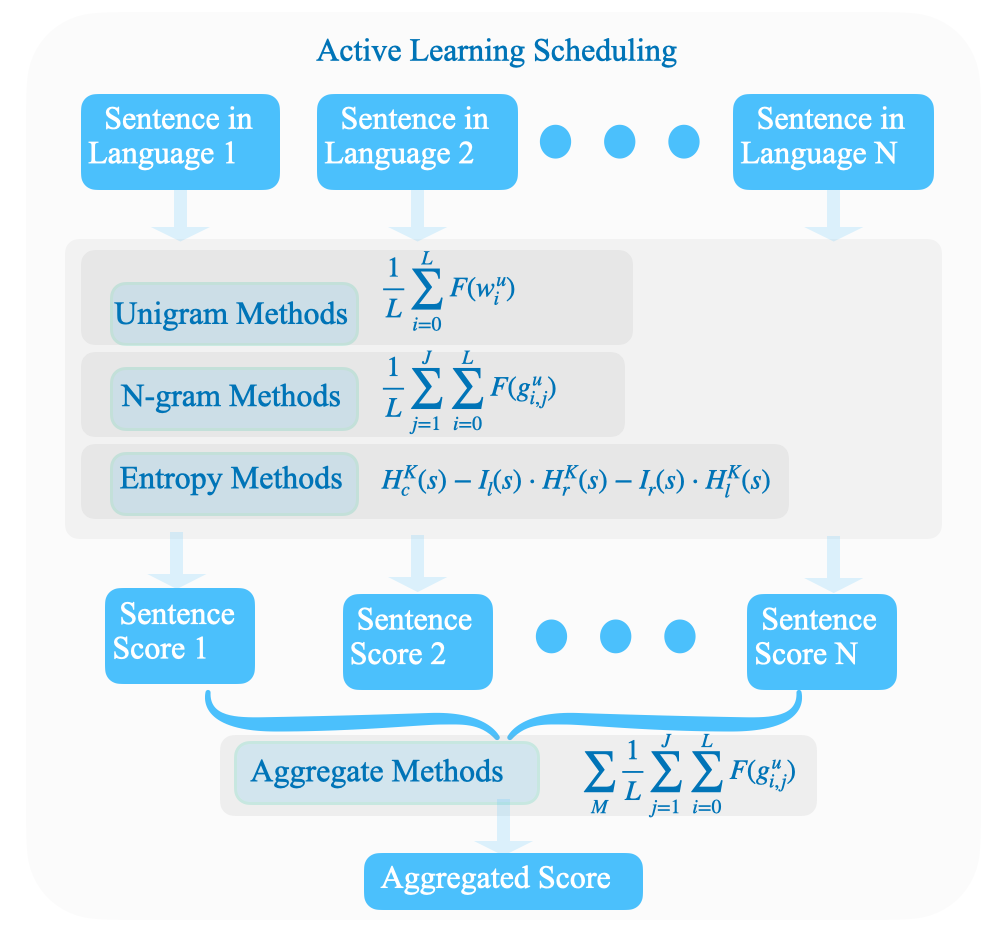}
  \caption{Visualizing different active learning methods. We score and rank each sentence in a text corpus. }
  \label{fig:active}
\end{figure}

To produce sentence ranking, traditional active learning approaches assume abundant data, but we have little to no data in the target severely low-resource language. We question this assumption and build seed corpora by ranking 
all sentences in existing translations from other languages to generalize to a new, severely low-resource language. This ranking is target-independent as we do not require any severely low-resource language data. To produce such a ranking, we explore active learning methods including random sampling, unigram, n-gram, entropy and aggregation methods (Table~\ref{table:ngramScore}). For each reference language, we build unigram, n-gram and entropy models (Figure~\ref{fig:active}). To prevent any language from overpowering the ranking, we aggregate sentence scores across multiple languages and rank the final aggregation. To select the pool of languages for aggregation, we build methods on different voting mechanisms. 

To curate a seed corpus in the new, severely low-resource language where we have no data initially, we pass the sentence ranking learned from known languages to human translators. Human translators take this ranking, and translate the top few ($\sim$1,000 or less) sentences, curating the seed corpus. This seed corpus is then used for training MT systems. 

Once the MT systems produces the translation draft, human translators post-edit the draft. Once a stage of post-editing is finished, it is being fed back to the MT systems for its training to produce better drafts. Over many iterations of MT drafting and human post-editing, the final translation of the entire text into the severely low-resource language is completed. 

\subsection{Different Active Learning Approaches}

\begin{table*}[t]
  \centering
  \small
  \resizebox{\textwidth}{!}{
    \begin{tabularx}{\textwidth}{p{4.5cm}p{4.4cm}p{1.0cm}p{1.2cm}p{1.0cm}p{1.2cm}}
    \toprule
    Book & Author & Books & Chapters & Pages & Languages\\    
    \midrule
    The Bible & Multiple & 66 & 1,189 & 1,281 & 689 \\
    The Little Prince & Antoine de Saint Exupéry & 1 & 27 & 96 & 382 \\
    Dao De Jing & Laozi & 1 & 81 & $\sim$10 & $>$250 \\
    COVID-19 Wiki Page & Multiple & 1 & 1 & $\sim$50 & 155 \\
    The Alchemist & Paulo Coelho & 1 & 2 & 163 & 70\\
    Harry Potter & J. K. Rowling & 7 & 199 & 3,407 & 60\\
    The Lord of the Rings & J. R. R. Tolkien & 6 & 62 & 1,037 & 57\\
    Frozen Movie Script & Jennifer Lee & 1 & 112 & $\sim$40 & 41\\
    The Hand Washing Song & Multiple & 1 & 1 & 1 & 28\\
    Dream of the Red Chamber & Xueqin Cao & 2 & 120 & 2500 & 23\\
    Les Misérables & Victor Hugo & 68 & 365 & 1,462 & 21\\
    \bottomrule
  \end{tabularx} }
  \caption{Examples of different texts with the
    number of languages translated to date \citep{unesco1932world, mayer2014creating, de2019principito, zi1993dao, covid19wiki, coelho2014alchemist, rowling2019harry, tolkien2012lord, jennifer2013frozen, thampi2020s, xueqin2011dream, hugo1863miserables}.}
    \label{table:exbooks}
\end{table*}

Having understood our translation workflow, we are ready to examine different active learning methods. One may ask, how does our translation task into low-resource languages warrant special study given there are many existing research on active learning. To understand the reason why our translation needs warrant a special study, it is important that we understand the concept of the Knapsack Problem. 

What is the Knapsack Problem? How is it relevant to translation? And how does it make active learning in our translation scenario different from existing active learning research? To answer these questions, let us define the Knapsack Problem first. 

\begin{definition}[Knapsack Problem]
Optimize the selection of items from a given set, considering their weights and values, with the aim of keeping the total weight within a given limit while maximizing the total value \citep{pisinger1997minimal,kellerer2004multidimensional}. 
\end{definition}

The Knapsack Problem finds practical applications in a myriad of decision-making scenarios, spanning a wide array of fields and industries, including Machine Translation \citep{eck2008developing}. If we compare sentences in a text with items in a set, and use translation costs as weights, and use sentence scores as values, then we can define the Knapsack Problem in Machine Translation. The Knapsack Problem is NP-hard, similarly, the Knapsack Problem in Machine Translation is also NP-hard.

\begin{definition}[Knapsack Problem in Machine Translation]
Optimize the selection of sentences from a given text, considering their translation costs and scores, with the aim of keeping the total cost within a given limit while maximizing the combined score \citep{eck2008developing}.
\end{definition}

With these definitions, we are ready to compare our active learning situations with existing active learning works in machine translation. Most of the existing research using active learning for machine translation can be framed as the Knapsack Problem. However, the difference lies mostly in what the sentence scores represent. Most of the existing work uses a coverage-based score with the goal of covering each item in vocabulary once \citep{eck2008developing, eck2005low, haffari2009activea}. This is very different when we are using active learning in translation into low-resource languages. Unlike existing work that covers the entire vocabulary, we use active learning to build a seed corpus that is as low as $\sim$600 sentences which only cover a small portion of the vocabulary. Our goal is to determine, with severely limited resources, which set of sentences, covering only a small subset of the vocabulary, is able to give the best translation draft in our translation efforts.   

Indeed, many researchers count the number of unknown n-grams as score functions to solve the Knapsack Problem, covering all vocabulary \citep{eck2008developing, eck2005low, haffari2009activea}.
Instead of solving the Knapsack Problem,  we choose sentences to partially cover the vocabulary and build an extremely small seed corpus. To cover the vocabulary strategically, we sum the frequency counts of the unknown n-grams to increase density. These frequency counts promote frequent words for learning to be meaningful under extremely low-resource scenarios. 

In Table~\ref{table:ngramScore} we denote frequency function by $F(\cdot)$, denote sequence length by $L$ and denote the highest n-gram order by $J$. 

We first examine random sampling, before we introduce 14 
more different active learning methods. 
Random sampling approach is different from the traditional 
portion-based approach of translation by field-linguists.   
The main difference of the two approaches is that
the portion-based approach focuses on preserving
coherence of the text locally, while the random-sampling
approach focuses on increasing coverage of the text
globally. Our results show that the random
sampling approach performs better. 
When training on a seed corpus of $\sim$1,000 lines
from the Bible and testing on the rest of the Bible
($\sim$30,000 lines), random sampling beats
the portion-based approach by +8.5 BLEU using
English as a simulated low-resource language training on
a family of languages built on the distortion measure, and by
+1.9 using a Mayan language, Eastern Pokomchi, training
on a family of languages based on the linguistic definition.
Using random sampling, machine translation
is able to produce a high-quality draft of the whole
text that expedites the subsequent iterations of translation efforts.

In addition to random sampling, we explore 14 more 
different active learning methods. 
To improve accuracy and effectiveness in building 
better optimized seed corpora than randomly sampled 
data so as to minimize the initial human effort, 
we propose explainable and robust active learning methods 
that perform as well as or better than random sampling; 
we transfer methods learned on data of known 
languages to the new, severely low-resource language. 
We also examine different training schedules and 
we find a strategic way of growing large 
multilingual models in a multilingual and 
multi-stage fashion with extremely small 
severely low-resource seed corpora. Our work 
aims to bridge the gap between human translators 
and machine translation systems for them to work together better. 

Moreover, we compare three different ways of incorporating
incremental post-edited data during 
the translation process. We find that
self-supervision using the whole translation draft
affects performance 
adversely, and is best to be avoided. We also show that adding
the newly post-edited text to
training with vocabulary update performs the best.

\section{Methodology}\label{method}
Our main goal is to translate a given text that is available in many languages to a new, severely low-resource language. In our translation workflow, we first develop active learning methods to transfer sentence ranking from known languages to a new, severely low-resource language. We then pass this ranking to human translators for them to translate the top few ($\sim$1,000 or less) sentences into the severely low-resource language, curating the seed corpus. We finally train on the seed corpus, either from scratch or from a pretrained model. 

To develop active learning methods to transfer sentence ranking from known languages to a new, severely low-resource language, we start with random sampling and comparing it with the traditional portion-based approach. In addition to random sampling, we propose and compare 14 active learning methods for machine translation into a new, severely low-resource language. To compare all active learning algorithms fairly, we use the same translation system unit as a control for all experiments, varying only the seed corpora built by different methods. We select the same number of words in all seed corpora as most translators are paid by the number of words
\citep{bloodgood2014bucking, eck2008developing, tomanek2009semi}.

\subsection{Training Schedule}
In our setup we have the new, severely low-resource language as the target language, and we have a few neighboring languages as the source languages that are either in the same linguistic language family or geographically close to facilitate linguistic transfer. In effect, we have $N$ source languages with full translations of the text and a new and severely low-resource language that has an extremely small seed corpus.

We train our models using a state-of-the-art multilingual transformer
by adding language labels to each source sentence
\citep{johnson2017google, ha2016toward, zhou2018massively, zhou2018paraphrases}.
We borrow the order-preserving named entity translation
method by replacing each named entity with \texttt{\_\_NE}s
\citep{zhou2018paraphrases} using a multilingual
lexicon table that covers 124 source languages and 2,939
named entities \citep{zhou2021family}. For example, the sentence ``Somchai calls Juan''
is transformed to ``\texttt{\_\_opt\_src\_en \_\_opt\_tgt\_ca} \texttt{\_\_NE0} calls \texttt{\_\_NE1}''
to translate to Chuj. 
We use families of close-by languages constructed by
ranking 124 source languages by distortion measure (\textit{FAMD}),
performance measure (\textit{FAMP})
and linguistic family (\textit{FAMO$^+$});
the distortion measure ranks languages by decreasing probability
of zero distortion, while the performance measure incorporates
an additional probability of fertility equalling one \citep{zhou2021family}.
Using families constructed, we pretrain our model first 
on the whole text of nearby languages,
then we train on the $\sim$1,000 lines
of low-resource data and the corresponding lines in other languages 
in a multi-source multi-target fashion.
We finally train on the $\sim$1,000 lines in a multi-source single-target
fashion \citep{zhou2021family}.

We combine translations
of all source languages into one. Let all $N$ translations
be $t_i, i = 1, \ldots, N$ and let similarity
between translations $t_i$ and $t_j$ be $S_{ij}$. 
We rank all translations according to how
centered it is with respect to other sentences by
summing all its similarities to the rest through $\sum_j S_{ij}$ for $i = 1, \ldots, N$.
We take the most centered translation
for every sentence, $\max_i \sum_j S_{ij}$, to build the combined
translation output. The expectation of the combined score is higher than that
of any of the source languages \citep{zhou2021family}.  

\begin{table}[t]
  \small
  \centering
  \begin{tabularx}{\textwidth}{p{1cm}p{8.2cm}p{6cm}}
    \toprule
    Name & Description & Score Function \\ 
    \midrule 
    \textit{S} & Frequency sum of unknown words & $ \sum\limits_{i=0}^L F(w^{u}_i)$ \\
    \textit{SN} & Normalized \textit{S} by $L$ & $\frac{1}{L} \sum\limits_{i=0}^L F(w^{u}_i)$ \\
    \textit{SNG$_{J}$} & Normalized Frequency sum of n-grams up to $J$ & $\frac{1}{L} \sum\limits_{j=1}^J \sum\limits_{i=0}^L F(g^{u}_{i,j})$ \\
    \textit{AGG$^{M}_{J}$} & Aggregation of n-gram scores up to $J$ with set $M$ & $\sum\limits_{M} \frac{1}{L} \sum\limits_{j=1}^J \sum\limits_{i=0}^L F(g^{u}_{i,j})$ \\
    \textit{ENT$^{K}$} & Entropy methods, $K$ is KenLM or not & $H_{c}^K(s) - I_{l}(s) \cdot H_{r}^K(s) - I_{r}(s) \cdot H_{l}^K(s)$ \\
    \bottomrule
  \end{tabularx}
    \caption{Summary of score functions. 
    }
\label{table:ngramScore}
\end{table}

\subsection{Active Learning Strategies} \label{2023methods}
\subsubsection{Random Sampling} 
Our initial approach is random sampling to increase coverage of the text. To show the effect of random sampling, we compare it with the baseline of the linguistic baseline of the excerpt-based approach, \textit{Luke}. The excerpt-based approach, which selects a portion of the text with consecutive sentences, preserves the text's formality, cohesion and context but lacks global coverage. Random sampling increases global coverage but sacrifices local coherence. 

\subsubsection{N-gram Approach}
In addition to random sampling, we devise different active learning methods based on n-gram methods, entropy methods, and aggregation methods. To show the effectiveness of these active learning methods, we compare them with two baselines: linguistic baseline of the excerpt-based approach, \textit{Luke}, and the statistical baseline of random sampling as our second baseline. Therefore, for the next section of active learning, we have 2 baselines: the linguistic baseline of the excerpt-based approach, \textit{Luke}, and the statistical baseline of random sampling, \textit{Rand}. 

N-gram methods have often been used in research. Many researchers count the number of unknown n-grams as score functions to solve the Knapsack Problem, covering all vocabulary \citep{eck2008developing, eck2005low, haffari2009activea}.
Instead of solving the Knapsack Problem,  we choose sentences to partially cover the vocabulary and build an extremely small seed corpus. To cover the vocabulary strategically, we sum the frequency counts of the unknown n-grams to increase density. These frequency counts promote frequent words for learning to be meaningful in the extremely low-resource scenario. In Table~\ref{table:ngramScore} we denote frequency function by $F(\cdot)$, denote sequence length by $L$ and denote the highest n-gram order by $J$. 

\subsubsection{Entropy Approach}
Many have worked on entropy methods in modelling density and diversity
\citep{ambati2011multi, eck2008developing, zeng2019empirical, haffari2009activea}.
Our problem defers from previous work in that our data is much smaller, and we rely on other languages to generalize to the severely low-resource language. Therefore, we combine the existing research in combining density and diversity metrics into score functions in ranking sentences together with frequency counts\footnote{In the entropy score function in Table~\ref{table:ngramScore}, we use highest n-gram order of 2 for NLTK's LM, we use highest n-gram order of 2 for the two halves ($H_{l}^K$ and $H_{r}^K$) and order of 5 for the sampled data ($H_{c}^K$) for KenLM. Since KenLM needs at least a few words to
start with, we use MLE as a warm start to select up to
5 sentences before launching KenLM. }. Frequency counts here helps us to promote frequent words in our use case of extremely small data and partial vocabulary. We would like to cover as many words as possible, but we also would like to give priority to more frequent words for learning to be meaningful in the extremely low-resource scenario.
We use traditional Language Models (LMs) instead of neural language models, as our data size is extremely small. 
For implementations of LMs, we use KenLM and NLTK's LM because of their simplicity and speed, especially KenLM \citep{heafield2011kenlm, loper2002nltk, gauss1816bestimmung, hagen1837grundzuge,edgeworth1909addendum}. 
In Table~\ref{table:ngramScore} we let $H(\cdot)$ be the cross entropy function, with the choice of KenLM (K) or NLTK (N). To separate training from testing in using language models, we divide the data into three portions, the sentences that we have chosen (\textit{c}), and the remaining that are split equally into two parts, left 
(\textit{l}) and right (\textit{r}). Let $I_{l}(\cdot)$ and $I_{r}(\cdot)$ be indicator functions to show whether a sentence belongs to the left or the right. We aim to maximize the diversity $H_{c}$ and optimize density by adjusting $H_{l}$ and $H_{r}$ \citep{koneru2022cost}. 

\subsubsection{Aggregation Approach}
To prevent any language from overpowering the ranking, we aggregate sentence scores across different languages (Figure~\ref{fig:active}). We investigate the use of a customized set of languages for each severely low-resource language, versus the use of a universal set of languages representing world languages. The former requires some understanding of the neighboring languages, the latter requires careful choices of the representative set \citep{blasi2021systematic}.
To build a universal ranking, we either aggregate over all existing languages, or create a representative pool for existing languages.

We have 4 aggregation methods: \textit{one-vote-per-language} (L), where we aggregate over all languages, \textit{one-vote-per-family} (F), where we aggregate over languages representing the top few families, \textit{one-vote-per-person} (P), where we aggregate over the top few most spoken languages, and \textit{one-vote-per-neighbor} (N), where we aggregate over a customized set of neighboring languages. For the world language distribution, L covers all, F samples across it, P covers the head, while N creates a niche area around the severely low-resource language.

Aggregation decreases variance and increases accuracy. Typical aggregation 
involve taking the sum or the average. Since they have the same
effect on sentence ranking, we take the sum for simplicity.

To implement \textit{one-vote-per-language}, we train on all available languages.
that have full translations of the text
, which calls for memory and time efficient algorithms. 
We use parallel algorithms, use hash maps for all data structures, 
skip disk-caching to save time and space, and reuse important 
data structures \citep{alon2022neuro}. 
To save space and time, we devise \textit{relaxed memoization}. 
We observe that only a few words are added to the vocabulary 
during each step, and therefore only a few sentences containing 
these new words have updated scores. 
At every step, we compute sentence score for each language, 
producing a score matrix of languages versus sentences. We 
update entries that are affected by the selected sentence, 
cache and reuse other entries. Further parallelism  reduce memory consumption and 
results in >360 times speedup, from $\sim$6.5 months 
to $\sim$13 hours. Our code is efficient in memory and time. 

\subsection{Joint Human Machine Translation}
Our work differs from the past research in that we put
low-resource translation into the broad collaborative
scheme of human machine translation. 
We compare the portion-based approach with 
the active learning approach in building seed corpora.
We also compare three methods of updating models with
increasing amount of human post-edited data.
We add the newly post-edited data to training in three ways: 
with vocabulary update, 
without vocabulary update, 
or incorporating the whole translation draft in a
self-supervised fashion additionally. 
For best performance,
we build the seed corpus by active learning, 
update vocabulary iteratively, 
and add newly post-edited data to 
training without self-supervision.
We also have a larger test set,
we test on $\sim$30,000 lines rather than
$\sim$678 lines from existing research \textsuperscript{\ref{fn1}}. 

We propose a joint human machine translation
workflow in Algorithm \ref{algo:proposedtrans}.
After pretraining on neighboring languages in Step 3,
we iteratively train on the randomly sampled
seed corpus of low-resource data in Step 4 and 5.
The reason we include both Step 4 and 5 in our algorithm
is because training both steps iteratively performs 
better than training either one \citep{zhou2021family}. 
Our model produces a translation draft of the whole text. 
Since the portion-based approach has the advantage
with formality, cohesion and contextual relevance, human
translators may pick and post-edit portion-by-portion iteratively. 
The newly post-edited data with updated vocabulary is added to
the machine translation models without self-supervision. 
In this way, machine translation systems
rely on quality parallel corpora that are incrementally produced
by human translators.
Human translators lean on machine
translation for quality translation draft to expedite translation.
This creates a synergistic collaboration between
human and machine. 

\subsection{Evaluation Metrics}
Existing multilingual systems produce multiple outputs from all source languages, rendering comparison messy. To simplify, we combine translations from all source languages into one by an existing \textit{centeredness method} \citep{zhou2021family}. Using this method, we score each translated sentence by the sum of its similarity scores to all others, and we take the highest ranked sentence by score as our final translation.  

To compare effectively, we control all test sets to be the same. Since different active learning strategies produce different seed corpora to be used as training and validation sets, the training and validation sets vary. Their complement, the test sets therefore also vary, rendering comparison difficult. To build the same test set, we devise an \textit{intersection method}. We take the whole text and carve out all seed corpora, that is, all training and validation sets from all experiments. The remaining is the final test set, which is the intersection of all test sets. 

Using our intersection method, we are able to build a common test set for all our experiments. One may ask, why not set aside a fixed amount of sentences for testing? The reason that we do not hold out a test set is because of our experiment setup. In each experiment, we use a different active learning algorithm to choose $\sim$3\% of the text to build the seed corpus, then use this $\sim$3\% of the text to translate $\sim$97\% of the text. We have more than 200 experiments in this chapter, and each has a different seed corpus. In each experiment, we compare all the sentences of the text to select sentences to build a seed corpus. If we do hold out a test set for all experiments, this bars those held-out test sentences from being selected into the seed corpus for each experiment. Consequently, all seed corpora will be changed as a result of holding out a test set. This is not what we intend in our experiment design. This is why we propose the intersection method so that every experiment has access to all sentences to select into its seed corpus and we can compare them most effectively.   

Our metrics are: chrF, characTER, BLEU, COMET score, and BERTscore \citep{popovic2015chrf, wang2016character, post-2018-call, zhang2019bertscore, stewart-etal-2020-comet, rei2021mt}. For random sampling, our metric used is mainly BLEU as we only show our result in two languages. However, for experiments on extended active learning strategies, we test on many more languages and we prioritize chrF over BLEU for better accuracy, fluency and expressive power in morphologically-rich languages \citep{papineni2002bleu}.

\begin{table*}[t]
  \small
  \centering
  \begin{tabularx}{\textwidth}{p{2cm}p{0.5cm}p{2.5cm}p{9.5cm}}
    \toprule 
    Target & L & Family & Source Languages \\
    \midrule
    Frisian & 0 & Germanic & English*, German, Dutch, Norwegian, Afrikaans, Swedish, French, Italian, Portuguese, Romanian \\  
    Hmong & 0 & Hmong–Mien & Komrem*, Vietnamese, Thai, Chinese, Myanmar, Haka, Tangsa, Zokam, Siyin, Falam \\
    Pokomchi & 0 & Mayan & Chuj*, Cakchiquel, Mam, Kanjobal, Cuzco, Ayacucho, Bolivian, Huallaga, Aymara, Guajajara \\
    Turkmen & 1 & Turkic & Kyrgyz*, Tuvan, Uzbek, Karakalpak, Kazakh, Azerbaijani, Japanese, Korean, Finnish, Hungarian \\
    Sesotho & 1 & Niger–Congo & Yoruba*, Gikuyu, Xhosa, Kuanyama, Kpelle, Fon, Bulu, Swati, Venda, Lenje \\
    Welsh & 1 & Celtic & English*, German, Danish, Dutch, Norwegian, Swedish, French, Italian, Portuguese, Romanian \\
    Xhosa & 2 & Nguni & Swati*, Gikuyu, Sesotho, Yoruba, Lenje, Gbaya, Afrikaans, Wolaitta, Kuanyama, Bulu \\
    Indonesian & 3 & Austronesian & Javanese*, Malagsy, Tagalog, Ilokano, Cebuano, Fijian, Sunda, Zokam, Wa, Maori \\	
    Hungarian & 4 & Uralic & Finnish*, French, English, German, Latin, Romanian, Swedish, Spanish, Italian, Portuguese \\
    Spanish & 5 & Romance & English*, German, Danish, Dutch, Norwegian, Swedish, French, Italian, Portuguese, Romanian \\
    \bottomrule
  \end{tabularx}
    \caption{Summary of different target languages used \citep{campbell2018cataloguing, collin2010ethnologue}.
    L, resource level, is from a scale of 0 to 5 \citep{joshi2020state}.
    Reference languages used for active learning methods except aggregate methods are starred. }
\label{table:ethnologue_languages}
\end{table*}

\section{Data}\label{data}
We show two sets of experiments: one on random sampling, 
the other on more extended active learning methods 
including n-gram methods, entropy methods, and aggregation methods. 

\subsection{Random Sampling}
To first test random sampling, we work on the Bible 
in 124 source languages \citep{mayer2014creating}, and
have experiments for English, a simulated
language, and Eastern Pokomchi, a Mayan language.
We train on $\sim$1,000 lines of low-resource data and on full texts for all
the other languages.
We aim to translate the
rest of the text ($\sim$30,000 lines) into the low-resource language.
In pretraining, we use
80\%, 10\%, 10\% split for training, validation
and testing. In training, we
use 3.3\%, 0.2\%, 96.5\% split for training, validation
and testing. Our test size is >29 times of the training size \textsuperscript{\ref{fn1}}.
We use the book ``Luke'' for the portion-based approach as suggested by many 
human translators. 

Training on $\sim$100 million parameters with Geforce RTX 2080 Ti,
we employ a 6-layer encoder and a 6-layer decoder with
512 hidden states, 8 attention heads,
512 word vector size, 2,048 hidden units,
6,000 batch size, 0.1 label smoothing,
2.5 learning rate, 0.1 dropout and attention dropout,
an early stopping patience of 5 after 190,000 steps,
``BLEU'' validation metric,
``adam'' optimizer and ``noam'' decay method \citep{klein2017opennmt, papineni2002bleu}. We increase
patience to 25 for larger data in the second stage of training in Figure
\ref{fig:curve_en_hmt} and \ref{fig:curve_ph_hmt}.

\subsection{N-gram, Entropy and Aggregation Methods}
In addition to random sampling, we are interested in comparing various n-gram methods, entropy methods and aggregation methods. To test these methods, we choose a more extensive set of target languages. When we choose target languages, we look at existing research, which classifies world languages into Resource 0 to 5, with 0 having the lowest resource and 5 having the highest  \citep{joshi2020state}. We choose 10 target languages ranging from Resource 0 to 5 (Table~\ref{table:ethnologue_languages}). For each target language we choose ten neighboring languages as source
languages (Table~\ref{table:ethnologue_languages}). These languages are Eastern Pokomchi, Hmong, and Frisian (Resource 0), Turkmen, Welsh and Sesotho (Resource 1), Xhosa (Resource 2), Indonesian (Resource 3), Hungarian (Resource 4), Chinese and Spanish (Resource 5) in Table~\ref{table:ethnologue_languages}.
We prioritize Resource 0 to 2 languages as real low-resource languages, and we use Resource 3 to 5 languages as hypothetical ones. 
It is surprising to us that a lot of the Resource 0 languages are not too far away from the rich-resource languages. Frisian, for example, are spoken near the Northern Sea near Netherlands and Germany, and is in close proximity with a few rich-resource European languages \cite{markey2011frisian}. However, because of the close proximity with rich-resource languages, low-resource languages like Frisian often suffer from lack of prestige and has a bigger threat to extinction as many younger people choose to speak the rich-resource languages nearby. This also suggests interesting research direction on low-resource languages and dialects that are in close proximity with rich-resource language communities.  

To translate into these languages, our text is the Bible in 125 languages \citep{mayer2014creating}. Each low-resource seed corpus contains $\sim$3\% of the text, while all other languages have full text. Our goal is to translate the rest of the text into the low-resource language. In pretraining, we use a 80/10/10 split for training, validation and testing, respectively. In training, we use approximately a 3.0/0.2/96.8 split for training, validation and testing, respectively. Our training data for each experiment is $\sim$1,000 lines. We use BPE with size of $\sim$3,000 for the low-resource language and $\sim$9,000 for the combined \citep{sennrich2016neural}. 

Training on $\sim$100 million 
parameters with Geforce RTX 2080 Ti and RTX 3090,
we use a 6-layer encoder and a 6-layer decoder with
512 hidden states, 8 attention heads,
512 word vector size, 2,048 hidden units,
6,000 batch size, 0.1 label smoothing,
2.5 learning learning rate and 1.0 finetuning learning rate, 
0.1 dropout and attention dropout,
a patience of 5 after 190,000 steps in {[N]}$^2$ with an update interval of 1000, 
a patience of 5 for {[N+1]}$^2$ with an update interval of 200, and 
a patience of 25 for {[N+1]} and {[1]}$^2$ with an update interval of 50, 
``adam'' optimizer and
``noam'' decay method \citep{klein2017opennmt, papineni2002bleu}. 

\begin{table*}[t]
  \centering
  \small
  \resizebox{\textwidth}{!}{
  \begin{tabular}{p{1.6cm}p{0.4cm}p{0.65cm}|p{0.4cm}p{0.6cm}|
      p{1.6cm}p{0.4cm}p{0.65cm}|p{0.4cm}p{0.6cm}|
      p{1.6cm}p{0.4cm}p{0.65cm}|p{0.4cm}p{0.6cm}}
    \toprule
    \multicolumn{15}{c}{Input Language Family} \\
    \midrule
    \multicolumn{5}{c|}{By Linguistics} & \multicolumn{5}{c|}{By Distortion} & \multicolumn{5}{c}{By Performance} \\
    \midrule
    \multicolumn{5}{c|}{\textit{FAMO$^+$}} & \multicolumn{5}{c|}{\textit{FAMD}} & \multicolumn{5}{c}{\textit{FAMP}} \\
    \midrule
    Training & \multicolumn{2}{c|}{\textit{Luke}} & \multicolumn{2}{c|}{\textit{Rand}} &
    Training & \multicolumn{2}{c|}{\textit{Luke}} & \multicolumn{2}{c|}{\textit{Rand}} &
    Training & \multicolumn{2}{c|}{\textit{Luke}} & \multicolumn{2}{c}{\textit{Rand}} \\
    \midrule
    Testing & \textit{Best} & \textit{All} & \textit{Best} & \textit{All} &
    Testing & \textit{Best} & \textit{All} & \textit{Best} & \textit{All} &
    Testing & \textit{Best} & \textit{All} & \textit{Best} & \textit{All} \\
    \midrule
    Combined & 37.9 & 21.9 & 42.8 & 28.6 &
    Combined & 38.6 & 22.9 & 44.8 & 31.4 &
    Combined & 40.2 & 23.7 & 44.6 & 30.6 \\
    \midrule
    German & 35.6 & 20.0 & 40.8 & 26.5 &
    German & 37.0 & 20.8 & 42.7 & 28.8 &
    German & 38.0 & 21.3 & 41.6 & 28.2\\
    Danish & 36.7 & 19.0 & 38.2 & 25.9 &
    Danish & 37.3 & 19.6 & 39.5 & 28.0 &
    Danish & 38.4 & 19.9 & 39.2 & 27.5 \\
    Dutch & 36.4 & 20.4 & 39.7 & 27.2 &
    Dutch & 36.4 & 21.1 & 41.9 & 29.6 &
    Dutch & 37.5 & 21.6 & 41.6 & 28.9 \\
    Norwegian & 36.5 & 20.2 & 40.0 & 26.9 &
    Norwegian & 37.2 & 20.8 & 41.4 & 29.1 &
    Norwegian & 37.5 & 21.1 & 41.0 & 28.4 \\
    Swedish & 34.9 & 19.7 & 39.9 & 26.2 &
    Afrikaans & 38.3 & 22.2 & 42.8 & 30.5 &
    Afrikaans & 39.5 & 22.9 & 42.3 & 29.8 \\
    Spanish & 36.8 & 21.5 & 39.8 & 27.6 &
    Marshallese & 35.1 & 21.6 & 41.4 & 28.8 &
    Spanish & 38.7 & 22.9 & 41.9 & 29.0 \\
    French & 36.0 & 19.7 & 39.6 & 26.1 &
    French & 36.2 & 20.3 & 41.1 & 28.3 &
    French & 37.3 & 20.7 & 40.5 & 27.5 \\
    Italian & 36.7 & 20.6 & 38.4 & 26.9 &
    Italian & 37.3 & 21.0 & 40.6 & 29.1 &
    Italian & 38.6 & 21.8 & 39.9 & 28.5 \\
    Portuguese & 32.4 & 15.8 & 30.1 & 21.3 &
    Portuguese & 33.2 & 16.5 & 33.6 & 24.0 &
    Portuguese & 33.7 & 16.3 & 33.1 & 22.9 \\
    Romanian & 34.9 & 19.3 & 37.1 & 26.0 &
    Frisian & 36.4 & 21.6 & 43.0 & 29.8 &
    Frisian & 37.8 & 22.3 & 42.2 & 29.1 \\
    \bottomrule
  \end{tabular}
  }
  \caption{Performance training on 1,093 lines of English data on \textit{FAMO$^+$}, \textit{FAMD} and \textit{FAMP}. We train using the portion-based approach in \textit{Luke}, and using random sampling in \textit{Rand}. During testing, \textit{Best} is the book with highest BLEU score, and \textit{All} is the performance on $\sim$29,000 lines of test data \textsuperscript{\ref{fn1}}.}
  \label{table:enRandCompare}
\end{table*}
\begin{table*}[t] 
  \centering
  \small
  \resizebox{\textwidth}{!}{
  \begin{tabular}{p{1.6cm}p{0.4cm}p{0.65cm}|p{0.4cm}p{0.6cm}|
  p{1.6cm}p{0.4cm}p{0.65cm}|p{0.4cm}p{0.6cm}|
  p{1.6cm}p{0.4cm}p{0.65cm}|p{0.4cm}p{0.6cm}}
    \toprule
    \multicolumn{15}{c}{Input Language Family} \\
    \midrule
    \multicolumn{5}{c|}{By Linguistics} & \multicolumn{5}{c|}{By Distortion} & \multicolumn{5}{c}{By Performance} \\
    \midrule
    \multicolumn{5}{c|}{\textit{FAMO$^+$}} & \multicolumn{5}{c|}{\textit{FAMD}} & \multicolumn{5}{c}{\textit{FAMP}} \\
    \midrule
    Training & \multicolumn{2}{c|}{\textit{Luke}} & \multicolumn{2}{c|}{\textit{Rand}} &
    Training & \multicolumn{2}{c|}{\textit{Luke}} & \multicolumn{2}{c|}{\textit{Rand}} &
    Training & \multicolumn{2}{c|}{\textit{Luke}} & \multicolumn{2}{c}{\textit{Rand}} \\
    \midrule       
    Testing & \textit{Best} & \textit{All} & \textit{Best} & \textit{All} &
    Testing & \textit{Best} & \textit{All} & \textit{Best} & \textit{All} &
    Testing & \textit{Best} & \textit{All} & \textit{Best} & \textit{All} \\
    \midrule
    Combined & 23.1 & 8.6 & 19.7 & 10.5 & 
    Combined & 23.3 & 8.5 & 17.7 & 9.5 & 
    Combined & 22.4 & 7.2 & 15.8 & 7.8 \\
    \midrule
    Chuj & 21.8 & 7.9 & 16.5 & 9.8 &
    Chuj & 22.0 & 7.9 & 15.4 & 8.9 & 
    Chuj & 21.8 & 7.0 & 13.2 & 7.3 \\ 
    Cakchiquel & 22.3 & 7.9 & 18.2 & 9.9 &
    Cakchiquel & 22.4 & 7.9 & 17.3 & 9.1 &
    Cakchiquel & 21.2 & 6.9 & 14.8 & 7.4 \\ 
    Guajajara & 19.9 & 7.1 & 14.7 & 8.9 &
    Guajajara & 19.2 & 6.9 & 14.2 & 8.2 &
    Guajajara & 18.9 & 5.9 & 10.6 & 6.6 \\
    Mam & 22.2 & 8.6 & 19.7 & 10.6 &
    Russian & 22.2 & 7.3 & 13.7 & 8.5 &
    Mam & 21.9 & 7.5 & 17.1 & 8.0 \\ 
    Kanjobal & 21.8 & 8.1 & 17.5 & 10.0 &
    Toba & 22.0 & 8.3 & 16.8 & 9.4 &
    Kanjobal & 21.6 & 7.1 & 13.8 & 7.6 \\ 
    Cuzco & 22.4 & 7.8 & 17.7 & 9.8 &
    Myanmar & 19.2 & 5.3 & 10.7 & 6.5 &
    Thai & 21.9 & 6.3 & 10.5 & 7.0 \\
    Ayacucho & 21.6 & 7.6 & 18.5 & 9.7 &
    Slovenský & 22.2 & 7.5 & 13.5 & 8.7 &
    Dadibi & 19.9 & 6.2 & 15.3 & 6.9 \\
    Bolivian & 22.3 & 7.8 & 17.4 & 9.8 &
    Latin & 22.0 & 7.8 & 14.8 & 9.0 &
    Gumatj & 19.2 & 3.8 & 8.9 & 3.3 \\
    Huallaga & 22.2 & 7.7 & 18.0 & 9.7 &
    Ilokano & 22.6 & 8.4 & 17.8 & 9.4 &
    Navajo & 21.4 & 6.5 & 13.5 & 7.3 \\
    Aymara & 21.5 & 7.5 & 18.6 & 9.6 &
    Norwegian & 22.6 & 8.3 & 16.7 & 9.4 &
    Kim & 21.6 & 7.0 & 13.9 & 7.5 \\
    \bottomrule
  \end{tabular}
  }
  \caption{Performance training on 1,086 lines of Eastern Pokomchi data on \textit{FAMO$^+$}, \textit{FAMD} and \textit{FAMP}. We train using the portion-based approach in \textit{Luke}, and using random sampling in \textit{Rand}. During testing, \textit{Best} is the book with highest BLEU score, and \textit{All} is the performance on $\sim$29,000 lines of test data \textsuperscript{\ref{fn1}}.}
    \label{table:phRandCompare}
\end{table*}
\begin{table}[t]
  \centering
  \footnotesize
  \begin{tabularx}{\columnwidth}{p{2.8cm}p{2.5cm}p{3.3cm}p{3.3cm}p{3.3cm}}
    \toprule
    Source & \textit{Seed} & \textit{Self-Supervised} & \textit{Old-Vocab} & \textit{Updated-Vocab} \\
    \midrule
    Combined & 30.8 & 24.4 (-6.4) & 32.1 (+1.3) & 32.4 (+1.6) \\
    \midrule
    Danish & 27.7 & 21.6 (-6.1) & 28.8 (+1.1) & 29.2 (+1.5) \\
    Norwegian & 28.6 & 22.5 (-6.1) & 29.8 (+1.2) & 30.2 (+1.6) \\
    Italian & 28.7 & 22.3 (-6.4) & 29.8 (+1.1) & 30.2 (+1.5) \\
    Afrikaans & 30.1 & 23.8 (-6.3) & 31.4 (+1.3) & 31.6 (+1.5) \\
    Dutch & 29.2 & 22.9 (-6.3) & 30.3 (+1.1) & 30.6 (+1.4) \\
    Portuguese & 23.8 & 18.3 (-5.5) & 24.6 (+0.8) & 25.0 (+1.2) \\
    French & 27.8 & 21.7 (-6.1) & 28.9 (+1.1) & 29.4 (+1.6) \\
    German & 28.4 & 22.4 (-6.0) & 29.5 (+1.1) & 29.9 (+1.5) \\
    Marshallese & 28.4 & 22.4 (-6.0) & 29.5 (+1.1) & 29.9 (+1.5) \\
    Frisian & 29.3 & 23.2 (-6.1) & 30.4 (+1.1) & 30.8 (+1.5) \\
    \bottomrule
    \end{tabularx}
  \caption{Comparing three ways of adding the newly post-edited book of 1 Chronicles \textsuperscript{\ref{fn1}}. \textit{Seed} is the baseline of training on the seed corpus alone, \textit{Old-Vocab} skips the vocabulary update while \textit{Updated-Vocab} has vocabulary update. \textit{Self-Supervised} adds the complete translation draft in addition to the new book. 
  }
  \label{table:1chron}
\end{table}

\section{Results}\label{results}
We first look at our results on random sampling, and then we compare it with n-gram, entropy and aggregation methods through extensive experiments in multiple severely low-resource languages. 

\subsection{Random Sampling}
\label{active:random}
We observe that random sampling performs 
better than the portion-based approach. Training on $\sim$3\% of the text, and testing on $\sim$97\% of the text, we see a sharp performance gain by random sampling. 
In Table \ref{table:enRandCompare} and \ref{table:phRandCompare},
random sampling gives a performance gain of +8.5
for English on FAMD and +1.9 for Eastern Pokomchi on FAMO$^+$ \footnote{
  \label{fn1}
  In the previous chapter, we test on $\sim$30,000 lines excluding
  the $\sim$1,000 lines of training and validation data. In this chapter, 
  we test on the intersection of different test sets. 
  In Table \ref{table:enRandCompare} and \ref{table:phRandCompare}, we test
  on $\sim$29,000 lines of data of the Bible excluding both the book of Luke
  and the randomly sampled $\sim$1,000 lines. In Table~\ref{table:1chron}, we
  evaluate on $\sim$29,000 lines of data of the Bible excluding both 
  the randomly sampled $\sim$1,000 lines and the book of 1 Chronicles. 
}.
The performance gain for
Eastern Pokomchi may be lower because Mayan languages are
morphologically rich, complex, isolated and opaque
\citep{aissen2017mayan, clemens2015ergativity, england2011grammar}.
English is closely related to many languages due to
colonization and globalization even though it is
artificially constrained in size \citep{bird2020decolonising}.
This may explain why Eastern Pokomchi
benefits less. 

In addition to evaluating all of the remaining text ($\sim$97\% of the text), we also evaluate the best book of the remaining text. The Bible has 66 books. The best performing book in our draft recommends human translators the easiest book to post-edit first. 

To simulate human translation efforts in Step 7 and 8 in Algorithm
\ref{algo:proposedtrans}, we rank 66 books of the Bible
by BLEU scores on English's FAMD and Eastern
Pokomchi's FAMO$^+$. We assume that BLEU ranking is available
to us to simulate human judgment.
In reality, this step is realized by human translators
skimming through the translation draft and comparing performances 
of different books by intuition and experience.
In Section \ref{conclusion},
we will discuss the limitation of this assumption.
Performance ranking of the simulated
low-resource language may differ from that of the
actual low-resource language.
But the top few may coincide
because of the nature of the text, independent of the language.
In our results, we observe that narrative books performs better than
philosophical or poetic books. The book of 1 Chronicles
performs best for both English and Eastern Pokomchi with random sampling.
A possible explanation is
that the book of 1 Chronicles is mainly narrative,
and contains many named
entities that are translated well by the
order-preserving lexiconized model.
We included the BLEU scores of the
best-performing book in Table \ref{table:enRandCompare} and \ref{table:phRandCompare}. 
Note that only scores of ``All'' are comparable across experiments
trained on the book of Luke with those trained by
random sampling as they evaluate on the same set \textsuperscript{\ref{fn1}}. 
For the best-performing book, it is the book of 1 Chronicles for random sampling,
and the book of Mark or the book of Matthew for experiments trained on the book of Luke.
Thus, we cannot compare BLEU scores for the best-performing books across experiments. 
We include them in the tables to show the quality of the translation draft human translators will
work on if they proceed to translate the best-performing book. 

In Table \ref{table:1chron}, we compare
three different ways of
updating the machine translation models by
adding a newly post-edited book that
human translators produced.
We call the baseline without addition of the new book \textit{Seed}.
\textit{Updated-Vocab} adds the new book to 
training with updated vocabulary
while \textit{Old-Vocab} skips the vocabulary update.
\textit{Self-Supervised} adds the whole translation draft
of $\sim$30,000 lines to pretraining in addition to the new
book. Self-supervision refers to using the small seed corpus to translate
the rest of the text which is subsequently used to train the model. 
We observe that the \textit{Self-Supervised} performs the
worst among the three. Indeed, \textit{Self-Supervised} performs even
worse than the baseline \textit{Seed}.
This shows that quality is much more important than
quantity in severely low-resource translation. It is better
for us not to add the whole translation draft to the
pretraining as it affects performance adversely.

\begin{figure*}
\centering
\begin{subfigure}{.45\textwidth}
  \centering
  \includegraphics[width=\linewidth]{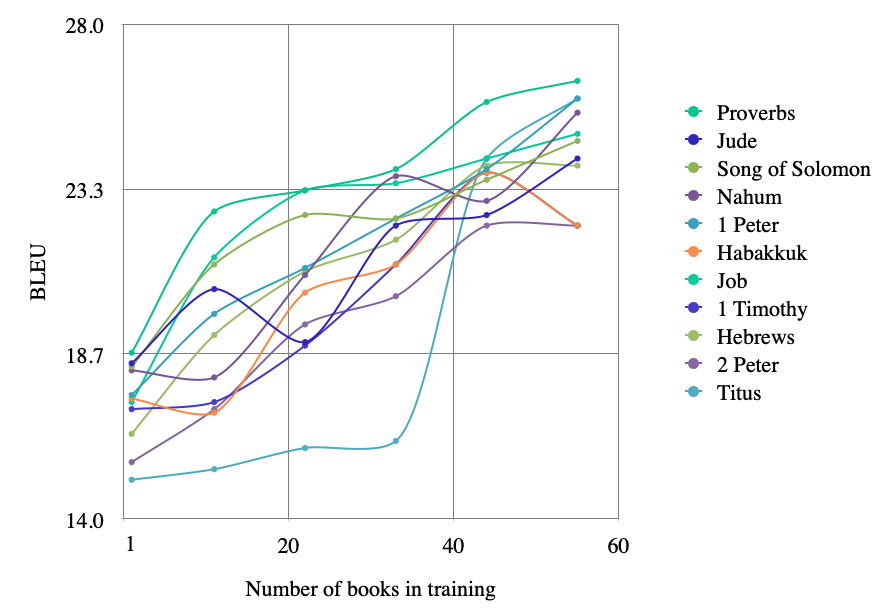}
  \caption{English}
  \label{fig:curve_en_hmt}
\end{subfigure}
\begin{subfigure}{.45\textwidth}
  \centering
  \includegraphics[width=\linewidth]{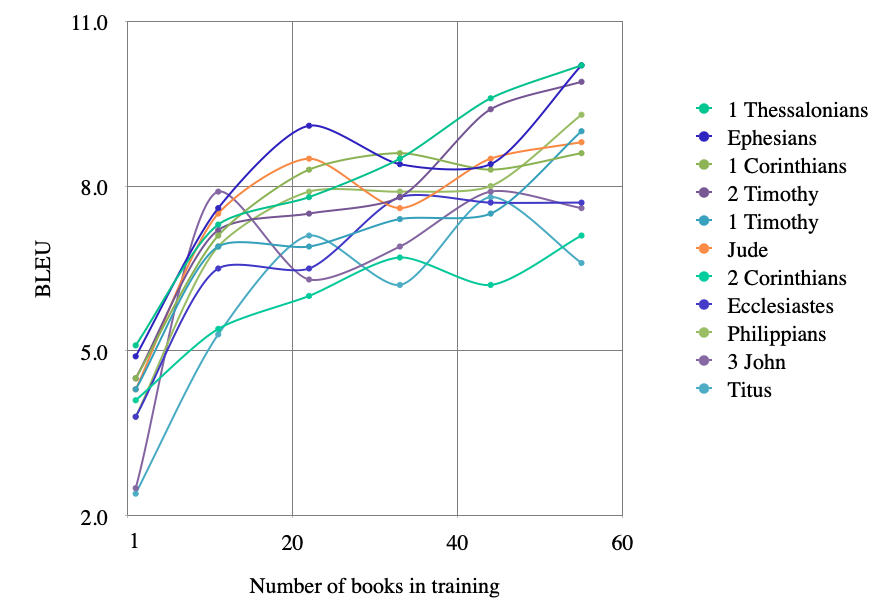}
  \caption{Eastern Pokomchi}
  \label{fig:curve_ph_hmt}
\end{subfigure}
\caption{Performance of the most difficult 11 books with increasing number of training books.}
\end{figure*}

On the other hand, we see that both \textit{Updated-Vocab} and
\textit{Old-Vocab} performs better than \textit{Seed} and
\textit{Self-Supervised}.
\textit{Updated-Vocab}'s performance is better than
\textit{Old-Vocab}. An explanation could be that \textit{Updated-Vocab}
has more expressive power with updated vocabulary.
Therefore, in our proposed algorithm,
we prefers vocabulary update in each iteration. If
the vocabulary has not increased, we may skip 
pretraining to expedite the process. 

We show how the algorithm is put into practice for English and Eastern Pokomchi 
in Figure \ref{fig:curve_en_hmt} and \ref{fig:curve_ph_hmt}.
After producing the first draft of the text by training a MT system on the seed corpus, we hold out the most difficult 11 books (the worst-performing 11 books) and set them aside as the test set for evaluating the entire iterative post-editing process. Taking the most difficult 11 books as the held-out test set,
we divide the other 55 books of the Bible into 5 portions to simulate 5 iterations of post-editing process. 
Using this setup, we translate the text by using the randomly sampled $\sim$1,000 lines of seed corpus
first, and then proceed with human machine translation
in Algorithm \ref{algo:proposedtrans}
in 5 iterations with increasing number of post-edited portions.
Each portion contains 11 books, serving as post-edited portion for each iteration. In each iteration, we simulate human post-editing process by feeding the actual translation of the given text portion to the MT system. MT system produces better and better drafts and we show the improvement using the most difficult 11 books. 

For English, we observe that philosophical books like ``Proverbs'' and poetry books like
``Song of Solomon'' perform very badly in the beginning, but begin to achieve
above 20 BLEU scores after adding 11 books of training data. 
This reinforces our earlier result that $\sim$20\% of the text
is sufficient for achieving high-quality translation \citep{zhou2018massively}.  
However, some books like ``Titus'' remains
difficult to translate even after adding 33 books of training data. This shows
that adding data may benefit some books more than the others.
A possible explanation is that there are 
multiple authors of the Bible, and books differ from each other
in style and content. Some books are closely related to each other, and may
benefit from translations of other books. But some may be
very different and benefit much less. 

\begin{table*}[t]
  \scriptsize
  \centering
  \begin{tabularx}{\textwidth}{p{1.5cm}p{0.7cm}p{0.7cm}p{1.1cm}p{0.9cm}p{0.9cm}p{0.5cm}p{0.5cm}p{1.1cm}p{1.1cm}p{0.8cm}p{0.9cm}} 
   \toprule
   $\uparrow$chrF & Frisian & Hmong & Pokomchi & Turkmen & Sesotho & Welsh & Xhosa & Indonesian & Hungarian & Spanish & Average \\
    \midrule
    \multicolumn{12}{l}{\textbf{Baselines:} }\\
    + \textit{Luke} & 47.5 & 41.6 & 39.4 & 34.9 & 41.2 & 41.2 & 32.0 & 43.3 & 34.4 & 46.7 & 40.2\\
    + \textit{Rand} & 50.5 & 43.9 & 42.8 & 38.9 & 43.2 & 46.0 & 34.9 & 47.2 & 37.4 & 50.1 & 43.5 \\
    \midrule 
    \multicolumn{12}{l}{\textbf{Our Models:} }\\
    + \textit{S} & 49.2 & 38.5 & 40.4 & 35.2 & 39.0 & 41.9 & 32.5 & 43.5 & 35.1 & 48.0 & 40.3 \\
    + \textit{SN} & 50.9 & 43.9 & 43.2 & 38.3 & 41.6 & 43.2 & 36.1 & 46.9 & 36.7 & 50.3 & 43.1 \\
    + \textit{SNG$_{2}$} & 53.2 & \bf{46.1} & 43.3 & 39.5 & 44.4 & 45.8 & 36.6 & 48.4 & 37.8 & 51.8 & 44.7 \\
    + \textit{SNG$_{3}$} & 52.7 & 46.0 & \bf{44.5} & 39.6 & \bf{45.5} & 47.5 & \bf{36.8} & 48.9 & \bf{39.2} & 52.3 & 45.3 \\
    + \textit{SNG$_{4}$} & \bf{53.6} & 45.7 & 44.4 & \bf{40.3} & 44.9 & 47.7 & 36.8 & \bf{49.1} & 39.0 & \bf{52.7} & \bf{45.4} \\
    + \textit{SNG$_{5}$} & 53.0 & 45.6 & 43.9 & 39.7 & 45.4 & 46.7 & 36.8 & 49.1 & 38.4 & 52.5 & 45.1 \\
    + \textit{ENT$^{N}$} & 50.9 & 43.7 & 38.1 & 37.2 & 42.5 & 44.5 & 34.7 & 46.7 & 36.0 & 49.9 & 42.4 \\
    + \textit{ENT$^{K}$} & 52.7 & 45.7 & 43.5 & 40.2 & 44.6 & 45.2 & 36.4 & 49.0 & 39.1 & 51.8 & 44.8 \\
    + \textit{AGG$^{L}_{5}$} & 47.1 & 41.5 & 39.8 & 34.0 & 39.9 & 42.1 & 31.4 & 43.5 & 33.7 & 45.2 & 39.8 \\
    + \textit{AGG$^{F}_{5}$} & 45.0 & 38.4 & 38.5 & 32.4 & 38.8 & 47.1 & 30.4 & 41.2 & 33.3 & 44.2 & 38.9 \\
    + \textit{AGG$^{P}_{5}$} & 45.5 & 38.8 & 38.0 & 32.0 & 38.8 & \bf{48.2} & 30.5 & 41.0 & 33.2 & 44.0 & 39.0 \\
    + \textit{AGG$^{N}_{5}$} & 45.4 & 39.1 & 38.3 & 32.4 & 38.8 & 48.0 & 30.7 & 41.2 & 33.2 & 44.3 & 39.1 \\
    \bottomrule
  \end{tabularx}
  \caption{140 experiments comparing 14 active learning methods translating into 10 different languages with Schedule \textit{B}.}
\label{table:14by10}
\end{table*}

For Eastern Pokomchi, though the performance
of the most difficult 11 books never reach
BLEU score of 20s like that of English experiments,
all books have BLEU scores that are steadily
increasing. Challenges remain for Eastern Pokomchi,
a Resource 0 language \citep{joshi2020state}. We hope to
work with native Mayan speakers to see ways we may improve the
results. 

\subsection{N-gram, Entropy, and Aggregation Methods}

In addition to active learning, we compare 14 active learning methods across 10 different target
languages (Table~\ref{table:14by10}). 

\textbf{\ul{Normalizing by sequence length improves density:}} Without normalization, the model chooses longer sentences with many rare words. Normalization improves density. For Sesotho, the chrF score is 39.0 without normalization and 41.6 with it. 

\textbf{\ul{Marginal benefit of increasing n-gram order wanes:}} Existing research shows bigrams suffice \citep{eck2008developing}. As the n-gram order increases, the data gets sparser and the marginal benefit 
subsides. Hmong has the best score (46.1) using bigrams.

\textbf{\ul{Tipping points vary with language:}} The optimal highest n-gram order may differ from language to language. 4-grams work best for Frisian while bigrams work best for Hmong. Hmong is an isolating language while Frisian is a fusional language. A possible explanation is that higher n-grams may have more impact on fusional languages.

\textbf{\ul{Entropy and n-gram methods both beat baselines and higher n-gram models perform best:}} KenLM is much faster and performs better than NLTK. The entropy method using KenLM beats both baselines. Frisian has a chrF score of 52.7 with the entropy method using KenLM. This is much higher than the baselines: \textit{Luke} (47.5) and \textit{Rand} (50.5). The 4-gram model (53.6) is higher because building LMs from a few lines of data may not be accurate. Simpler n-gram models work better than more evolved entropy models with small data. 

\textbf{\ul{Aggregation over all languages serves as a universal ranking:}}  The first 10 active learning methods are based on learning from one reference language and generalizing to the low-resource language, while the last 4 focus on aggregation over multiple languages (Table~\ref{table:14by10}).  
For Welsh, aggregation over multiple languages (48.2 with most spoken languages) performs better than those that rely on one reference language; but for other languages aggregation performs worse. Aggregation over all languages performs better than other aggregation methods for all languages except Welsh. 
This hinges on the reference language. For Frisian, choosing English (a Germanic language) as a reference language, performs better than aggregation. For Welsh (a Celtic language), choosing a reference language that is not as close, performs worse. But we often do not have such information for the low-resource languages. In such cases, universal ranking by aggregating over all languages is useful. 


\textbf{\ul{We recommend n-gram methods as the most efficient in computational cost:}} 
Among all of our active learning methods, random sampling does not require much computation, and therefore costs the least in terms of computational time. However, random sampling does not perform as well as n-gram models. Comparing n-gram models with entropy models and aggregation models, n-gram models is the most effective, costing the least in terms of computational time, and performs the best. For example, a 4-gram active learning model takes around 38 minutes to train, while an entropy model typically takes 14 hours to train. Since aggregation methods by design take multiple reference languages, they take longer time than than any models based on a single reference language. This is the reason we highly recommend n-gram methods as the most efficient and the least expensive. 

\textbf{\ul{Our active learning methods mimic curriculum learning:}} Our models pick short and simple sentences first, 
emulating curriculum learning and helping human translators \citep{bengio2009curriculum, graves2017automated, jiang2015self}. 

\textbf{\ul{All active learning methods cover different genres:}} Our methods pick a mix of sentences from different genres, sentence
lengths and complexity levels. Moreover, our methods pick narrative sentences first, which is helpful for human translators.


\textbf{\ul{Our model captures some language subtleties:}} Apart from the metrics, we showed our translation to native speakers (Table \ref{table:quali_srclanguages}). We translate ``He sees that it is good'' to ``Lug ca rua huv nwg lu sab''(``He puts it in the liver'') in Hmong, which uses liver to express joy. This increases lexical choice. 

\section{Conclusion}\label{conclusion}
To serve our main goal of minimizing human translation and post-editing efforts, our contribution in this chapter is that we minimize the seed corpus to be $\sim$3\% of the text. Indeed, we use $\sim$3\% of the text to translate the $\sim$97\% of the text in the low-resource language we want to translate to. Having minimized the training data to be $\sim$3\% of the text in the given low-resource language, we optimize this process with active learning on which $\sim$3\% of the text to translate first to produce the seed corpus. To determine which $\sim$3\% of the text, we show 
results from random sampling, as well as n-gram, entropy, and 
aggregation methods in addition to the portion-based approach to build seed corpus without any low-resource language data. Of all these methods, we find that the n-gram method (in particular, the 4-gram method) is sufficient for producing high translation performance when we are given complete information of languages close to 
the given low-resource language. However, when we are not given complete information of the close-by languages, we recommend the aggregation method proposed in this chapter as a universal ranking to use. This minimizes human translation efforts in the production of the seed corpus. Additionally, we also compare three different ways
of updating the machine translation models by
adding newly post-edited data iteratively.
We find that vocabulary update is necessary, but 
self-supervision by pretraining with whole translation draft 
is best to be avoided. 

However, we still face challenges with the lack of local coherence and context.
The excerpt-based approach
enjoys advantages with formality, cohesion and contextual
relevance. Active learning methods, on the contrary,
do not have consecutive sentences and therefore lose local coherence and pose challenges to human
translators \citep{muntes2012context, denkowski2015machine, sperber2017transcribing, maruf2019survey, webster2020gutenberg, zhou2021active, salunkhe2016hybrid}. 
Human translators may not be receptive to active learning methods like random sampling. Additionally, 
it may take longer for human translators to translate a set of non-consecutive sentences, because each sentence may be from a different chapter with a different context and it is harder for human translators to recycle translations. 
Indeed, the lack of local coherence and global context is a real challenge 
when human translators work with machine translation systems. 
Moreover, improving local coherence is important and is also an active research area.

One limitation of our work is that in real life scenarios, we do not have the reference
text in low-resource languages to 
produce the BLEU scores to decide the 
post-editing order. Consequently, field linguists
need to skim through and decide the post-editing order 
based on intuition. However, computational models
can still help. One potential way to tackle it 
is that we can train on $\sim$1,000 lines from
another language with available text and
test on the 66 books. Since our results show that
the literary genre plays important role in the performance ranking,
it would be reasonable to determine the order
using a ``held-out language'' and then using that
to determine order in the target low-resource language. 
In the future, we
would like to work with human translators who understand and speak 
low-resource languages. 

Another concern human translators may have is the creation
of randomly sampled seed corpora. To gauge the amount of interest
or inertia, we have interviewed some
human translators and many are interested. However,
it is unclear whether human translation quality of randomly sampled data
differs from that of the traditional portion-based approach. 
We hope to work with human translators closely to determine whether the translation 
quality difference is manageable. 

We are also curious how our model will perform with
large literary works like ``Lord of the Rings'' and ``Les Misérables''. 
We would like to see whether it will translate well
with philosophical depth and literary complexity. However,
these books often have copyright
issues and are not as easily available as the Bible data.
We are interested in collaboration with teams who have multilingual
data for large texts, especially multilingual COVID-19 data. 

Having examined various active learning methods in building seed corpora that optimize 
machine translation, we have established the first step of the human machine translation 
workflow as show in Figure \ref{fig:workflow}. In the next chapter,  we will focus on the next few 
steps in the workflow by focusing on how to use large multilingual models to most effectively train 
on such a small seed corpus in the low-resource language. 

\begin{table*}[t]
  \scriptsize
  \centering
  \begin{tabularx}{\textwidth}{p{1.1cm}p{6.6cm}p{7.8cm}} 
    \toprule
    Target & System Translation & Reference \\
    \midrule    
    Frisian & mar Ruth sei: Ik scil dy net forlitte, en ik scil fen dy net weromkomme; hwent hwer ``tstû hinnegeane, den scil ik hinnegean, en dêr scil ik dy fornachtsje. dyn folk is myn folk, en dyn God is myn God. & mar Ruth sei: Sit net tsjin my oan, dat ik jo forlitte en weromtsjen scil; hwent hwer ``t jo hinne geane, dêr scil ik hinne gean, en hwer ``t jo fornachtsje, dêr scil ik fornachtsje; jins folk is myn folk en jins God is myn God;\\
    Hmong & Lauj has rua nwg tas, ``Tsw xob ua le ntawd, kuv yuav moog rua koj lub chaw kws koj moog, hab kuv yuav nyob huv koj haiv tuabneeg. koj yog kuv tug Vaajtswv.'' & tassws Luv has tas, ``Tsw xob has kuas kuv tso koj tseg ncaim koj rov qaab moog. koj moog hovtwg los kuv yuav moog hab, koj nyob hovtwg los kuv yuav nyob hov ntawd hab, koj haiv tuabneeg los yog kuv haiv tuabneeg hab, koj tug Vaajtswv los yog kuv tug Vaajtswv. \\
    Pokomchi & eh je' wili i xq'orarik reh i Rut: Maacanaa' chih taj i hin. re' hin naa nub'anam aweh chupaam i ye'aab' naa nuk'achariik ayu'. re' hin naa nuk'achariik awuuk', eh re' hin naa nukahniik chi nuDios, inki. & re' Rut je' wili i chaq'wik xub'an: Maa pahqaaj aakuyariik weh re' hin ma' jaruuj nee tinukanaa' kahnoq, xa aha' pa' nee tiooj i hat, nee wo' kinooj chawiij, xa aha' pa' nee ti k'achariik i hat ar nee kink'acharik i hin. eh re' aatinamiit re' wo' re' nutinamiit i hin, eh re' aaDios re' wo' re' nuDios i hin. \\
    Turkmen & Rut: oňa: ``Sen nirä gitseň, men hem seniň ýanyňa gitmerin. Sen nirä gitseň, men hem seniň halkym bolaryn. Men seniň Hudaýym bolaryn. & emma Rut: ``Seni terk edip ýanyňdan gitmegi menden haýyş etme. sen Nirä gitseň, Menem şol ýere gitjek. sen nirede bolsaň, Menem şol ýerde boljak. seniň halkyň - meniň halkym, seniň Hudaýyň meniň Hudaýym bolar. \\
    Sesotho & yaba Ruthe o re ho yena: ``O se ke wa tloha ho wena, hobane ke tla ya le wena, ke tla ya le wena, mme ke tla ya hona moo. setjhaba sa ka, le Modimo wa hao.'' & empa Ruthe a re: ``O se ke wa nqobella hore ke kgaohane le wena, kapa hore ke se ke ka tsamaya le wena, hobane'' moo o yang teng ke tla ya teng, moo o phelang teng ke tla phela teng; tjhaba sa heno e be tjhaba sa heso, Modimo wa hao e be Modimo wa ka. \\   
    Welsh & a Ruth a ddywedodd, Nuw gael arnaf fi, atolwg, atolwg, oddi wrthyt: canys lle yr wyt yn myned, ac yno yr wyt yn myned, y byddaf fy hun. dy bobl yw fy bobl, a'th Dduw yw fy Duw. & a Ruth a ddywedodd, Nac erfyn arnaf fi ymado â thi, i gilio oddi ar dy ôl di: canys pa le bynnag yr elych di, yr af finnau; ac ym mha le bynnag y lletyech di, y lletyaf finnau: dy bobl di fydd fy mhobl i, a'th Dduw di fy Nuw innau:\\
    Xhosa & URute waphendula wathi: ``Undiyekeli ukuba ndixhamle, kuba ndiza kuhlala apho uthanda khona. mna ndiza kuba ngabantu bam, abe nguThixo wam.'' & Waphendula uRute wathi: ``Sukundinyanzela usithi mandikushiye. apho uya khona, nam ndiya kuya, ndiye kuhlala nalapho uhlala khona, amawenu abe ngamawethu, noThixo wakho abe nguThixo wam. \\
    Indonesian & tetapi Rut: menjawab: ``Janganlah engkau meninggalkan aku dan pulang ke tempat kediamanmu, sebab aku akan pergi dan berdiam di mana engkau diam, sebab orang-orangmu akan menjadi umat-Ku dan Allahmu.'' & tetapi kata Rut: ``Janganlah desak aku meninggalkan engkau dan pulang dengan tidak mengikuti engkau; sebab ke mana engkau pergi, ke situ jugalah aku pergi, dan di mana engkau bermalam, di situ jugalah aku bermalam: bangsamulah bangsaku dan Allahmulah Allahku; \\
    Hungarian & Ruth így felelt: Nem kérlek téged, hogy gondolj meg téged, mert csak hozzád megyek, és én otthagytam, hogy legyenek hozzád. a te népem az én, és az én Istenem az én. & de Ruth azt felelte: Ne unszolj engem, hogy elhagyjalak és visszatérjek tőled. mert ahová te mégy, odamegyek, ahol te megszállsz, ott szállok meg. Néped az én népem, és Istened az én Istenem. \\
    Spanish & y Rut: dijo a David: No me permite de ti, y me quitaré de ti; porque donde vayas, yo iré a donde vayas, y habitaré; y tu pueblo es mi pueblo, y tu Dios es mi Dios. & respondió Rut: No me ruegues que te deje, y me aparte de ti; porque a dondequiera que tú fueres, iré yo, y dondequiera que vivieres, viviré. tu pueblo será mi pueblo, y tu Dios mi Dios. \\
    \bottomrule
  \end{tabularx}
    \caption{Qualitative evaluation using \textit{SNG$_{5}$} 
    to translate into each target language. }
\label{table:quali_srclanguages}
\end{table*}
\removelabelprefix

\chapter{Optimizing with Large Pretrained Models}\label{big:large}
\addlabelprefix{6}
\epigraph{``The deepest connection you have with someone and their culture, is through learning their language.''}{\textit{Marisa J Taylor}}

\lettrine{H}{aving examined various active learning methods} in building a seed corpus that optimizes machine translation, we are interested in increasing effectiveness of how to best train on such a small 
seed corpus. 
To serve our main goal of minimizing human translation and post-editing efforts, having minimized the seed corpus to $\sim$3\% of the text, and optimized which $\sim$3\% of the text is to be translated first, we are interested in how to best train on such a small sample of the text and produce the best translation performance to minimize post-editing efforts. In this chapter, we will show how to most effectively train using large pretrained models to minimize human post-editing efforts. We find it most useful to adapt the large pretrained models to the domain first, and then to the target low-resource language. 

\section{Introduction}

A language dies when no one speaks it. A language needs attention when 
it is spoken by enough people that it could survive under favorable
conditions but few or no children are learning it
\citep{crystal2002language, kincade1991decline, wurm2001atlas}. 
More than half of the ~7,139 languages will die in the next 80 years \citep{austin2011cambridge, eberhard2021ethnologue}. 
Many of these languages are low-resource. 
These languages may survive and thrive if they gain prestige, power and visibility \citep{crystal2002language}. 
Frisian, for example, struggles to gain prestige in Germany, 
and is low-resource even though it has a large number of speakers. 
Hebrew, conversely, has been revived as a spoken language because 
it is critical to the development and identity of the Jewish community.  
We empower these language communities by exercising a language\footnote{The material in this chapter was originally 
published in LoResMT at ACL, 2023 \citep{zhou2023train}.}. 
This can be achieved by translating important texts to their language so that 
these communities can gain information, knowledge, power and visibility in 
their own language. 

\begin{figure*}[t]
  \centering
  \includegraphics[width=0.8\linewidth]{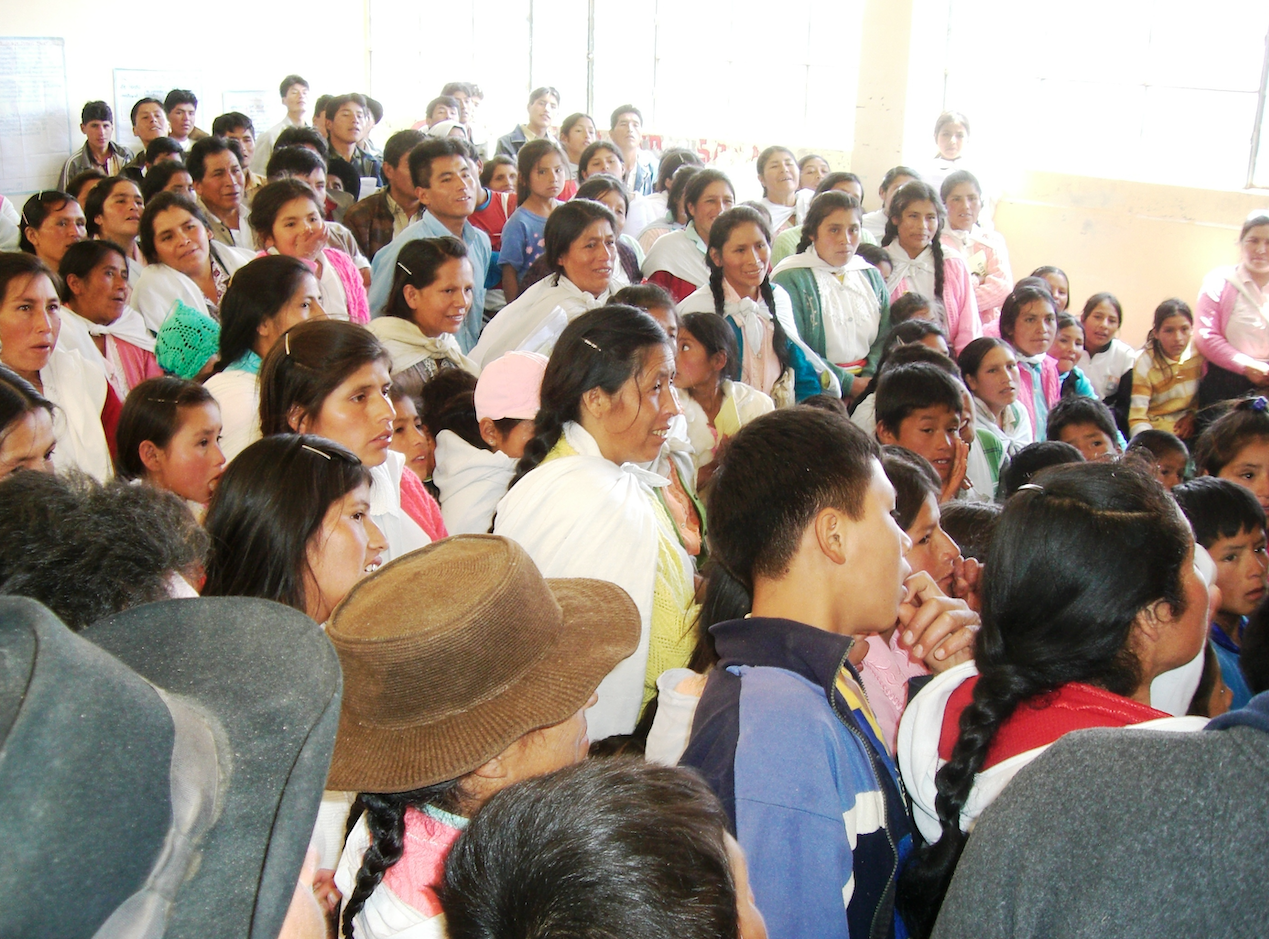}
  \caption{A community in Peru that speaks Panao Quechua. Photograph provided by Mark Bean.} 
  \label{fig:peru_read}  
\end{figure*}

\begin{figure}
  \centering
  \includegraphics[width=0.9\linewidth]{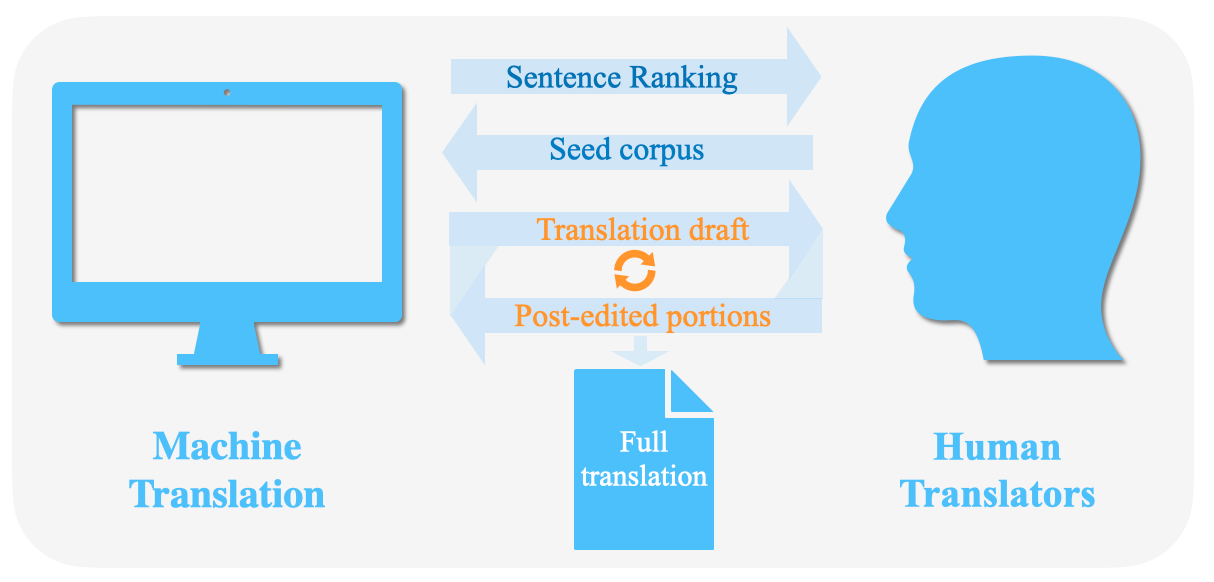}
  \caption{Translation workflow for low-resource languages, focusing on training on the seed corpus followed by iterations of post-editing and updated training. }
  \label{fig:workflow}
\end{figure}

The problem in these scenarios, therefore, is not to build a high accuracy translation engine for \textit{any texts} using huge data corpora, but rather to build a good translation for a \textit{known} text (for which translations in many other languages exist), but in a new language with only extremely little seed data (a few hundred sentences).  We assume there is little to no low-resource language data and few human translators. 
To produce high quality translation, existing methods rely on a seed corpus produced by human translators. Previous work has shown progress in using extremely small seed corpora with as small as $\sim$1,000 lines of data and has found that active learning performs better than choosing a fixed portion of the text to build a seed corpus \citep{zhou2021family, lin2020pre, qi2018and}. 
However, researchers have yet to completely solve the problem of using large multilingual models for representational learning to train (or adapt) them to a new, low-resource language by training on an extremely small seed corpus.

To solve this problem, we examine different training schedules and we find a strategic way of growing large multilingual models in a multilingual and multi-stage fashion with extremely small low-resource seed corpora. 

In our translation workflow, human translators are informed by machine sentence ranking to produce a seed corpus. To curate a seed corpus in the new, low-resource language where we have no data initially, we pass the sentence ranking learned from known languages to human translators. Human translators take this ranking, and translate the top few ($\sim$1,000 or less) sentences, curating the seed corpus. 

Using the seed corpus created by active learning, machine systems then train on the seed corpus to produce a full translation draft. Human translators post-edit the draft, and feed new data to machines each time they finish post-editing a portion of the text. In each iteration, machines produce better and better drafts with new data, and human translators find it easier and faster to post-edit. Together they complete the translation of the whole text into an low-resource language (Figure~\ref{fig:workflow}).  

To train on such small seed corpus, we find pretraining to be key. For the pretrained model, we either create our own pretrained model by training on known languages, or use an existing pretrained model. We explore both paths in our work, with and without activating the knowledge in existing large pretrained models. We observe an average increase of 28.8 in chrF score over the baselines. 

Our contribution is therefore: we activate the knowledge of large multilingual models by proposing multilingual and multi-stage adaptations through 24 different training schedules; we find that adapting pretrained models to the domain and then to the low-resource language works best.  
Our model can stand on its own and can also be boosted by large multilingual models. Our model 
works on many languages spanning different resource levels.

\section{Methods}
We translate a fixed text that is available in many languages to a new, low-resource language. In our translation workflow, we first develop active learning methods to transfer sentence ranking from known languages to a new, low-resource language. We then pass this ranking to human translators for them to translate the top few ($\sim$1,000 or less) sentences into the low-resource language, curating the seed corpus. We finally train on the seed corpus, either from scratch or from a pretrained model. 

We build training schedules on an extremely small seed corpus. We propose and compare 24 training schedules for machine translation into a new, low-resource language. To compare all experiments fairly, we use the same translation system unit as a control for all experiments, varying only the seed corpora built by different methods. We select the same number of words in all seed corpora as most translators are paid by the number of words they translate 
\citep{bloodgood2014bucking, eck2008developing, tomanek2009semi}.

\begin{figure}
  \centering
  \includegraphics[width=0.6\linewidth]{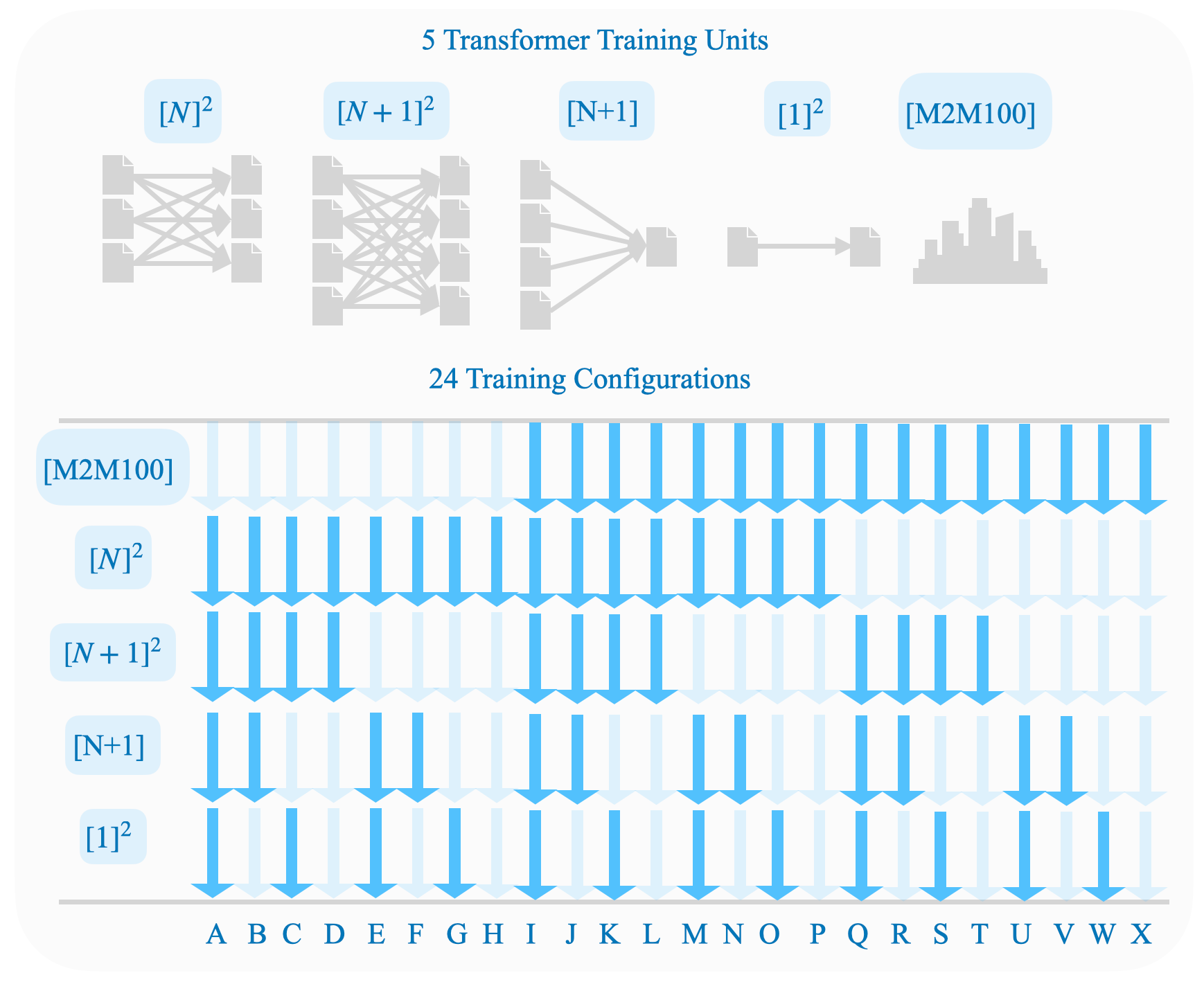}
  \caption{24 different training schedules.\\
  {[N]}: multilingual model on N neighboring languages\\
  {[N+1]}$^2$: multi-target model with low-resource language \\
  {[N+1]}: single-target model with low-resource language\\
  {[1]}$^2$: autoencoder in low-resource language.}
  \label{fig:config13}
\end{figure}

\subsection{Training Schedules}
In our setup we have the new, low-resource language as the target language, and we have a few neighboring languages as the source languages that are either in the same linguistic language family or geographically close to facilitate linguistic transfer. In effect, we have $N$ source languages with full translations of the text and a new and low-resource language that has an extremely small seed corpus, which could be a few hundred lines of data. 

We use the state-of-the-art multilingual transformer prepending both source and target language labels to each source sentence \citep{johnson2017google, ha2016toward}. For precise translation for all named entities, we use an existing method of \textit{order-preserving named entity translation} by masking each named entity with ordered \texttt{\_\_NE}s using a parallel multilingual lexicon table in 125 languages \citep{zhou2021family, wu2018creating}. 

Using this multilingual transformer architecture as a base, we build 5 training units on the small seed corpus of the new, low-resource language and the existing translations of known languages. 
We let {[N]}$^2$ denote the training of all source languages in a N-by-N multilingual transformer. We let {[N+1]}$^2$ denote the training of all languages including the low-resource language in a (N+1)-by-(N+1) multilingual transformer. We let {[N+1]} denote the (N+1)-by-1 multilingual transformer that focuses on translating into the low-resource language. We let {[1]}$^2$ be the autoencoder on the low-resource language. 

Our translation system is built on these 5 training units: an optional {[M2M100]} \citep{fan2021beyond}, {[N]}$^2$, {[N+1]}$^2$, {[N+1]} and {[1]}$^2$. 
These 5 stages increase in specificity while they decrease in data size. Building on them, we show 24 different training schedules, among which 8 are pretrained with in-domain data and 16 are pretrained with out-of-domain large multilingual models (Figure~\ref{fig:config13}). We only consider models with pretraining and therefore do not exhaust all 32 training schedules. 

\subsection{Active Learning Strategies} \label{2023methods} 
We have two baselines: the linguistic baseline of the excerpt-based approach, \textit{Luke}, and the statistical baseline of random sampling, \textit{Rand}. The excerpt-based approach, which selects a portion of the text with consecutive sentences, preserves the text's formality, cohesion and context but lacks global coverage. Random sampling increases global coverage but sacrifices local coherence.

In addition to random sampling, we also explore n-gram, entropy and aggregation methods as introduced in the previous chapter. 
\begin{table*}[t]
  \small
  \centering
  \begin{tabularx}{\textwidth}{p{2cm}p{0.5cm}p{2.5cm}p{9.5cm}}
    \toprule 
    Target & L & Family & Source Languages \\
    \midrule
    Frisian & 0 & Germanic & English*, German, Dutch, Norwegian, Afrikaans, Swedish, French, Italian, Portuguese, Romanian \\  
    Hmong & 0 & Hmong–Mien & Komrem*, Vietnamese, Thai, Chinese, Myanmar, Haka, Tangsa, Zokam, Siyin, Falam \\
    Pokomchi & 0 & Mayan & Chuj*, Cakchiquel, Mam, Kanjobal, Cuzco, Ayacucho, Bolivian, Huallaga, Aymara, Guajajara \\
    Turkmen & 1 & Turkic & Kyrgyz*, Tuvan, Uzbek, Karakalpak, Kazakh, Azerbaijani, Japanese, Korean, Finnish, Hungarian \\
    Sesotho & 1 & Niger–Congo & Yoruba*, Gikuyu, Xhosa, Kuanyama, Kpelle, Fon, Bulu, Swati, Venda, Lenje \\
    Welsh & 1 & Celtic & English*, German, Danish, Dutch, Norwegian, Swedish, French, Italian, Portuguese, Romanian \\
    Xhosa & 2 & Nguni & Swati*, Gikuyu, Sesotho, Yoruba, Lenje, Gbaya, Afrikaans, Wolaitta, Kuanyama, Bulu \\
    Indonesian & 3 & Austronesian & Javanese*, Malagsy, Tagalog, Ilokano, Cebuano, Fijian, Sunda, Zokam, Wa, Maori \\	
    Hungarian & 4 & Uralic & Finnish*, French, English, German, Latin, Romanian, Swedish, Spanish, Italian, Portuguese \\
    Spanish & 5 & Romance & English*, German, Danish, Dutch, Norwegian, Swedish, French, Italian, Portuguese, Romanian \\
    \bottomrule
  \end{tabularx}
    \caption{Summary of different target languages used \citep{campbell2018cataloguing, collin2010ethnologue}.
    L, resource level, is from a scale of 0 to 5 \citep{joshi2020state}.
    Reference languages used for active learning methods except aggregate methods are starred. }
\label{table:ethnologue_languages}
\end{table*}

\begin{table*}[t]
  \footnotesize
  \centering
  \begin{tabularx}{\textwidth}{p{1.5cm}p{0.65cm}p{0.75cm}p{1.1cm}p{0.95cm}p{0.8cm}p{0.56cm}p{0.56cm}p{1.25cm}p{1.25cm}p{0.8cm}p{0.8cm}} 
   \toprule
   $\uparrow$chrF & Frisian & Hmong & Pokomchi & Turkmen & Sesotho & Welsh & Xhosa & Indonesian & Hungarian & Spanish & Average \\
    \midrule
    \multicolumn{12}{l}{\textbf{Baselines:} }\\
    + Bilingual & 23.1 & 25.0 & 28.7 & 18.9 & 25.2 & 22.2 & 21.4 & 27.2 & 20.1 & 22.1 & 23.4 \\
    + Multilingual & 28.0 & 28.1 & 31.9 & 22.6 & 28.3 & 26.5 & 23.9 & 29.7 & 22.3 & 26.8 & 26.8 \\
    \midrule
    \multicolumn{12}{l}{\textbf{Our Models:} }\\
    + Schedule \textit{B} & 50.5 & 43.9 & 42.8 & 38.9 & 43.2 & 46.0 & 34.9 & 47.2 & 37.4 & 50.1 & 43.5 \\
    + Active (AL) & 53.6 & 45.7 & 44.4 & 40.3 & 44.9 & 47.7 & 36.8 & 49.1 & 39.0 & 52.7 & 45.4 \\
    \bottomrule
  \end{tabularx}
  \caption{Results for translation into 10 languages that are 
  new and severely low-resource to the system, independent of M2M100. }
\label{table:4by10}
\end{table*}

\begin{table}[t]
  \footnotesize
  \centering
  \begin{tabularx}{\columnwidth}{p{3cm}p{2cm}p{2cm}p{2.5cm}p{2cm}p{2cm}} 
   \toprule
   $\uparrow$chrF & Frisian & Welsh & Hungarian & Spanish  & Average \\
    \midrule
    \multicolumn{6}{l}{\textbf{Baselines:} }\\
    + Bilingual & 23.1 & 22.2 & 20.1 & 22.1 & 21.9 \\
    + Multilingual & 28.0 & 26.5 & 22.3 & 26.8 & 25.9 \\
    + M2M100 & 26.0 & 9.9 & 38.8 & 47.5 & 24.9\\ 
    \midrule
    \multicolumn{6}{l}{\textbf{Our Models:} }\\
    + Schedule \textit{I} & 53.5 & 49.5 & 42.2 & 53.2 & 49.6 \\ 
    + Active (AL)  & 54.9 & 49.8 & 43.2 & 54.9 & 50.7 \\
    \bottomrule
  \end{tabularx}
  \caption{Results for translation into 4 languages that are 
  new and severely low-resource to the system, 
  activating knowledge in M2M100 and leveraging active learning.    
  }
\label{table:5by10}
\end{table}

\begin{table*}[t]
  \footnotesize
  \centering
  \begin{tabularx}{\textwidth}{p{1.2cm}XXXXXXXXXXXXXXXX} 
    \toprule
     Network & I & J & K & L & M & N & O & P & Q & R & S & T & U & V & W & X \\
    \midrule
    {[M2M100]} & $\Downarrow$ & $\Downarrow$ & $\Downarrow$ & $\Downarrow$  & $\Downarrow$ & $\Downarrow$ & $\Downarrow$ & $\Downarrow$  & $\Downarrow$ & $\Downarrow$ & $\Downarrow$ & $\Downarrow$  & $\Downarrow$ & $\Downarrow$ & $\Downarrow$ & $\Downarrow$ \\
    {[N]}$^2$  & $\Downarrow$ & $\Downarrow$ & $\Downarrow$ & $\Downarrow$  & $\Downarrow$ & $\Downarrow$ & $\Downarrow$ & $\Downarrow$  &  &  &  & &  &  &  \\
    {[N+1]}$^2$  & $\Downarrow$ & $\Downarrow$ & $\Downarrow$ & $\Downarrow$ &  &  &  &   & $\Downarrow$ & $\Downarrow$ & $\Downarrow$ & $\Downarrow$ &  &  &  \\ 
    {[N+1]} & $\Downarrow$ & $\Downarrow$ & & & $\Downarrow$ & $\Downarrow$ & & & $\Downarrow$ & $\Downarrow$ & & & $\Downarrow$ & $\Downarrow$ &\\
    {[1]}$^2$  & $\Downarrow$ & & $\Downarrow$ & & $\Downarrow$ & &  $\Downarrow$ & & $\Downarrow$ & & $\Downarrow$ & & $\Downarrow$ & & $\Downarrow$ &\\
    \midrule
    $\uparrow$chrF & \bf{52.9} & \bf{51.8} & 49.5 & \bf{52.8} & \bf{52.7} & \bf{51.9} & 27.4 & 16.9 & 49.6 & 48.5 & 39.6 & 48.7 & 48.5 & 45.7 & 27.8 & 26.3 \\
    $\downarrow$cTER & 0.492 & 0.508 & 0.482 & 0.488 & 0.493 & 0.502 & 0.654 & 0.800 & 0.530 & 0.546 & 0.553 & 0.539 & 0.538 & 0.579 & 0.650 & 0.667 \\
    $\uparrow$BLEU & 28.8 & 27.9 & 24.2 & 28.9 & 28.8 & 28.2 & 3.0 & 0.6 & 24.8 & 24.2 & 13.9 & 24.3 & 24.5 & 22.0 & 3.4 & 3.3 \\ 
    $\uparrow$COMET & \text{-}0.56 & \text{-}0.59 & \text{-}0.63 & \text{-}0.53 & \text{-}0.56  & \text{-}0.57 & \text{-}1.28 & \text{-}1.75 & \text{-}0.67  & \text{-}0.70 & \text{-}0.89 & \text{-}0.68 & \text{-}0.69  & \text{-}0.80 & \text{-}1.21 & \text{-}1.30 \\
    $\uparrow$BERTS &  0.891 & 0.889 & 0.886 & 0.892 & 0.891 & 0.890 & 0.813 & 0.775 & 0.883 & 0.881 & 0.861 & 0.882 & 0.880 & 0.873 & 0.823 & 0.819 \\
    \bottomrule
\end{tabularx}
    \caption{Comparing 16 training schedules with M2M100. BERTS is BERTScore, cTER is characTER and LRatio is length ratio.  
    }
\label{table:config2}
\end{table*}

\begin{table}[t]
  \footnotesize
  \centering
  \begin{tabularx}{\columnwidth}{p{1.5cm}XXXXXXXX} 
    \toprule
    Network & A & B & C & D & E & F & G & H \\
    \midrule
    {[N]}$^2$  & $\Downarrow$ & $\Downarrow$ & $\Downarrow$ & $\Downarrow$  & $\Downarrow$ & $\Downarrow$ & $\Downarrow$ & $\Downarrow$ \\
    {[N+1]}$^2$  & $\Downarrow$ & $\Downarrow$ & $\Downarrow$ & $\Downarrow$ &  &  &   \\ 
    {[N+1]} & $\Downarrow$ & $\Downarrow$ & & & $\Downarrow$ & $\Downarrow$ & \\
    {[1]} $^2$  & $\Downarrow$ &  & $\Downarrow$ &  & $\Downarrow$ & &  $\Downarrow$ &  \\ 
    \midrule
    $\uparrow$chrF & 38.7 & \bf{51.1} & 35.6 & 50.8 & 43.4 & \bf{51.2} & 25.6 & 24.1 \\
    $\downarrow$cTER & 0.555 & 0.517 & 0.572 & 0.515 & 0.523 & 0.507 & 0.650 & 0.682  \\
    $\uparrow$BLEU & 12.5 & 24.9 & 9.2 & 24.5 & 17.5 & 26.2 & 2.5 & 2.1 \\ 
    $\uparrow$COMET & \text{-}0.87 & \text{-}0.66 & \text{-}0.91 & \text{-}0.65 & \text{-}0.81 & \text{-}0.63 & \text{-}0.99 & \text{-}1.02 \\
    $\uparrow$BERTS & 0.850 & 0.882 & 0.839 & 0.884 & 0.865 & 0.885 & 0.801 & 0.794 \\
    \bottomrule
\end{tabularx}
    \caption{Comparing 8 training schedules without M2M100. \\
     {[N]}$^2$: multilingual model on N neighboring languages\\
     {[N+1]}$^2$ : multi-target model with low-resource language \\
     {[N+1]}: single-target model with low-resource language\\
     {[1]}$^2$: autoencoder in low-resource language.  
    }
\label{table:config1}
\end{table}

\subsection{Evaluation Method and Metrics}
Existing multilingual systems produce multiple outputs from all source languages, rendering comparison messy. To simplify, we combine translations from all source languages into one by an existing \textit{centeredness method} \citep{zhou2021family}. Using this method, we score each translated sentence by the sum of its similarity scores to all others. We rank these scores and take the highest score as our combined score. The expected value of the combined score is higher than that of each source. 

To compare effectively, we control all test sets to be the same. Since different experiments use different seed corpora as training and validation sets, the training and validation sets vary. Their complement, the test sets therefore also vary, rendering comparison difficult. To build the same test set, we devise an \textit{intersection method}. We take the whole text and carve out all seed corpora, that is, all training and validation sets from all experiments. The remaining is the final test set, which is the intersection of all test sets across all experiments. 

Our metrics are: chrF, characTER, BLEU, COMET score, and BERTscore \citep{popovic2015chrf, wang2016character, post-2018-call, zhang2019bertscore, stewart-etal-2020-comet, rei2021mt}. We prioritize chrF over BLEU for better accuracy, fluency and expressive power in morphologically-rich languages \citep{papineni2002bleu}.

\section{Data}
Existing research classifies world languages into Resource 0 to 5, with 0 having the lowest resource and 5 having the highest \citep{joshi2020state}. We choose 10 target languages ranging from Resource 0 to 5 (Table~\ref{table:ethnologue_languages}). For each target language we choose ten neighboring languages as source
languages (Table~\ref{table:ethnologue_languages}). These languages\footnote{For simplicity, in Table \ref{table:ethnologue_languages} 
Pokomchi is Eastern Pokomchi, Hmong is Hmong Hoa, 
Kanjobal is Eastern Kanjobal, Mam is Northern Mam, 
Cuzco is Cuzco Quechua, Ayacucho is Ayacucho Quechua, 
Bolivian is South Bolivian Quechua, and Huallaga is
Huallaga Quechua, Chinese is Traditional Chinese, 
Haka is Haka Chin, Siyin is Siyin Chin, Falam is 
Falam Chin, Kpelle is Kpelle Guinea. } are Eastern Pokomchi, Hmong, and Frisian (Resource 0), Turkmen, Welsh and Sesotho (Resource 1), Xhosa (Resource 2), Indonesian (Resource 3), Hungarian (Resource 4), Chinese and Spanish (Resource 5) in Table~\ref{table:ethnologue_languages}.
We prioritize Resource 0 to 2 languages as real low-resource languages, and we use Resource 3 to 5 languages as hypothetical ones to show the spectrum.
It is surprising to us that a lot of the Resource 0 languages are not too far away from the rich-resource languages. Frisian, for example, are spoken near the Northern Sea near Netherlands and Germany, and is in close proximity with a few rich-resource European languages \cite{markey2011frisian}. However, because of the close proximity with rich-resource languages, low-resource languages like Frisian often suffer from lack of prestige and has a bigger threat to extinction as many younger people choose to speak the rich-resource languages nearby. This also suggests interesting research direction on low-resource languages and dialects that are in close proximity with rich-resource language communities.  

Among these ten languages, Frisian, Welsh, Hungarian and Spanish are on the list of languages trained in M2M100 \citep{fan2021beyond}. We apply a multi-stage adaptation of large pretrained models to these four languages. During training, we use the large sentence-piece vocabulary from M2M100 which contains 128k unique tokens for optimized performance. 

To translate into these languages, our text is the Bible in 125 languages \citep{mayer2014creating}. Each low-resource seed corpus contains $\sim$3\% of the text, while all other languages have full text. Our goal is to translate the rest of the text into the low-resource language. In pretraining, we use a 80/10/10 split for training, validation and testing, respectively. In training, we use approximately a 3.0/0.2/96.8 split for training, validation and testing, respectively. Our training data for each experiment is $\sim$1,000 lines. We use BPE with size of $\sim$3,000 for the low-resource language and $\sim$9,000 for the combined \citep{sennrich2016neural}. 

\begin{table*}[t]
  \scriptsize
  \centering
  \begin{tabularx}{\textwidth}{p{1.5cm}p{0.6cm}p{0.6cm}p{1.1cm}p{0.9cm}p{0.8cm}p{0.5cm}p{0.6cm}p{1.1cm}p{1.1cm}p{0.8cm}p{0.9cm}} 
   \toprule
   $\uparrow$chrF & Frisian & Hmong & Pokomchi & Turkmen & Sesotho & Welsh & Xhosa & Indonesian & Hungarian & Spanish & Average \\
    \midrule
    \multicolumn{12}{l}{\textbf{Baselines:} }\\
    + \textit{Luke} & 47.5 & 41.6 & 39.4 & 34.9 & 41.2 & 41.2 & 32.0 & 43.3 & 34.4 & 46.7 & 40.2\\
    + \textit{Rand} & 50.5 & 43.9 & 42.8 & 38.9 & 43.2 & 46.0 & 34.9 & 47.2 & 37.4 & 50.1 & 43.5 \\
    \midrule 
    \multicolumn{12}{l}{\textbf{Our Models:} }\\
    + \textit{S} & 49.2 & 38.5 & 40.4 & 35.2 & 39.0 & 41.9 & 32.5 & 43.5 & 35.1 & 48.0 & 40.3 \\
    + \textit{SN} & 50.9 & 43.9 & 43.2 & 38.3 & 41.6 & 43.2 & 36.1 & 46.9 & 36.7 & 50.3 & 43.1 \\
    + \textit{SNG$_{2}$} & 53.2 & \bf{46.1} & 43.3 & 39.5 & 44.4 & 45.8 & 36.6 & 48.4 & 37.8 & 51.8 & 44.7 \\
    + \textit{SNG$_{3}$} & 52.7 & 46.0 & \bf{44.5} & 39.6 & \bf{45.5} & 47.5 & \bf{36.8} & 48.9 & \bf{39.2} & 52.3 & 45.3 \\
    + \textit{SNG$_{4}$} & \bf{53.6} & 45.7 & 44.4 & \bf{40.3} & 44.9 & 47.7 & 36.8 & \bf{49.1} & 39.0 & \bf{52.7} & \bf{45.4} \\
    + \textit{SNG$_{5}$} & 53.0 & 45.6 & 43.9 & 39.7 & 45.4 & 46.7 & 36.8 & 49.1 & 38.4 & 52.5 & 45.1 \\
    + \textit{ENT$^{N}$} & 50.9 & 43.7 & 38.1 & 37.2 & 42.5 & 44.5 & 34.7 & 46.7 & 36.0 & 49.9 & 42.4 \\
    + \textit{ENT$^{K}$} & 52.7 & 45.7 & 43.5 & 40.2 & 44.6 & 45.2 & 36.4 & 49.0 & 39.1 & 51.8 & 44.8 \\
    + \textit{AGG$^{L}_{5}$} & 47.1 & 41.5 & 39.8 & 34.0 & 39.9 & 42.1 & 31.4 & 43.5 & 33.7 & 45.2 & 39.8 \\
    + \textit{AGG$^{F}_{5}$} & 45.0 & 38.4 & 38.5 & 32.4 & 38.8 & 47.1 & 30.4 & 41.2 & 33.3 & 44.2 & 38.9 \\
    + \textit{AGG$^{P}_{5}$} & 45.5 & 38.8 & 38.0 & 32.0 & 38.8 & \bf{48.2} & 30.5 & 41.0 & 33.2 & 44.0 & 39.0 \\
    + \textit{AGG$^{N}_{5}$} & 45.4 & 39.1 & 38.3 & 32.4 & 38.8 & 48.0 & 30.7 & 41.2 & 33.2 & 44.3 & 39.1 \\
    \bottomrule
  \end{tabularx}
  \caption{140 experiments comparing 14 active learning methods translating into 10 different languages with Schedule \textit{B}.}
\label{table:14by10}
\end{table*}

\begin{table}[t]
  \footnotesize
  \centering
  \begin{tabularx}{\columnwidth}{p{2.5cm}p{2.5cm}p{2.5cm}p{2.5cm}p{2.5cm}p{2.5cm}} 
    \toprule
    $\uparrow$chrF & Frisian & Welsh & Hungarian & Spanish  & Average \\
    \midrule
    \multicolumn{6}{l}{\textbf{Baselines:} }\\
    + \textit{Luke} & 49.3 & 44.3 & 38.8 & 48.4 & 45.2 \\
    + \textit{Rand} & 53.5 & 49.5 & 42.2 & 53.2 & 49.6 \\
    \midrule 
    \multicolumn{6}{l}{\textbf{Our Models:} }\\
    + \textit{S} & 51.9 & 45.9 & 40.4 & 51.1 & 47.3 \\
    + \textit{SN} & 54.8 & 47.4 & 42.3 & 53.2 & 49.4 \\
    + \textit{SNG$_{2}$} & 54.5 & 49.5 & 43.5 & 54.2 & 50.4 \\
    + \textit{SNG$_{3}$} & 54.4 & 50.4 & \bf{43.9} & 54.5 & \bf{50.8} \\
    + \textit{SNG$_{4}$} & \bf{54.9} & 49.8 & 43.2 & \bf{54.9} & 50.7 \\
    + \textit{SNG$_{5}$} & 54.5 & 50.1 & 43.5 & 54.1 & 50.6 \\
    + \textit{ENT$^{N}$} & 52.7 & 47.2 & 40.9 & 52.9 & 48.4 \\
    + \textit{ENT$^{K}$} & 54.6 & 49.4 & 43.5 & 53.8 & 50.3 \\
    + \textit{AGG$^{A}_{5}$} & 49.4 & 44.2 & 37.3 & 48.2 & 44.8\\
    + \textit{AGG$^{S}_{5}$} & 46.5 & 49.8 & 36.4 & 46.4 & 44.8\\
    + \textit{AGG$^{M}_{5}$} & 48.6 & 50.4 & 36.5 & 46.9 & 45.6\\
    + \textit{AGG$^{T}_{5}$} & 48.8 & \bf{50.8} & 36.4 & 46.9 & 45.7\\
    \bottomrule
  \end{tabularx}
    \caption{56 experiments activating the knowledge in M2M100 with Schedule \textit{I}. }
\label{table:14by4}
\end{table}

Training on $\sim$100 million 
parameters with Geforce RTX 2080 Ti and RTX 3090,
we use a 6-layer encoder and a 6-layer decoder with
512 hidden states, 8 attention heads,
512 word vector size, 2,048 hidden units,
6,000 batch size, 0.1 label smoothing,
2.5 learning learning rate and 1.0 finetuning learning rate, 
0.1 dropout and attention dropout,
a patience of 5 after 190,000 steps in {[N]}$^2$ with an update interval of 1000, 
a patience of 5 for {[N+1]}$^2$ with an update interval of 200, and 
a patience of 25 for {[N+1]} and {[1]}$^2$ with an update interval of 50, 
``adam'' optimizer and
``noam'' decay method \citep{klein2017opennmt, papineni2002bleu}. 

For finetuning from a M2M100 Model, 
training on $\sim$418 million 
parameters with Geforce RTX 3090,
we use a 12-layer encoder and a 12-layer decoder with
1024 hidden states, 16 attention heads,
1024 word vector size, 4,096 hidden units, 
0.2 label smoothing,
0.0002 training learning rate and finetuning 0.00005 learning rate, 
0.1 dropout and attention dropout,
a patience of 10, ``BLEU'' validation metric, 
``adam'' optimizer and
``noam'' decay method \citep{fan2021beyond, schwenk2019ccmatrix, el2019massive}. 

\section{Results}
For simplicity, we use the centeredness method to combine translations 
from all source languages and have one score per metric. To compare across 
different methods, all experiments have the same test set (3,461 lines), 
the intersection of all test sets.

\textbf{\ul{Our models improve over the baselines:}} With Schedule \textit{I}, we observe an average improvement of 24.7 in chrF score over the M2M100 baseline (Table~\ref{table:5by10}). By active learning with 4-gram model, we observe an increase of 28.8 in chrF score over the bilingual baseline. 

\textbf{\ul{Our strategic training schedule improves the translation further by activating the knowledge of M2M100 :}} With Schedule \textit{B} and the 4-gram model, we observe an average improvement of 18.6 in chrF score over the multilingual baseline (Table~\ref{table:4by10}). For Schedule \textit{I}, the increase is 24.8 over the multilingual baseline (Table~\ref{table:5by10}). Indeed, the increase with the activation of M2M100 is greater. This shows that our strategic schedules\footnote{In Table~\ref{table:4by10}, our model with training scheduling uses Schedule \textit{B}, our model with active learning uses \textit{SNG$_{4}$}.  In Table~\ref{table:5by10}, our model with training scheduling uses Schedule \textit{I}, our model with active learning uses \textit{SNG$_{4}$}.  } improve translation performance by activating the knowledge of M2M100. 

\subsection{Training Schedules}
We compare 24 training schedules using a randomly sampled seed corpus ($\sim$1,000 lines) to translate into Frisian (Table \ref{table:config2} and \ref{table:config1}). 

\textbf{\ul{Pretraining with {[N]}$^2$ works well without M2M100:}} We compare 8 training schedules without M2M100 (Table~\ref{table:config1}). We find that Schedule \textit{B} (pretraining on {[N]}$^2$ and training on {[N+1]}$^2$ and {[N+1]}) and Schedule \textit{F} (pretraining on {[N]}$^2$ and training on {[N+1]}) work well without M2M100. Schedule \textit{B} gives a chrF score of 51.1 and Schedule \textit{F} gives a chrF score of 51.2.  

M2M100 is useful when a target language and its corresponding source languages are in the M2M100 list and the test set does not overlap with the M2M100 training set. However, we strongly advise discretion, as training data for large pretrained models is usually not clearly specified and most are not trained with low-resource languages in mind. M2M100 training data may very likely contain the Bible data, so it only serves as a comparison and provides an alternative view to show that our model is robust with large models. When M2M100 does not apply, our models pretrained with {[N]}$^2$ suffice.

\textbf{\ul{Full stage training increases robustness:}} For models without M2M100 we can use Schedule \textit{B} (Table~\ref{table:14by10}) or F (Table~\ref{table:14by10_old}). Though the results for Frisian are similar, B is much better than F for morphologically rich languages like Pokomchi, Turkmen and Xhosa. Indeed, B with full training is more robust than F, which skips {[N+1]}$^2$. Similarly, for models with M2M100, we can use Schedule \textit{I} (Table~\ref{table:14by4}) or \textit{L} (Table~\ref{table:14by4_old}). Again, Schedule \textit{I} with full training stages perform better than Schedule \textit{L}.  

\textbf{\ul{Applying M2M100 alone gives poor results:}} Schedule \textit{X} produces poor results (Table~\ref{table:config2}). Problems include catastrophic forgetting, bias towards rich-resource languages, and unclean data. Existing research shows some released models mislabel their English data as Welsh \citep{radfordrobust}. 

\textbf{\ul{Mixed models with M2M100 perform well:}} A few training schedules beat those pretrained with {[N]}$^2$ (Table~\ref{table:config1}). Schedule \textit{I} (training on 5 stages) gives a chrF score of 52.9, L (training 3 stages skipping {[N+1]} and {[1]}$^2$) gives 52.8, M (training 4 stages skipping {[N+1]}$^2$) gives 52.7, J (training 4 stages skipping {[1]}$^2$) gives 51.8, and N (training 3 stages skipping {[N+1]}$^2$ and {[1]}$^2$) gives 51.9. All are higher than 
those without M2M100. 

\textbf{\ul{Adapting M2M100 to the domain and then to the low-resource language works best:}} 
Schedule \textit{I} (training on 5 stages) with score 52.9 performs best. These models first adapt M2M100 to the domain by doing another pretraining on N$^2$. After adapting M2M100 to the domain, we adapt the model to the low-resource language by training on {[N+1]}$^2$. The final two stages {[N+1]} and {[1]}$^2$ are optional.  


\textbf{\ul{Our models and mixed models perform better than M2M100 alone:}} We examine the selected sentences from different active learning algorithms and gain insight into how sentences are chosen. M2M100 often produces extremely short sentences or repetition. Our models do not have those issues. 

\section{Conclusion and Future Work}
Our key contributions is that we compare 24 schedules with large pretrained models in translation to low-resource languages. Our model is robust with large multilingual models. We find that adapting large pretrained models first to the domain by training on all text translations in existing source languages (N$^2$), followed by adapting it to the low-resource language by training on all translations including the low-resource data ({[N+1]}$^2$) works best. These two stages are the most essential while the rest is optional, and we recommend Schedule \textit{I} that trains on all 5 stages introduced in this chapter. This helps to minimize human post-editing efforts during the subsequent iterations after translation of the seed corpus.

While the industry trend is to move towards bigger models with bigger data, our minimalist approach not only uses fewer languages, but we also aggregate over fewer languages. 
Our vocabulary size is $\sim$3000 for low-resource languages; this is in sharp comparison with large multilingual models like M2M100 with vocabulary size 128,108 for 100 languages. 
This saves computation power and resources, and therefore time and money, while improving translation performance. 

Evaluation is still a challenge. It is difficult to find native
speakers and establish 
long-term collaborations. There is also much variety 
among low-resource languages. Some are
more accessible than others and these might provide earlier, realistic
evaluations of our method. 
Hmong and Eastern Pokomchi are harder to assess while 
Frisian and Welsh, and many Eastern dialects in southern 
China and Indonesia, offer easier access and  evaluation. 
Once we widen the set of low-resource languages by including 
more accessible ones, there are more possibilities to 
evaluate less accessible ones.
Empowering low-resource languages is not just a technology
problem. It requires much effort in communication and collaboration
with local communities. 
We welcome collaboration with native speakers 
to broaden our research perspective and to deepen mutual 
understanding of its diversity and complexity. 
Through our technologies, we would like to work with local 
communities to revive low-resource languages 
by bringing more young people to speak and use those languages. This will 
empower local communities and help them to flourish. 

In the next chapter, we 
will focus on working with human translators in the 
field of low-resource languages, and deploy the human machine translation 
workflow to evaluate its effectiveness in the real-world. 

\begin{table}[b!]
  \scriptsize
  \centering
  \begin{tabularx}{\columnwidth}
  {p{2.5cm}p{2.5cm}p{2.5cm}p{2.5cm}p{2.5cm}p{2.5cm}} 
    \toprule
    $\uparrow$chrF & Frisian & Welsh & Hungarian & Spanish  & Average \\
    \midrule
    \multicolumn{6}{l}{\textbf{Baselines:} }\\
    \textit{Luke} & 49.1 & 41.7 & 38.3 & 48.7 & 44.5 \\
    \textit{Rand} & 52.8 & 46.8 & 41.9 & 52.9 & 48.6 \\
    \midrule
    \multicolumn{6}{l}{\textbf{Our Models:} }\\
    \textit{S} & 51.6 & 44.8 & 40.7 & 52.0 & 47.3 \\
    \textit{SN} & 53.2 & 45.8 & 42.2 & 52.9 & 48.5 \\
    \textit{SNG$_{2}$} & 54.2 & 47.6 & 42.5 & 53.8 & 49.5 \\
    \textit{SNG$_{3}$} & 53.7 & 47.9 & \bf{43.3} & \bf{54.5} & 49.9 \\
    \textit{SNG$_{4}$} & \bf{54.3} & 48.5 & 43.2 & 54.4 & \bf{50.1} \\
    \textit{SNG$_{5}$} & 53.9 & 48.6 & 43.2 & 54.5 & 50.1 \\
    \textit{ENT$^{N}$} & 52.1 & 44.8 & 40.7 & 52.4 & 47.5 \\
    \textit{ENT$^{K}$} & 53.7 & 46.7 & 43.1 & 53.7 & 49.3 \\
    \textit{AGG$^{A}_{5}$} & 48.4 & 43.2 & 37.1 & 48.4 & 44.3 \\
    \textit{AGG$^{S}_{5}$} & 47.3 & 48.1 & 36.1 & 47.1 & 44.7 \\
    \textit{AGG$^{M}_{5}$} & 46.9 & 47.8 & 36.3 & 47.2 & 44.6 \\
    \textit{AGG$^{T}_{5}$} & 47.1 & \bf{48.8} & 36.1 & 46.8 & 44.7 \\
    \bottomrule
  \end{tabularx}
    \caption{56 experiments integrated with M2M100 on Schedule \textit{L}. }
\label{table:14by4_old}
\end{table}

\begin{table*}[b]
  \scriptsize
  \centering
  \begin{tabularx}{\textwidth}{p{1.1cm}p{0.7cm}p{0.8cm}p{1.1cm}p{1.0cm}p{0.8cm}p{0.6cm}p{0.6cm}p{1.25cm}p{1.25cm}p{0.8cm}p{0.9cm}} 
    \toprule
   $\uparrow$chrF & Frisian & Hmong & Pokomchi & Turkmen & Sesotho & Welsh & Xhosa & Indonesian & Hungarian & Spanish & Average \\
    \midrule
    \multicolumn{12}{l}{\textbf{Baselines:} }\\ 
    \textit{Luke} & 47.5 & 38.2 & 37.4 & 33.8 & 38.5 & 38.5 & 29.2 & 41.7 & 31.5 & 46.3 & 38.3\\
    \textit{Rand} & 51.3 & 38.9 & 41.5 & 36.4 & 39.0 & 43.1 & 32.1 & 45.3 & 34.8 & 50.2 & 41.3 \\
    \midrule 
    \multicolumn{12}{l}{\textbf{Our Models:} }\\
    \textit{S} & 48.7 & 35.8 & 39.8 & 27.6 & 36.1 & 38.1 & 29.4 & 41.5 & 32.5 & 47.5 & 37.7 \\
    \textit{SN} & 50.9 & 38.4 & 41.5 & 36.9 & 38.7 & 41.1 & 32.5 & 44.8 & 33.1 & 49.2 & 40.7 \\
    \textit{SNG$_{2}$} & 52.9 & 40.9 & 42.4 & 37.3 & 41.0 & 44.3 & 33.4 & 45.8 & 35.8 & 51.2 & 42.5 \\
    \textit{SNG$_{3}$} & 53.1 & 41.8 & \bf{43.2} & \bf{38.4} & 41.9 & 45.6 & \bf{34.0} & 47.0 & 36.4 & 52.2 & \bf{43.4} \\
    \textit{SNG$_{4}$} & \bf{53.6} & 41.8 & 42.2 & 38.1 & 41.7 & 44.5 & 33.5 & \bf{47.5} & 36.7 & \bf{52.5} & 43.2 \\
    \textit{SNG$_{5}$} & 53.0 & 41.5 & 42.0 & 38.1 & \bf{42.3} & 45.1 & 33.5 & 47.3 & 36.4 & 52.2 & 43.1 \\
    \textit{ENT$^{N}$} & 50.7 & 39.5 & 34.0 & 34.8 & 39.4 & 42.5 & 32.4 & 44.4 & 33.9 & 48.6 & 40.0 \\
    \textit{ENT$^{K}$} & 52.5 & \bf{42.4} & 42.3 & 38.5 & 41.6 & 43.4 & 33.6 & 47.1 & \bf{37.1} & 51.7 & 43.0 \\
    \textit{AGG$^{L}_{5}$} & 47.4 & 38.8 & 38.9 & 33.2 & 37.3 & 40.1 & 28.9 & 41.6 & 31.7 & 45.7 & 38.4 \\
    \textit{AGG$^{F}_{5}$} & 44.6 & 36.0 & 37.1 & 30.9 & 35.8 & 44.3 & 27.8 & 39.2 & 30.7 & 43.9 & 37.0 \\
    \textit{AGG$^{P}_{5}$} & 45.2 & 36.6 & 36.9 & 30.8 & 35.6 & 44.9 & 27.9 & 39.0 & 30.5 & 43.8 & 37.1 \\
    \textit{AGG$^{N}_{5}$} & 45.4 & 36.8 & 37.1 & 31.3 & 35.7 & \bf{46.0} & 28.0 & 39.2 & 30.2 & 43.8 & 37.4 \\
    \bottomrule
  \end{tabularx}
    \caption{140 experiments comparing 14 active learning methods translating into 10 different languages on Schedule \textit{F}.  
    }
\label{table:14by10_old}
\end{table*}

\begin{table*}[t]
  \scriptsize
  \centering
  \begin{tabularx}{\textwidth}{p{1.2cm}p{0.75cm}p{0.8cm}p{1.2cm}p{1.0cm}p{0.8cm}p{0.6cm}p{0.6cm}p{1.25cm}p{1.2cm}p{0.8cm}p{0.9cm}} 
   \toprule
   Seed Size & Frisian & Hmong & Pokomchi & Turkmen & Sesotho & Welsh & Xhosa & Indonesian & Hungarian & Spanish & Average \\
    \midrule
    Word count & 25695 & 31249 & 36763 & 17354 & 25642 & 25786 & 15017 & 22318 & 18619 & 22831 & 24127\\
    \midrule
    \multicolumn{6}{l}{Line count for each experiment}\\
    \multicolumn{6}{l}{\textbf{Baselines:} }\\
    \textit{Luke} & 1151 & 1151 & 1151 & 1151 & 1151 & 1151 & 1151 & 1151 & 1151 & 1151 & 1151\\
    \textit{Rand} & 1022 & 1001 & 1101 & 1045 & 976 & 1117 & 988 & 1065 & 1066 & 1023 & 1040 \\
    \midrule
    \multicolumn{6}{l}{\textbf{Our Models:} }\\
    \textit{S} & 692 & 654 & 832 & 689 & 657 & 771 & 598 & 634 & 644 & 682 & 685 \\
    \textit{SN} & 1522 & 1399 & 1522 & 1524 & 1434 & 1595 & 1501 & 1601 & 1545 & 1488 & 1513 \\
    \textit{SNG$_{2}$} & 1484 & 1350 & 1490 & 1454 & 1369 & 1557 & 1418 & 1513 & 1468 & 1463 & 1457 \\
    \textit{SNG$_{3}$} & 1385 & 1319 & 1468 & 1416 & 1317 & 1439 & 1368 & 1451 & 1415 & 1365 & 1394 \\
    \textit{SNG$_{4}$} & 1327 & 1295 & 1419 & 1367 & 1279 & 1409 & 1309 & 1426 & 1374 & 1310 & 1352\\
    \textit{SNG$_{5}$} & 1289 & 1289 & 1397 & 1311 & 1280 & 1381 & 1256 & 1359 & 1334 & 1273 & 1317 \\
    \textit{ENT$^{N}$} & 1796 & 1721 & 1769 & 1840 & 1761 & 1914 & 1839 & 1967 & 1884 & 1805 & 1830\\
    \textit{ENT$^{K}$} & 1340 & 1287 & 1507 & 1266 & 1132 & 1405 & 1128 & 1358 & 1264 & 1327 & 1301 \\
    \textit{AGG$^{A}_{5}$} & 984 & 1025 & 1060 & 998 & 967 & 1031 & 1016 & 1018 & 993 & 958 & 1005 \\
    \textit{AGG$^{S}_{5}$} & 1049 & 1084 & 1152 & 1043 & 1025 & 1182 & 1147 & 1093 & 1076 & 1019 & 1087\\
    \textit{AGG$^{M}_{5}$} & 1058 & 1097 & 1159 & 1109 & 1025 & 1232 & 1159 & 1101 & 1087 & 1018 & 1105\\
    \textit{AGG$^{T}_{5}$} & 1048 & 1094 & 1153 & 1101 & 1020 & 1274 & 1141 & 1101 & 1087 & 1014 & 1103\\
    \bottomrule
  \end{tabularx}
    \caption{Seed Corpus Size for different target languages. The seed corpus gives rise to both training data and validation data, therefore the training size is smaller than the above. Note that all experiments for a given target language share the same number of words, although they have different number of lines. Since each language use different number of words to express the same meaning of a given text, we choose the number of words in the given book ``Luke'' as the standard reference for each target language. For example, ``Luke'' in Xhosa contains 15,017 words while ``Luke'' in Frisian contains 25,695 words. }
\label{table:datasize}
\end{table*}

\removelabelprefix

\chapter{A Quechuan Case Study}\label{big:confidence}
\addlabelprefix{7}
\epigraph{``The difference between the right word and the almost right word is really a large matter – it’s the difference between lightning and a lightning bug.''}{\textit{Mark Twain}}

\lettrine{H}{aving examined various active learning methods} to build a seed corpus in the low-resource language, and having explored multiple scheduling methods to use large pretrained multilingual models to train on such a small seed corpus, we dive into the real-life applications of translation into low-resource languages by working closely with field linguists and human translators in Peru. In this chapter, we evaluate our progress from all previous chapters through a case study of the Quechuan language family by working with field linguist Mark Bean and his translation team. We show effectiveness of our model and our proposed human machine translation workflow for translation of a multi-source text into a new, low-resource language. We find that translation performance is significantly positively correlated with language similarity. The more connected a language is, the higher the translation performance. Furthermore, we find decluttering poorly-connected languages improves performance. Using this finding, we show our results in translating into a new, low-resource language called Sihuas Quechua. 

\section{Introduction} 
Machine Translation is not only a core scientific problem for Computer Science, but also a cost-cutting business tool. The market for language services is estimated to be US \$72.2 billions in 2023 \citep{depalma2020language, services2023market}. Many governments, businesses, international organizations like the European Union, and numerous global missions will accelerate in their spending on translation services by US \$20.83 billions during 2022-2026 in a recent forecast \citep{technavio2022global}. And the total market for language services is forecast to climb to US \$98.1 billions in 2028 \citep{services2023market}.

Although the current market for translation into low-resource languages are relatively small compared with the world's spending on translation services, its potential market is substantial. Indeed, translation into low-resource languages carries immense value in serving the cultural goal of saving and reviving low-resource languages and the humanitarian goal of assisting the everyday needs of local communities. 
 
To serve the goal of cultivating and expanding potential translation markets in low-resource languages, machine translation cannot stand alone. Large MT system outputs have problems with repetition, mislabelled data, hallucination, inaccuracy, underproduction, over-fitting and inconsistencies \citep{marcus2022deep,bommasani2021opportunities, thai2022exploring}. Neural MT systems often requires human post-editing to produce readily publishable materials, independent of how evolved the models are \citep{denkowski2015machine}. Systems adapted from large MT systems may have some improvements in translation, but still have problems and need post-editing to produce publishable translations. Indeed, post-editing and human checking is crucial in producing publishable materials \citep{thai2022exploring}.

Given the importance of post-editing and human checking, symbiosis is necessary between Machine Translation and human translation to accelerate and benefit the translation industry. In our translation framework, our goal is to translate a text that is available in multiple source languages to a new, severely low-resource language. To accomplish this goal, we rely on the close collaboration between MT systems and human translators to finish a high quality translation of the given text. We begin by performing active learning methods to rank sentences in the given text on available languages. Once we finish ranking, we inform human translators. After human translators obtain the ranking, they prepare the translation of the top few ranked sentences ($\sim$1,000 lines) to build our seed corpus. Once human translators finish building the seed corpus, they pass it to MT systems. Using this seed corpus, MT systems train and produce the first translation draft of the entire text in the severely low-resource language. Working on this draft, human translators post-edit and complete translation for a few portions of the text. These newly translated text is then fed back to the MT systems to improve training and produce improved version of the draft. Iterations of post-editing not only produce new data to improve translation quality by MT systems, but also serve as a positive feedback loop and expedite the entire process. In this way, human translators and MT systems work together to finish a high quality translation of the text. 

\removelabelprefix
\begin{table*}[t]
  \scriptsize
  \centering
  \begin{tabularx}{\textwidth}{p{2.3cm}p{0.6cm}p{0.8cm}p{0.95cm}p{0.8cm}p{0.8cm}p{0.45cm}p{0.45cm}p{1.0cm}p{1.0cm}p{0.7cm}p{0.7cm}} 
   \toprule
   $\uparrow$chrF & Frisian & Hmong & Pokomchi & Turkmen & Sesotho & Welsh & Xhosa & Indonesian & Hungarian & Spanish & Average \\
    \midrule
    Baseline & 23.1 & 25.0 & 28.7 & 18.9 & 25.2 & 22.2 & 21.4 & 27.2 & 20.1 & 22.1 & 23.4 \\
    \midrule
    \multicolumn{12}{l}{\textbf{Our Models:} }\\
    + Chapter \ref{big:family}, \ref{big:paraphrase}, \ref{big:ipml}  & 28.0 & 28.1 & 31.9 & 22.6 & 28.3 & 26.5 & 23.9 & 29.7 & 22.3 & 26.8 & 26.8 \\
    + Chapter \ref{big:large} & 50.5 & 43.9 & 42.8 & 38.9 & 43.2 & 46.0 & 34.9 & 47.2 & 37.4 & 50.1 & 43.5 \\
    + Chapter \ref{big:active} & 53.6 & 45.7 & 44.4 & 40.3 & 44.9 & 47.7 & 36.8 & 49.1 & 39.0 & 52.7 & 45.4 \\
    \bottomrule
  \end{tabularx}
  \caption{Result summary for translation into 10 languages that are 
  new and severely low-resource to the system, independent of M2M100. }
\label{table:4by10}
\end{table*}
 
\begin{table}[t]
  \scriptsize
  \centering
  \begin{tabularx}{\columnwidth}{p{3cm}p{2cm}p{2cm}p{2.5cm}p{2cm}p{2cm}} 
   \toprule
   $\uparrow$chrF & Frisian & Welsh & Hungarian & Spanish  & Average \\
    \midrule
    Baseline & 23.1 & 22.2 & 20.1 & 22.1 & 21.9 \\
    \midrule
    \multicolumn{6}{l}{\textbf{Our Models:} }\\
    + Chapter \ref{big:family}, \ref{big:paraphrase}, \ref{big:ipml}  & 28.0 & 26.5 & 22.3 & 26.8 & 25.9 \\
    + Chapter \ref{big:large} & 53.5 & 49.5 & 42.2 & 53.2 & 49.6 \\ 
    + Chapter \ref{big:active} & 54.9 & 49.8 & 43.2 & 54.9 & 50.7 \\
    \bottomrule
  \end{tabularx}
  \caption{Result summary for translation into 4 languages that are 
  new and severely low-resource to the system, 
  leveraging knowledge in M2M100 and using active learning.    
  }
\label{table:5by10}
\end{table}

Having understood this translation workflow between MT systems and human translators, we would like to know how it works in real life and evaluate the progress from all previous chapters. In Table~\ref{table:4by10} and Table~\ref{table:5by10}, we show the summary of results from the previous chapters. We see that multilingual training by carefully selecting similar languages to train with (Chapter \ref{big:family}, Chapter \ref{big:paraphrase} and Chapter \ref{big:ipml}), active learning (Chapter \ref{big:active}) and staged finetuning with large pretrained models (Chapter \ref{big:large}) improve translation performance in our experiments. To evaluate improvements made from Chapter \ref{big:family} to \ref{big:large}, and to translate our progress in academic settings into the real-world applications, we work with field linguist Mark Bean and his team in Peru on a case study of the Quechuan language family.  

Working on this Quechuan case study, we show in this chapter that for well-connected languages, our finding of multilingual training by carefully selecting similar languages to train with (Chapter \ref{big:family}, Chapter \ref{big:paraphrase} and Chapter \ref{big:ipml}), active learning (Chapter \ref{big:active}) and staged finetuning with large pretrained models (Chapter \ref{big:large}) improve translation performance. For poorly-connected languages, due to the low language similarity available for cross-lingual transfer, the impact of our method could not be tested. 

\begin{figure*}[t]
  \centering
  \includegraphics[width=0.7\linewidth]{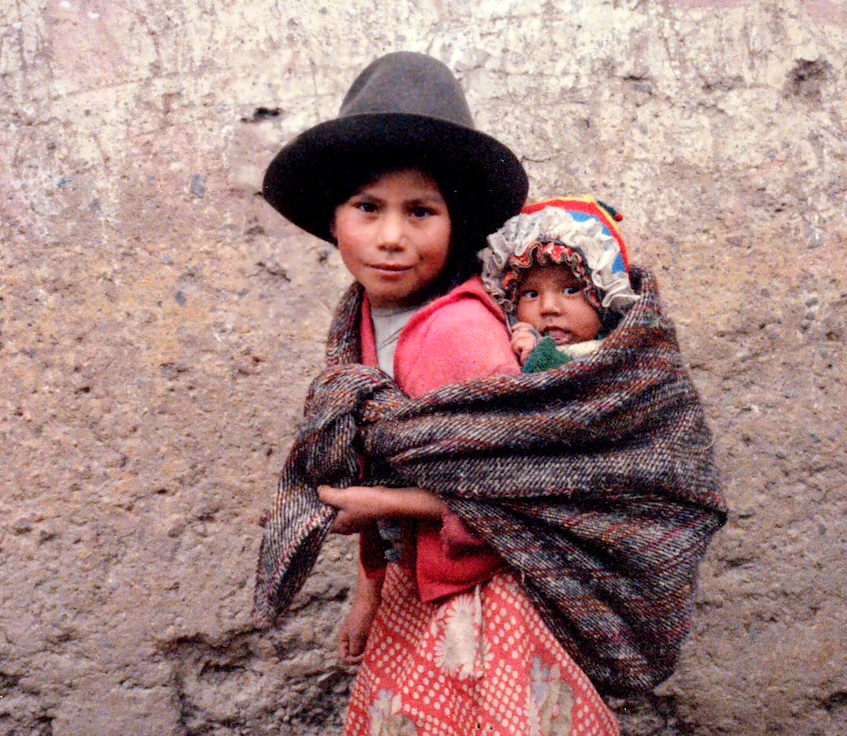}
    \caption{Sisters who speak Margos Quechua in Peru. Photograph by Mark Bean. }
    \label{fig:peru_girl}  
\end{figure*}

\begin{table}[t]
\centering
  \small
  \begin{tabularx}{\columnwidth}{p{2cm}|p{8cm}p{1cm}p{1cm}p{1cm}p{1cm}}
    \toprule
    ISO Code & Language & Total & Books & OT & NT \\ 
    \midrule
    quz	& Quechua, Cuzco & 31099 & 66 & 23142 & 7957 \\
    quy & Quechua, Ayacucho & 31099 & 66 & 23142 & 7957 \\
    quh & Quechua, South Bolivian & 31099 & 66 & 23142 & 7957 \\
    qub & Quechua, Huallaga	& 31099 & 66 & 23142 & 7957 \\ 
    qxo (qxoc) & Quechua, Southern Conchucos & 31099 & 66 & 23142 & 7957 \\ 
    qxo (qxoh) & Quechua, Huacaybamba & 31099 & 66 & 23142 & 7957 \\ 
    qve & Quechua, Eastern Apurímac	& 31099 & 66 & 23142  & 7957\\
    qvh & Quechua, Huamalíes-Dos de Mayo Huánuco & 31099 & 66 & 23142 & 7957\\ 
    qvm & Quechua, Margos-Yarowilca-Lauricocha & 31099 & 66 & 23142 & 7957\\ 
    qvw & Quechua, Huaylla Wanca & 31099 & 66 & 23142  & 7957\\
    qwh & Quechua, Huaylas Ancash & 31099 & 66 & 23142 & 7957\\ 
    qxn & Quechua, Northern Conchucos Ancash & 31099 & 66 & 23142 & 7957\\ 
    inb	& Quechua, Inga & 7957 & 27 & 0 & 7957\\
    quf & Quechua, Lambayeque & 7957 & 27 & 0 & 7957 \\
    qul & Quechua, North Bolivian & 7957 & 27 & 0 & 7957 \\
    qup & Quechua, Southern Pastaza	& 7957 & 27 & 0 & 7957 \\
    qvc & Quechua, Cajamarca & 7957 & 27 & 0 & 7957 \\
    qvn & Quechua, North Junín & 7957 & 27 & 0 & 7957 \\
    qvs & Quechua, San Martín & 8308 & 28 & 351 & 7957 \\ 
    qvz & Quechua, Northern Pastaza	& 7957 & 27 & 0 & 7957 \\
    qxh & Quechua, Panao & 7957 & 27 & 0 & 7957 \\
    qws	& Quechua, Sihuas & 7373 & 23 & 1494 & 5879 \\
     \bottomrule
    \end{tabularx}
  \caption{Quechuan Family. ``Total'' is the total number of lines in the text, ``OT'' is the number of lines in Old Testament while ``NT'' is that in New Testament, and ``Books'' is the number of books translated. To differentiate Southern Conchucos from Huacaybamba, we use ``c'' and ``h''. }
\label{table:quechuan}
\end{table}

\section{A Case Study on Quechuan Languages}
To show effectiveness of our methods, we collaborate with field linguist, Mark Bean, and his team who work closely 
with the low-resource Quechuan language communities in South America, especially in the region of Peru. Mark and his team have been spending their lives working closely with the Quechuan language community to translate the Bible text into each of the target Quechuan languages. 

Our goal in this case study on the Quechuan language family is three-fold. Firstly, we aim to find out whether our method works in real-world translation efforts. Furthermore, we want to determine under which conditions it is most favorable to apply our method to translate into the low-resource languages successfully. Lastly, in the instances where those conditions are not met, we suggest a few future research directions to improve translation quality. 

\subsection{History and Geography}
The Quechuan language family is a varied group of languages covering a wide Andean region of South America, stretching from Colombia, Ecuador, Peru, Bolivia to northwest Argentina \citep{luykx2016communicative, howard2011quechua}. There is broad spectrum of sociolinguistic diversity among Quechuan languages throughout the Spanish colonial history of the area \citep{durston2007pastoral}. 

Quechua is widely spoken in a wide range of Peruvian Andes before the expansion of the Inca Empire \citep{cobo2010history}. The Inca empire enforced Quechua as the official language during its rule where many diverse dialects are developed, influenced by local languages \citep{king2006quechua}. For example, Quechua is influenced by Aymara in the area of Cuzco, which is the old Inca capital \citep{turino2017quechua}. In addition to the support of the Inca Empire, the Spanish rulers also helped Quechuan languages to grow and flourish \citep{escobar2011spanish}. The Catholic church encouraged and facilitated the first written form of Quechua \citep{de1951lexicon}. 

Ethnologue has 45 Quechuan languages which are then divided into two groups: central (Quechua I) and peripheral (Quechua II) \citep{eberhard2021ethnologue, adelaar2013quechua}. Within the categories, they are part of the dialect continuum as they mostly contain dialects spoken across the Peruvian Andean area where they could be mutually intelligible if they are close enough \citep{blum2023phylolinguistic}. However, languages are not mutually intelligible across categories \citep{landerman1991quechua}.  

\subsection{Key Languages in Analysis}
We show the Quechuan family of languages\footnote{Note that Huacaybamba Quechua and Southern Conchucos Quechua both have the ISO code ``qxo'', in other tables of this thesis, we will differentiate the two as ``qxoh''(Huacaybamba) and ``qxoc''(Southern Conchucos). } we are working on in Table \ref{table:quechuan}. We have Quechuan languages that differ in resource levels: we have 12 languages that have complete Bible text, and 9 languages that have at least New testament text, and 1 language that only has a few stories translated. There are twelve Quechuan languages that has complete translation of the Bible text: Cuzco Quechua, Ayacucho Quechua, South Bolivian Quechua, Huallaga Quechua, Southern Conchucos Quechua, Huacaybamba Quechua, Eastern Apurímac Quechua, Huamalíes-Dos de Mayo Huánuco Quechua, Margos-Yarowilca-Lauricocha Quechua, Huaylla Wanca Quechua, Northern Conchucos Ancash Quechua, Huaylas Ancash Quechua. These languages have complete Bibles in both Old Testament and New Testament. We also have nine other languages that have only partial translations of the Bible, having at least the entire translation of the New Testament. They are: Inga Quechua, Lambayeque Quechua, North Bolivan Quechua, Southern Pastaza Quechua, Cajamarca Quechua, North Junín Quechua, San Martín Quechua, Northern Pastaza Quechua, and Panao Quechua. The language that does not have the entire translation of the New Testament is Sihuas Quechua. Sihuas has partial translations of the New Testament and a few chapters from the Old Testament.   	 

\section{Data} 
We use both the partial and complete translations across 22 languages in Quechuan family. We have the complete translations of the Bible data in 12 languages. These data include both the Old Testament and the New Testament data. We also use the 9 languages that has New Testament data. We have a language, Sihuas, that has only partial New Testament and partial Old Testament data. There are three languages that we are focusing on in our detailed qualitative evaluation, and they are: Margos-Yarowilca-Lauricocha Quechua (\textit{Margos}), Panao Quechua (\textit{Panao}) and Sihuas Quechua (\textit{Sihuas}) \citep{van2020highland, espichan2018language, smith2009panao, durston2007pastoral}. There is an existing translation of the complete Bible in Margos, however, there is no existing translation of the Old Testament in Panao. Finally, no translation of the Old Testament or the New Testament exists in Sihuas.

\section{Results Analysis}
We evaluate the effectiveness of our model by investigating the following questions: 1.) Does our method presented in this thesis work in real-world translation? 2.) Under which conditions does our method work, and under which conditions does it not work? 3.) When our method works, how well does it work? 4.) When our method does not work, which future research directions could we pursue to improve translation performance? 

To answer these questions, we would like to examine the core focus of this thesis by looking at the relationship between translation performance and language similarity. In this section, we first look at language similarity and determine its correlation with MT performance in three different perspectives. After determining their correlation, we conclude this analysis in translation into Sihuas Quechua, a new, low-resource language. 

\begin{figure*}[t]
  \centering
  \includegraphics[width=0.85\linewidth]{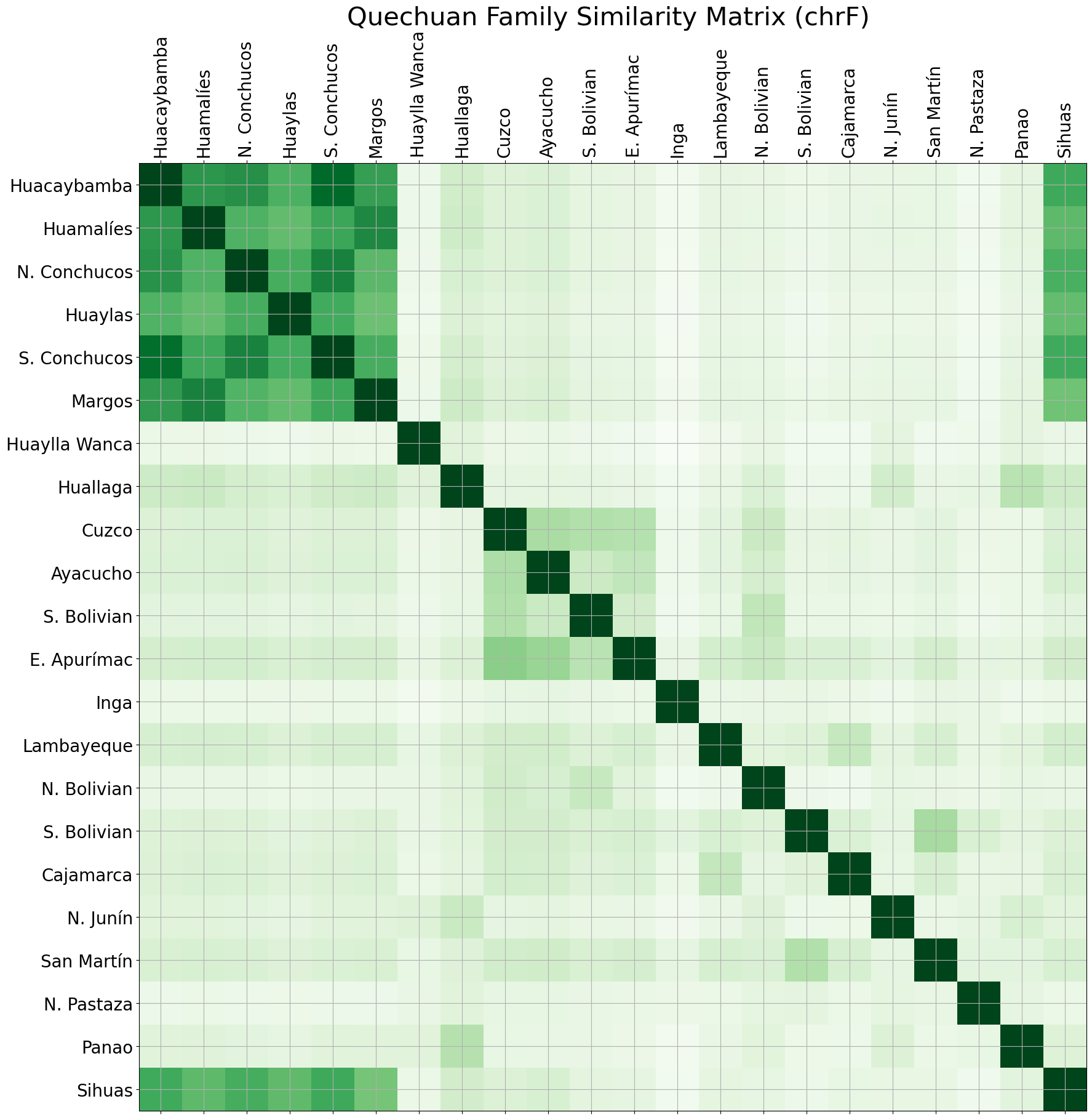}
    \caption{Similarity matrix on Quechuan family (chrF). }
    \label{fig:similarity_chrf_22lan}  
\end{figure*}

\begin{figure*}[t]
    \centering
    \begin{subfigure}{0.48\linewidth}
        \centering
        \includegraphics[width=\linewidth]{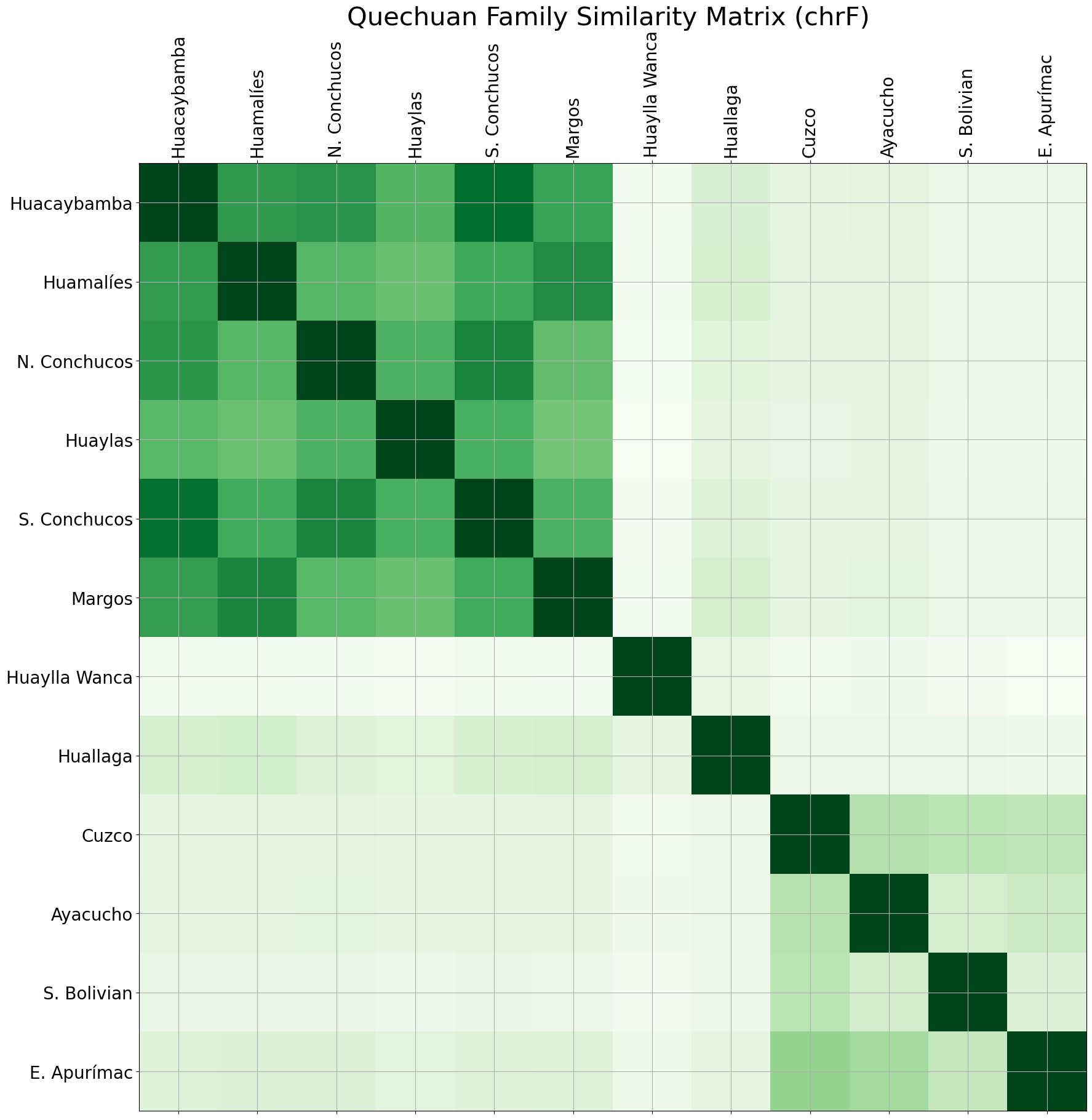}
        \caption{Similarity matrix on 12 Quechuan languages.}
        \label{fig:similarity_chrf_12lan}
    \end{subfigure}
    \hfill
    \begin{subfigure}{0.48\linewidth}
        \centering
        \includegraphics[width=\linewidth]{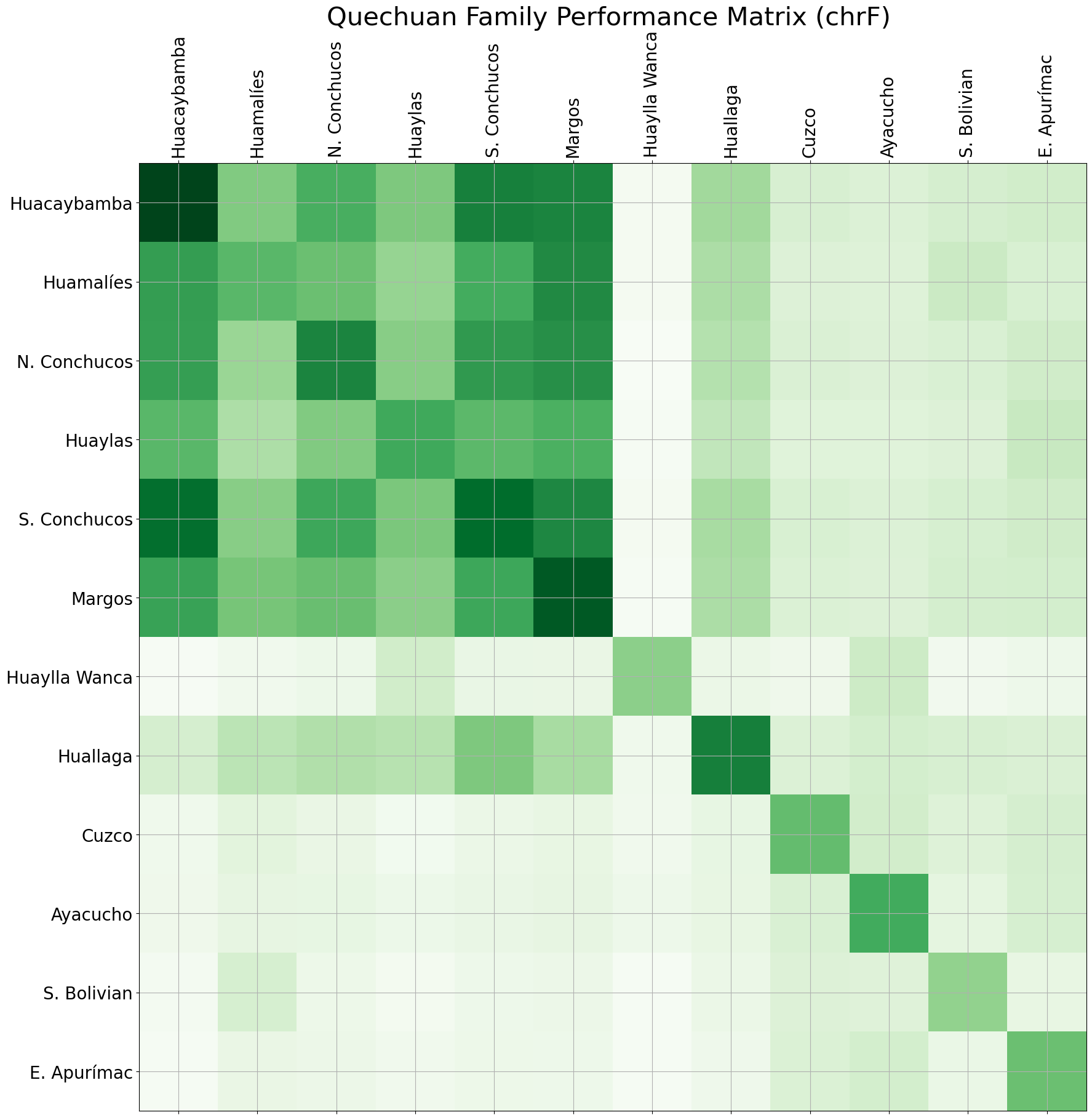}
        \caption{Performance matrix on 12 Quechuan languages.}
        \label{fig:performance_chrf_12lan}
    \end{subfigure}
    \caption{Comparison of similarity and performance matrices (chrF).}
    \label{fig:correlation_matrices}
\end{figure*}

\subsection{Similarity Analysis}
We begin our analysis by examining language similarity. There are many methods to measure language  \cite{hajic2000machine, oncevay2020bridging, chung2020improving}. Most methods require information on typology \citep{chowdhury2020understanding, rama2012good, pienemann2005processability, svalberg1998english, hansen2012beginning, comrie2005world}, World Atlas of Language Structures \cite{comrie2005world}, and the Swadesh list \cite{huang2007towards}. Given severely low-resource scenarios in our use case, we do not make assumptions on external resources, and we aim to measure language similarity in a light-weight manner by only using the text we are translating. 

Using the given text, we measure language similarity between any two given languages by looking at the text translations in them. We analyse sentence overlap (percentage of matching sentences in both languages) and word overlap (percentage of matching words in both languages). We also use metrics including characTER, chrf and 1-gram and 4-gram BLEU \citep{popovic2015chrf, papineni2002bleu, wang2016character} on translations in both languages. 

Using these pair-wise similarity measures, we show a 22-by-22 similarity matrix for all Quechuan languages in our case study using chrF metric in Figure \ref{fig:similarity_chrf_22lan} and a 12-by-12 close-up version for the 12 Quechuan languages with full text translation in Figure~\ref{fig:similarity_12lan_all}. Margos and Sihuas are in the dark green area (well-connected), Panao is in the almost white area (poorly-connected). We clearly see that there is a strong language cluster among these 6 Quechuan languages: Huacaybamba Quechua, Huamalíes-Dos de Mayo Huánuco Quechua, Northern Conchucos Ancash Quechua, Huaylas Ancash Quechua, Southern Conchucos Quechua, and Margos-Yarowilca-Lauricocha Quechua. There is a less obvious language cluster among 4 other Quechuan languages: Cuzco Quechua, Ayacucho Quechua, South Bolivian Quechua, and Eastern Apurimac Quechua. Apart from these two language clusters, the similarity matrix is mostly light green in other areas except for the language Sihuas Quechua. 

Sihuas Quechua, represented as last row and column in the similarity matrix in Figure \ref{fig:similarity_chrf_22lan}, stands out as deep green. Even though Sihuas is the least resourced among the 22 languages, it is well connected to the main cluster we identified earlier and is similar to each of the 6 languages in said cluster. 

In addition to chrF metric, we show the full set of language similarity matrices using chrF, characTER, 1-gram BLEU, 4-gram BLEU, sentence
overlap and word overlap in Figure \ref{fig:similarity_all_22lan}. We see that all 6 matrices are very similar, though having different levels of granularity. The sentence overlap metric, for example, is the coarsest measure as every pair is mostly light green. This is very intuitive because it is indeed rare that the same sentence is expressed in the exactly the same way in two languages. For example, Margos has 5\% sentence match and 88\% word match. The low percentage of sentence match could be caused by alignment issues or close-to-zero edit distances that are not captured by this metric. Indeed, other metrics like chrF, characTER, 1-gram BLEU, 4-gram BLEU offer more fine-grained analysis portfolios for each language. The most fine-grained analysis is done through character-level metrics like chrF and characTER. 

\subsection{How Similarity Affect Performance}
Having explored various similarity measures on 22 Quechuan languages, we would like to examine how similarity influences translation performance. We show three approaches in this section: the round robin experiment, and the effects of adding and removing similar languages. We examine each approach closely in the remaining section. 

\begin{table*}[t]
    \centering
    \small
    \begin{tabularx}{\textwidth}{p{7.2cm}p{0.6cm}p{1.7cm}p{2.5cm}p{2.5cm}}
    \toprule
    Target Language & chrf & characTER & 1-gram BLEU & 4-gram BLEU \\ 
    \midrule
    Quechua Huacaybamba & 81.4 & 0.858 & 75.0 & 49.4 \\ 
    Quechua Huamalíes-Dos de Mayo Huánuco & 60.0 & 0.518 & 56.2 & 23.0 \\ 
    Quechua Northern Conchucos Ancash & 70.3 & 0.666 & 66.0 & 30.9 \\ 
    Quechua Huaylas Ancash & 61.9 & 0.555 & 59.6 & 22.7 \\ 
    Quechua Southern Conchucos & 80.4 & 0.793 & 77.4 & 46.8 \\
    Quechua Margos-Yarowilca-Lauricocha & 83.2 & 0.835 & 80.1 & 54.7 \\
    Quechua Huaylla Wanca & 27.6 & 0.231 & 32.1 & 3.9 \\
    Quechua del Huallaga Huánuco & 52.5 & 0.417 & 45.5 & 12.8 \\ 
    Quechua Cuzco & 39.9 & 0.328 & 33.7 & 4.7 \\ 
    Quechua Ayacucho & 39.4 & 0.320 & 29.8 & 2.9 \\ 
    Quechua South Bolivian & 41.4 & 0.317 & 37.2 & 6.2 \\ 
    Quechua Eastern Apurímac & 43.2 & 0.310 & 33.9 & 5.7 \\ 
    \bottomrule
    \end{tabularx}
    \caption{Key result summary of the round robin experiments. }
    \label{table:roundrobin}
\end{table*}

\subsubsection{Round Robin Experiment}
Firstly, we conduct a set of round robin experiments. To simplify our visualization and assure data symmetry, we choose 12 languages with complete text translations for this set of experiments: Cuzco Quechua, Ayacucho Quechua, South Bolivian Quechua, Huallaga Quechua, Southern Conchucos Quechua, Huacaybamba Quechua, Eastern Apurímac Quechua, Huamalíes-Dos de Mayo Huánuco Quechua, Margos-Yarowilca-Lauricocha Quechua, Huaylla Wanca Quechua, Northern Conchucos Ancash Quechua, Huaylas Ancash Quechua. 

\begin{table*}[t]
    \centering
    \small
    \begin{tabularx}{\textwidth}{XXXXX} 
        \toprule
        Spearman & chrf & characTER & 1-gram BLEU & 4-gram BLEU \\
        \midrule
        Correlation & 0.771 & 0.729 & 0.703 & 0.741 \\
        P-value & $4.25 \times 10^{-25}$ & $1.31 \times 10^{-29}$ & $2.51 \times 10^{-26}$ & $9.65 \times 10^{-23}$ \\
        \bottomrule
    \end{tabularx}
    \caption{Key result summary of correlation between performance and similarity.}
    \label{table:correlation}
\end{table*}
For each round of the round robin experiments, we take one of these 12 languages as the hypothetical target low-resource language, and assume that we only have New Testament data for this language. Since all the other 11 languages have complete translations of the whole Bible including both the Old Testament and the New Testament, we train on all the 11 complete Bibles and test on the Old Testament of the chosen target language. We use DeltaLM as our large pretrained model for multi-stage adaptations \citep{deltalm}.

For DeltaLM, we use the sentence-piece vocabulary that comes directly with it. This vocabulary contains 250k unique tokens. We have tried to build our own vocabulary that is morpheme-based; however, training with existing vocabulary works much better than morpheme-based self-constructed vocabulary. Therefore, we use the default vocabulary with 250 unique tokens. 

For training schedules, we train using both Schedule \textit{B} (without using large pretrained models) and Schedule \textit{J} (using large pretrained models). For example, if we choose Margos as the target low-resource language, we reach a BLEU score of 53.0 with Schedule \textit{B} and we have a BLEU score of 54.7 with Schedule \textit{J}. Indeed, Schedule \textit{J} is also preferred by the field linguists through qualitative evaluation. For simplicity, we show results of Schedule \textit{J} in this section. 

We show key results from the round robin experiments in Table \ref{table:roundrobin}. Each row represents the combined score from all 11 source languages into the given target language specified in the first column. We see that there is a wide spectrum of machine translation performance. The experiment translating into Margos, for example, reaches a BLEU score of 54.7 and a chrF score of 83.2. However, the experiment translating into Cuzco reaches a BLEU score of 4.7 and a chrF score of 39.9. To understand this large difference in performance, we want to establish the correlation of performance and similarity to the target language. 

\subsubsection{Correlation between Performance and Similarity}
To understand the correlation between translation performance and language similarity, we see that they are positively correlated in Figure \ref{fig:correlation_matrices}. To visualize how translation performance is driven by language similarity, we show similarity matrices and performance matrices for the 12 Quechuan languages we have shown earlier in Figure \ref{fig:correlation_matrices_all} and Figure \ref{fig:correlation_matrices}. Comparing them side-by-side, we see their resemblance, especially at the 6-language cluster that is dark green (well-connected area). 

\begin{figure}
    \includegraphics[width=.4\textwidth]{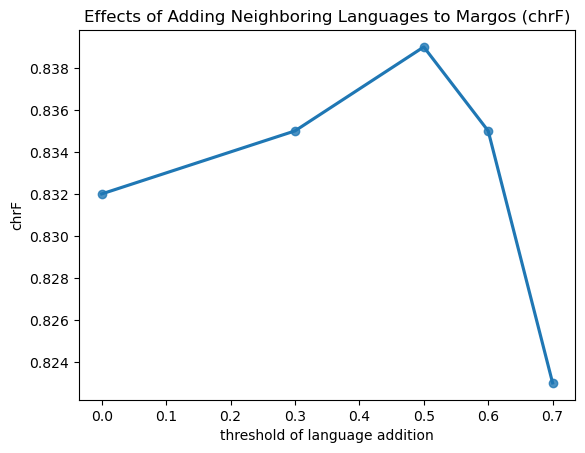}\hfill
    \includegraphics[width=.4\textwidth]{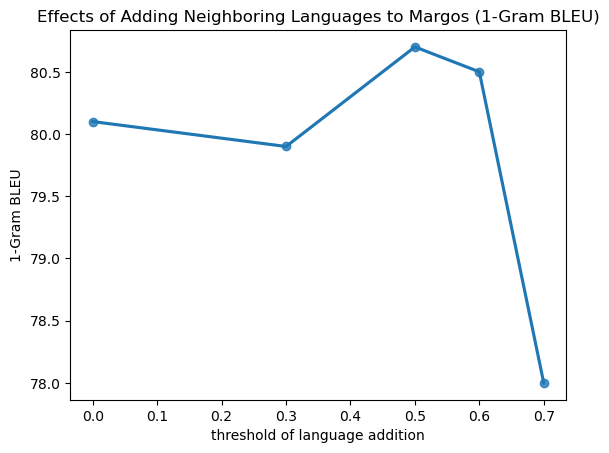}
    \\[\smallskipamount]
    \includegraphics[width=.4\textwidth]{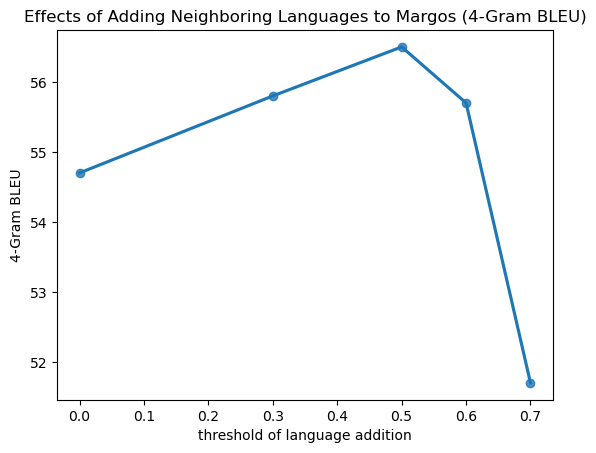}\hfill
    \includegraphics[width=.4\textwidth]{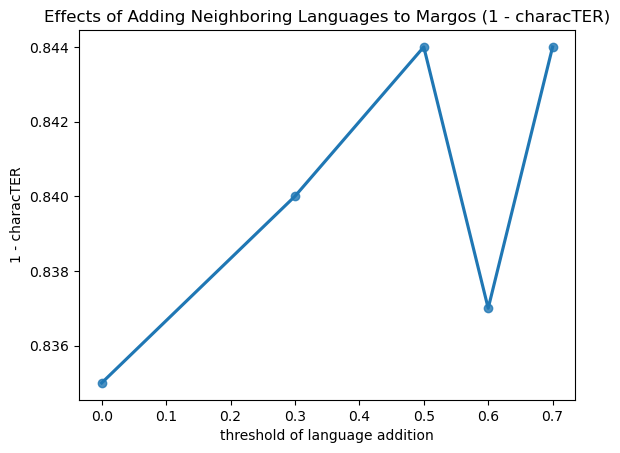}\hfill
    \caption{Effects of adding similar languages in translation into Margos}\label{fig:addlan}
\end{figure}

\begin{figure}
    \includegraphics[width=.4\textwidth]{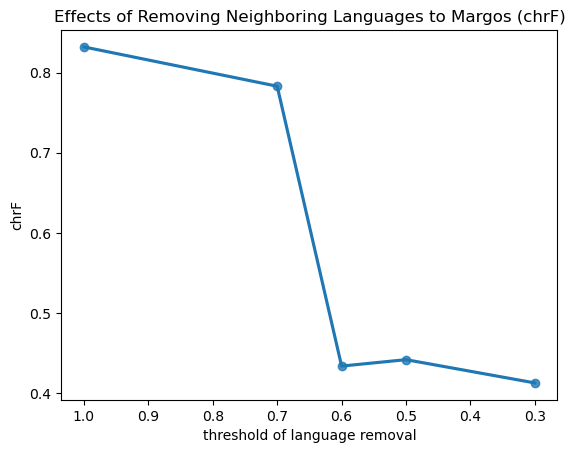}\hfill
    \includegraphics[width=.42\textwidth]{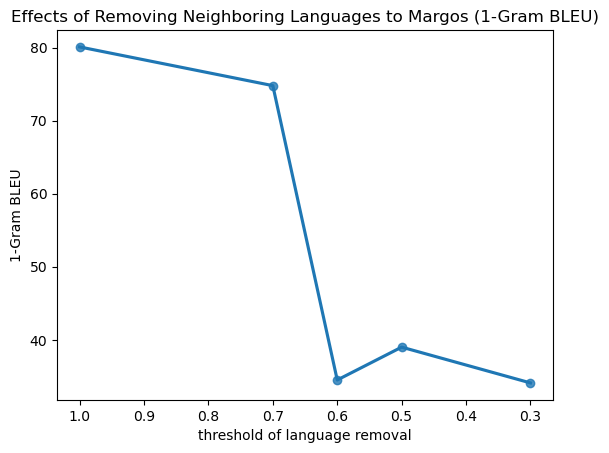}
    \\[\smallskipamount]
    \includegraphics[width=.41\textwidth]{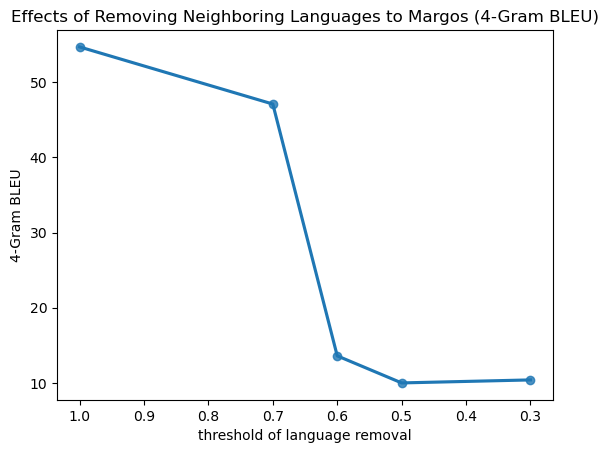}\hfill
    \includegraphics[width=.41\textwidth]{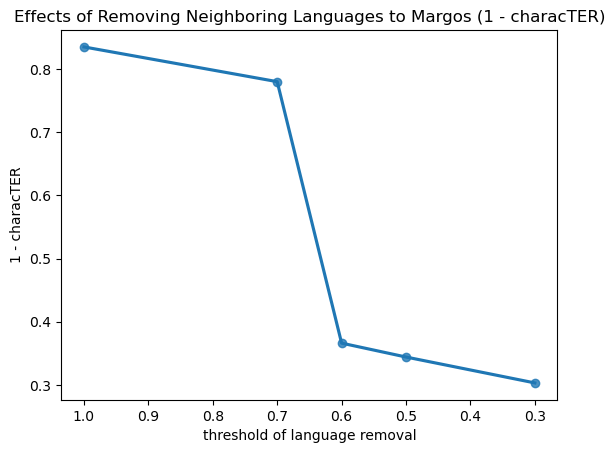}\hfill
    \caption{Effects of removing similar languages in translation into Margos}\label{fig:droplan}
\end{figure}

To determine the statistical significance, we take the column-wise Spearman correlation between the performance matrix and the similarity matrix and show key results in Table \ref{table:correlation}. For the chrF metric, we have a correlation of 0.771 between the performance matrix and the similarity matrix. The associated p-value is $4.25 \times 10^{-25}$, which shows high statistical significance. A detailed analysis of Spearman correlation in all 4 metrics (chrF, characTER, 1-gram BLEU, 4-gram BLEU) shows extremely significant p-values and high correlation scores in Table \ref{table:correlation}. Indeed, this shows that the more well-connected the target language is, the higher the translation performance. 

Having established the strong positive correlation between performance and similarity, we show a more fine-grained comparison in Figure \ref{fig:correlation_matrices_all}. The last column of this figure shows the fine-grained correlation and p-value by each source language. We see that there is strong positive correlation for languages close to the 6-language cluster while there is little correlation for languages that are already far away from the other languages. This means the more connected a language is, the more language similarity correlates with translation performance. Moreover,  the more connected a language is, the better the performance.

\subsubsection{Effects of Adding and Removing Similar Languages}
Having established that the more connected a language is, the better the translation performance, we now examine the relationship between translation performance and language similarity from another perspective. In Figure \ref{fig:similarity_chrf_12lan}, we show the degree of connection between languages in varying shades of green. The darker green represents high similarity and the lighter green represents low similarity. Using the varying similarity scores, we visualize how similarity correlates with performance by measuring performance while adding or removing similar languages. 

We choose Margos as the target language to serve as the control, and rank other languages according to their similarity to Margos. As we drop or add similar languages, we measure the performance of our method with a varying number of source languages. We show the effects of adding languages in Figure \ref{fig:addlan} and that of removing languages in Figure \ref{fig:droplan}. The x-axis is the chrF similarity threshold. For example, a threshold of 0.5 in Figure \ref{fig:addlan} means that the corresponding experiment only trains on source languages that achieves at least a chrF score of 0.5. On the contrary, a threshold of 0.5 in Figure \ref{fig:droplan} means that the corresponding experiment only trains on source languages that achieve a chrF score less than 0.5. 

The optimal threshold for translating into Margos by adding languages is noticeably 0.5 chrF as shown in Figure \ref{fig:addlan}, meaning that it is optimal to train on source languages that are closer to the target low-resource language. Moreover, adding more distant languages hurts translation performance. 

Additionally, we find similar results by removing languages in Figure \ref{fig:droplan}. If we were to train on source languages that achieve a chrF score less than 0.7, we still maintain good performance. However, once we move to 0.6, the performance plunges. This shows how crucial source languages that are closer to the low-resource language are. Therefore, it is important to keep these languages instead of removing them. 

Both sets of experiments point to the same conclusion that choosing languages that are very similar to the target language is key in achieving high translation performance. Furthermore, decluttering poorly-connected languages helps to improve translation. 

\subsubsection{Understanding Poorly-Connected Languages} 
Having established that the more connected a language is, the higher the translation performance, it is easy to understand why our multi-stage adaptation method works very well with highly similar Quechuan languages as shown in Table \ref{table:roundrobin}. For example, the experiment translating into Margos reaches a BLEU score of 54.7 and a chrF score of 83.2. However, in the same round robin experiments in Table \ref{table:roundrobin}, we also observe that 
our multi-stage adaptation method does not work as well with a few poorly-connected Quechuan languages that are very different from the rest. For instance, the experiment translating into Cuzco reaches a BLEU score of 4.7 and a chrF score of 39.9. 

To understand this low performance with poorly-connected languages, let us focus on a more detailed comparison between the performance matrix and similarity matrix in Figure \ref{fig:correlation_matrices_all}. The last column of this figure shows the fine-grained correlation and p-value by each source language. \ref{fig:correlation_matrices_all}. The p-value is close to 0 for strongly-connected languages while the p-value is very high for poorly-connected languages. In other words, there is a significant positive correlation for strongly-connected languages while there is little correlation for poorly-connected languages. This means the more connected a language is, the more language similarity correlates with translation performance. Consequently, for poorly-connected languages, we cannot deduce correlation between performance and similarity, and their performance is therefore uncertain. 

One potential reason for such poor performance is that the advantage of multilinguality does not lend any leverage to the poorly-connected languages because generalization and cross-lingual learning is hard. Another reason is very related to the practice of the human translation team. In the past, translations are done through years. Each translation team may translate a set of languages together using the same conventions and notations. When a new translation team is formed, the new team may choose to use different notations and conventions. This may result in a lot changes as it may affect the language documentation process and could potentially add more difficulty in translating into languages that are already poorly-connected. 

One may ask, what could help translation into such poorly-connected languages? Does adding more data help? In our experiments with one of the poorly-connected languages, Panao Quechua, we initially have poor results. In order to help us have more data for training,  human translators translate more sentences and provide us with much more data for training. With more data, our translation result remains poor. This could be because when languages are far apart, multilinguality may not be realistically useful due to the limited cross-lingual transfer learning. 

\subsubsection{Beyond the Quechuan Family}
We have considered languages in the Quechuan family so far. However, we have not considered languages outside of the Quechuan family. Are there languages outside the Quechuan family that are similar to the target language that we may use for training? To visualize this, we measure similarity of 142 source languages to a given target language. We show language similarity rankings for Margos (Figure \ref{fig:similarityqm}), Panao (Figure \ref{fig:similarityqx}) and Sihuas (Figure \ref{fig:similarityqd}). 

We find that Aymara \citep{hardman1985aymara}, which is a South American native language spoken in a nearby region (on the Altiplano) close to Peru, has relatively visible similarity with Margos. However, the closest languages are still from the Quechuan language family. This means that even though some Quechuan languages are very different from one another, they are still closer to each other than those outside the Quechuan language family. Therefore, we still recommend finding similar languages within the Quechuan family for translating into any of the Quechuan low-resource languages. 

\begin{table*}[t]
  \small
  \centering
  \begin{tabularx}{\textwidth}{p{7cm}p{7cm}p{2cm}} 
    \toprule
    System Translation & Human Post-edits & Edits \\
    \midrule    
Tayta Diostam päyakö qamkuna kaqćhö kaykar Crisputa y Gayullata bautizanqäpita.
& Tayta Diostam\textbf{\textcolor{blue}{i}} päyakö qamkuna kaqćhö kaykar Crisputa\textbf{\textcolor{blue}{wan}} y Gayullata bautizanqäpita. & 2 \\
\textbf{\textcolor{blue}{Ts}}aymi mayqëkipis niyankimanku noqapa shutëćhö bautizakuyanqëkita.
& \textbf{\textcolor{blue}{Ch}}aymi mayqëkipis niyankimanku noqapa shutëćhö bautizakuyanqëkita. & 1  \\
\textbf{\textcolor{blue}{Ts}}aynöllami Estéfanas castallata bautizash\textbf{\textcolor{blue}{q}}ä. Tsaypita mastaqa manami pitapis bautizanqäta yarpäku.
& \textbf{\textcolor{blue}{Ch}}aynöllami Estéfanas castallata bautizash\textbf{\textcolor{blue}{k}}ä. Tsaypita mastaqa manami pitapis bautizanqäta yarpäku. & 2 \\
    \bottomrule
  \end{tabularx}
    \caption{Qualitative evaluation in translation into Sihuas. }
\label{table:quali_sihuas}
\end{table*}

\begin{figure}
    \centering
    \includegraphics[width=.6\textwidth]{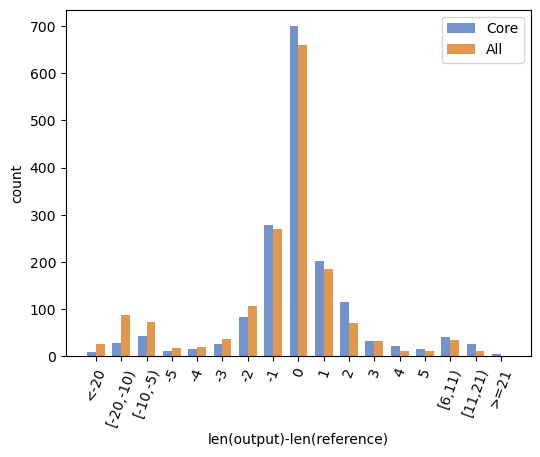}
    \caption{Output and reference length difference for two systems translating to Sihuas using compare-MT \citep{neubig2019compare}. The blue system translates using languages that are at least 0.6 chrF with Sihuas, while the orange system trains on all. }
    \label{fig:compareMT}
\end{figure}

\begin{figure*}[t]
    \centering
    \begin{subfigure}{0.3\textwidth}
        \includegraphics[height=.95\textheight]{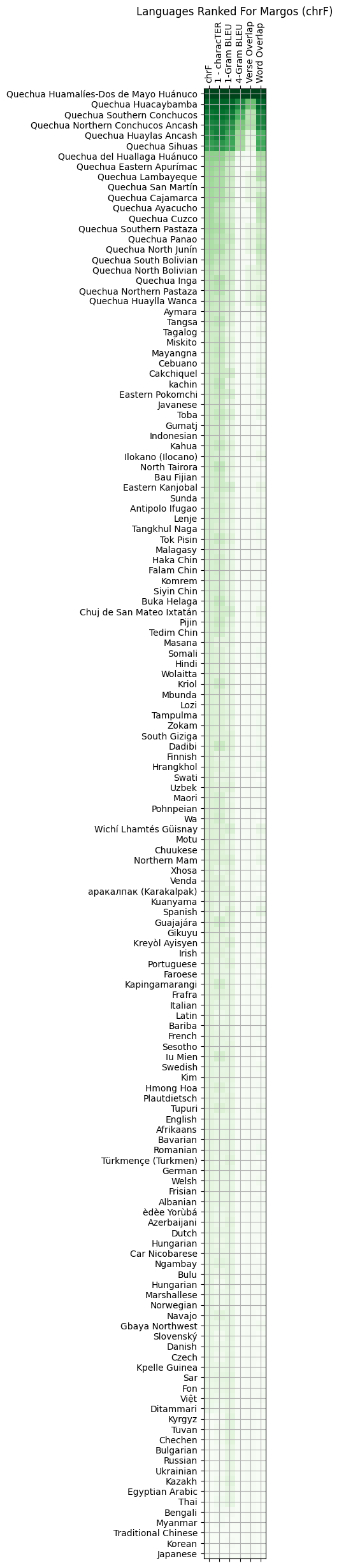}
        \caption{Margos.}
        \label{fig:similarityqm}
    \end{subfigure}
    \hfill
    \begin{subfigure}{0.3\textwidth}
        \includegraphics[height=.95\textheight]{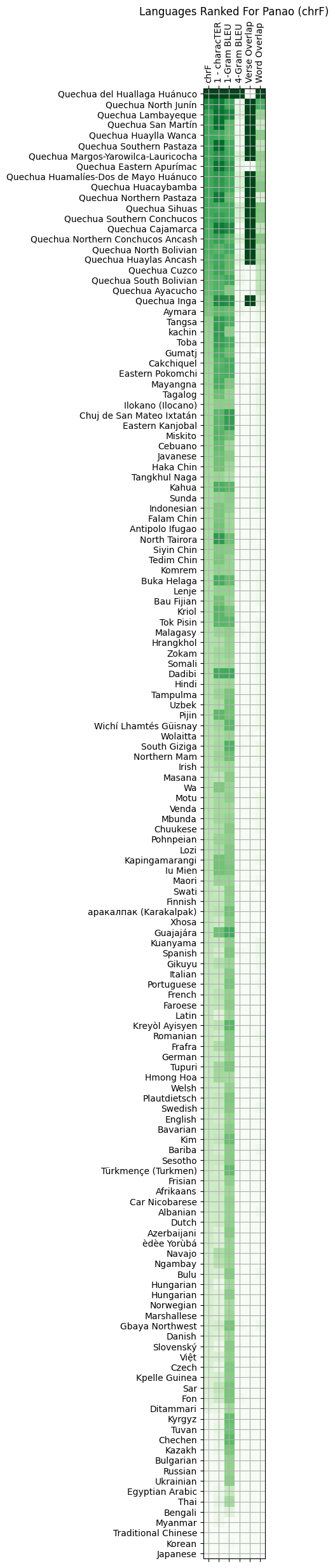}
        \caption{Panao.}
        \label{fig:similarityqx}
    \end{subfigure}
    \hfill
    \begin{subfigure}{0.3\textwidth}
        \includegraphics[height=.95\textheight]{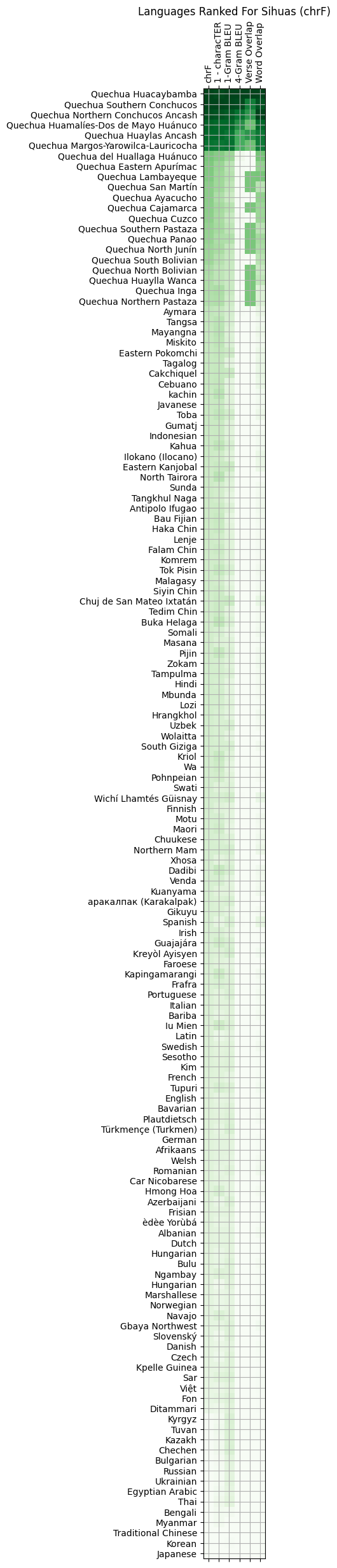}
        \caption{Sihuas.}
        \label{fig:similarityqd}
    \end{subfigure}
    \caption{Language rankings by similarity}
    \label{fig:similarity_combined}
\end{figure*}

\begin{figure*}[t]
    \centering
    \begin{subfigure}{0.3\textwidth}
        \includegraphics[height=.95\textheight]{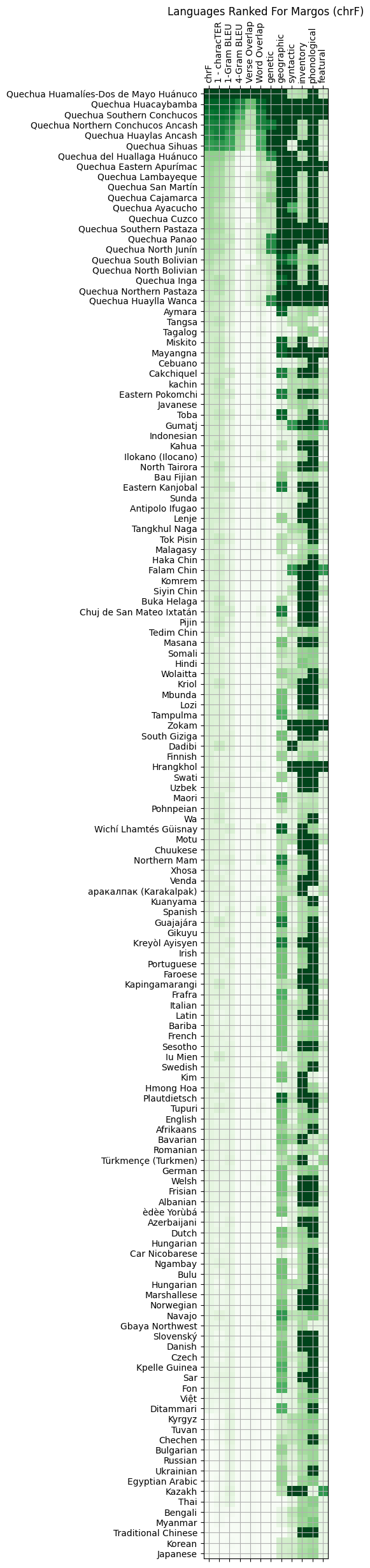}
        \caption{Margos.}
        \label{fig:similarity_withlan2vec_qm}
    \end{subfigure}
    \hfill
    \begin{subfigure}{0.3\textwidth}
        \includegraphics[height=.95\textheight]{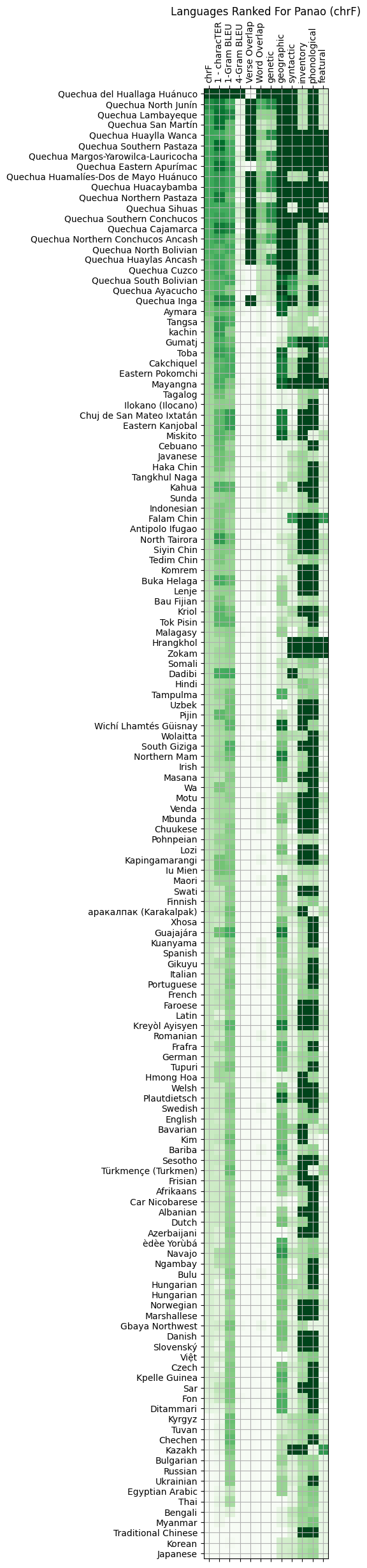}
        \caption{Panao.}
        \label{fig:similarity_withlan2vec_qx}
    \end{subfigure}
    \hfill
    \begin{subfigure}{0.3\textwidth}
        \includegraphics[height=.95\textheight]{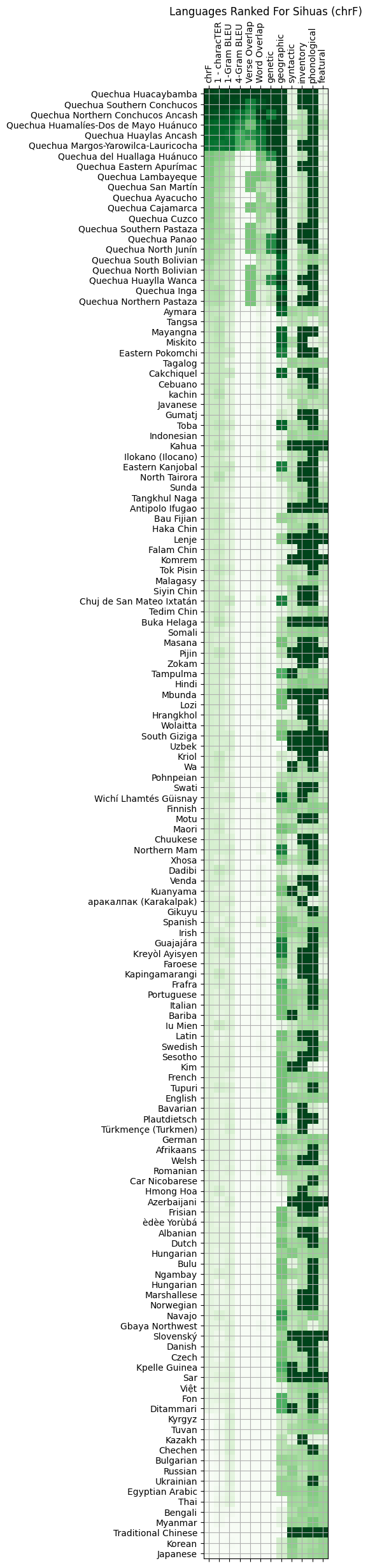}
        \caption{Sihuas.}
        \label{fig:similarity_withlan2vec_qd}
    \end{subfigure}
    \caption{Language rankings by similarity with typological features. }
    \label{fig:similarity_combined_withlan2vec}
\end{figure*}

\begin{figure}
    \includegraphics[width=.43\textwidth]{similarity_22lan_chrfs.png}\hfill
    \includegraphics[width=.43\textwidth]{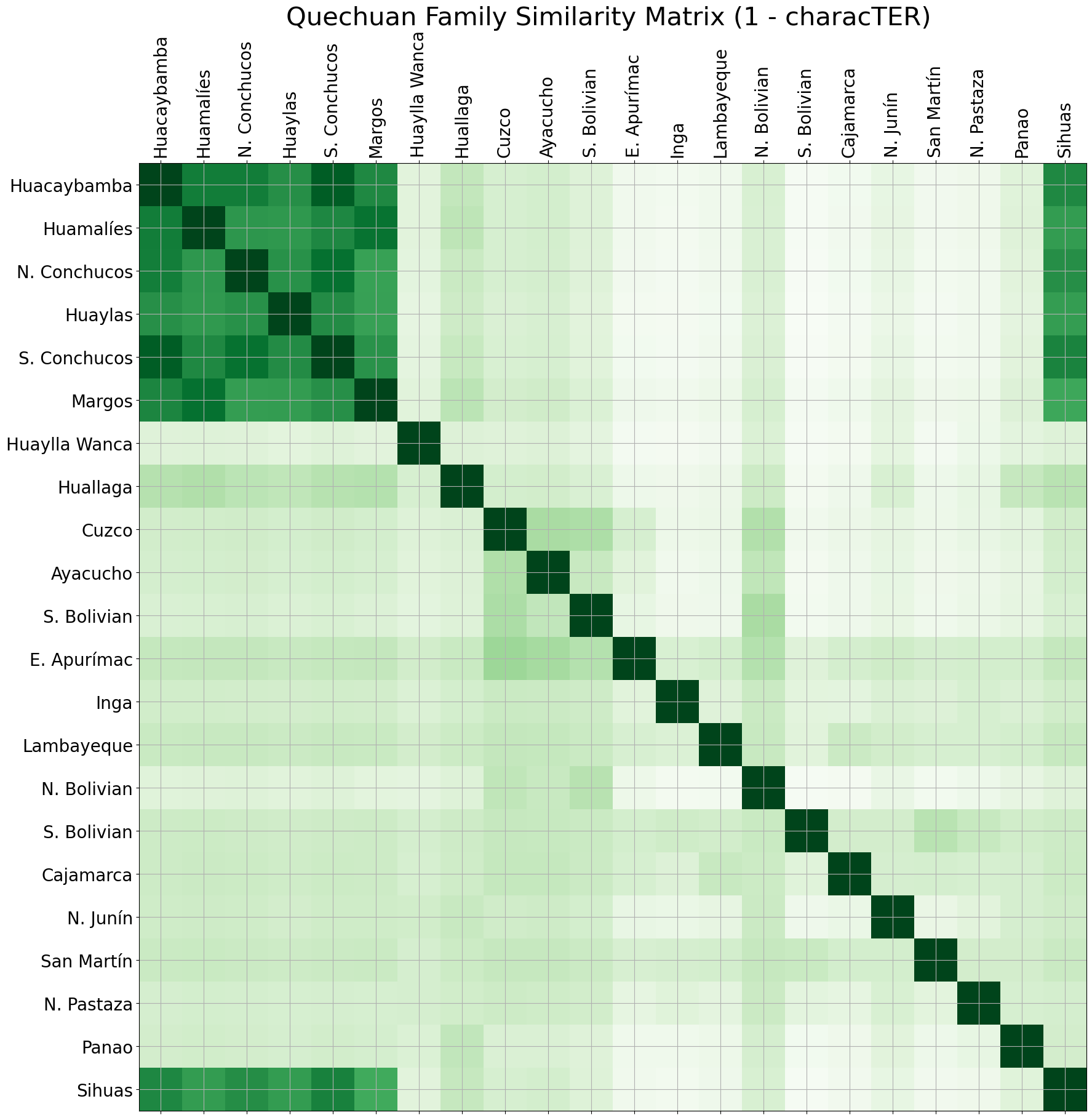}
    \\[\smallskipamount]
    \includegraphics[width=.43\textwidth]{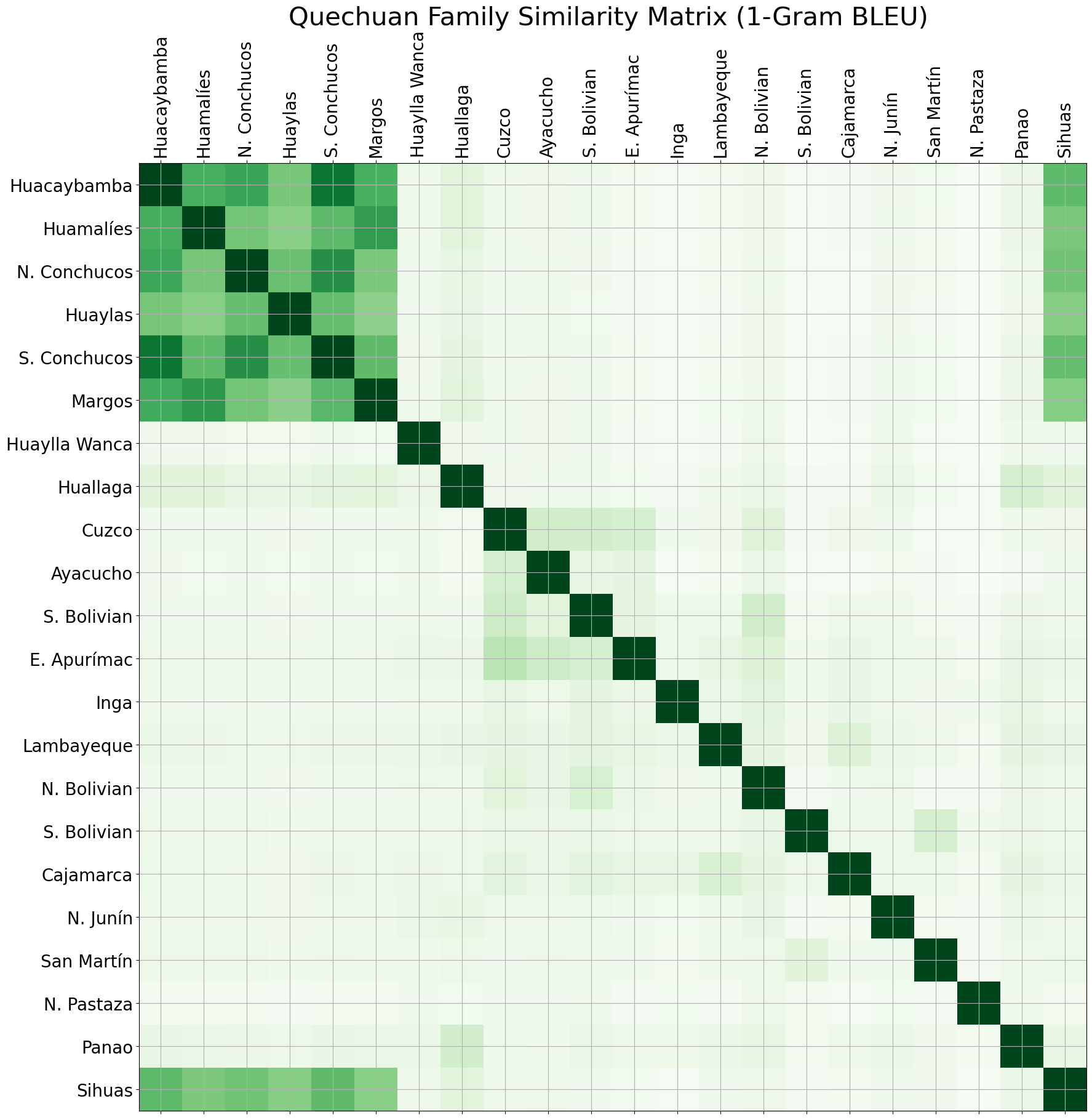}\hfill
    \includegraphics[width=.43\textwidth]{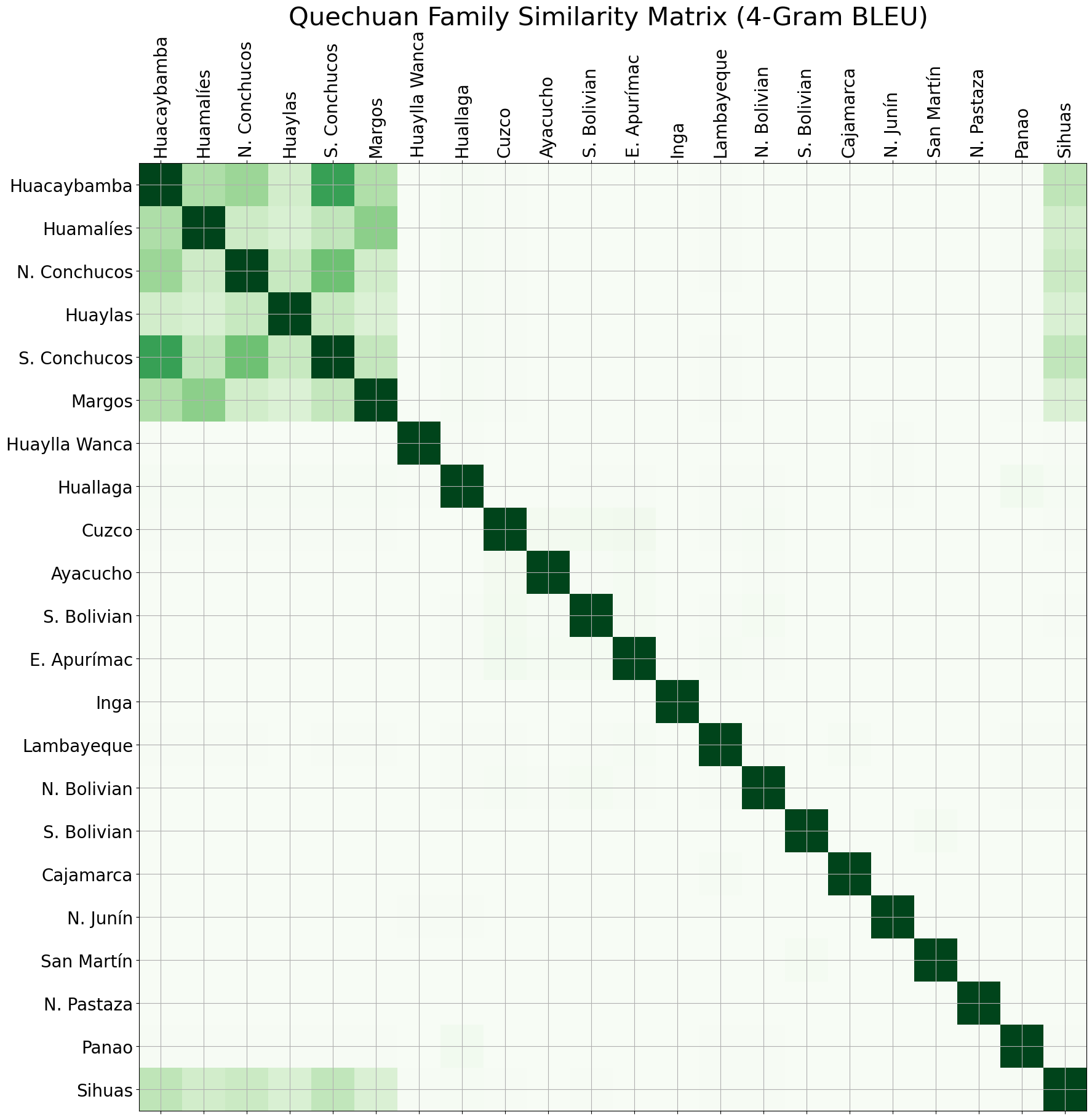}\hfill
    \\[\smallskipamount]
    \includegraphics[width=.43\textwidth]{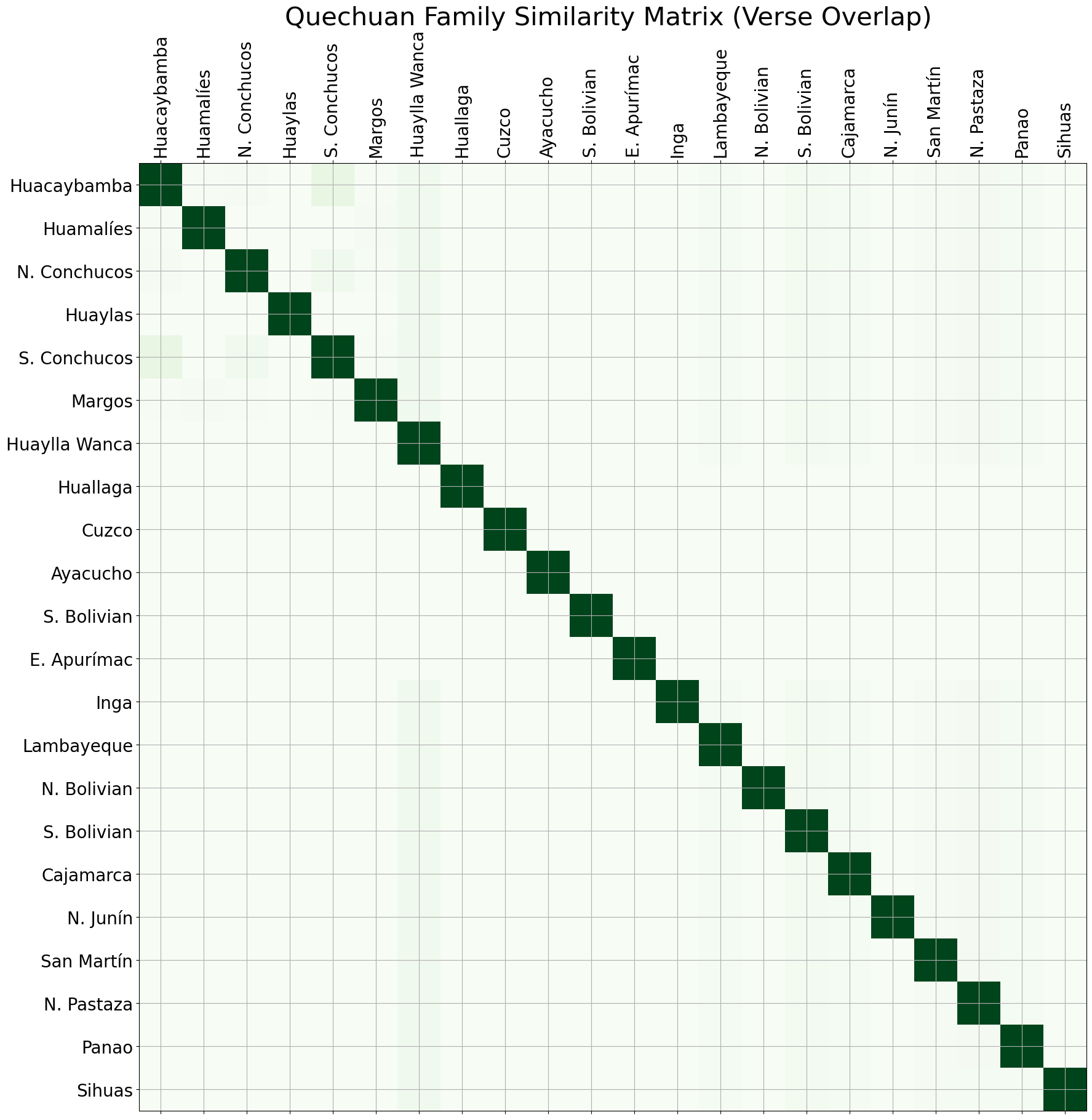}\hfill
    \includegraphics[width=.43\textwidth]{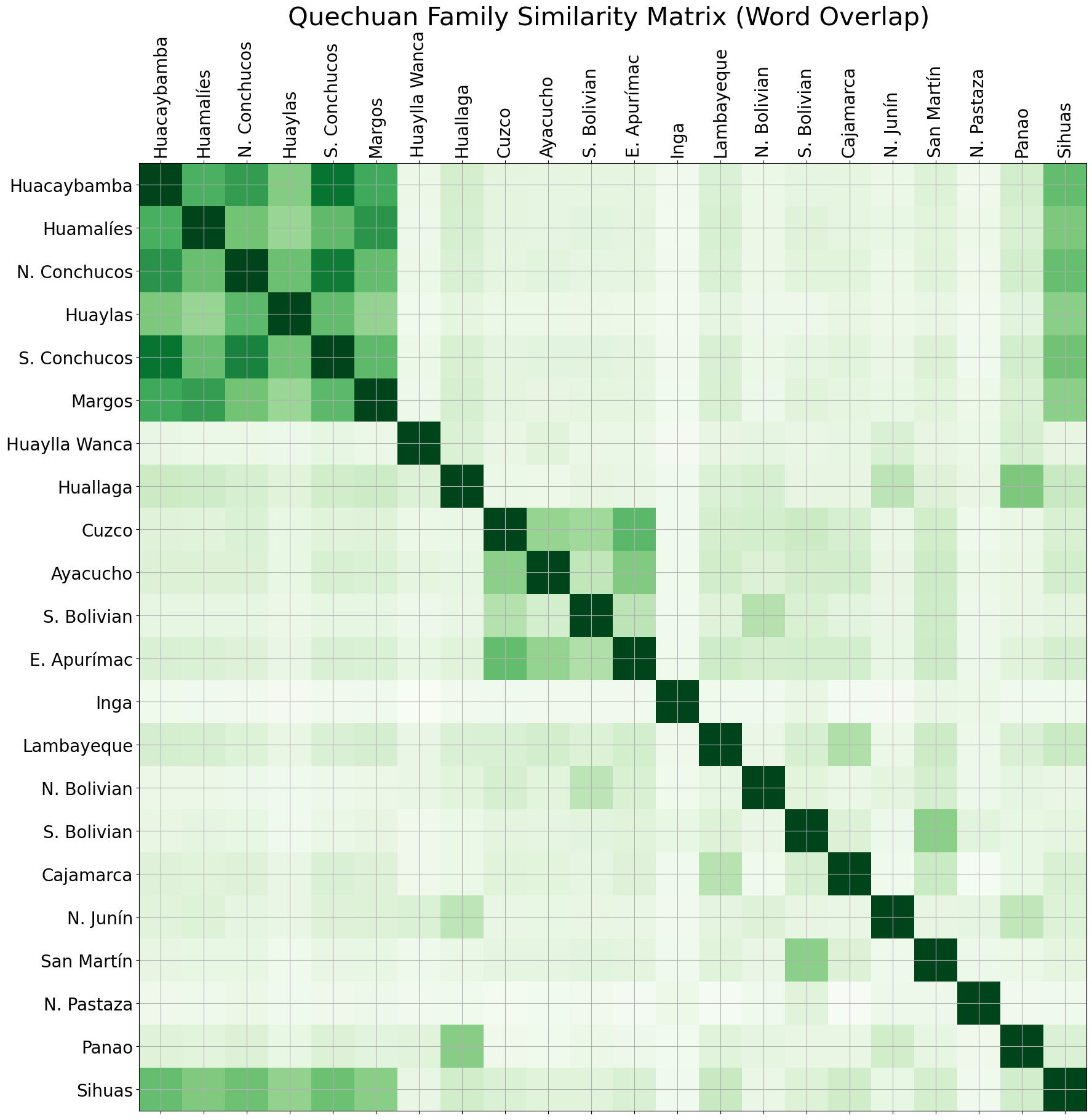}
    \caption{Similarity matrix based on chrF, characTER, 1-gram BLEU, 4-gram BLEU, sentence overlap and word overlap. }\label{fig:similarity_all_22lan}
\end{figure}
\begin{figure}
    \includegraphics[width=.43\textwidth]{similarity_chrfs.png}\hfill
    \includegraphics[width=.43\textwidth]{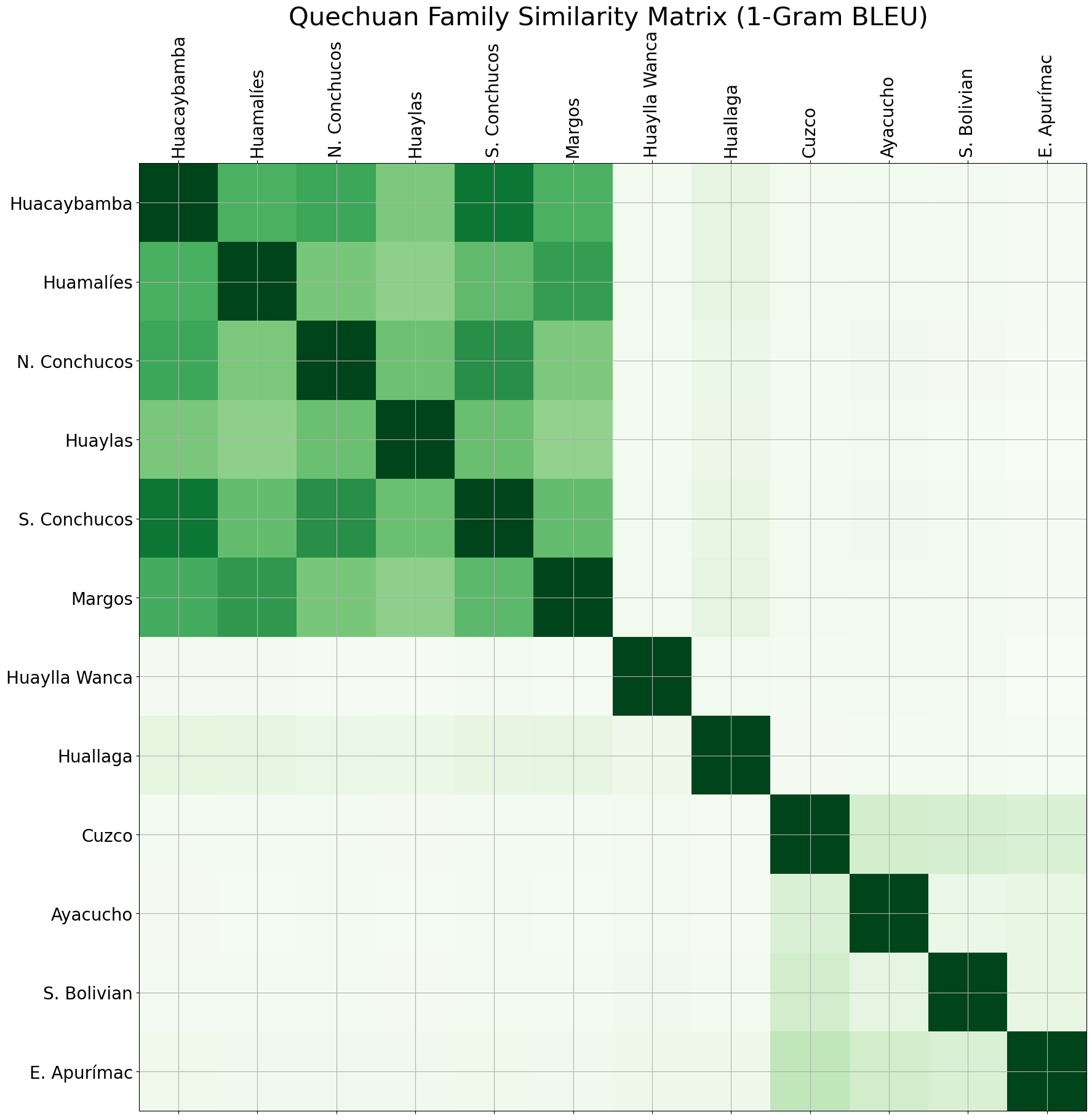}
    \\[\smallskipamount]
    \includegraphics[width=.43\textwidth]{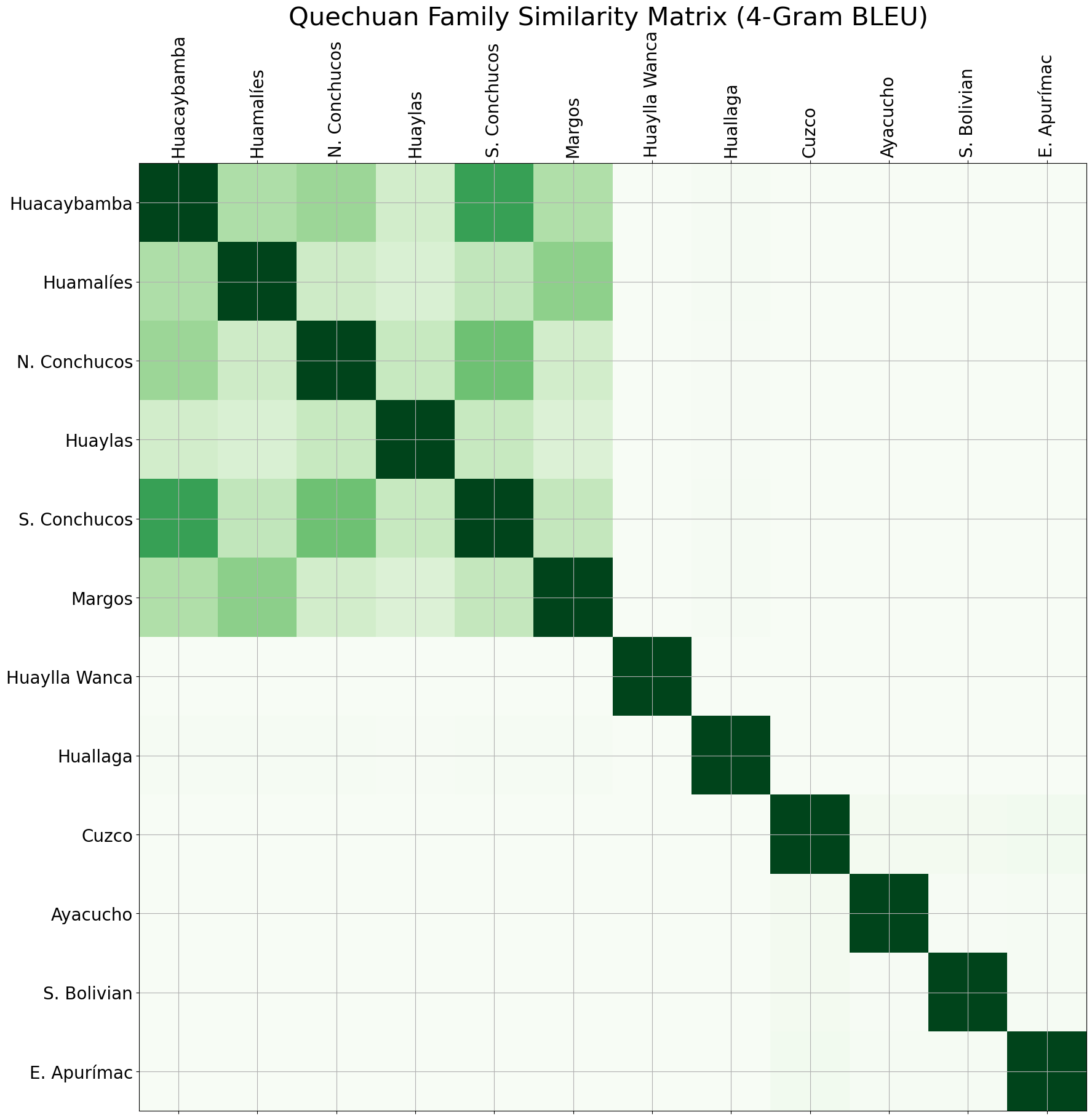}\hfill
    \includegraphics[width=.43\textwidth]{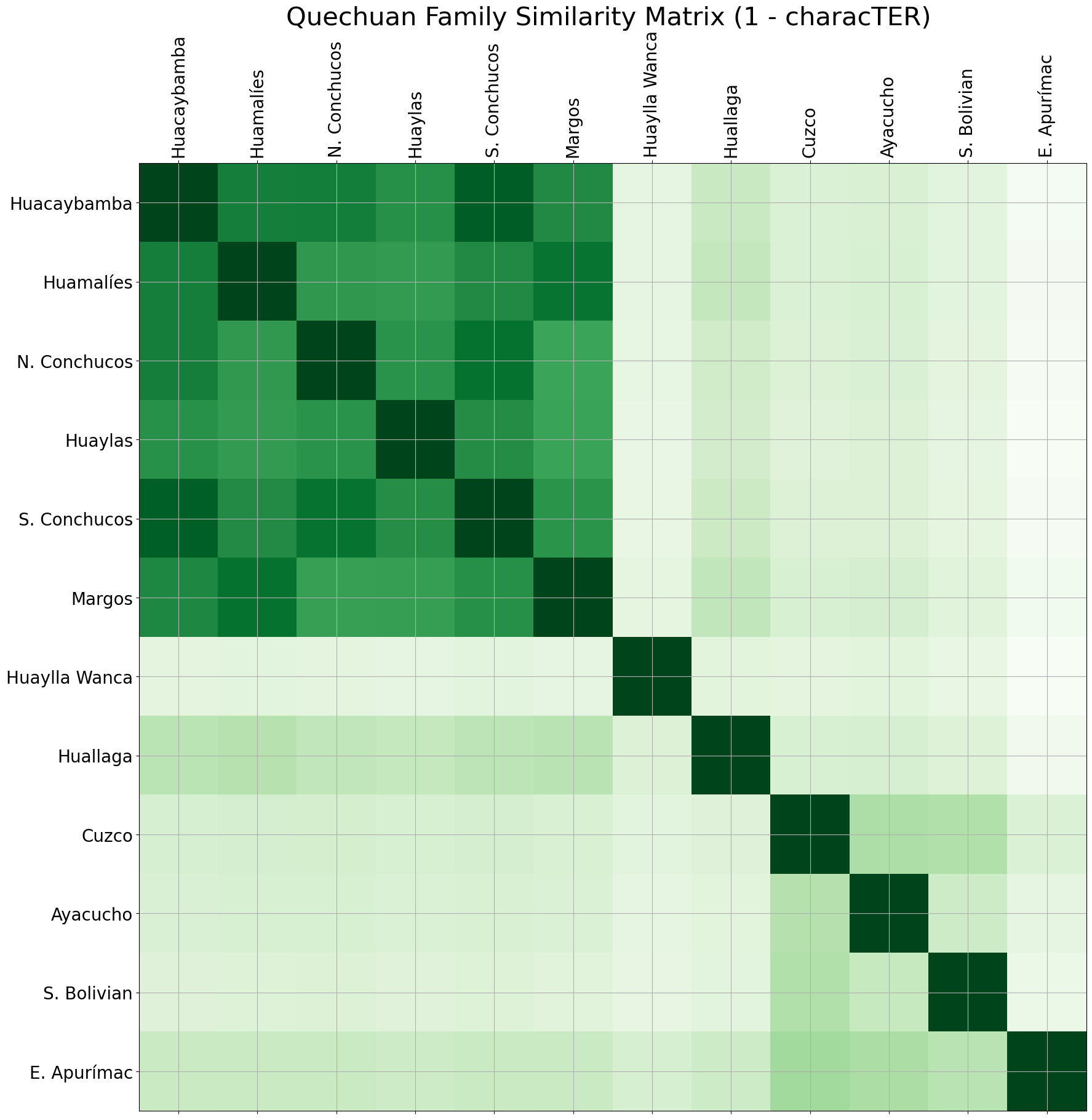}
    \\[\smallskipamount]
    \includegraphics[width=.43\textwidth]{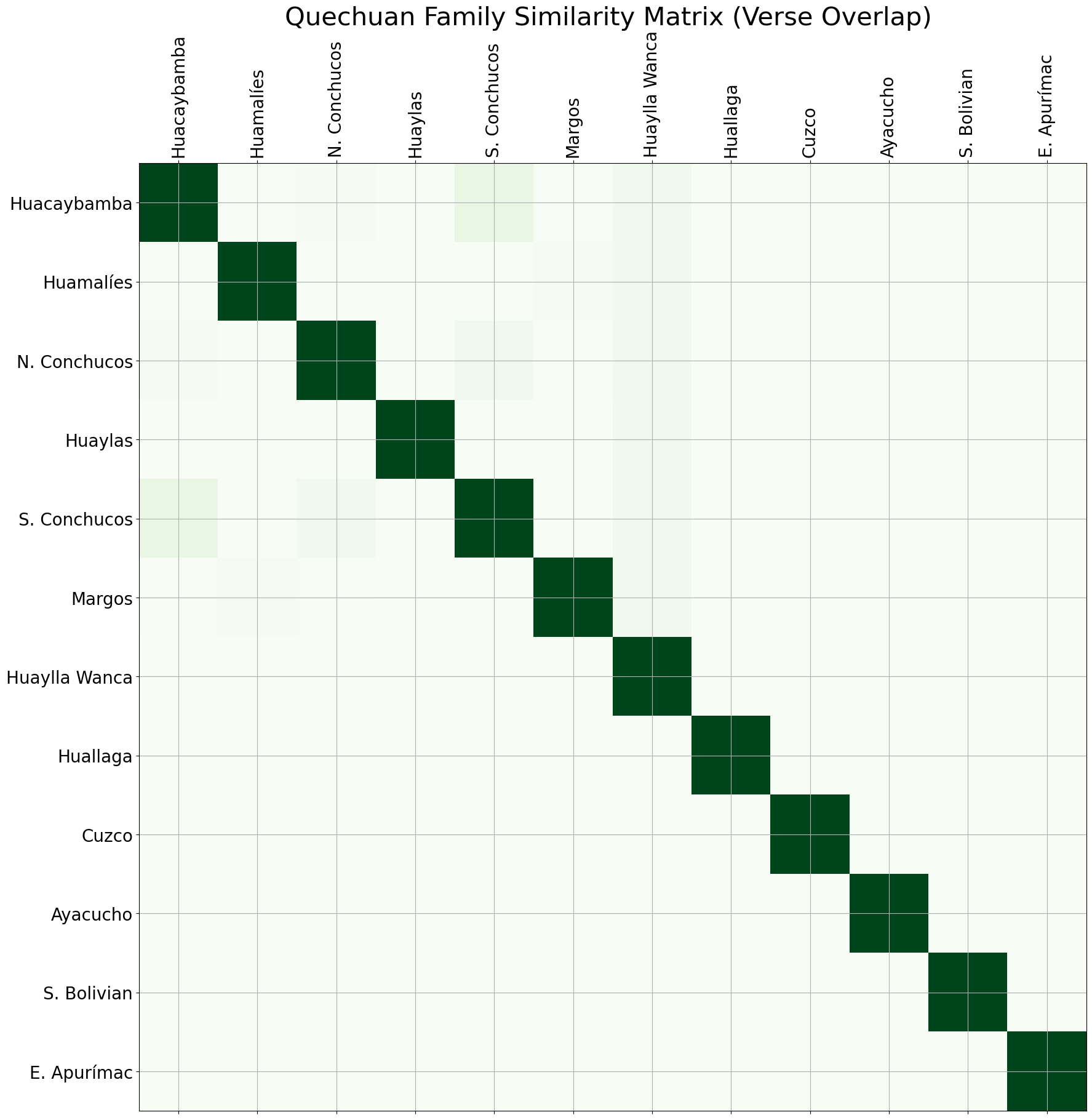}\hfill
    \includegraphics[width=.43\textwidth]{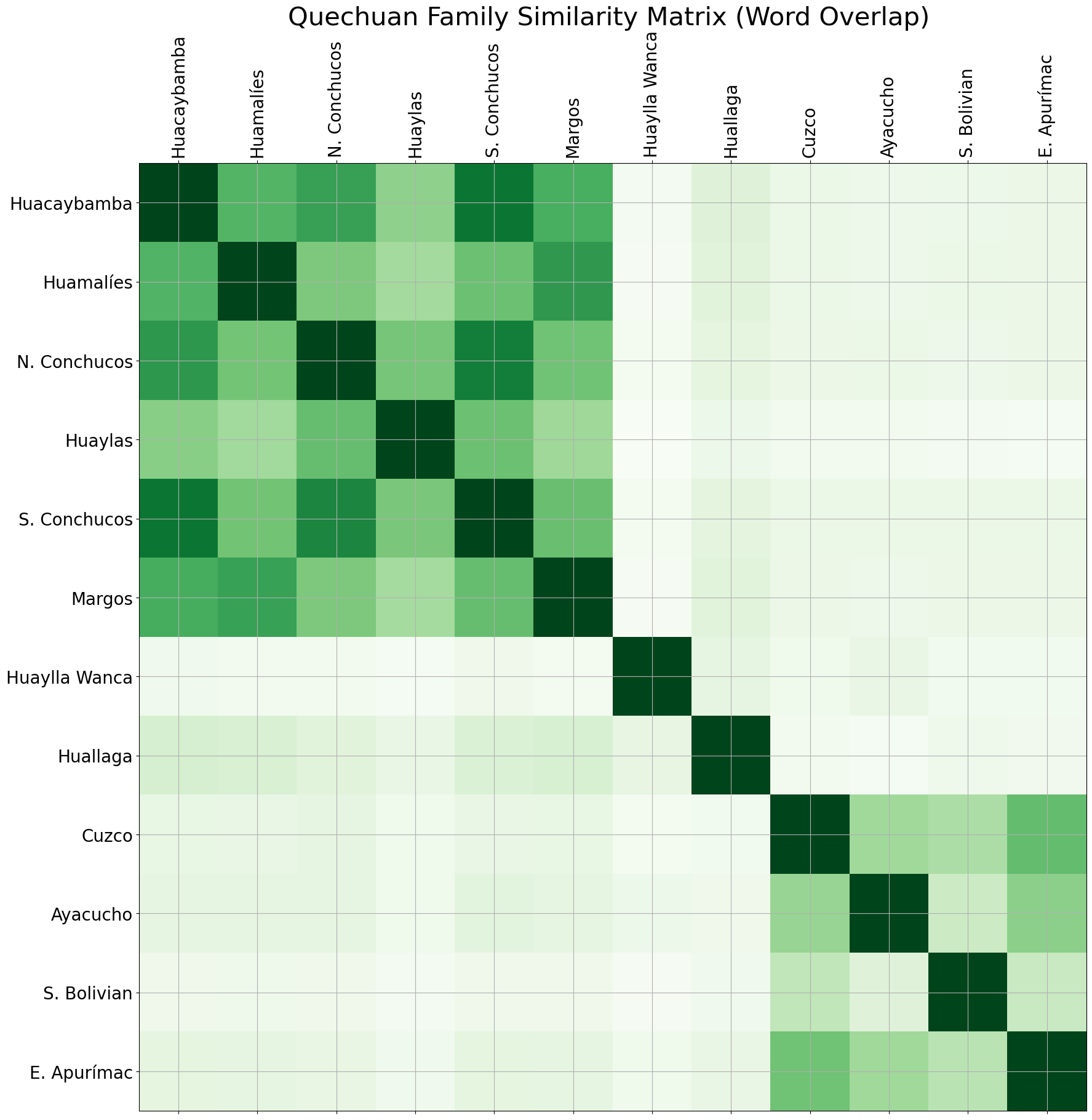}
    \caption{Similarity Matrix for 12 Quechuan languages (zooming into 12 main Quechuan languages in Figure~\ref{fig:similarity_all_22lan}). }\label{fig:similarity_12lan_all}
\end{figure}
\begin{figure}
    \includegraphics[width=.24\textwidth]{similarity_chrfs.png}\hfill
    \includegraphics[width=.24\textwidth]{performance_chrfs.png}\hfill
    \includegraphics[width=.14\textwidth, height=.15\textheight]{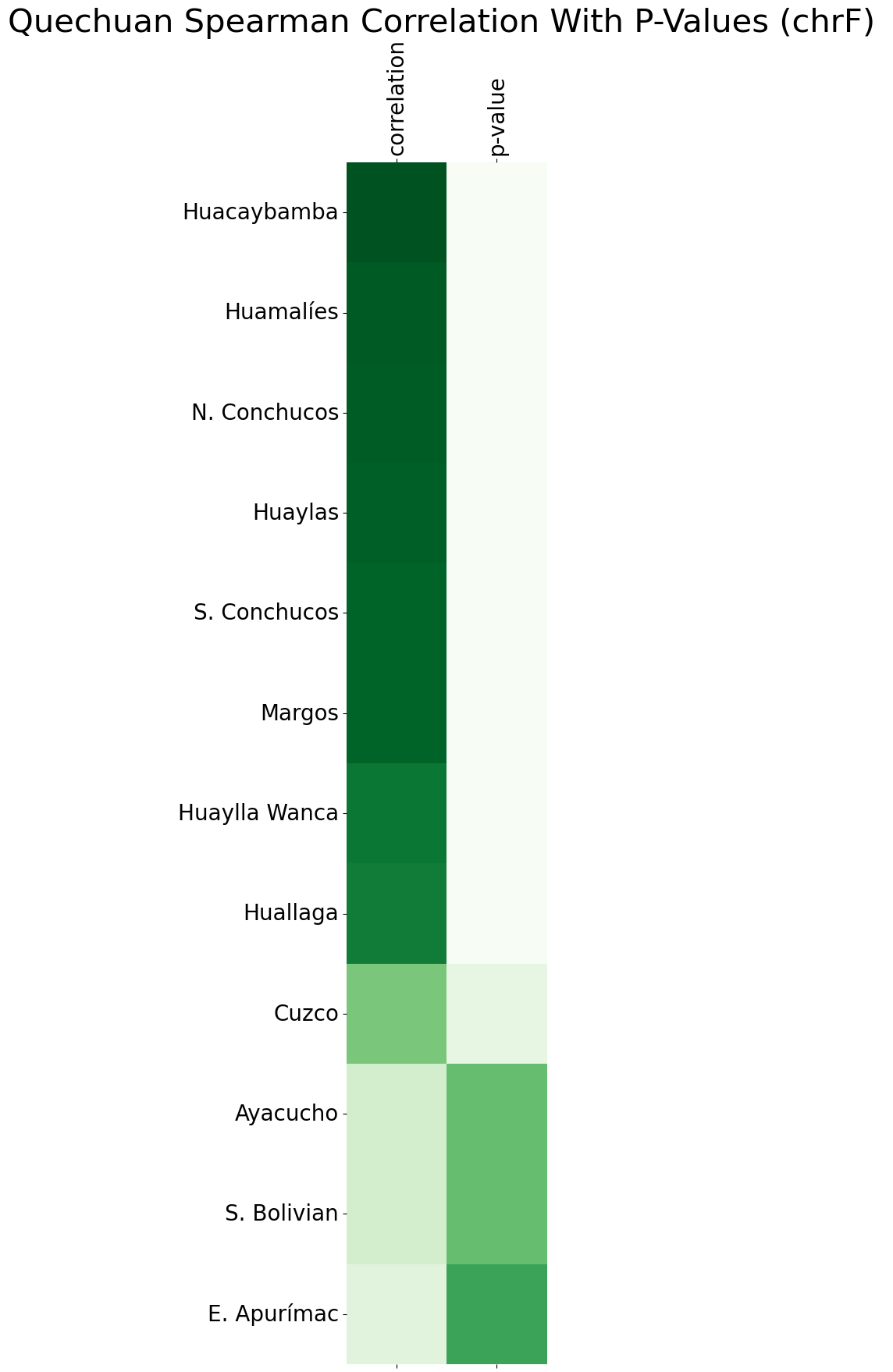}
    \\[\smallskipamount]
    \includegraphics[width=.24\textwidth]{similarity_characTER.png}\hfill
    \includegraphics[width=.24\textwidth]{similarity_characTER.png}\hfill
    \includegraphics[width=.14\textwidth, height=.15\textheight]{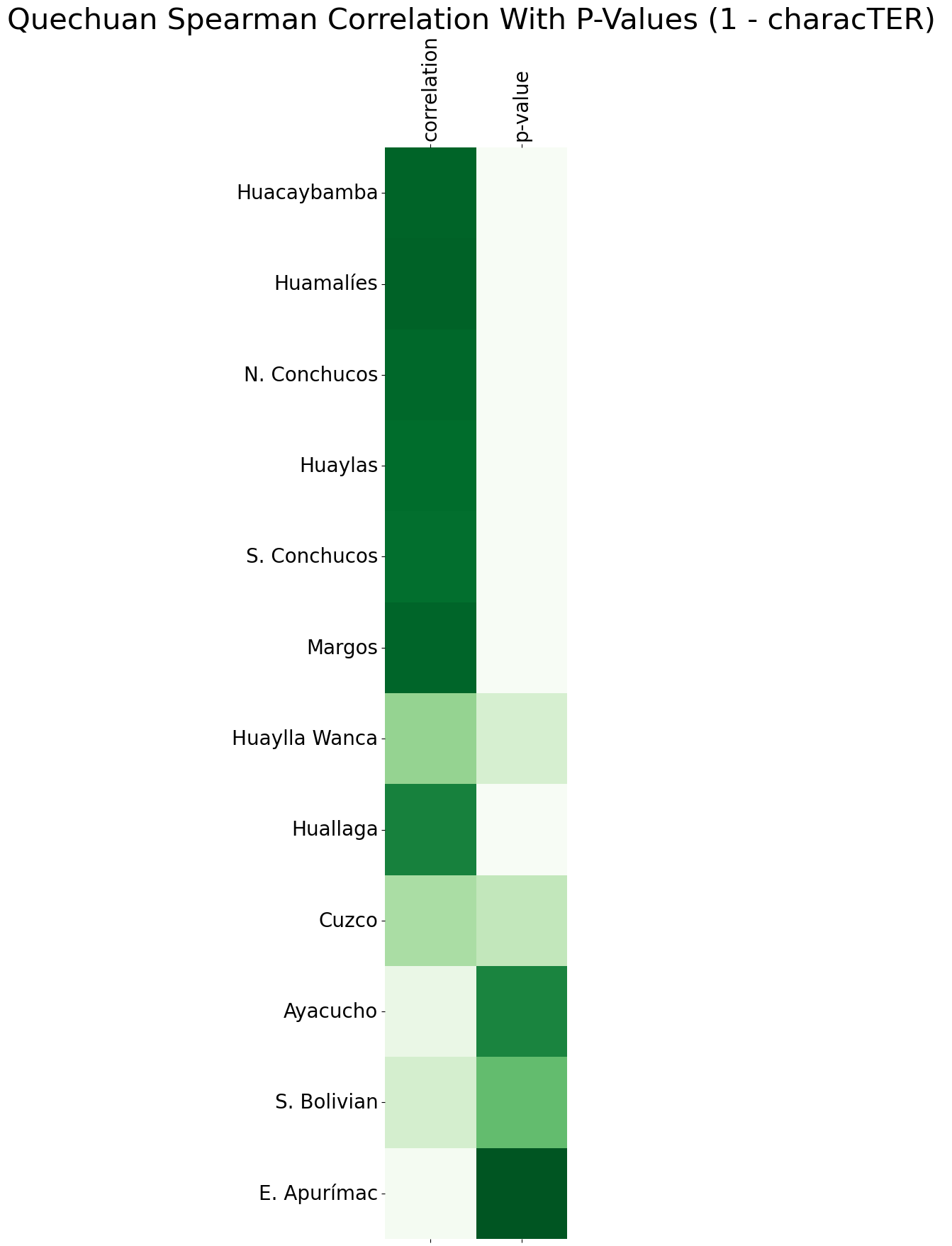}
    \\[\smallskipamount]   
    \includegraphics[width=.24\textwidth]{similarity_1bleus.png}\hfill
    \includegraphics[width=.24\textwidth]{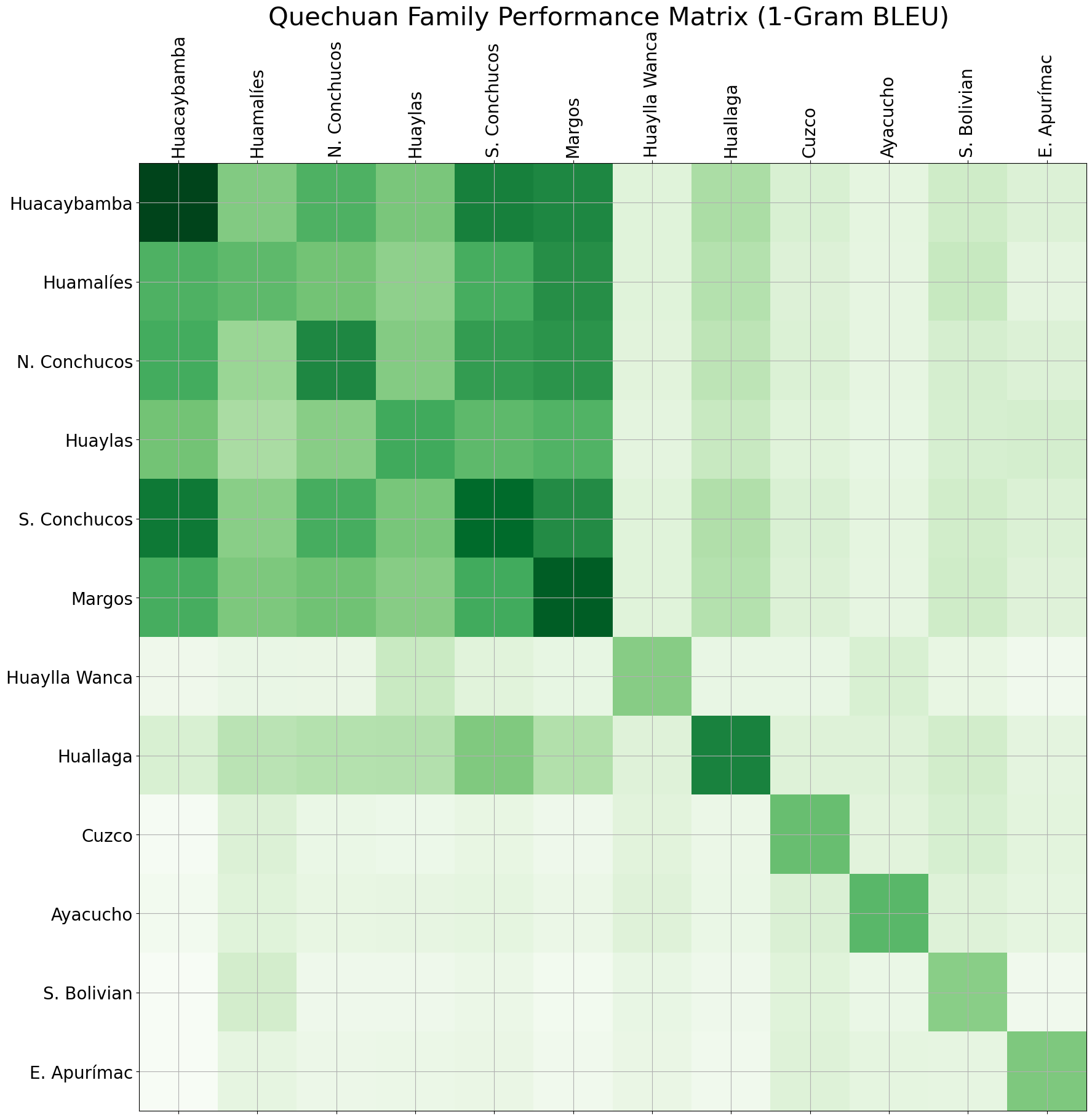}\hfill
    \includegraphics[width=.14\textwidth, height=.15\textheight]{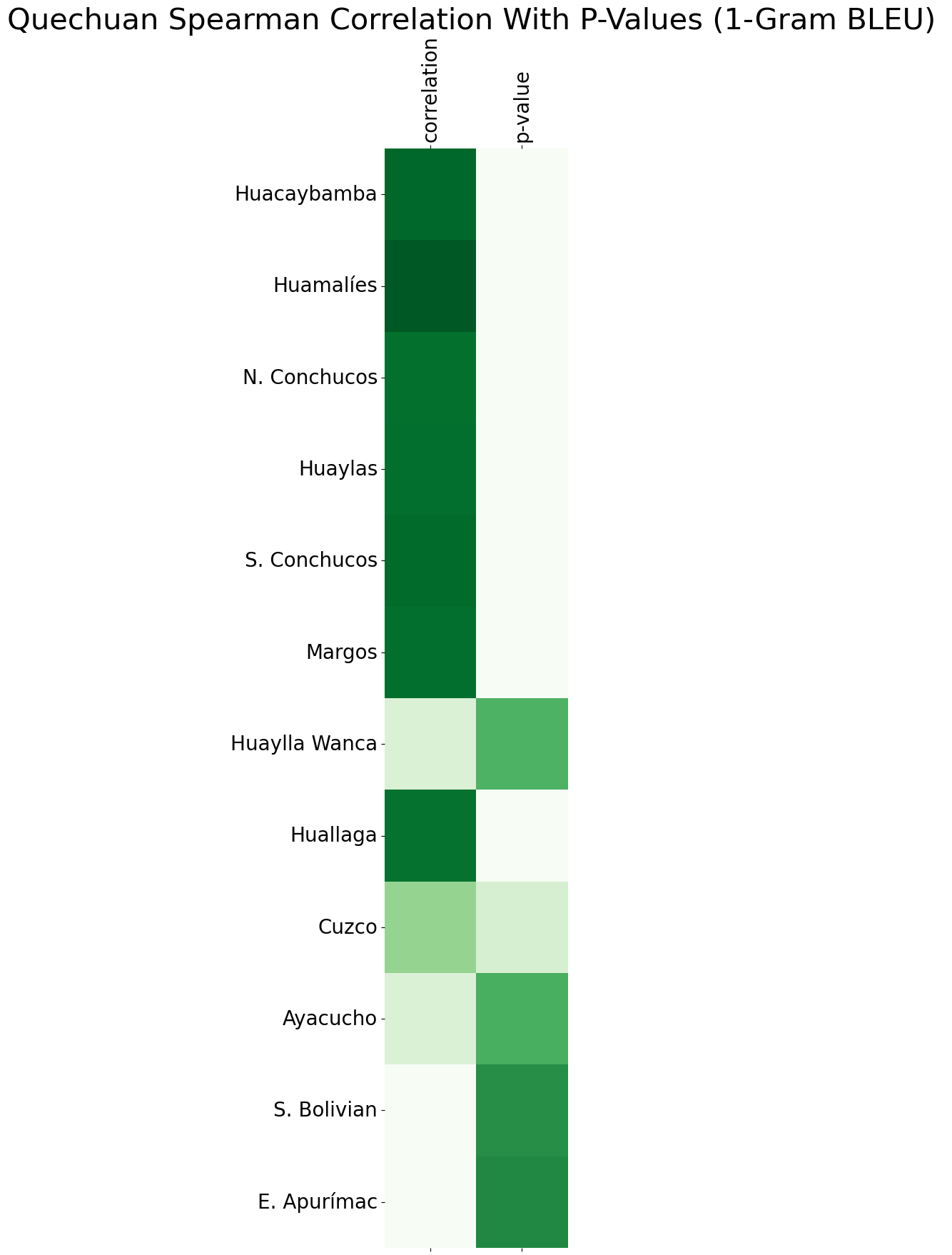}
    \\[\smallskipamount]
    \includegraphics[width=.24\textwidth]{similarity_4bleus.png}\hfill
    \includegraphics[width=.24\textwidth]{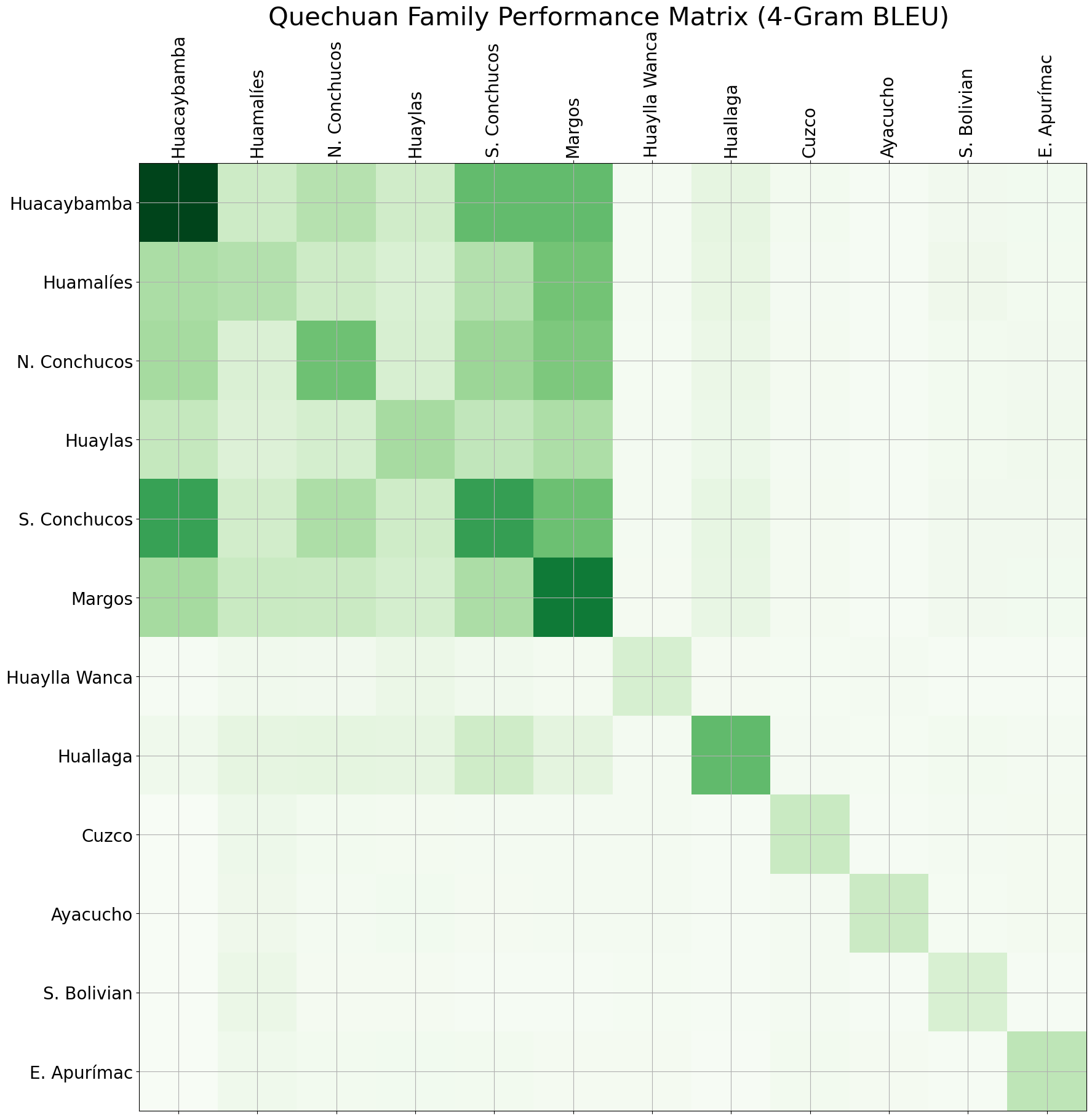}\hfill
    \includegraphics[width=.14\textwidth, height=.15\textheight]{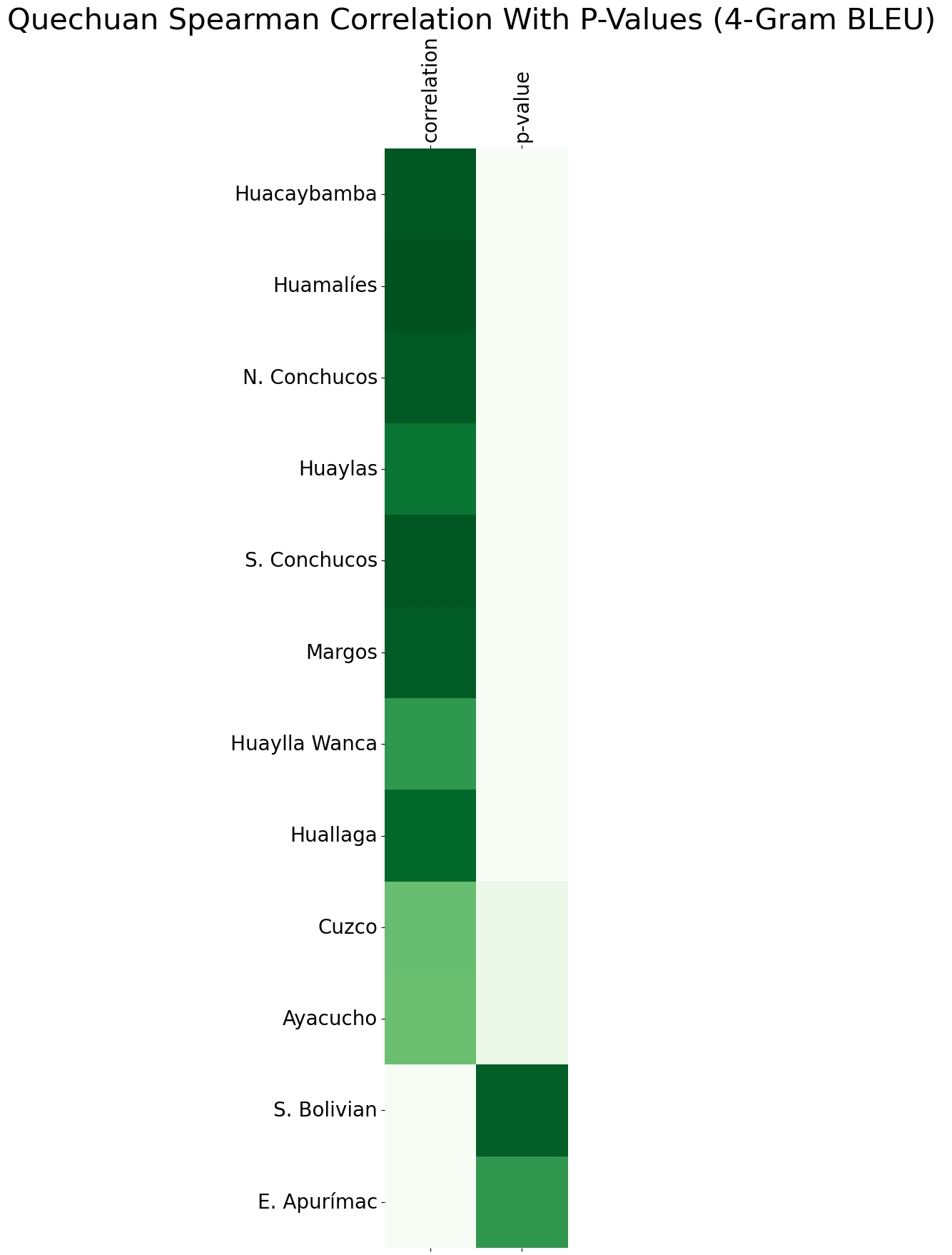} 
    \caption{Complete comparison similarity and performance matrices using chrF, characTER, 1-gram BLEU and 4-gram BLEU. The last column shows fine-grained correlation and p-value for each source language. }\label{fig:correlation_matrices_all}
\end{figure}

\begin{figure}
    \includegraphics[width=.43\textwidth]{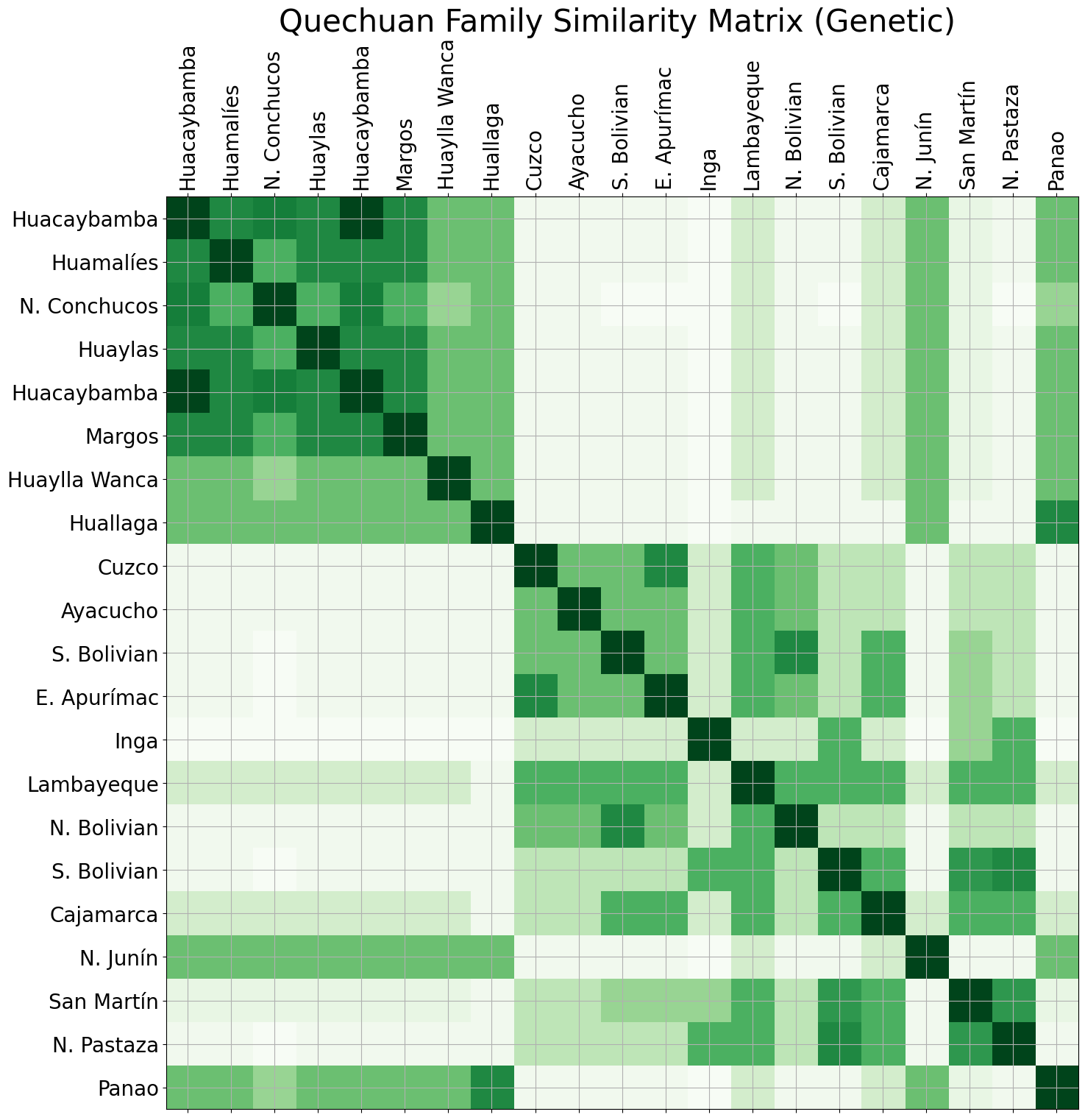}\hfill
    \includegraphics[width=.43\textwidth]{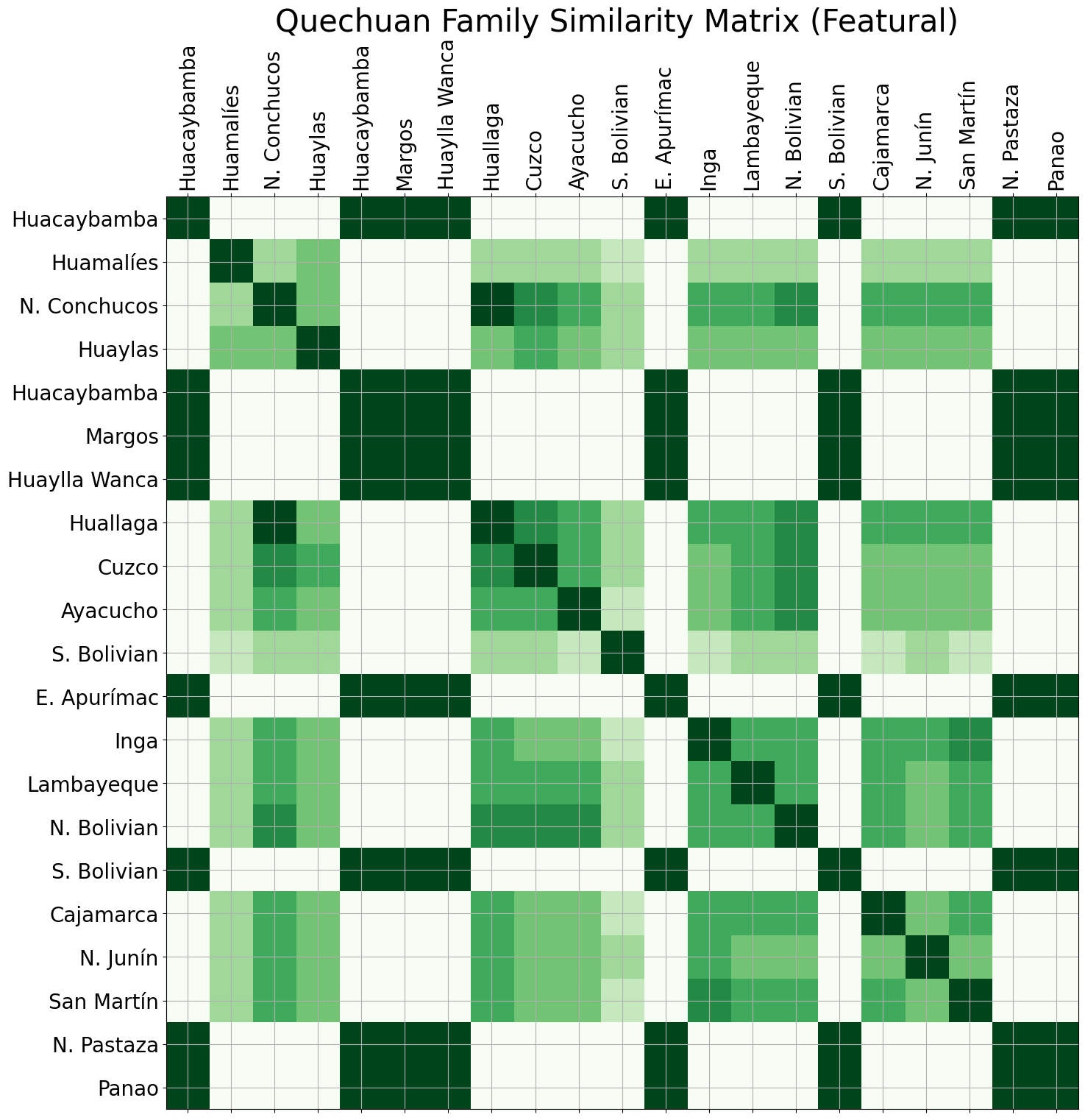}\hfill
    \\[\smallskipamount]
    \includegraphics[width=.43\textwidth]{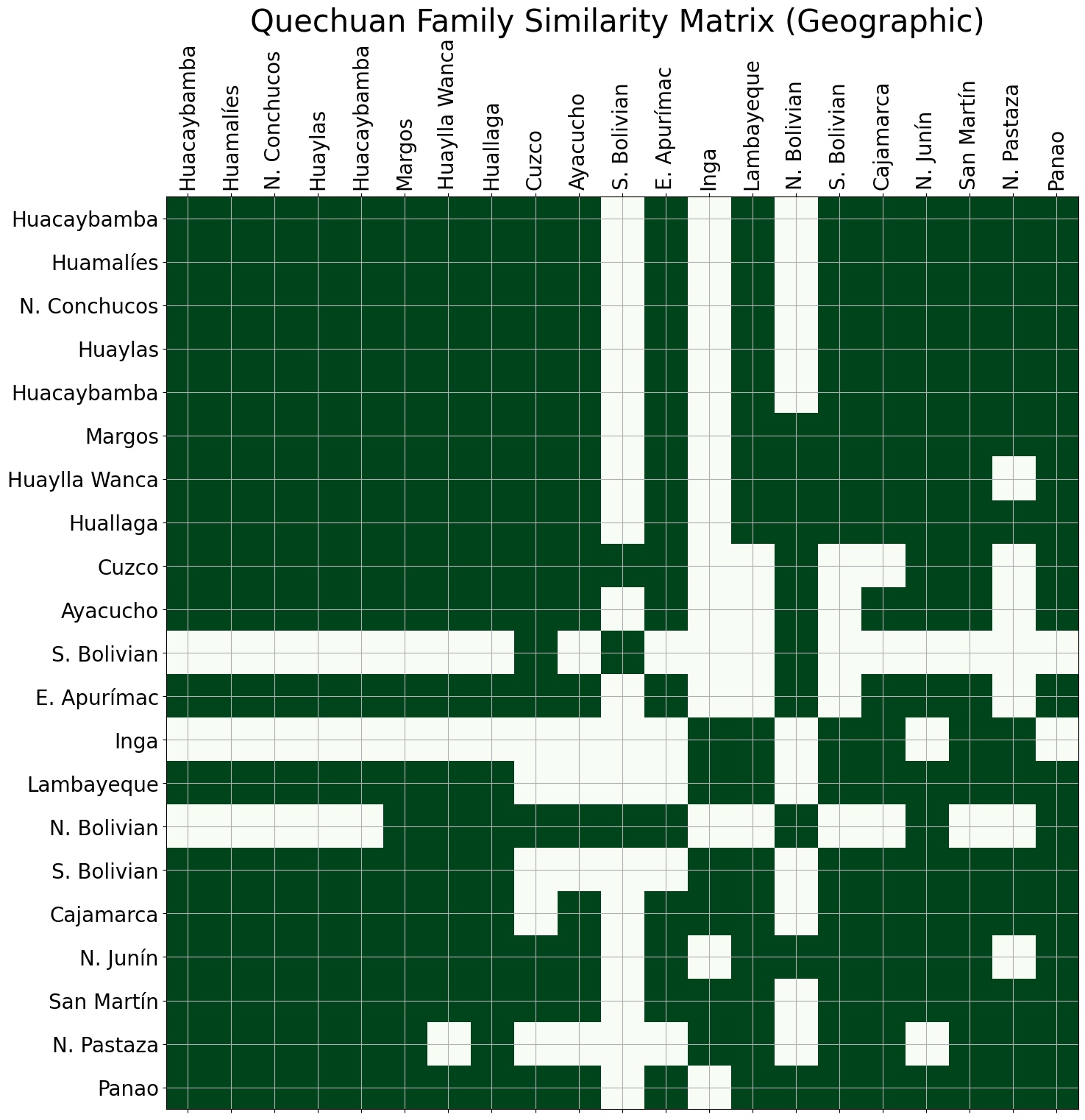}\hfill
    \includegraphics[width=.43\textwidth]{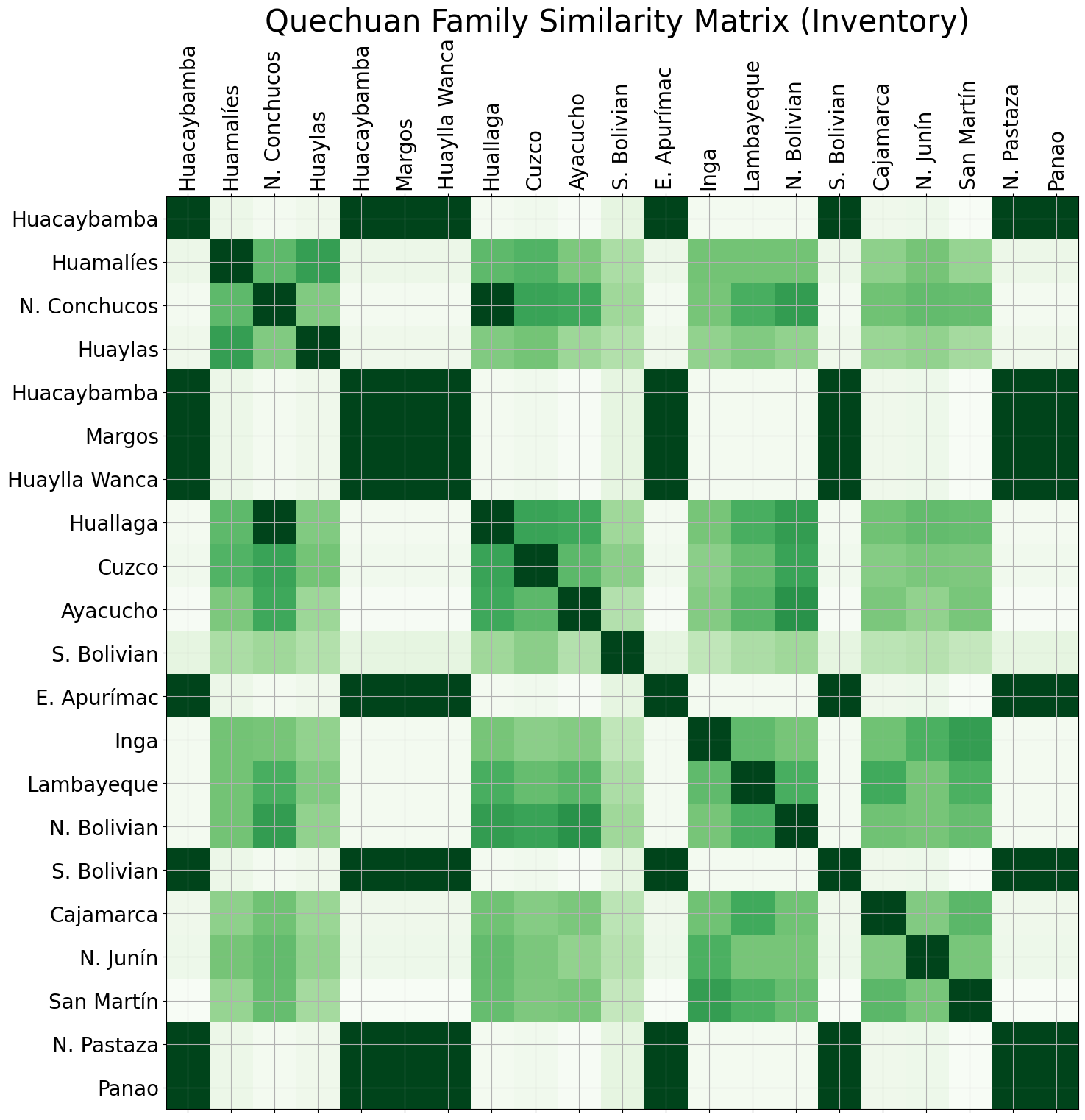}
    \\[\smallskipamount]
    \includegraphics[width=.43\textwidth]{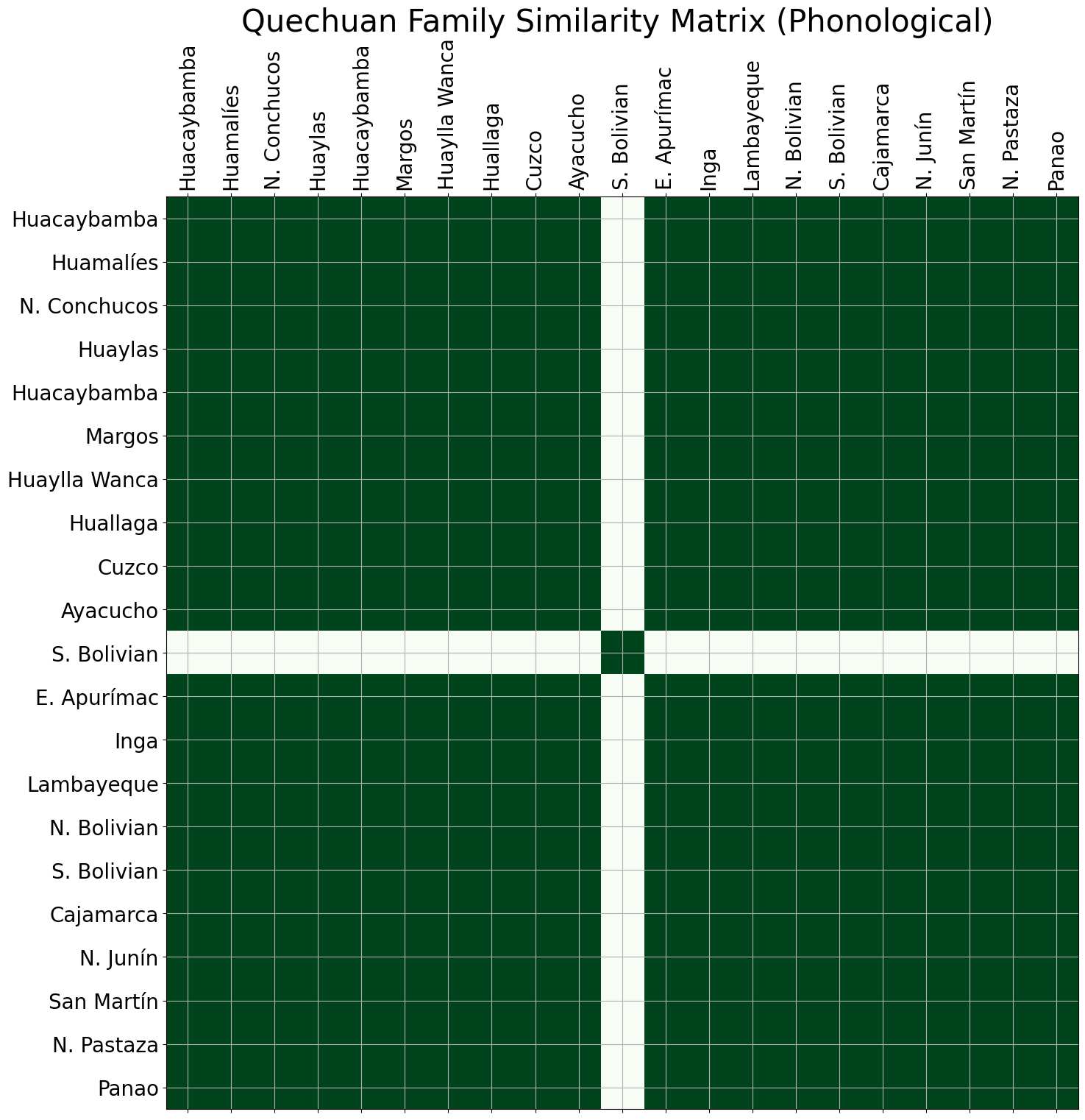}\hfill
    \includegraphics[width=.43\textwidth]{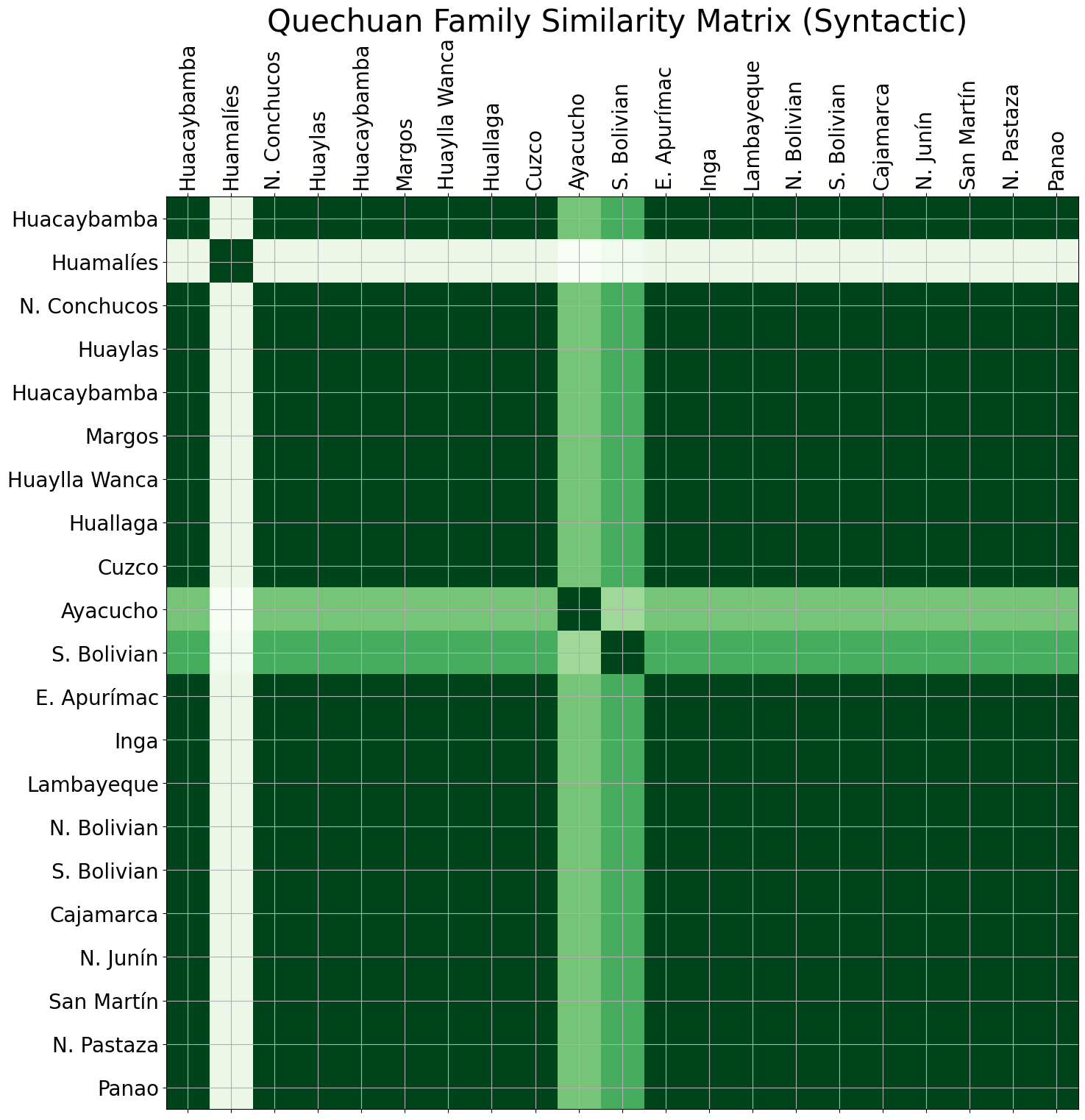}
    \caption{Similarity matrix based on genetic, featural, geographic, inventory, phonological and syntactic similarities \citep{littell2017uriel, malaviya2017learning}. }\label{fig:similarity_all_21lan}
\end{figure}

\begin{figure}
    \includegraphics[width=.43\textwidth]{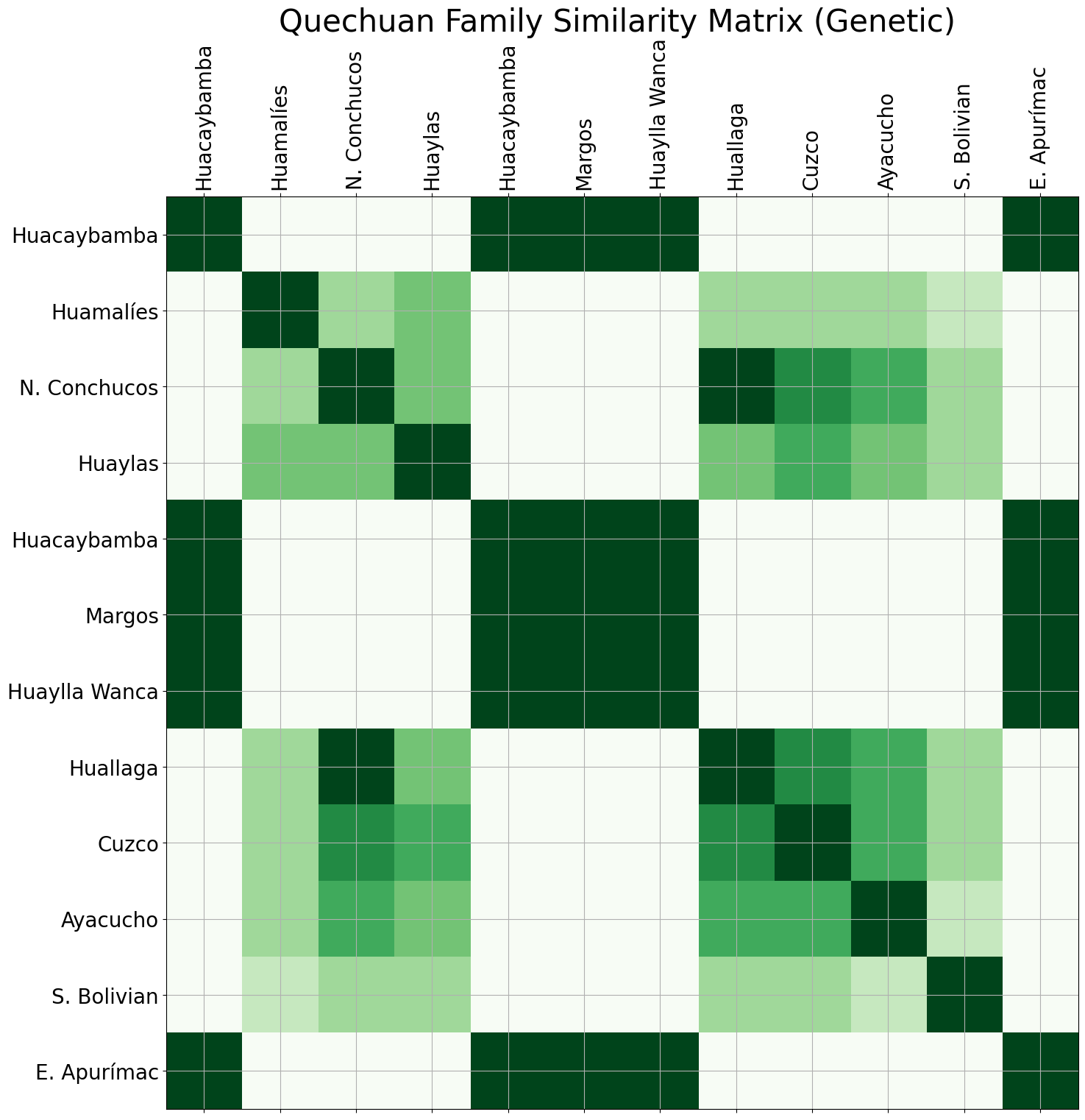}\hfill
    \includegraphics[width=.43\textwidth]{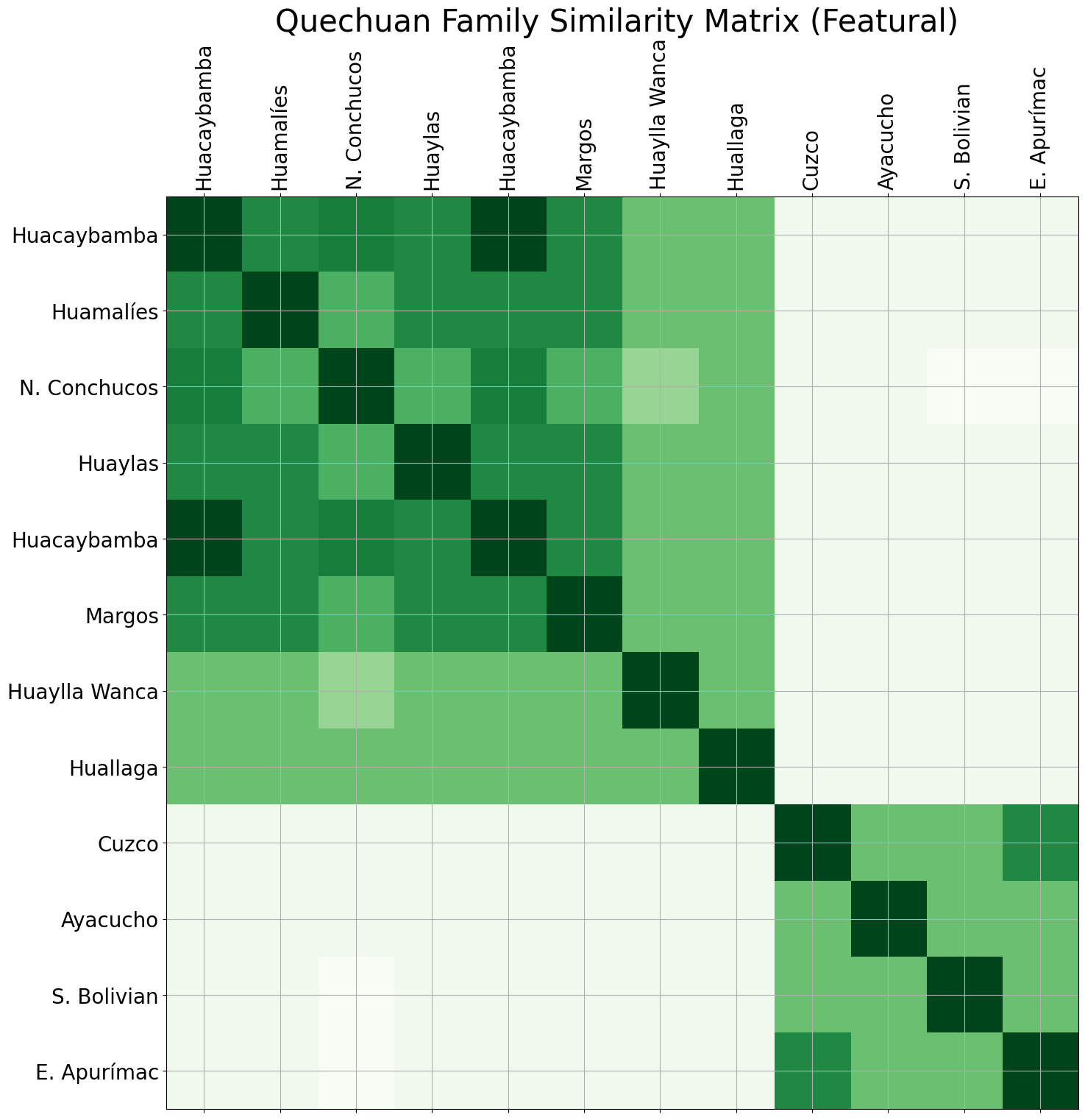}\hfill
    \\[\smallskipamount]
    \includegraphics[width=.43\textwidth]{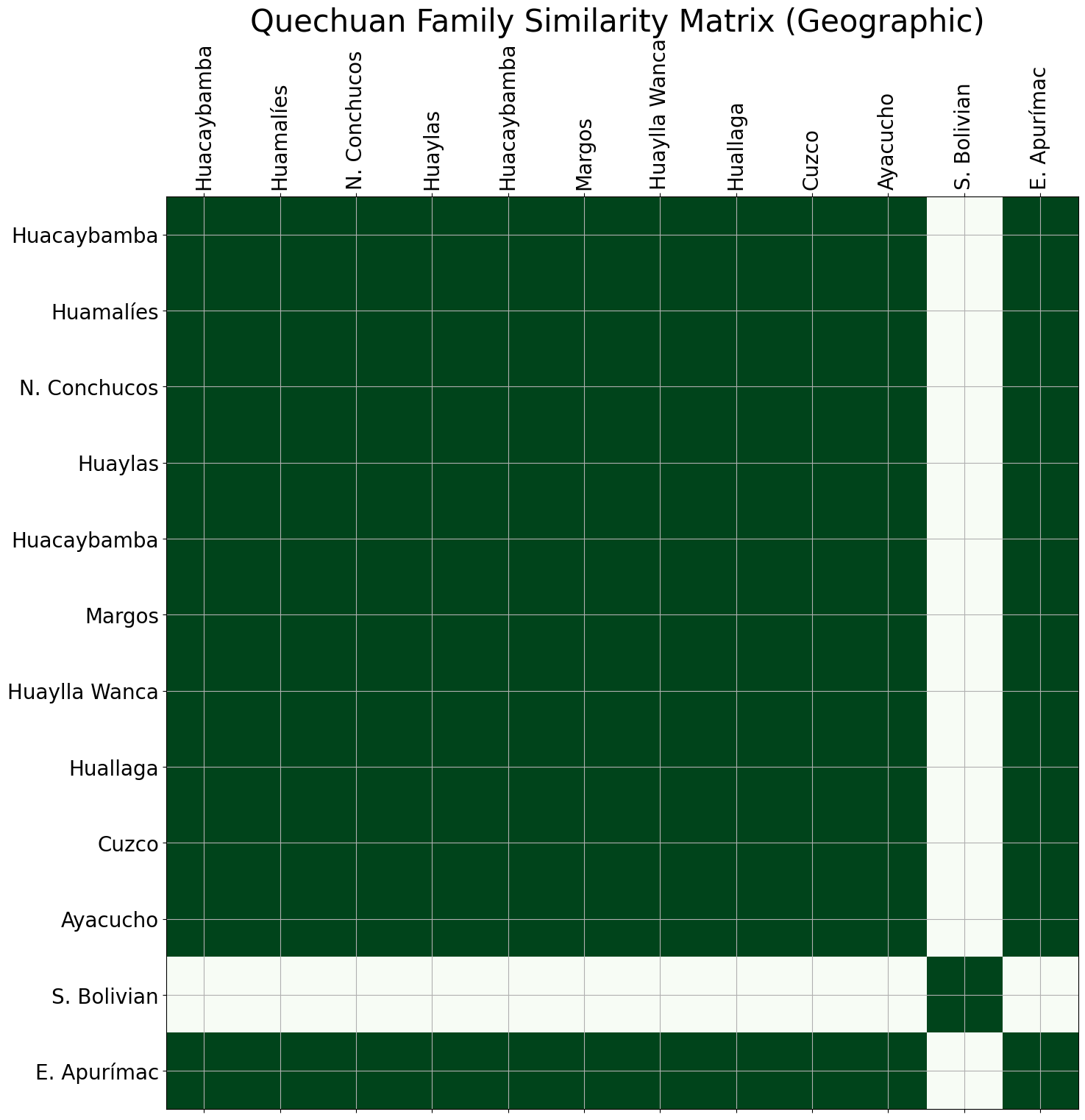}\hfill
    \includegraphics[width=.43\textwidth]{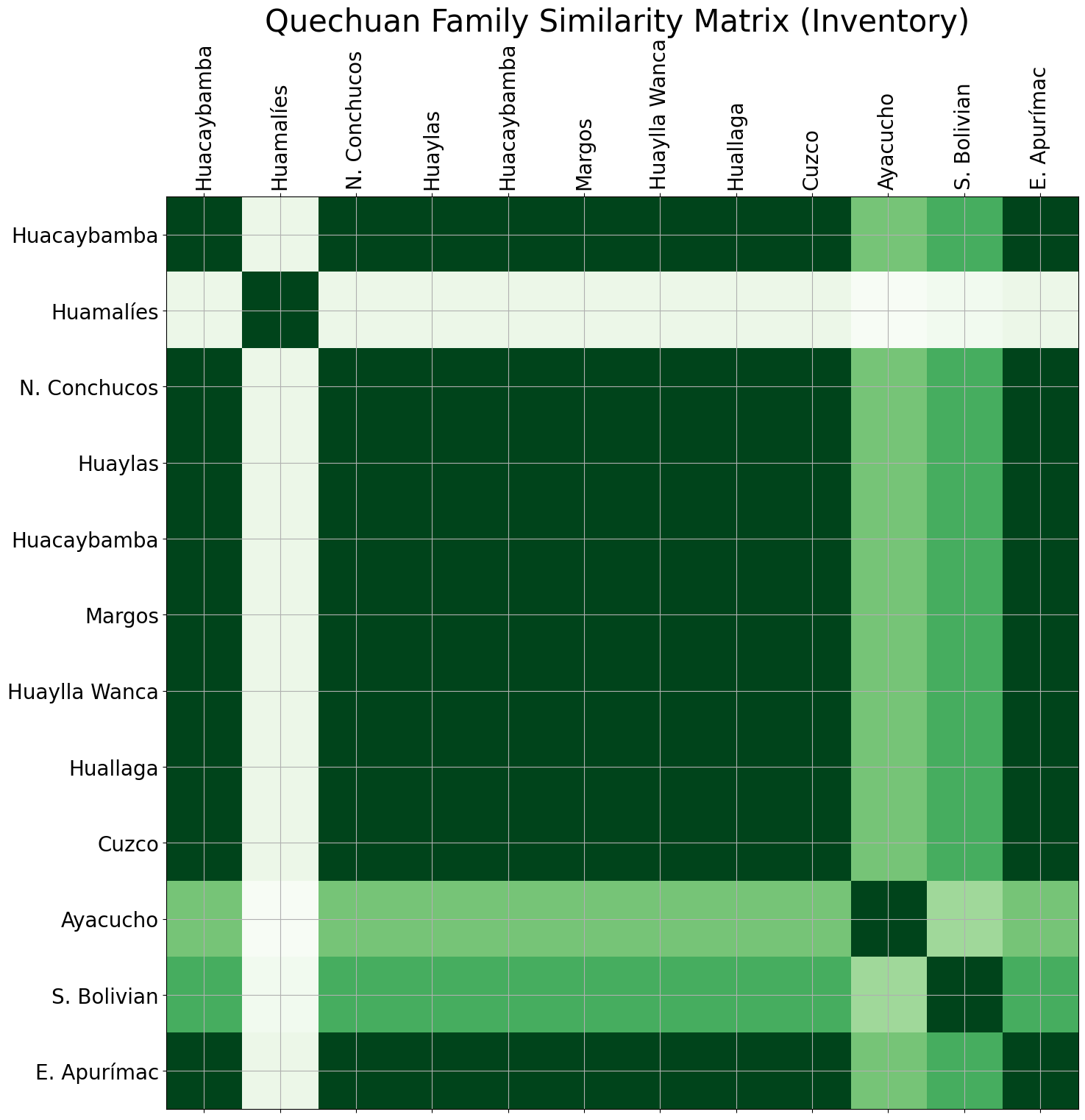}
    \\[\smallskipamount]
    \includegraphics[width=.43\textwidth]{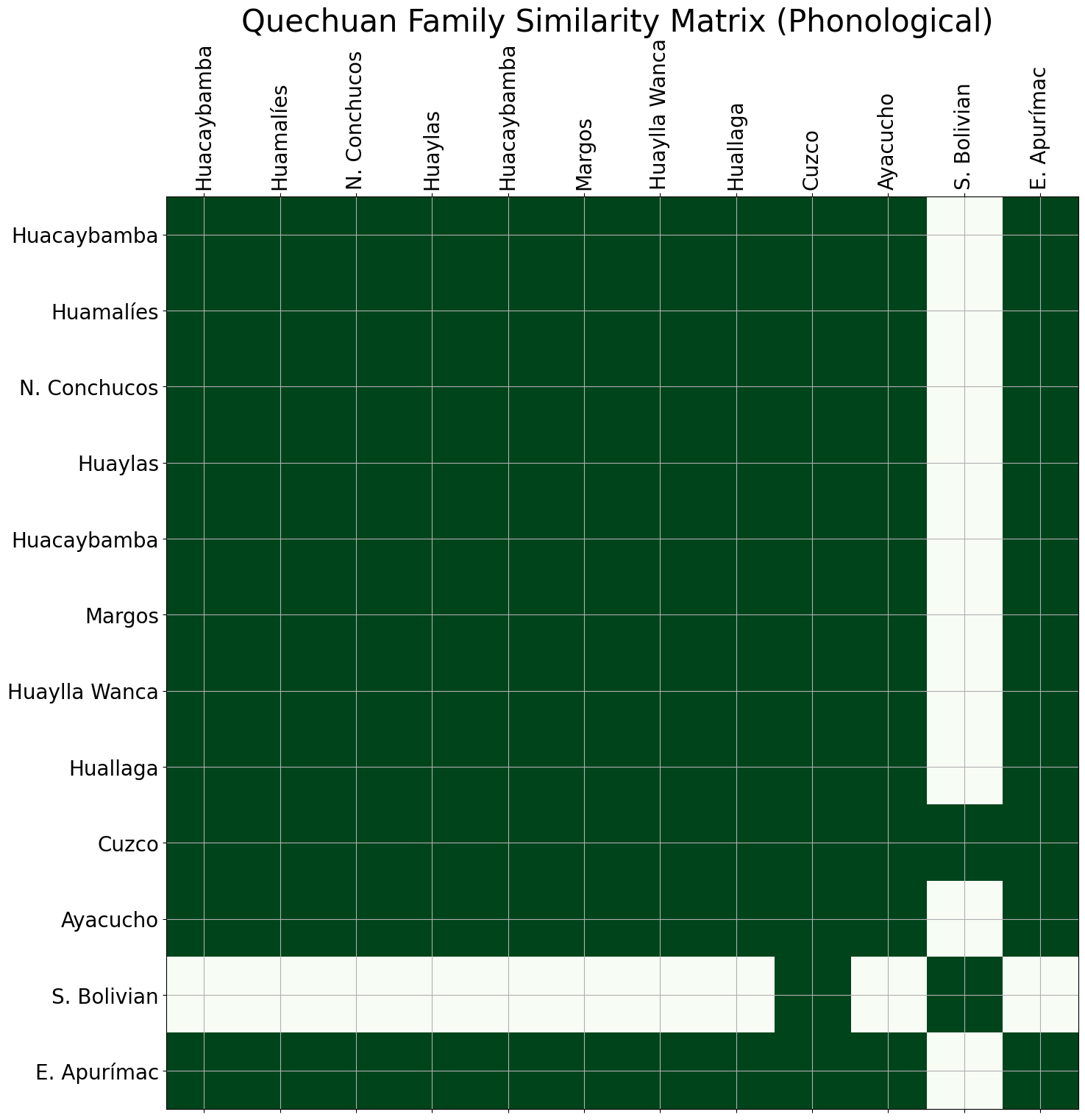}\hfill
    \includegraphics[width=.43\textwidth]{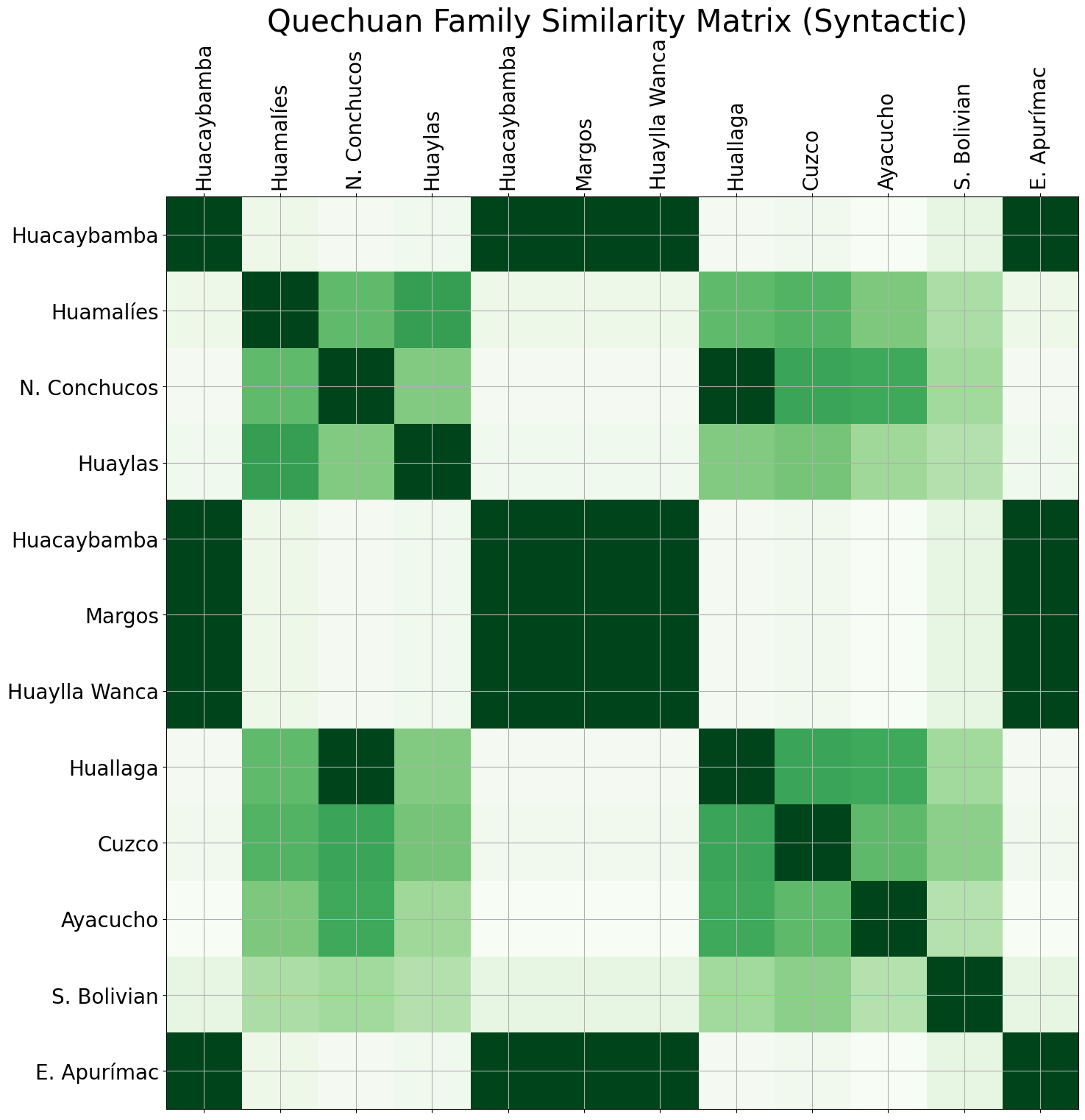}
    \caption{Similarity matrix based on genetic, featural, geographic, inventory, phonological and syntactic similarities for 12 Quechuan languages \citep{littell2017uriel, malaviya2017learning}.}
    \label{fig:similarity_all_21lan_lan2vec}
\end{figure}

\begin{figure}
    \includegraphics[width=.43\textwidth]{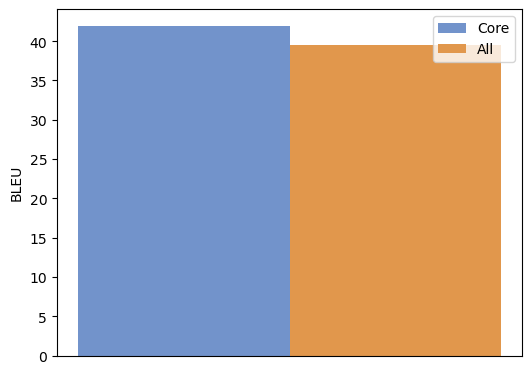}\hfill
    \includegraphics[width=.43\textwidth]{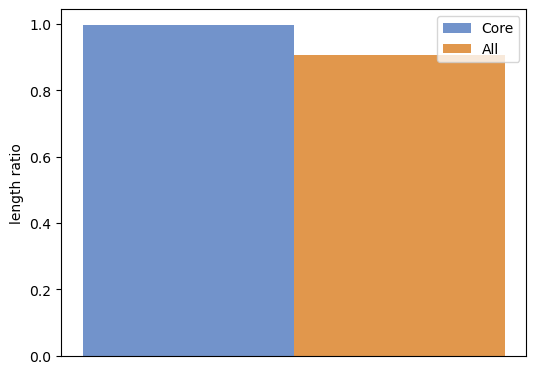}\hfill
    \\[\smallskipamount]
    \includegraphics[width=.43\textwidth]{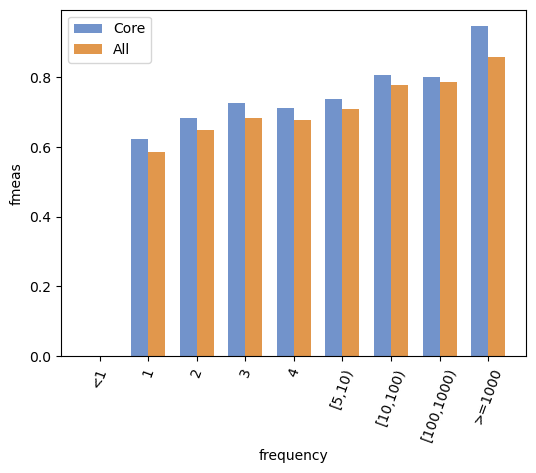}\hfill
    \includegraphics[width=.43\textwidth]{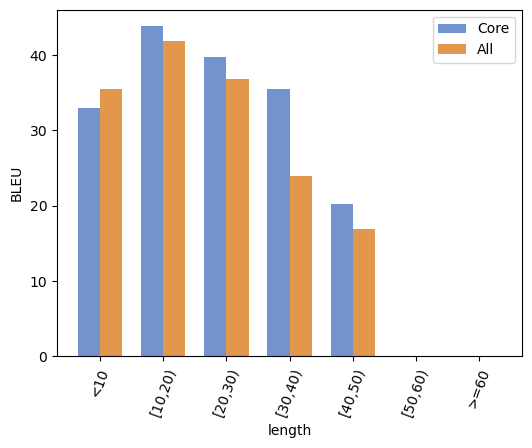}
    \\[\smallskipamount]
    \includegraphics[width=.43\textwidth]{005-sent-lengthdiff-count.png}\hfill
    \includegraphics[width=.43\textwidth]{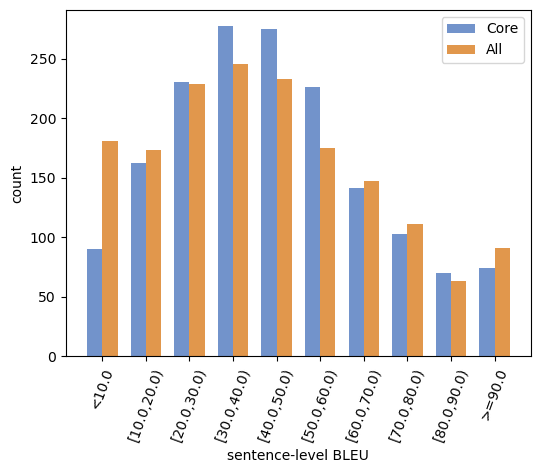}
    \caption{Detailed comparison between the system trained on only close languages that have similarity scores above 0.6 chrF (blue) versus the system trained on all (orange) using compare-MT \citep{neubig2019compare}.}\label{fig:compareMTall}
\end{figure}

\subsection{Translation into Sihuas}
Having understood that the more well-connected a language is, the higher the translation performance,  we identified Sihuas. Sihuas, though having the least resources, is very close to the other Quechuan languages, especially to the 6 language cluster shown in Figure \ref{fig:similarity_chrf_12lan}. We therefore conduct experiments to translate into Sihuas. 

\subsubsection{Translation Results}

When we train on the incomplete translations of the New Testament and test on partial translations of the Old Testament, we reach a BLEU score of 43.4 and a chrF score of 75.8 and a characTER score of 0.279. This indeed shows that the well-connected language is easier to translate into. 

Furthermore, applying what we learned about the threshold of closeness by adding and removing similar languages from training, we conduct an experiment by training only on languages that have similarity scores of at least 0.5 chrF with Sihuas. We improve our translation score to 44.7 of BLEU, 75.2 of chrF and 0.248 of characTER. However, this is not the most optimal. This shows that we improve translation performance by decluttering poorly-connected source languages.

Finally, after working further with field linguists who speak Sihuas, we decide to train only on languages that have similarity scores of at least 0.6 chrF with Sihuas, and we improve our translation score further to 46.3 of BLEU, 76.3 of chrF and 0.242 of characTER. This is our final translation system for Sihuas. In Figure \ref{fig:compareMT} and \ref{fig:compareMTall}, we show that the system that trains only on close languages performs better than the system that trains on all available languages. 

\subsubsection{Comparing with Post-Editing Existing Source Texts}
Having shown the high performance of our MT system that translates to Sihuas Quechua, we compare our method with post-editing on existing source texts directly. Post-editing existing source texts is what human translators could do without help from any MT systems. Comparing these two is very meaningful to understand our contribution.  

Our MT system is more effective than just post-editing existing books. If we post-edit from the translation of the text in Huacaybamba, the closest language to Sihuas, we have a chrF score of 68.3 and a BLEU score of 25.7 between Huacaybamba and Sihuas. Our machine translation has a chrF score of 76.3 and a BLEU score of 46.3. Even though there is a $\sim$8-point difference in chrF, there is a $\sim$20-point difference in terms of BLEU. Given the $\sim$20-point BLEU difference, there could be a lot of words in Huacaybamba that is similar to Sihuas, but not exactly the same. Our translation system is able to learn the sub-word level morphology changes, and therefore make it easier for the human translators.

In addition to the $\sim$20-point BLEU difference, human translators benefit more from a better post-editable draft of the text from our MT system, rather than post-editing from source texts directly. Our MT system lessens the chance of errors and reduces the amount of deletion, insertions and changes that human translators need to do to post-edit well. 

\section{Limitations and Future Work}
From this Quechuan case study, we demonstrate that for well-connected languages, our finding of multilingual training by carefully selecting similar languages to train with (Chapter \ref{big:family}, Chapter \ref{big:paraphrase} and Chapter \ref{big:ipml}), active learning (Chapter \ref{big:active}) and staged finetuning with large pretrained models (Chapter \ref{big:large}) improve translation performance. For poorly-connected languages, due to the low language similarity available for cross-lingual transfer, the impact of our method could not be tested. 

Our work is limited by the text and the translations of this text in different languages in Quechuan family that is available to us in this chapter. However, the relationship between similarity and performance is generalizable to other languages and language families. In Chapter \ref{big:ipml}, we have examined the translation performance using our method for the European medical EMEA dataset, have shown good performance. In Chapter \ref{big:active} and Chapter \ref{big:large}, we have shown our method is generalizable to different target languages. However, all of our results in this thesis is limited by the text and the amount of data that is available to us. 

This limitation points us to many opportunities to extend this work. We are keen to explore more varied target low-resource languages and language families. We are also open to collaborate with more human translators working in other parts of the world to extend our work to other domains including legal, literary and educational documents. 

In addition to wider domains and more languages, another area of interest is to find languages that may not be close to each other in the textual domain but in other domains that we could use to improve translation. One potential method is to transform our data from the textual space to the phonological space or other non-textual domains. In Figure \ref{fig:similarity_all_21lan}, Figure \ref{fig:similarity_all_21lan_lan2vec} and Figure \ref{fig:similarity_combined}, we show typological features based on genetic, featural, geographic, inventory, phonological and syntactic similarities. Comparing these features with our similarity graphs in Figure \ref{fig:similarity_all_22lan}, we see much richer connections among the Quechuan languages. If we can find a way to transform our data into the phonological space and find related languages to improve performance score, that will be very promising.

Furthermore, it is very important to explore what we can do when there are very few similar languages. In the case for translating into Panao Quechua, translation performance suffers as Panao is more distant from other languages in the textual form. We have tried to build a Quechuan-specific morpheme vocabulary. However, once we change the vocabulary from the large pretrained model to incorporate this Quechuan-specific morpheme vocabulary, training diverges. It is difficult to train large pretrained models using the new vocabulary in academic settings, but this gives room for creative solutions in future research. 

Finally, there are continued conversations with the field linguists and human translators working in the field. Understanding and working with their needs are key in building long-term collaborative relationships. This process is a continued dialogue. It is through this continued dialogue that we respect the dignity of both the indigenous low-resource language communities as well as the field linguists working with the natives. It is also through the same dialogue that the low-resource language community and field linguists learn to trust machine translation systems. We would like to continue this process of mutual trust and mutual understanding through continued conversations. 

Having closely examined our case study of the Quechuan language family, we conclude this thesis in the next chapter. 

\removelabelprefix

\chapter{Conclusion}\label{big:conclude}
\addlabelprefix{8}
\epigraph{``Words travel worlds. Translators do the driving.''}{\textit{Anna Rusconi}}

\lettrine{E}{arlier}, we defined our thesis statement as the following: 
\begin{myindentpar}{1cm} 
    \noindent \textbf{\MakeUppercase{Thesis Statement}}
    \textit{In translating a closed text that is known in advance and available in multiple source languages into a new and severely low-resource language,
    we argue that generalization to out-of-domain texts is not necessary, but
    generalization to new languages is necessary.
    Translation performance gain comes from massive source parallelism by careful choice of
    close-by language families, style-consistent corpus-level paraphrases within the same language and strategic adaptation of existing large pretrained multilingual models to the domain first and then to the language. Such performance gain makes it possible for machine translation systems to collaborate with human translators to expedite the translation process into new, low-resource languages.}
\end{myindentpar}  

To conclude, we summarize our contributions and describe how they support our thesis statement. Moreover, we show the limitations of our research and propose different ways to further this work in the future. We also discuss the broader impact of our work from academic research to the real-world machine translation field.   

\removelabelprefix 
\section{Summary of Contributions}
\begin{figure*}[t]
  \centering
  \includegraphics[width=1\linewidth]{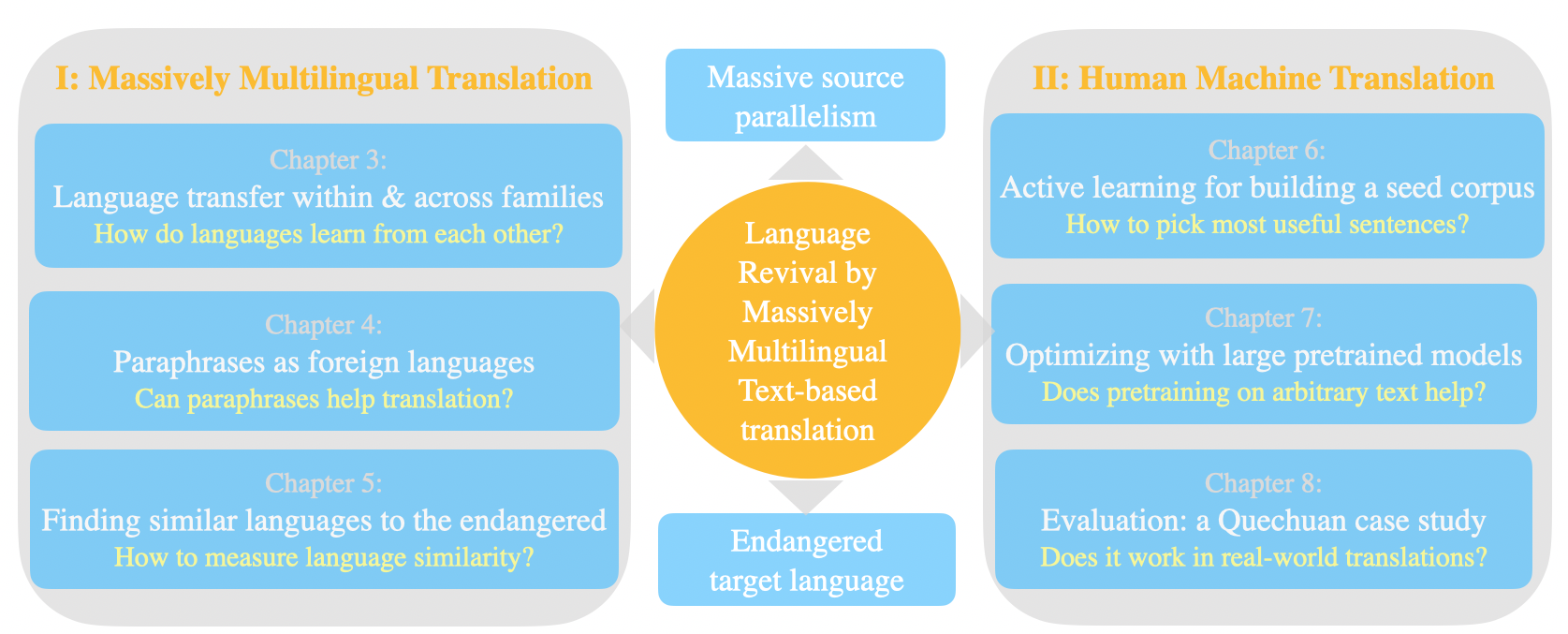}
    \caption{Recap of the work done as part of this thesis. }
    \label{fig:overview1}  
\end{figure*}

While we examine all of the following topics in the rest of the thesis, we summarize our contributions to the research community through two main parts: 1.) massively multilingual translation, and 2.) human machine translation as shown in Figure~\ref{fig:overview1}. 

In severely low-resource scenarios, we explore ways to effectively learn from massive source parallelism in Part \ref{part:part1}. In Part \ref{part:part2}, we build a human machine translation workflow for machine translation systems and human translators to work together seamlessly through active learning and large pretrained models. We show proof of concept that it is possible to produce a quality translation draft of the whole text through as little as a few hundred lines ($\sim$3\% of the text) of the low-resource data. On top of demonstrating that it is possible to translate using little resource, we build multiple mechanisms to improve effectiveness and accuracy. 

\subsection{Key contributions}
As discussed in Chapter \ref{big:intro}, our goal is to minimize human translation and post-editing  efforts required to generate a full publishable-standard translation of the given text. Ideally we want to hire a large number of human translators, measure and compare the resources (time and money) used to translate the same text into a target low-resource language that does not have any translations of the text under two scenarios: with and without the using this thesis. However, this ideal solution is unrealistic especially in large translation projects. Large translation projects in real-life usually takes decades, if not centuries of work, which is beyond the scope of this thesis. This is why we transform our goal of minimizing human translation efforts required to generate a full translation of the given text into 
two practical proxy sub-goals as the following: 
\begin{enumerate}
\item 
Optimizing and minimizing the amount of sentences to be used to construct seed corpus.
\item 
Maximizing the quality and utility of MT-generated translation of the full text and optimizing translation efficiency.
\end{enumerate}

\begin{figure*}[t]
  \centering
  \includegraphics[width=0.7\linewidth]{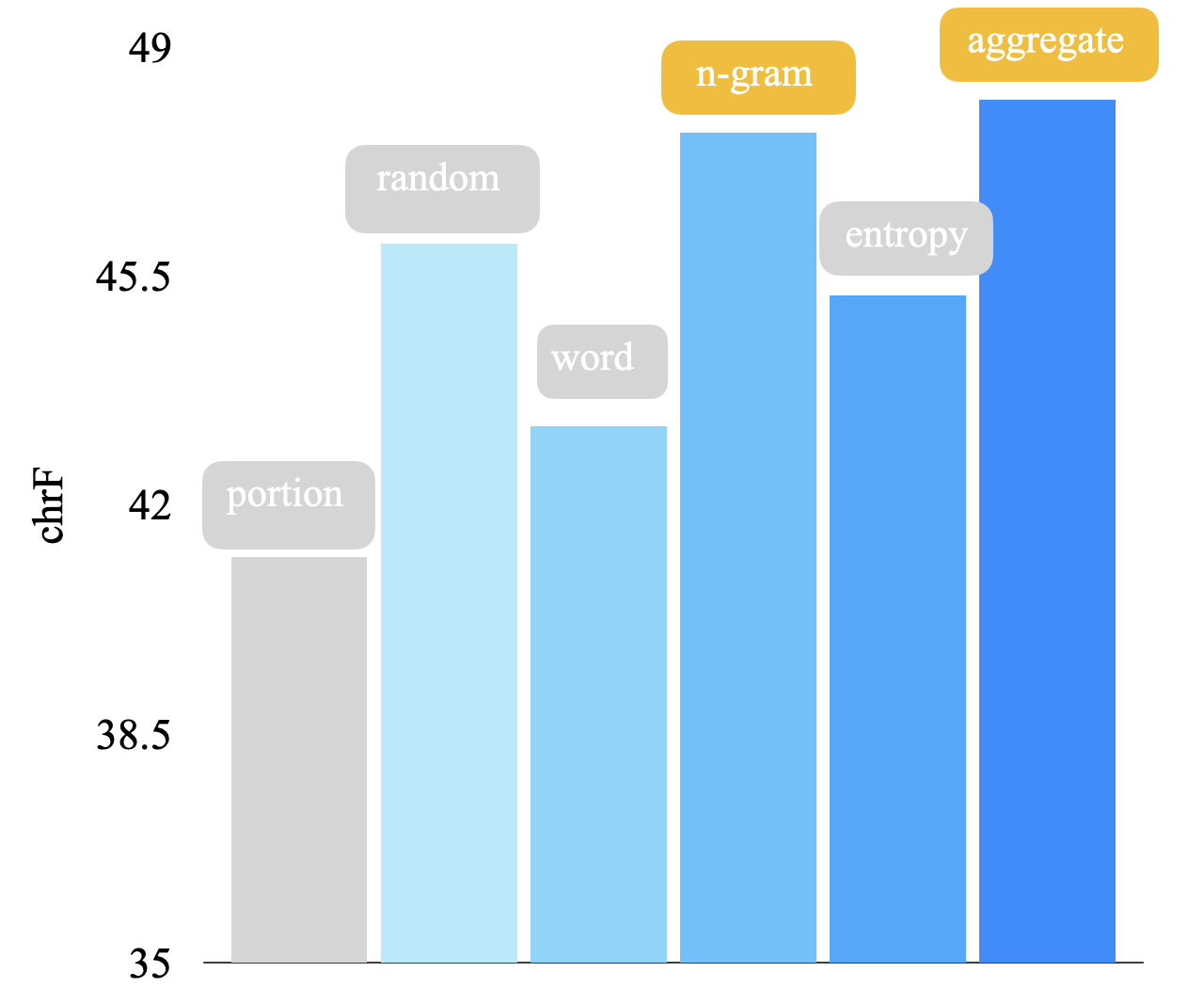}
    \caption{Key result of minimizing the amount of sentences to be used to construct seed corpus for translation into Welsh. }
    \label{fig:key1}  
\end{figure*}

\begin{figure*}[t]
  \centering
  \includegraphics[width=0.6\linewidth]{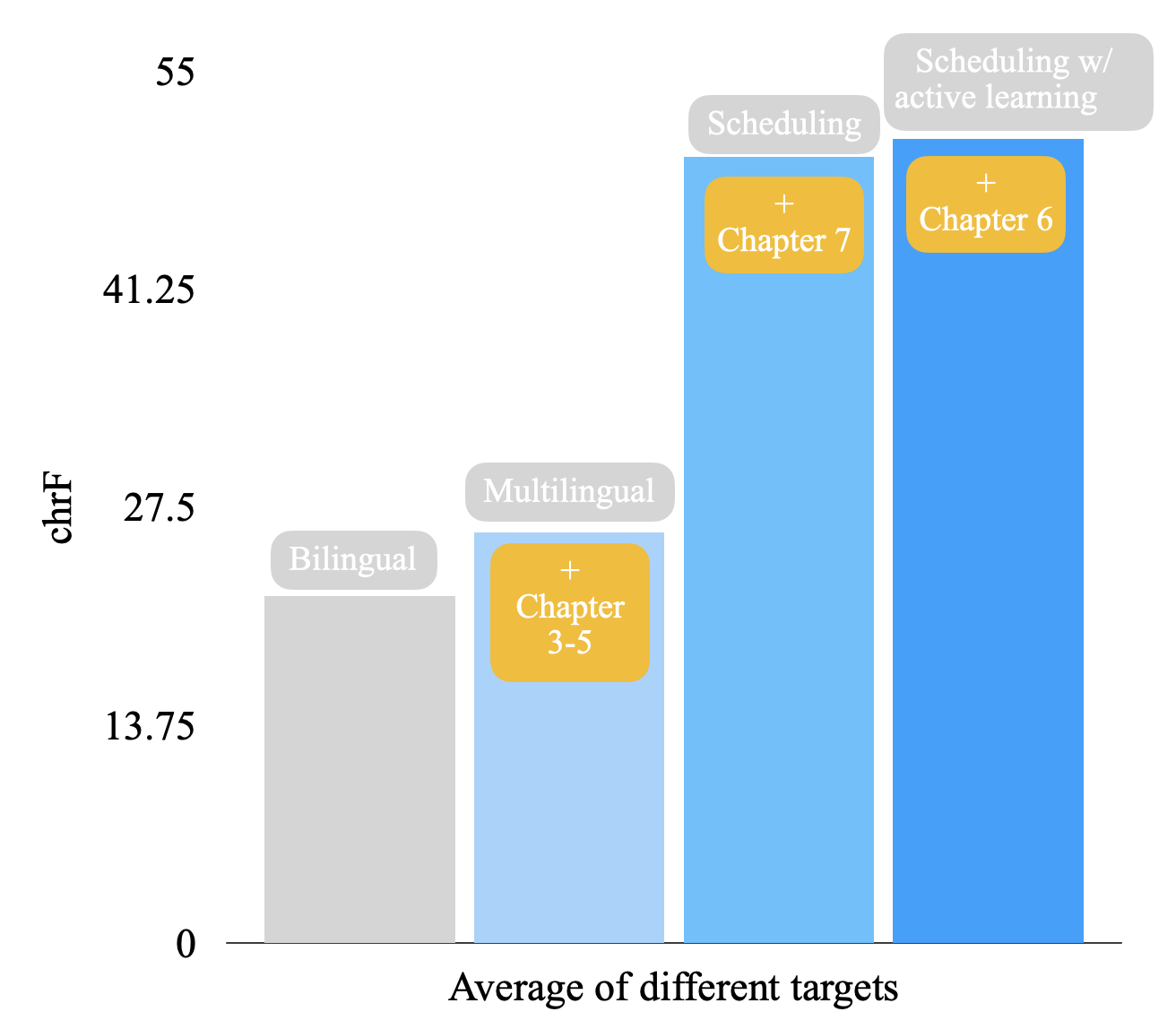}
    \caption{Key result of maximizing the quality and utility of MT-generated translation of the full text.}
    \label{fig:key2}  
\end{figure*}

The first sub-goal of minimizing the seed corpus serves as a proxy in Chapter \ref{big:active} to minimize the human translation efforts in the creation of the seed corpus, while the second sub-goal of maximizing translation performance serves as a proxy in Chapter \ref{big:large} to minimize human translation efforts in the post-editing process during the subsequent iterations. 

To measure translation performance, our primary automatic metric in this thesis is chrF \citep{popovic2015chrf}. We choose chrF for accuracy, fluency and expressive power in morphologically-rich languages \citep{papineni2002bleu}. We use the metric chrF in Chapter \ref{big:intro} and Chapter \ref{big:conclude} of this thesis to motivate and summarize our main contributions of this paper. 

Using chrF, we summarize the key contributions based on these two sub-goals in Figure~\ref{fig:key1} and Figure~\ref{fig:key2}. Figure~\ref{fig:key1} shows our key results for the first sub-goal of minimizing the seed corpus that serves as a proxy in Chapter \ref{big:active} to minimize the human translation efforts in the creation of the seed corpus. Figure~\ref{fig:key2} shows the second sub-goal of maximizing translation performance that serves as a proxy in Chapter \ref{big:large} to minimize human translation efforts in the post-editing process during the subsequent iterations.

In Figure~\ref{fig:key1}, our main contribution is that we minimize the seed corpus to be $\sim$3\% of the text, and we use $\sim$3\% of the text to translate the $\sim$97\% of the text in the low-resource language we want to translate to. Having minimized the training data to be $\sim$3\% of the text in the given low-resource language, we optimize with active learning on which $\sim$3\% of the text to translate first to produce the seed corpus. To determine which $\sim$3\% of the text, we find that n-gram method (in particular, 4-gram method) is sufficient for producing high translation performance when we do not have access to complete information about languages close to 
the low-resource language we want to translate to. However, when we do not have access to such information, we recommend the aggregation method proposed in Chapter \ref{big:active} as a universal ranking to use. This minimizes human translation efforts in the production of the seed corpus. These are the key contributions for our first sub-goal. 

Moreover, in Figure~\ref{fig:key2}, we further show our key contributions for our second sub-goal. We show the scheduling with large pretrained models using 4-gram method as our active learning method gives the best translation performance. This helps to minimize human post-editing efforts during the subsequent iterations after translation of the seed corpus.

The detailed contributions towards these two sub-goals by different chapters are described in the following two sections. We also show the limitations of this work followed by potential future research to accomplish the main goal beyond this thesis. 

\subsection{Massively Multilingual Translation}
In Part \ref{part:part1}, we explore source parallelism in translation of a given text into new, low-resource languages through massively multilingual training. In Chapter \ref{big:family}, we build cross-lingual transfer both within a given language family and also across different language families. In Chapter \ref{big:paraphrase}, we treat paraphrases within the same language as foreign languages, and train on corpus-level paraphrases to improve translation performance. 
In Chapter \ref{big:ipml}, we build our own linguistic distance metric based on translation distortion, fertility and performance. 
 
\textbf{Massively Parallel Intra-family and Inter-family Learning:} Our contribution in building cross-lingual transfer, resolving the variable binding problem and producing high quality translations under severely low-resource data scenario is three-fold, extending from multi-source multi-target attentional Neural MT (NMT).

Firstly, to examine intra-family and inter-family influences, we add source and target 
language family labels in training.
Training on multiple families improves
BLEU score significantly; moreover, we find training on two 
neighboring families closest to the 
low-resource language gives reasonably
good BLEU scores. 

Secondly, we conduct an 
ablation study to explore
how generalization changes with 
different amounts of data and
find that we only need a small amount of
low-resource language data to produce 
reasonably good BLEU scores. We use full data	
except for the ablation	study.

Finally, to address the variable-binding 
problem, we
build a parallel lexicon table across 
twenty-three European
languages and devise a novel order-preserving
named entity translation method. Our method works
in translation of any text with a fixed set of
named entities known in advance. 
Our goal is to minimize manual labor, but
not to fully automate the manual process to ensure
the correct translation of named entities and their
ordering. 

\textbf{Paraphrases as Foreign Languages:} We treat paraphrases, rewordings of texts with  
preserved semantics, as foreign languages,
and train a unified NMT model on paraphrase-labeled data
with a shared attention in the style of multilingual NMT. Our main findings in harnessing paraphrases in NMT are the following. 

Our multi-paraphrase NMT results show significant improvements in BLEU scores over all baselines. 
In addition, our paraphrase-exploiting NMT uses only two languages, the source and the target languages, and achieves higher BLEUs than the multi-source and multi-target NMT that incorporates more languages. 

Furthermore, we find that adding the source paraphrases helps better than adding the target paraphrases, and find that adding paraphrases at both the source and the target sides is better than adding at either side. We also find that adding paraphrases with additional multilingual data yields mixed performance; its performance is better than training on language families alone, but is worse than training on both the source and target paraphrases without language families. 

Moreover, adding paraphrases improves the sparsity issue of rare word translation and diversity in lexical choice.

\textbf{Family of Origin and Family of Choice: Massively Parallel Lexiconized Iterative Pretraining for Severely low-resource Machine Translation:} We have five contributions in building customized set of languages that are close to the severely low-resource language. 

Firstly, we rank the
124 source languages to determine their
closeness to the low-resource language and choose the top few. 
We call the linguistic definition of
language family \textit{Family of Origin} (FAMO), and
we call the empirical definition of higher-ranked 
languages using our metrics 
\textit{Family of Choice} (FAMC).
They often overlap, but may not coincide.

Secondly, we build an
\textit{Iteratively Pretrained Multilingual Order-preserving Lexiconized Transformer} (IPML)
training on $\sim$1,000 lines
of low-resource data. 
Using iterative pretraining, 
we get a +23.9 BLEU increase over
a multilingual order-preserving lexiconized transformer
baseline (MLc)
using English as a hypothetical low-resource
language, and a +10.3 BLEU increase over our asymmetric baseline. 
Training with the low-resource
language on both the source and target sides
boosts translation into the target side.
We have a 42.8 BLEU score for Portuguese-English translation
on the medical EMEA dataset.

Thirdly, we use a real-life severely
low-resource Mayan language, Eastern Pokomchi,
a Class 0 language \cite{joshi2020state}
as one of our experiment
setups. In addition, we also use English
as a hypothetical low-resource language for easy 
evaluation. 

Fourthly, we also add an order-preserving lexiconized
component to translate named entities well. To
solve the variable-binding
problem to distinguish ``Ian calls Yi''
from ``Yi calls Ian'' \cite{fodor1988connectionism, graves2014neural, zhou2018massively},
we build a lexicon table for
2,939 Bible named entities in 124 source languages
including more than 66 severely low-resource languages. 

Finally, we combine translations from all source
languages by using a novel method. For every sentence,
we find the translation that is closest to
the translation cluster center.
The expected value of the BLEU score for our combined translation
is higher than translation from any of the individuals. 

\subsection{Human Machine Translation}
In Part \ref{part:part2}, having examined source parallelism, we build a human machine translation workflow algorithm for machine translation systems to collaborate with human translators to expedite the process.  
In Chapter \ref{big:active}, we develop various active learning methods on known languages and transfer ranking to the new, low-resource language. In Chapter \ref{big:large}, we activate the knowledge of large multilingual models by proposing multilingual and multi-stage adaptations through different training schedules; we find that adapting pretrained models to the domain and then to the low-resource language works best. In Chapter \ref{big:confidence}, we evaluate our work by translating academic progress to the real-world translation process in a case study of the Quechuan language family. 

\textbf{Active Learning for Human Machine Translation: } 
We use a random sampling approach
to build seed corpora when resources are extremely
limited. 
We recognize that the portion-based translation is very helpful in 
producing quality translation with
formality, cohesion and contextual relevance.
Thus, our proposed way is not to
replace the portion-based approach, but instead,
to get the best of both worlds and
to expedite the translation process. 
The two approaches differ in that
the portion-based approach focuses on preserving
coherence of the text locally, while the random-sampling
approach focuses on increasing coverage of the text
globally. 

Moreover, we compare three different ways of incorporating
incremental post-edited data during
the translation process. We find that
self-supervision using the whole translation draft
affects performance 
adversely, and is best to be avoided. 

Furthermore, we also show that adding
the newly post-edited text to
training with vocabulary update performs the best. 

\textbf{Train Global, Tailor Local: Minimalist Multilingual Translation into Low-Resource Languages: } 
We push the limits of random sampling and explore more active learning methods.
Our contribution is three-fold. 

Firstly, we develop 14 active learning methods on known languages and transfer ranking to the new, low-resource language. 

Secondly, we activate the knowledge of large multilingual models by proposing multilingual and multi-stage adaptations through 24 different training schedules; we find that adapting pretrained models to the domain and then to the low-resource language works best. 

Thirdly, we aggregate scores from 115 languages to provide a universal ranking and increase robustness by the \textit{relaxed memoization} method.

\textbf{Evaluation: } 
We focus on the case study of the Quechuan language family. We find that machine translation performance is significantly positively correlated with language similarity. The more connected a language is, the better it is to translate into this language. Furthermore, we find that decluttering poorly-connected languages improves translation scores. Using this result, we demonstrate our methods in translating into a new, low-resource language called Sihuas and achieve high quality translation performance. 

\section{Limitations}
Having shown our contributions to the scientific community, we discuss 
the limitations of our work based on system-level constraints, data-level constraints, task-level constraints, evaluation-level constraints and machine-level constraints. 

\subsection{System-Level Constraints}

\subsubsection{Incompleteness}
From Gödel's incompleteness theorems \citep{godel1931formal, godel1986kurt}, we understand that there is no all-encompassing ``Theory of Everything'' that unifies all that is provable and true. Gödel argues that there exists mathematical statements that are undecidable within a formal system, and such systems cannot prove its own consistency. This theory not only shakes the field of mathematics, physics, philosophy, Computer Science theory, and linguistics, it also offers profound insights into the limitation of Machine Translation. 

Applying Gödel's incompleteness theorems, Machine Translation, a formal system, regardless of its depth and complexity, may not be able to solve every translation need. However, we may still produce models that produce good enough translations that are useful to humans. Indeed, 25 years after Gödel presented incompleteness theorem, Turing asked the question ``Can Machines Think?'' and paved the road for Machine Translation \citep{turing1956can}. Turing didn't propose that machines could think in the same way humans do, but he believed they could simulate intelligent behavior to a point where it could be hard to distinguish humans from machines. 

Following Gödel's and Turing's thoughts, we could build machine translation systems that produce good enough translations that are useful for communication. However, such machine translation systems, like any formal system, are always incomplete and limited. 

\subsubsection{End-to-End Learning}
In addition to Gödel's incompleteness theorems, we also face limitations particular to end-to-end systems. There are a few known weaknesses of end-to-end systems. For example, most of the end-to-end networks depends on stochastic gradient descent where slow convergence or getting stuck at local optima could be a real problem \citep{glasmachers2017limits}. Moreover, end-to-end systems suffer from interpretability \citep{von2021transparency}, racial/gender bias \citep{prates2020assessing}, composability \citep{srivastava2018towards}, resource-demanding nature \citep{moorkens2022ethics} and many other shortcomings \citep{glasmachers2017limits}. 

Indeed, our work is limited by limitations inherent to all Machine Translation systems, especially end-to-end learning systems. 

\subsubsection{Large Pretrained Models}
Moreover, our work is also limited to the large pretrained models that are available to us at the time of research for each chapter. In Chapter \ref{big:large}, we use large pretrained models like M2M100. And in Chapter \ref{big:confidence}, we further use DeltaLM. The method of adapting large pretrained models to the domain and then to the low-resource language is generalizable to other large pretrained models. However, the effect of the translation performance gain we present in this thesis is limited to the large pretrained models. 

\subsection{Data-Level Constraints}
On top of limitations at the system level, we are also limited by 
the following data-level constraints including data representation, data accessibility, and data design decisions. 

\subsubsection{Data Representation}
Our work is limited to the representation of our data in text form. From the extended language similarity measures that we have shown in the previous chapter, even though many Quechuan languages are phonologically close, they may be very far apart in written form. Many local languages did not have a written form until a team of field linguists started the documentation and translation process. The initial team of field of linguists who started the process is therefore pivotal in determining the written form of the given low-resource language. Different teams may have different conventions, methodologies, and beliefs in how to best document and translate a language. Indeed, there are many factors in determining the differences between written forms of two languages even though they might be relatively close in the phonological space. 

Without a multi-faceted data representation in many non-textual spaces including the phonological space, our work is limited to the similarities and features machines can learn from the written form. Indeed, our work is limited by the type of data. However, there is a wider range of data beyond text that is helpful for us, that includes audio, video, pictures and different varieties of multi-modal data. 

Indeed, our research will benefit from more variety of data that includes audio, and 
video data that covers wide range of domains. 

\subsubsection{Data Accessibility}
In addition to the limitation of data representation, our work is 
limited to data accessibility and copyrights. 
This work can be applied to any text, including large literary 
texts, instruction menus, infectious disease prevention brochures, 
immigrant welcome booklets. However, most of these texts are copyrighted 
and not easily accessible. For those that are more accessible, its translations in multiple source languages are usually non-existent, scattered and not normalized. Training and testing on the Bible dataset 
is therefore a viable way for testing our methods, as the Bible dataset is 
relatively accessible and consolidated. 

However, even with the Bible dataset, different translations of the Bible 
carry different copyrights. There are hundreds of completed Bibles, but we 
only have $\sim$125 that are completed and available to us. A large number of completed 
Bibles are copyrighted and not available to us. This limits the number of source languages 
we can train on. It also limits the number target languages that we can test on as the chance of 
finding similar languages for extremely low-resource and isolated languages in a 
relatively small dataset is low. Both of the these limit the number of experiments 
we can do in this space. 

Indeed, our work will thrive on more accessible data from a wider
range of domains. 

\subsubsection{Data Design Decisions}
In addition to data representation and data accessibility constraints, we also face constraints in our problem setup and data design decisions. In our problem setup, given a text that is multilingually available in many languages, we are interested in translating it into a new, low-resource language. The problem has three unique aspects that are different from traditional MT problems: 1.) our text is closed, not arbitrary 2.) our text has complete translations in all source languages 3.) our text has little to no translation in the target low-resource language. 

In this unique setup, we make a few design decisions that suit the purpose of automatic evaluation needs. In designing experiments from Chapter \ref{big:family} to Chapter \ref{big:confidence}, for each target low-resource language that already has full translations of the text, we train on $\sim$3\% of the data, and test on $\sim$97\% of the data. Since we are training on all source languages that have complete translations of the text, we indirectly have access to the $\sim$97\% of the data in source languages as well. This is intended in our problem setup because we are translating a closed text. Since machine translation systems and human translators have seen the full text in many existing translations, we intend to encode the meaning of all of the sentences in the text and decode it into the target low-resource language. This design decision is limited by our unique problem setup and is suitable for this task. However, the training and testing setup will be different if we work on open-domain and open text in the future. 

\subsection{Task-Level Constraints}
\label{limitation:task}
In addition to limitations at the system level and the data level, we also face limitations at the task level. 

As we have discussed in Chapter \ref{big:intro}, ideally we want to measure time taken and money spent for the entire text translation project. However, this is beyond the scope of this thesis. In order to set practical goals for the scope of this thesis, we use two practical proxy sub-goals to transform our goal of minimizing human translation efforts required to generate a full translation of the given text into tangible forms. These two sub-goals are: 

\begin{enumerate}
\item 
Optimizing and minimizing the amount of sentences to be used to construct seed corpus.
\item 
Maximizing the quality and utility of MT-generated translation of the full text and optimizing translation efficiency.
\end{enumerate}

As a result, we are not directly measuring the exact time and money used for completion of the text translation with and without our methods, and we lack realistic meta-data of the entire text translation process to complete the whole text. The translation process with the ideas in this thesis includes seven stages: 

\begin{enumerate}
\item 
\textbf{\ul{Language discovery stage:}} This is the hardest stage that researchers do not usually focus on. There are many low-resource language communities that are very isolated from the rest of the world. Some of these low-resource communities are extremely suspicious and hostile towards outsiders \citep{faught2008john, schonhuth2019dead, samson2021translation}. Therefore, it is very important to build a friendly first contact with the low-resource community, and discern  whether a language is a new language or a dialect. This is the language discovery process, which may take a long time. 
\item 
\textbf{\ul{Language learning stage:}} This stage is also important as human translators learn and formalize the orthography of the language. The orthography system is important for standardization not just for the given target language, but also for future work on similar languages. 
\item 
\textbf{\ul{Language literacy stage:}} After human translators have learned the language and formalized the orthography, human translators need to teach local communities these language tools if human translators want to involve native speakers in the subsequent translation stages.
\item 
\textbf{\ul{Seed corpus translation stage:}} This is what we focused on in detail in Chapter \ref{big:active} on active learning. Machine systems provide different data selection methods for human translators to produce the seed corpus. 
\item 
\textbf{\ul{Iterative post-editing stage:}} This is what we focused on in detail in Chapter \ref{big:large} on large pretrained models. Machine systems optimize translation drafts for human translators to post-edit on, while human translators provide more post-edited data for machine translation systems to train on. Together, they complete a translation of the whole text. 
\item 
\textbf{\ul{Quality control stage:}} This stage may involve multiple levels of focus groups, including human translators, quality control teams and native low-resource language speakers. The translated text needs to be checked to pass multiple standards before it is released. In the case of Bible Translation, multiple focus groups performing linguistic and theological checks are employed for accurate final translation of the given text. 
\item 
\textbf{\ul{Document production stage:}} For any translation of a text, the final presentation of the document may include chapter headers,
footnotes, paragraph numbers, section headers, titles and many various forms of formatting components. The document production process is therefore much more than the content
translation and is paramount in the final publication of the translated text in the low-resource language. 
\end{enumerate}

Among the above seven stages of the text translation process, Stage 4 and 5 can be accelerated by models and algorithms introduced in this thesis. However, since the entire text translation process contains five other stages that are beyond the scope of this thesis, we face constraints of time and money estimation of the entire process in the real-world. 

\subsection{Evaluation-Level Constraints} 
In addition to limitations at the system level, the data level and the task level, we also face limitations at the evaluation level. 

\subsubsection{Qualitative Evaluation}
On top of automatic evaluation metrics like chrF, characTER that we 
may use for low-resource languages that are often morphological-rich, our work is limited 
by human evaluation by native speakers. 
Most members of the research community do not speak the low-resource 
languages, and it is difficult to find native
speakers and establish 
long-term collaborations. 

There is also a lot of diversity across all 
low-resource languages. Some are more accessible than others. 
Hmong and Eastern Pokomchi are harder to assess while 
Frisian and Welsh, and many Eastern dialects in southern 
China and Indonesia, are easier to access. These languages could potentially 
provide easier access and evaluation opportunities. 

Empowering and reviving these languages is not just 
a scientific problem, it requires communication, building 
trust with field linguists and native speakers and building 
long-term connections with low-resource language communities. 
In this thesis, we collaborated with a group of field linguists 
working on Quechuan language families. And in the future, we are 
looking to broaden the collaboration to include more human translators, field 
linguists, and native speakers. 

\subsubsection{Cultural-Aware Evaluation} 
Broadening our collaboration with native speakers and human translators is also important 
for understanding the target low-resource language community culture. This work is limited 
by the extent of our understanding of the native community culture. 
Cultural-aware translations and evaluation are key, especially in 
expressing subtleties in 
rhetoric \cite{levin1998interlingua, larson1984meaning}. 
This is especially relevant in non-Western communities 
where expressions are implicit rather than explicit, 
and true meanings in communication might hinge on what is not said rather 
than what is said. If we return to one of our earlier 
examples, when a host is asking ``Is your sake cold?'', 
what the host is actually saying is that ``Would you like me to warm the sake for you?''. If we translate the sentence 
without understanding the culture behind this language, 
we may do a good job translating the sentence verbatim but completely 
miss the main message that the speaker wants to get across. 
Therefore, we would like to model culture-specific
subtleties in the future work. 

\subsection{Machine-Level Constraints}
In addition to the limitations at the system level, the data level, the task level and the evaluation level, we are also limited at the machine level. We use machines in academic settings. 
All our research is done with 2 cards of Geforce RTX 1080 Ti, 
2 cards of Geforce RTX 2080 Ti and 1 card of RTX 3090.  
Our limitation is not just on the type of graphic cards (especially 
the memory), but also on the number of graphic cards. 

As we have mentioned in the previous few chapters, in situations when 
computing power is a constraint, we usually devise parallel methods 
like \textit{relaxed memoization} to compensate for our limitation. Though clever algorithm and parallel computing could help us to 
achieve good performance with limited resources, it could only help to a 
limited extent. In the future, if we have computing power to train a text-specific 
pretrained large model using our dedicated vocabulary on our own, we could achieve higher performance.  Indeed, our research would benefit with more accessible and stronger 
large scale computing power.  

\section{Future Directions}

Given the limitations we face at the system level, the data level, the evaluation level and the machine level, we are interested in pushing our research in the following future directions. These future directions include: overcoming data-level constraints, broadening applications and tasks, improving post-editing user experience, moving beyond limitations on large pretrained models, and overcoming evaluation-level constraints. 

\subsection{Overcoming Data-Level Constraints}
Firstly, to push the limitations of data representation, we are interested in venturing into multi-modal learning, and translation into sign languages. 

\subsubsection{Multi-Modal Learning}
Our work will benefit from a multi-faceted data representation in many non-textual spaces including the phonological space. By venturing into speech and visual domain, our work will no longer be limited to similarities and features machines can learn from the written form. When two languages are not close in the written form, they might be close in the phonological space or other non-textual spaces. By representing our data in multiple spaces beyond text is helpful for us to identify and learn from other languages to better translate into a new, low-resource language. 

Looking beyond text, there is a wider range of data representations that is helpful for us. It includes audio, video, pictures and different varieties of multi-modal data. Indeed, our research will benefit from more variety of data that covers wider range of domains.

\subsubsection{Sign Language}
Many researchers have worked on sign languages. Translation into sign languages in the low-resource language communities is a very meaningful and important work. For example, researchers find that a substantial portion of children in a native Nicaraguan community are genetically inclined to be born deaf; and these deaf children learn and create sign languages \citep{prates2020assessing, senghas2001children, senghas1995children, senghas2005emergence, morgan2006nicaraguan}. Translating into these low-resource languages where many children are deaf is therefore centered on translating into sign languages. Many researchers have ventured into sign language \citep{kezar2023exploring, thomason2020vision}, this is indeed a promising research direction.  

\subsubsection{Real-life Data Encoding}
Working with real-life severely low
resource languages presents a series of challenges that we may not expect. For example, 
among the languages that have complete Bible translations in our dataset,  we have 
more than 66 real-life
severely low-resource languages. Many of such languages 
have ``latin-1'' encoding rather than ``utf-8''
encoding. And many of them do not even have
``latin-1'' encoding. However, most transformer
platforms work with only ``utf-8'' encoding.
Preprocessing takes a lot of effort. In the future, when we expand to other datasets or other source
languages, we will face more situations similar to this and we will overcome them. 

\subsection{Broadening Applications and Tasks}

\subsubsection{Wider Range of Applications}
On top of exploring more diverse data representation, we also would like to explore a wider 
range of applications and datasets. In addition to the datasets that are used in this thesis, we are interested in collaborating with more diverse teams to translate medical and literary texts, 
including infectious disease prevention 
brochures, immigrant welcoming booklet, instruction menus, 
large literary texts, movie scripts, and song lyrics. These texts are helpful for low-resource language communities to understand and improve their own welfare and healthcare and communication with the outside world. 

Our method can be applied to the medical and healthcare domains which is immensely 
valuable to the low-resource communities. 
In Chapter \ref{big:ipml}, we have applied our method to the European medical 
dataset EMEA and achieved high translation performance \cite{zhou2021family}. 
EMEA dataset is built by the European Medicines Agency and
has a lot of medical information that may be beneficial to the
low-resource communities. However, existing dataset in EMEA is limited 
to the European languages. Unlike European languages, a lot of severely 
low-resource languages do not receive a lot of attention and have extremely 
limited budgets in curating such useful datasets. In the future, 
if we can work with human translators to curate multilingual 
dataset that includes many low-resource languages in the medical domain, 
this will be very helpful. One specific use case is the translation of 
COVID-19 guidelines. In an event of a global pandemic, many low-resource communities 
are affected and there is immense value in creating a multilingual healthcare and medical dataset as in the 
COVID-19 case. With such dataset, we could help to translate into many 
low-resource communities. Indeed, use cases of our model in the healthcare and medical 
domains has significant value.  

In addition to the medical and healthcare domains, we are also keen in exploring 
the education domain in all subject areas and levels that are relevant to the low-resource communities. There are many textbooks and course materials that are in English, and many 
are not in low-resource languages. There is tremendous value in translating and providing 
access for such texts to help the low-resource communities to learn and flourish. This is 
especially relevant in language preservation and revival for the young generations. A low-resource 
language will have a better chance at flourishing when young children are learning and speaking 
the low-resource languages rather than English. Indeed, there is immense value in translating 
textbooks and educational materials into low-resource languages in al subject areas and levels that are relevant. 

Moreover, another example of future work that is very close to our research community is improving communication between mainstream rich-resource language communities and immigrants/refugees. Pittsburgh, like numerous European and North American cities, is increasingly welcoming a growing number of refugees and immigrants from various countries as the world becomes more globalized. While a lot of channels have been established to help these refugees and immigrants, most of these channels are in English. It would be much more helpful if such channels are in their own languages. For example, translating driving manuals into their languages would immensely help them to transition into cities like Pittsburgh. This not only helps the elderly population among these refugees and immigrants who may not speak or understand English, but also helps those that do to feel home. This will increase inclusivity and diversity. 

\subsubsection{Comprehensive Text Translation Tasks}
As we have discussed in Section \ref{limitation:task}, there are at least seven stages in the text translation process, among which 5 that we have not focused on in this thesis. 

\begin{enumerate}
\item 
\textbf{\ul{Language discovery stage:}} Building the friendly first contact, language discovery, and the discernment process to determine whether a language is a new language or a dialect, is very important.
\item 
\textbf{\ul{Language learning stage:}} Human translators learns and formalizes the orthography of the language.
\item 
\textbf{\ul{Language literacy stage:}} Human translators teach local communities to involve native speakers in the subsequent translation stages.
\item 
\textbf{\ul{Seed corpus translation stage:}} This is what we focused in detail in Chapter \ref{big:active} on active learning. 
\item 
\textbf{\ul{Iterative post-editing stage:}} This is what we focused in detail in Chapter \ref{big:large} on large pretrained models. 
\item 
\textbf{\ul{Quality control stage:}} This stage may involve multiple levels of focus groups, including human translators, quality control teams and native low-resource language speakers. 
\item 
\textbf{\ul{Document production stage:}} For any translation of a text, the final presentation of the document may include chapter headers,
footnotes, paragraph numbers, section headers, titles and many various forms of formatting components, and is important.
\end{enumerate}
This thesis accelerates Stage 4 and 5. In the future, we are interested in building the entire seven stages to complete full text translation process. This requires many number of years and resources, and is a very important process. 

\subsection{Improving Post-Editing User Experience}
In addition to exploring wider range of applications and datasets, 
we could also work with human translators in the 
post-editing process to improve translation. There are a number of ways that the machine translation 
system could optimize post-edits and use post-edits to expedite translation. 

\begin{figure*}
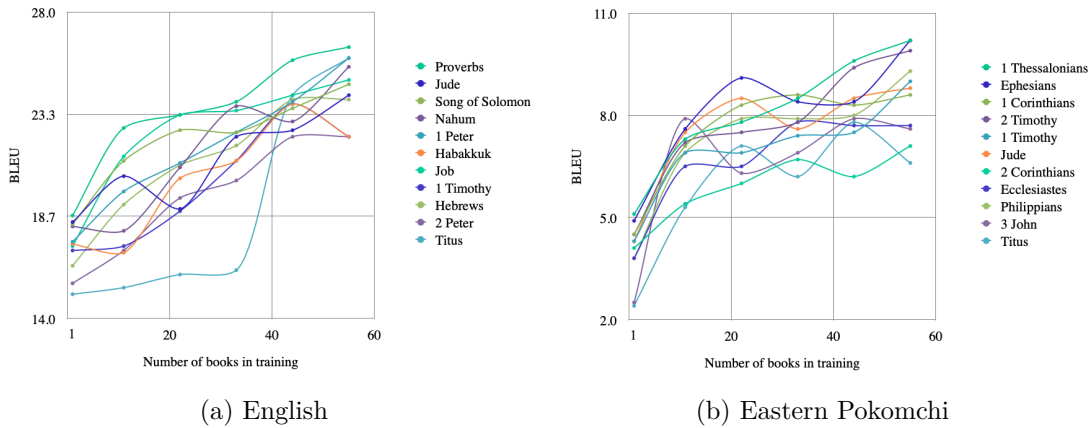

\centering
\begin{subfigure}{.45\textwidth}
  \centering
  \includegraphics[width=\linewidth]{curve_en.png}
  \caption{English}
  \label{fig:curve_en_hmt}
\end{subfigure}
\begin{subfigure}{.45\textwidth}
  \centering
  \includegraphics[width=\linewidth]{curve_ph.png}
  \caption{Eastern Pokomchi}
  \label{fig:curve_ph_hmt}
\end{subfigure}
\caption{Performance of the most difficult 11 books with increasing number of training books.}
\end{figure*}

\subsubsection{Iterative Post-Editing}
As discussed in Chapter \ref{big:active}, we show simulation of the iterative post-editing process for translation into English and Eastern Pokomchi 
in Figure \ref{fig:curve_en_hmt} and \ref{fig:curve_ph_hmt}.
After producing the first draft of the text by training a MT system on the seed corpus, we hold out the most difficult 11 books (the worst-performing 11 books) and set them aside as the test set for evaluating the entire iterative post-editing process. Taking the most difficult 11 books as the held-out test set, we divide the other 55 books of the Bible into 5 portions to simulate 5 iterations of post-editing process. Using this setup, we use Schedule \textit{B} in 5 iterations with increasing number of post-edited portions. Each portion contains 11 books, serving as post-edited portion for each iteration. In each iteration, we simulate human post-editing process by adding the actual translation of the given text portion to the MT system. MT system produces better and better drafts and we show the improvement using the most difficult 11 books. 

This simulation gives an understanding of the iterative post-editing, however, it is just a simulation. Going beyond simulation, we are interested in measuring and recording the real post-editing process by working with real-life text translation process. In real-life text translation process, there is coordination between different teams and translation stages. This coordination process may influence the improvement of the translation over the entire iterative process, and cannot be simulated. In the future, we look forward to see more work between teams in completing the iterative post-editing process in real-life.  

\subsubsection{Iterative Learning From Post-Edits}
The key of our research on Human Machine Translation is the 
close collaboration between human translators and machine translation. 
In our work, once human translators receive the machine translated draft, they 
post-edits a portion of the text. The machine then take this newly post-edited 
portions and add to the training data, and therefore produce a better draft. Iterations 
of post-editing and training improve and expedite translation. 

In addition to using post-edited data as additional training data, we could also 
learn from post-edits so that human translators do not need to repeat the same edits. 
For example, in the case of Sihuas Quechua, the correct spelling is ``chaymi'' instead of the machine translated ``tsaymi''. The edits involves changing ``ts'' to ``ch''. 
We could automate edits like this to save time for the human translators so that they do not have 
to repeat their edits. 

\subsubsection{Comprehensive Document Production}
In addition to learning from post-edits from human translators, machine systems could produce 
a comprehensive document translation rather than content translation. For any translation of a 
text, the final presentation of the document may include chapter headers, footnotes, paragraph numbers, 
section headers, titles and many various forms of formatting components. The document production process 
is therefore much more than the content translation. In this thesis, all of the input data are stripped of the formatting information including titles, headers, and footnotes. Consequently, training with such inputs produces outputs that also does not contain such information. In real-world 
document translation productions, these formatting issues are important. 
Therefore, we could aim to produce a comprehensive document translation in the future. 

To produce a comprehensive document translation, we could train on the completely labeled document with all titles, headers and footnotes as inputs and aim to produce outputs that contain such information. However, while some title and headers are well aligned, footnotes may not be. Indeed, including such information may render data alignment and structure incomplete. Incomplete data alignment and structure may affect the performance of our multilingual training. Additionally, neural network beyond structured data is still an active area of research \citep{tayefi2021challenges, sha2018order, bianchini2018deep}. Therefore, directly adding formatting information to training may not be the best solution. 

Instead of directly adding formatting information to training, we could add a pre-processing stage and a post-processing stage where we create an external data structure for the given document, and align titles and headers. For these aligned titles and headers, they could follow the same translation mechanism as the main content of the document. For footnotes, we can create a partial aligned data for training for the best possible translation. This method has a strong potential of maintaining accurate structured prediction. 

Furthermore, there are successes in industries where companies have successfully produced comprehensive document translations. In the future, we are open to collaborate with industry leaders to learn and to better our system so as to create a comprehensive document production process for human translators. 

\subsubsection{Adapting to Different Writing Styles}
In addition to comprehensive document production, our work could benefit from adapting to more 
varied writing styles. 
We notice the writing style of the test data is not uniform. For example, 
the text in the book of Mark
usually starts with conjunctives. ``en'', a Dutch conjunctive
word, is carried over to the translation output as
almost every test sentence begins with conjunctives. This is a unique 
writing style used by the author Mark, but is not shared with other test data 
which is written by different authors. 

When the writing styles of the test data differs from
the training data, machine translation systems encounter challenges.
We would like to improve the translation performance
of testing on entirely different text with different
writing styles. This is another way to minimize the post-editing efforts and improve user experience for human translators. 

\subsection{Moving Beyond Pretrained Models Limits}
As we have discussed in our limitations, the performance gain in this thesis is limited by the large pretrained models we use. However, our finding of adapting large pretrained models to the domain and then to the language is universal and generalizable to other tasks, domains and large pretrained models. 

\subsubsection{Current and Future Large Language Models}
Based on this thesis, in the future we want to focus on adapting or fine-tuning a current or future state-of-the-art multilingual Large Language Model to the task of translating a known text into a new, and low-resource target language. This task is largely three-fold: 
\begin{enumerate}
    \item Learning vocabulary from the target low-resource language.
    \item Learning to generate grammatical and coherent text in the target low-resource language.
    \item Learning to translate appropriate meaningful content into the target low-resource language.
\end{enumerate}

For the first sub-task, we could benefit from collaboration with human translators and working with any available monolingual data in the given new, low-resource language. In the case of Sihuas Quechua that we studied in Chapter \ref{big:confidence}, our human translators provides us a list of morphemes in Sihuas Quechua that they learned during the language discovery process before we work on text translation. This information is very valuable. In addition, if there is any existing monolingual data in the given new, low-resource language, it will contribute greatly to discovery of vocabulary and morphemes. There are large number of low-resource languages that are morphologically rich, having monolingual data is not just helpful for learning word vocabulary but also helpful for learning sub-word level morphemes. 
Indeed, morpheme-related research in machine translation is difficult but important. From our research, we find that if we replace the vocabulary from the BPE-based Large Pretrained Models (which all the Large Pretrained Models present currently are) with morpheme-based vocabulary, the results are very poor. In the future, this problem may be solved in multiple ways including training our own domain-specific pretrained models with morpheme-based vocabulary in the target new, low-resource language, which we will discuss in the next section. Morpheme-level research is a very important future direction. 

Moreover, the second and third sub-tasks could be tackled jointly. There is much we could do based on the result of this thesis. Our method of adapting to the domain and then to the language can still be used in today's swiftly-evolving and fast-growing field of Large Language Models. In the future, we are interested in working with current and future large pretrained models and perform multi-stage adaptation to the domain first and then to the new, low-resource language. 

Resolving these three sub-tasks in the future is very important. Additionally, we also would like to discuss training our own pretrained models suited to our task. 

\subsubsection{Training Domain-Specific Pretrained Models}
In addition, we will also benefit from training our own pretrained model in the future. 
If we have computing power to train a domain-specific 
pretrained large model on all languages in our dataset ($\sim$145 languages) 
using our dedicated vocabulary, we could achieve higher performance. If our dataset grows 
and we could train on more languages like a few key companies, we could contribute 
to the community by translating the text into any given language much more 
easily. 
When we train more future experiments, we want to work with 
more massively parallel systems. Indeed, we will benefit from more computing power and training our own domain-specific pretrained models. 

\begin{table}[t]
  \centering
  \small
  \begin{tabularx}{\textwidth}{p{2.8cm}p{1.0cm}p{1.8cm}XXX}
    \toprule
    Book & chrF & characTER & 4-gram BLEU & 1-gram BLEU & Number of Lines \\
    \midrule
    Esther & 0.868 & 0.142 & 57.4 & 83.2 & 167 \\
    1 Chronicles & 0.864 & 0.133 & 60.9 & 84.2 & 958 \\
    Haggai & 0.862 & 0.111 & 59.9 & 83.2 & 38 \\
    2 Chronicles & 0.859 & 0.14 & 57.1 & 81.7 & 824 \\
    Joshua & 0.85 & 0.154 & 56.2 & 81.6 & 658 \\
    2 Kings & 0.85 & 0.157 & 57.0 & 82.0 & 719 \\
    Ezra & 0.843 & 0.143 & 58.9 & 82.6 & 280 \\
    Jeremiah & 0.84 & 0.141 & 54.3 & 79.2 & 1365 \\
    1 Kings & 0.835 & 0.165 & 58.0 & 81.7 & 831 \\
    Habakkuk & 0.834 & 0.149 & 48.7 & 75.8 & 56 \\
    \bottomrule
    \end{tabularx}
  \caption{Top 10 ranked Old Testament books translating into Quechua Margos.}
    \label{table:top20}
\end{table}

\subsection{Overcoming Evaluation-Level Constraints} 

\subsubsection{More Fine-Grained Evaluation}
In addition to improving post-editing user experience for human translators, we are interested in more fine-grained evaluation. In this work, we have used a few different automatic evaluation metrics. In addition to the varying evaluation metrics, there are many different ways to create a more fine-grained 
evaluation mechanism. 

To build a more fine-grained evaluation system, we first need to understand that our test data is not uniform as we have discussed before. For example, the Bible has 66 books, covering different topics, genres and writing styles. The book of ``Psalms'' is poetry-based while Paul's letters like ``Romans'' are very philosophical. 
So far, we have looked at the the entire text as a whole. 
We would like to work with more fine-grained translation 
evaluation on all 66 books of the Bible.
In Table~\ref{table:top20}, we show the top 20 performing Old Testament books when we test on the Bible based on a small portion of Margos data. 
Choosing books that are best helped by Machine Translation could 
inform human translators the sequence of post-editing steps. We have explored this briefly in Chapter \ref{big:ipml}. In the future, we would like to work with human translators iteration-by-iteration as rankings produced by each iteration differs from each another. 

\subsubsection{Continued Dialogue with Native Speakers}

In translations into low-resource languages, oftentimes we as researchers 
do not speak these languages. This makes qualitative evaluation really hard. We may be 
able to understand the named entities. 
For example, in the case of translation into Eastern Pokomchi, 
we can read some of the named entities ``Jesús'', ``Galilea'', ``Simón''
and ``Andres'' in the machine translated text
``Eh noq ojik i rub{\textquotesingle}an i 
Jesús juntar i k{\textquotesingle}isa palaw i Galilea, xrilow 
reje i Simón ruch{\textquotesingle}ihil i Andres, re{\textquotesingle} i 
rutuut i k{\textquotesingle}isa palaw, ruum jinaj i k{\textquotesingle}isa 
palaw barco''. However, we cannot read and speak Eastern Pokomchi
which is a Mayan language and we do not know anybody
who speaks it. If we can work with native speakers
of Eastern Pokomchi, it would greatly help with
qualitative evaluation. 

There are continued conversations with field linguists and human translators working in the field. Understanding their needs is pivotal in building a long-term collaborative relationship that benefits both sides. This process is a continued dialogue. It is through this continue dialogue that we respect the dignity of both the indigenous low-resource language communities as well as the field linguists working in the field, while they learn to trust machine translation systems through time. We would like to continue this process of mutual trust and mutual understanding of both sides through continued conversations. 

\section{Broader Impact}
Our work is done with the intention of reviving and empowering low-resource languages, and more importantly, reviving the communities who speak low-resource languages. We want to lift up people and communities who are otherwise not in the spotlight of the world's attention. Lifting up low-resource language communities with this work, we aim to bring diversity, inclusivity, and equity to different language communities. And we want to provide NLP solutions that are inclusive and accessible to people across the world. 

This could bring voices to low-resource language communities, 
understanding needs of the elderly population among the low-resource language speakers, 
dissemination of infectious disease prevention information, 
welcoming immigrants and refugees from low-resource language communities with books in their own language, educating deaf children in the low-resource language communities, 
communicating and bridging the low-resource communities with the world through 
translating large literary texts, movies and songs into the low-resource language. 

\section{Key takeaways}

We have a few key takeaways from this thesis. Given data scarcity in the low-resource scenarios, we explore ways to effectively learn from massive source parallelism in Part \ref{part:part1}. We then build a human machine translation workflow for machine translation systems and human translators to work together seamlessly through active learning and large pretrained models in Part \ref{part:part2}. We show proof of concept that it is possible to produce quality translation draft of the whole text through as little as a few hundred lines ($\sim$3\% of the text) of the low-resource data. In addition to demonstrate that it is possible to translate using little resource, we show various ways to improve effectiveness and accuracy. 

Firstly, in Part \ref{part:part1}, we examine how source parallelism benefit translation of a given text into new, low-resource languages through massively multilingual training. In Chapter \ref{big:family}, we build cross-lingual transfer both within a given language family and also across different language families; we showed that training with two close-by families typically builds sufficiently good cross-lingual transfer in multilingual training. We also propose an order-preserving lexiconized machine translation model to resolve the variable binding problem and producing high quality lexiconized translations under severely low-resource scenarios. In Chapter \ref{big:paraphrase}, we treat paraphrases within the same language as foreign languages, and train on corpus-level paraphrases to improve translation performance. We find that our multi-paraphrase translation models improve performance better than multilingual models and improves the sparsity issue of rare word translation as well as diversity in lexical choice. 
In Chapter \ref{big:ipml}, we build our own linguistic distance metric based on translation distortion, fertility and performance. We propose a method, \textit{Iteratively Pretrained Multilingual Order-preserving Lexiconized Transformer} (IPML), to train on low-resource language data. We push the limit by using only $\sim$1,000 lines ($\sim$3.5\% of the entire text) to translate the whole text and achieved good translation performance using IPML. 

In Part \ref{part:part2}, having examined source parallelism, we build a human machine translation workflow algorithm for machine translation systems to collaborate with human translators to expedite the process.  
In Chapter \ref{big:active}, we first develop various active learning methods on known languages and transfer ranking to the new, low-resource language. Secondly, we activate the knowledge of large multilingual models by proposing multilingual and multi-stage adaptations through different training schedules in Chapter \ref{big:large}; we find that adapting pretrained models to the domain and then to the low-resource language works best. Thirdly, we aggregate scores from 115 languages to provide a universal ranking and increase robustness by \textit{relaxed memoization} method. In Chapter \ref{big:confidence}, having examined both source parallelism and human machine translation workflow, we evaluate our work by translating academic progress to the real-world translation process in a case study of the Quechuan language family. We collaborate extensively with a translation group with in-depth knowledge of various Quechuan languages and focus on evaluation. We find that machine translation performance is significantly positively correlated with language similarity. The more connected a language is, the better it is to translate into this language. Furthermore, decluttering poorly-connected languages improves translation performance. Using this finding, we show our results in translating into a new, low-resource language called Sihuas Quechua.

\removelabelprefix

\backmatter

\renewcommand{\bibsection}{\chapter{\bibname}}
\bibliographystyle{plainnat}                                                                                          
\bibliography{thesis} 

\end{document}